\documentclass{article}

\usepackage{microtype}
\usepackage{graphicx}
\usepackage{subcaption}
\usepackage{booktabs} 
\usepackage{ifthen}   
\usepackage{xcolor}   

\usepackage{hyperref}

\usepackage{xcolor}
\usepackage{ifthen}

\newcommand{\ShowComments}{no} 

\newcommand{\LEI}[1]{}
\newcommand{\shenyang}[1]{}
\newcommand{\yaoqing}[1]{}
\newcommand{\addressedshenyang}[1]{}
\newcommand{\tianyu}[1]{}
\newcommand{\PY}[1]{}
\newcommand{\PYB}[1]{}
\newcommand{\addressedyaoqing}[1]{}
\newcommand{\addressedvigk}[1]{}
\newcommand{\Oli}[1]{}
\newcommand{\vigk}[1]{}

\ifthenelse{\equal{\ShowComments}{yes}}{
  \renewcommand{\LEI}[1]{\textcolor{orange}{LEI: #1}}
  \renewcommand{\shenyang}[1]{\textcolor{brown}{shenyang: #1}}
  \renewcommand{\yaoqing}[1]{\textcolor{red}{[Yaoqing: #1]}}
  \renewcommand{\addressedshenyang}[1]{\textcolor{cyan}{[(Addressed) shenyang: #1]}}
  \renewcommand{\tianyu}[1]{\textcolor{blue}{TIANYU: #1}}
  \renewcommand{\PY}[1]{\textcolor{blue}{PY: #1}}
  \renewcommand{\PYB}[1]{\textcolor{blue}{[PY: #1]}}
  \renewcommand{\addressedyaoqing}[1]{\textcolor{cyan}{[(Addressed) Yaoqing: #1]}}
  \renewcommand{\addressedvigk}[1]{\textcolor{cyan}{[(Addressed)vigk: #1]}}
  \renewcommand{\Oli}[1]{\textcolor{purple}{[Oli: #1]}}
  \renewcommand{\vigk}[1]{\textcolor{violet}{[VK: #1]}}
}{}

\usepackage[accepted]{icml2026/icml2026}

\usepackage{amsmath}
\usepackage{amssymb}
\usepackage{mathtools}
\usepackage{amsthm}

\usepackage{amsmath,amsfonts,bm}

\def\eqref#1{equation~\ref{#1}}

\def\1{\bm{1}}

\def\vzero{{\bm{0}}}
\def\vone{{\bm{1}}}

\def\vxi{{\bm{\xi}}}
\def\va{{\bm{a}}}

\def\vx{{\bm{x}}}
\def\vy{{\bm{y}}}

\def\mA{{\bm{A}}}
\def\mB{{\bm{B}}}

\def\mE{{\bm{E}}}

\def\mG{{\bm{G}}}

\def\mI{{\bm{I}}}

\def\mM{{\bm{M}}}

\def\mO{{\bm{O}}}

\def\mQ{{\bm{Q}}}
\def\mR{{\bm{R}}}
\def\mS{{\bm{S}}}

\def\mW{{\bm{W}}}
\def\mX{{\bm{X}}}
\def\mY{{\bm{Y}}}

\DeclareMathAlphabet{\mathsfit}{\encodingdefault}{\sfdefault}{m}{sl}
\SetMathAlphabet{\mathsfit}{bold}{\encodingdefault}{\sfdefault}{bx}{n}

\def\gN{{\mathcal{N}}}

\def\sN{{\mathbb{N}}}

\def\sR{{\mathbb{R}}}
\def\sS{{\mathbb{S}}}

\usepackage{mathrsfs}

\newcommand{\norm}[1]{\left\lVert#1\right\rVert}

\def\vzero{{\mathbf{0}}}
\def\vone{{\mathbf{1}}}

\def\vbeta{{\boldsymbol{\beta}}}

\newtheorem{theorem}{Theorem}[section]

\newtheorem{proposition}[theorem]{Proposition}
\newtheorem{lemma}[theorem]{Lemma}
\newtheorem{corollary}[theorem]{Corollary}
\newtheorem{definition}[theorem]{Definition}
\newtheorem{assumption}[theorem]{Assumption}

\newtheorem{remark}[theorem]{Remark}

\usepackage[capitalize,noabbrev]{cleveref}

\usepackage[textsize=tiny]{todonotes}

\icmltitlerunning{Balancing Learning Rates Across Layers: Exact Two-Step Dynamics and Optimal Scaling in Linear Neural Networks}

\begin{document}

\twocolumn[
\icmltitle{Balancing Learning Rates Across Layers: Exact Two-Step Dynamics and Optimal Scaling in Linear Neural Networks}



  \icmlsetsymbol{equal}{*}

  \begin{icmlauthorlist}
    \icmlauthor{Tianyu Pang}{Dartmouth}
    \icmlauthor{Vignesh Kothapalli}{Stanford}
    \icmlauthor{Shenyang Deng}{Dartmouth}
    \icmlauthor{Haohui Wang}{VT}
    \icmlauthor{Dawei Zhou}{VT}
    \icmlauthor{Yaoqing Yang}{Dartmouth}
  \end{icmlauthorlist}

  \icmlaffiliation{Dartmouth}{Dartmouth College}
  \icmlaffiliation{Stanford}{Stanford University}
  \icmlaffiliation{VT}{Virginia Tech}

  \icmlcorrespondingauthor{Yaoqing Yang}{Yaoqing.Yang@dartmouth.edu}

  \icmlkeywords{Machine Learning, ICML}

  \vskip 0.3in
]



\printAffiliationsAndNotice{}  

\begin{abstract}
We study optimal learning-rate selection in two-layer and three-layer linear neural networks trained to learn linear
target functions. In particular, we derive the exact closed-form expressions for the gradients and test loss after one and two steps of gradient descent, enabling a precise characterization of early training dynamics. 
We characterize how learning rates should scale under the gradient approximation in the first two steps, and prove that performing updates with this approximation yields a tractable surrogate loss with a tight, small approximation error. This formulation enables the theoretical analysis of layer-wise learning rates and reveals a distinct early-training regime: test loss can be minimized by unequal learning rates at the initial step, while equal learning rates become optimal in subsequent steps. Our numerical experiments validate the theory
and demonstrate the importance of balancing layer-wise learning rates early during training. The code is available at: \href{https://github.com/TDCSZ327/Layer-Balancing}{\texttt{TDCSZ327/Layer-Balancing}}.

\end{abstract}

\section{Introduction}

The dynamics of gradient descent in deep neural networks are shaped not only by the architecture and initialization ~\citep{saxe2019mathematical, Goldt_2020, kunin2024get} but also by the choice of learning rates for individual layers. In practice, networks often use separate learning rates across layers, or adopt layer-wise adaptive schedules ~\citep{you2017large, zhou2023temperature, wang2025sharpness, he2026one} to accelerate convergence and improve generalization. Even in simpler linear overparameterized models, however, it is unclear how the relative scale of layer-wise learning rates affects training trajectories and resulting test performance. Small deviations in early gradient updates can propagate through layers
and affect learned representations in ways that are difficult to quantify.

In neural networks, the gradient updates tend to couple dominant, signal-aligned components with smaller residual terms~\citep{ba2022high, wang2023spectral, kothapalli2025spikes}. Since the norms of these components vary across layers, the learning rates play a crucial role in determining the training dynamics. A theoretical analysis of these components requires a layer-by-layer training assumption that is atypical of practical settings. On the other hand, linear networks offer a rich alternative where the linear interaction between layers 
naturally determines the structure of the gradient~\citep{saxe2014exactsolutionsnonlineardynamics}. In particular, each layer's update depends on the product of the weights in the other layers and the data matrix. However, the effects of layer-wise learning rates on the learning dynamics are still not well understood.

Previous works on continuous-time gradient flow analyses in linear networks suggest that the weight norms across layers may balance over time~\citep{du2018algorithmic, ye2021global,wang2021large}. Whereas kernel-based approximations predict nearly linear evolution of outputs~\citep{jacot2018neural, hu2022universality}. However, these approaches do not study the effects of learning rate selection in discrete, finite-step settings on the signal-residual coupling in gradient updates.
This coupling effect is further complicated by network depth. Even for three-layer linear networks, the gradient of the output with respect to an intermediate layer contains products of multiple weight matrices and the data matrix, leading to higher-order interactions that influence both the magnitude and direction of updates. Approaches that consider layers independently fail to capture these effects~\citep{arora2018convergence}, and conventional mean-field~\citep{Mei_2018} or maximal-update~\citep{yang2021tensor,yang2022tensor} analyses tend to rely on infinitesimal step sizes.
Consequently, predicting how the choice of layer-wise learning rates affects early generalization requires a framework that can both isolate the leading components of the gradient and quantify their contribution to test performance.

Previous work on layer-wise adaptation has focused primarily on heuristic or asymptotic regimes. Methods such as per-layer decay, adaptive optimizers~\citep{zhou2023temperature,liu2024model}, or normalization-based rescaling~\citep{you2018imagenet,yang2022tensor} are motivated by empirical improvements but do not offer explicit formulas linking learning rates to test loss. Analyses of implicit bias~\citep{arora2019implicit,gidel2019implicitregularizationdiscretegradient} or norm balancing describe certain asymptotic trajectories, yet they do not address the finite-step dynamics where early layer-wise interactions are critical. In multi-layer settings, these interactions determine whether early updates align with the signal or are dominated by cross-layer interference, and small differences in learning rates can have a disproportionate effect on generalization.

In this paper, we develop a framework for analyzing layer-wise learning rates in two-layer and three-layer linear networks under random orthogonal initialization. Central to our approach is a gradient decomposition that separates the dominant, label-aligned component of each layer's update from smaller residual terms, allowing closed-form expressions for the test loss after one and two gradient steps. Our contributions include: 

\begin{itemize}
    \item Characterizing the dominant components of gradients and rigorously bounding residual terms in operator norm, establishing conditions under which approximate gradients accurately capture test loss dynamics.
    \item Showing that in two-layer networks, symmetric learning rates across layers are suboptimal after a single update due to the distinct roles of representation and readout, but become locally optimal after two updates in sufficiently wide networks, revealing a transition from asymmetric to balanced learning rates.
    \item Extending the analysis to three-layer networks with a scalar output,\addressedyaoqing{I thought the three-layer case uses a scalar output?} capturing richer cross-layer interactions, identifying distinct scaling regimes for admissible learning rates, and providing explicit test loss expressions that include higher-order interactions between layer updates.
    \item Identifying critical thresholds for learning rates (\(\eta = O(h\sqrt{h})\) for two layers, \(\eta = O(h)\) for three layers) beyond which gradient dynamics and test loss behavior qualitatively change, connecting with maximal-update and mean-field scaling regimes.
\end{itemize}

While our analysis is restricted to linear networks with orthogonal initialization, it provides a principled foundation for understanding how layer-wise learning rates shape early generalization, and offers insights that can guide the design of learning rate schedules in more complex architectures.

\section{Related Work}
\label{sec:related_work}


\subsection{Layer-wise Hyperparameter Tuning}
When training and fine-tuning deep learning models, layer-wise hyperparameter tuning serves as a lightweight and memory-efficient tuning paradigm~\citep{yao2024layer}. It has shown great potential to reconcile the coarse granularity of global tuning~\citep{loshchilov2016sgdr,hu2024minicpm} and the high memory demands of parameter-wise tuning~\citep{kingma2014adam, liu2019variance,yao2021adahessian}.
For example,~\citet{howard2018universal, long2015learning} have shown layer-wise learning rate strategies can enhance test accuracy in both transfer learning and domain adaptation tasks. $\texttt{LARS}$ and $\texttt{LAMB}$~\citep{you2017large,you2018imagenet} propose ``trust ratio'' to assign layer-wise learning rates and mitigate gradient divergence in large-batch training. 
accelerating the training of large models on computer vision (CV) and natural language processing (NLP) tasks.
$\texttt{AutoLR}$~\citep{ro2021autolr}  automatically tunes its layer-wise learning rates according to the ``role'' of each layer to balance layer-wise weight variations.
$\texttt{Adam-mini}$~\citep{zhang2024adam} and $\texttt{Blockwise-LR}$~\citep{wang2025sharpness} \addressedyaoqing{Either also say the name of the second paper or use citet. The first way sounds better.} assign layer-wise learning rate based on the different Hessian block structures in Transformers~\citep{vaswani2017attention}. Complementing these algorithmic heuristics, in this work we provide an exact two-step characterization in two- and three-layer linear networks that links layer-wise learning-rate allocation directly to test loss, yielding a principled prescription for when asymmetric versus balanced learning rates are optimal. We include more discussion of other layer-wise parameters, such as the pruning ratio and weight decay, in Appendix~\ref{app:more_realted_work}.



\subsection{Layer-balancing Phenomenon}
\label{sec:related_work_layer_balancing}
Prior work like~\citet{du2018algorithmic, wang2021large, ye2021global} shows that the norm difference between adjacent layers of deep homogeneous models stays constant or vanishes during training, they term this as one kind of automatic layer balancing. More recently, ~\citet{zhou2023temperature, liu2024model} find that balancing weight spectra across layers helps model training, and they propose a layer-wise learning rate scheduler, called \texttt{TempBalance}, that allocates learning rates by assessing the heavy-tailness of each layer (a property that correlates with layer quality). 

In addition, \citet{kunin2024get} study the training dynamics of linear neural networks and find that when all layers learn at similar rates, linear neural networks exhibit rapid feature learning. Likewise, \citet{yang2022tensor} and \citet{yang2024tensor} show that for square matrices, balanced learning-rates can be optimal under the maximal-update parameterization. Both results are consistent with our findings.

\section{Preliminaries and Setup}
\label{sec:setup}


\paragraph{Notation.} For $n \in \sN$, we denote $[n] = \{1, \cdots, n\}$. We use $O(\cdot)$ to denote the standard big-O notation and the subscript $O_d(\cdot)$ to denote the asymptotic limit of $d \to \infty$. Formally, for two sequences of real numbers $x_d$ and $y_d$, $x_d = O_d(y_d)$ represents $\lim_{d \to \infty} |x_d| \leq C_1 |y_d|$ for some constant $C_1$. Similarly, $x_d = O_{d, \mathbb{P}}(y_d)$ denotes that the asymptotic inequality almost surely holds under a probability measure $\mathbb{P}$. The definitions can be extended to the standard $\Omega(\cdot), \Theta(\cdot)$  notations analogously.
For two sequences of real numbers $x_d$ and $y_d$, $x_d\asymp y_d$ represents $ C_2 |y_d| \le |x_d| \le C_1 |y_d|$, for constants $C_1, C_2 > 0$
For a real matrix $\mB=(B_{ij})_{n\times m} \in \sR^{n \times m}$, $\mB^{\circ p}$ represents an element-wise $p$-power transformation such that $\mB^{\circ p}=(B_{ij}^{p})_{n\times m}$. $\odot$ is the matrix Hadamard product, $\operatorname{sign}(.)$ denotes the element-wise sign function. $\norm{\cdot}$ denotes the $\ell_2$ norm for vectors and the operator norm for matrices. $\norm{\cdot}_{F}$ denotes the Frobenius norm.
$\vzero_{h \times d}, \vone_{h \times d} \in \sR^{h \times d}$ represent the all-zero and all-ones matrices, respectively. 

\begin{definition}[Orthogonal initialization]
      We say a random matrix $\mO_1   \in \mathbb{R}^{h \times h}$  is random orthogonal if it is uniformly distributed on the orthogonal group with respect to the Haar measure, i.e. $\mO_1\mO_1^{\top}=\mO_1^{\top}\mO_1 =\mI_h$.  We say  a  random vector $\va_1   \in \mathbb{R}^{h }$  is random orthogonal if it is uniformly distributed on the orthogonal group with respect to the Haar measure, i.e. $\va_1^{\top}\va_1 =1,$ and $\mathbb{E}[\va_1\va_1^{\top}]=\frac{1}{h}\mI_h$.
\end{definition}
\begin{assumption}
\label{sec:assumption-whiten}
For random orthogonal initialization of the NN weights, we assume  data number $n =$ model width $h =$ data dimension $d$.
\end{assumption}




\paragraph{Two-layer and three-layer NNs.} 
For the two-layer case, we consider a linear NN  as our \textit{student} model $f(\cdot) : \sR^h \to \sR^h$. For an input $\vx_i \in \sR^h$, its prediction is formulated as:
\begin{align}
    \label{update_formula_1}
    f(\vx_i) &= \frac{1}{h} \vx_i^{\top}\mW_1 \mW_2.
\end{align}
For the three-layer case, we consider a linear NN  as our \textit{student} model $f^*(\cdot) : \sR^h \to \sR$. For an input $\vx_i \in \sR^h$, its prediction is formulated as:
\begin{align}
    \label{update_formula_2}
    f^*(\vx_i) &= \frac{1}{\sqrt{h}} \vx_i^{\top}\mW_1 \mW_2 \va.
\end{align}
Here $\mW_1 \in \sR^{h \times h},\mW_2 \in \sR^{h \times h}, \va \in \sR^h$ are the first two hidden layers and last layer weights, respectively, with random orthogonal initialization. To keep the loss well-scaled,  we use a scaling coefficient of $\frac{1}{h}$ for the two-layer network and $\frac{1}{\sqrt{h}}$ for the three-layer network.
 \addressedyaoqing{Give a one-sentence explanation of the scaling coefficients.}

\paragraph{Dataset.}
We use linear \textit{teacher} models to generate the training data of both two-layer and three-layer \textit{student} networks under random orthogonal initialization.
We sample input data $\mX \in \mathbb{R} ^{h\times h}$ as $h$ data points $\{\vx_1, \cdots, \vx_h\}$, where $\frac{1}{\sqrt{h}}\mX$ is a random orthogonal matrix.  To simplify the analysis, we do not consider label noise.

\begin{itemize}
    \item \textbf{Two-layer NN Case. } For a given $\vx_i \in \sR^h$, we use a linear teacher model $F:\sR^h \to \sR^h$ to generate the corresponding label $\vy_i \in \sR^h$ \citep{du2018algorithmic} as:
\begin{equation}
\label{eq:target_2_layer_nn}
    \vy_i = F(\vx_i)  = \mM^{\top} \vx_i .
\end{equation}
Here, $\mM \in \sR^{h \times h}$ is the \textit{target matrix}, where $\sqrt{h}\mM$ is a random orthogonal matrix. We represent $\mX \in \sR^{h \times h}, \mY \in \sR^{h\times h}$ as the input matrix and the label matrix, respectively.
    \item \textbf{Three-layer NN Case. } For a given $\vx_i \in \sR^h$, we use a linear teacher model $F^*:\sR^h \to \sR$ to generate the corresponding scalar label $y_i \in \sR$ as:
\begin{equation}
    y_i = F^*(\vx_i)  = \vbeta^{*\top} \vx_i.
\end{equation}
Here random orthogonal vector $\vbeta^* \in \sR^h$  is the \textit{target direction}.
We represent $\mX \in \sR^{h \times h}, \vy \in \sR^h$ as the input matrix and the label vector, respectively.\addressedyaoqing{It's OK to say $h\times h$. It's also OK to say $h\times n$. But you have to make it consistent. You can either say $h\times h$ in all places, or you can say $h\times n$ and then say $h=n$. Currently, the two ways are mixed.} \tianyu{ will revise the notation to only say  $h\times h$.}
\end{itemize}

\paragraph{Training procedure.}
We adopt GD as the optimizer for training. Here we consider a simple training procedure: For the two-layer NN, we apply GD updates on both the layers simultaneously.
For the three-layer NN,  each GD update  only  simultaneously trains the first two hidden layers and we fix the last layer weights $\va \in \sR^h$. In both settings, we employ the mean-squared error as our training loss function: 
\begin{align}
\label{eql2}
    \hat{L}_{\text{two-layer}}(f,\mX, \mY)=\frac{1}{2h}\norm{\mY-f(\mX)}_F^2 \\
\label{eql3}
    \hat{L}_{\text{three-layer}}(f^*,\mX, \vy)=\frac{1}{2h}\norm{\vy-f^*(\mX)}^2.
\end{align}
Where $\mX \in \mathbb{R}^{h\times h}, \mY \in \mathbb{R}^{h \times h}, \vy \in \mathbb{R}^{h}$ are training data, and labels for two-layer and three-layer NN, respectively.

\begin{assumption}
\label{sec:assumption}
We consider a non-asymptotic setting, where $h,n \leq C$ and $C$ is a large constant. 
\end{assumption}
\begin{assumption}
\label{assumption:lr}
We aim to determine whether using the same learning rates across layers leads to minimal test loss for networks trained with a one-step or two-step GD update when $\eta_1+\eta_2=2h^{\alpha}$, where $h^{\alpha}\leq$ \textit{critical threshold} (\( O(h\sqrt{h})\) for two-layer NN, \( O(h)\) for three-layer NN).
\addressedyaoqing{Define "critical threshold".}
\end{assumption}

\section{Overview of Main Results: Balancing Layer-wise Learning Rates}

In this work, we study how layer-wise learning rates influence early training dynamics and generalization in linear neural networks. Our analysis reveals how the interplay between network width, training depth, and learning-rate scale drives the emergence of asymmetric versus balanced updates across layers.

\paragraph{Gradient decomposition and leading-order approximation.} 
For both two-layer and three-layer networks, we decompose the exact gradients into leading-order, signal-aligned terms and smaller residual terms:
\[
\mG_i^t = \mB_i^t-\mA_i^t  , \quad i=1,2,
\]
where $\mA_i^t$ captures the primary contribution from the labels, and $\mB_i^t$ is higher-order in  $\frac{1}{\sqrt{h}}$. For learning rates below the critical thresholds ($\eta_1, \eta_2 \le O(h\sqrt{h})$ for two-layer, $\eta_1, \eta_2 \le O(h)$ for three-layer), $\mB_i^t$ is negligible in norm. This decomposition justifies replacing exact gradients with their leading-order components when computing the test loss, simplifying analysis and revealing the dominant factors that govern learning-rate balance.

\begin{figure}[tb]
    \centering
    \begin{subfigure}[t]{0.48\linewidth}
        \centering
        \includegraphics[width=\textwidth]{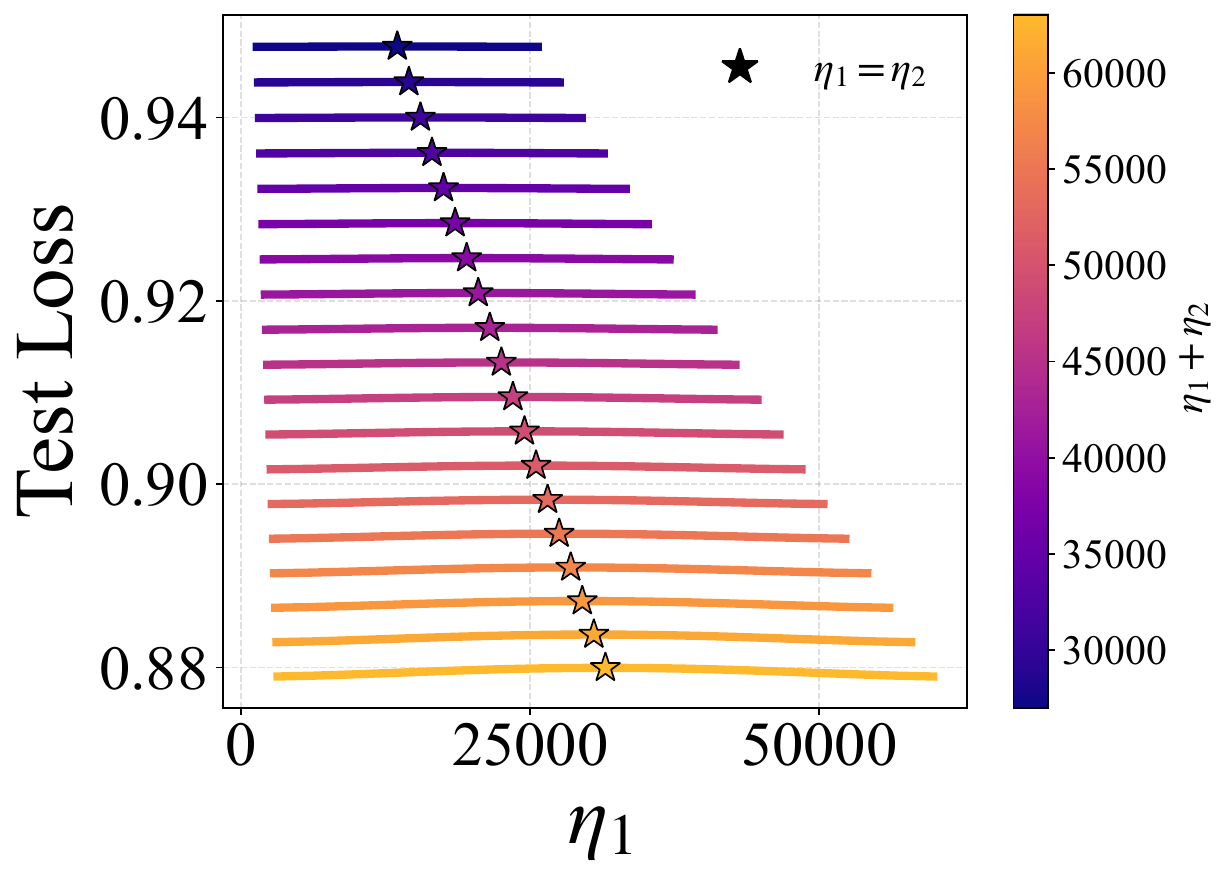}
        \caption{  1-step (theory) }
    \end{subfigure}
    \hfill
        \begin{subfigure}[t]{0.48\linewidth}
        \centering
        \includegraphics[width=\textwidth]{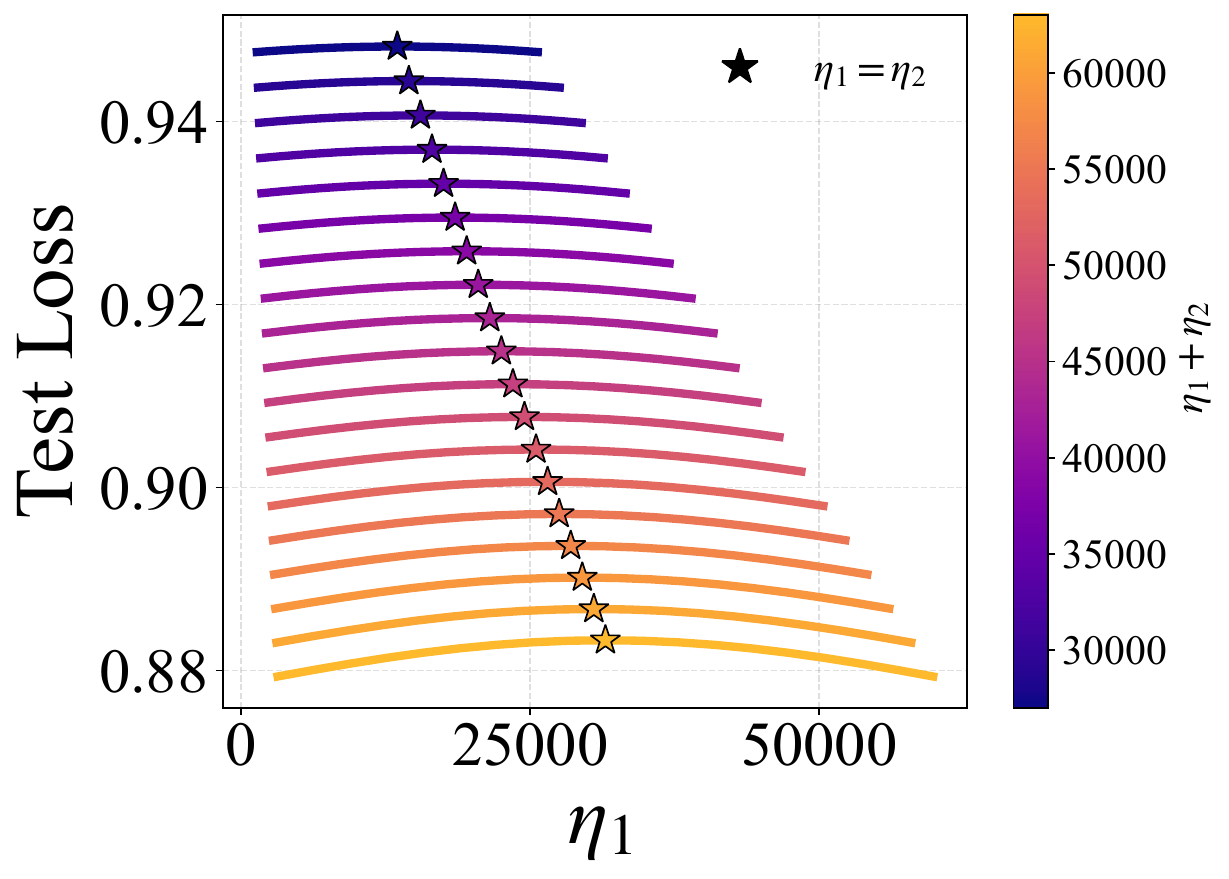}
        \caption{  1-step (experiment) }
    \end{subfigure}
        \hfill
    \begin{subfigure}[t]{0.48\linewidth}
        \centering
        \includegraphics[width=\textwidth]{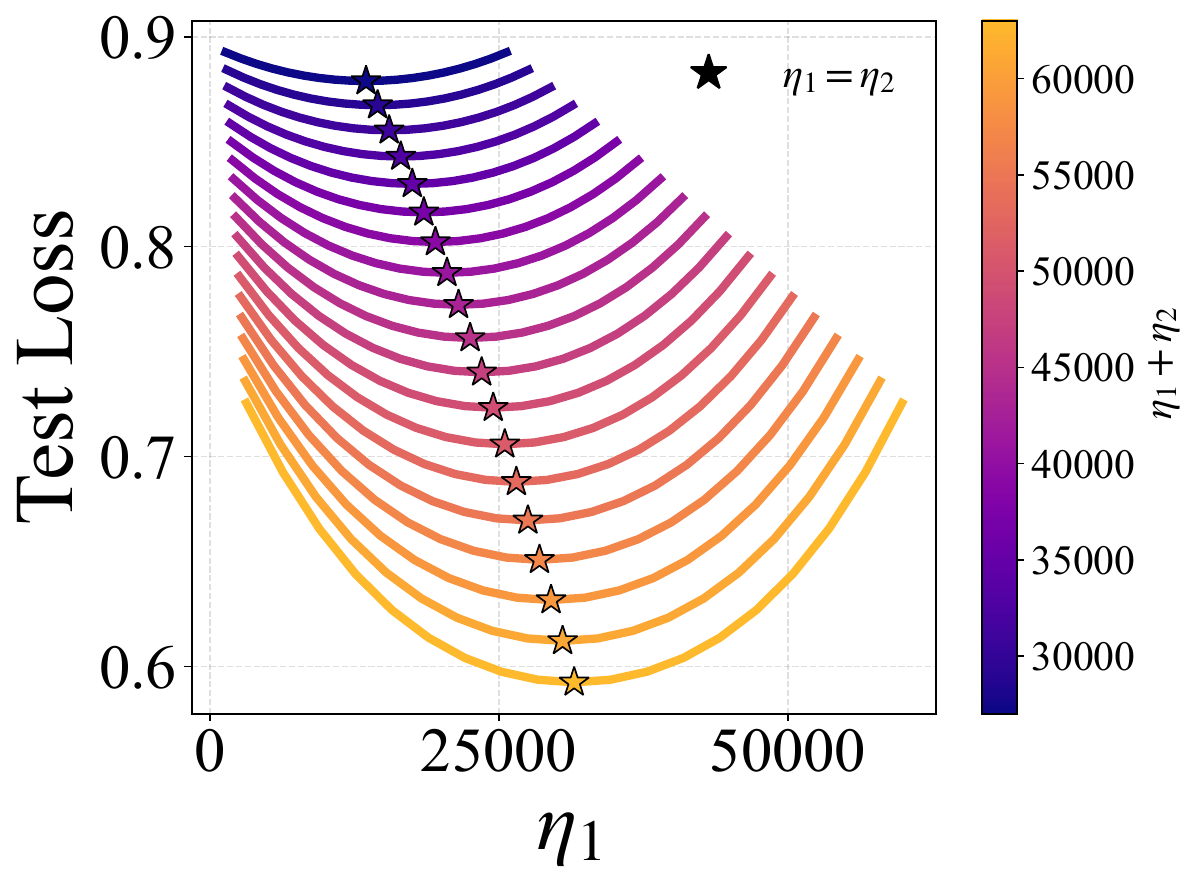}
        \caption{   2-step (theory)  }
    \end{subfigure}
    \hfill
    \begin{subfigure}[t]{0.48\linewidth}
        \centering
        \includegraphics[width=\textwidth]{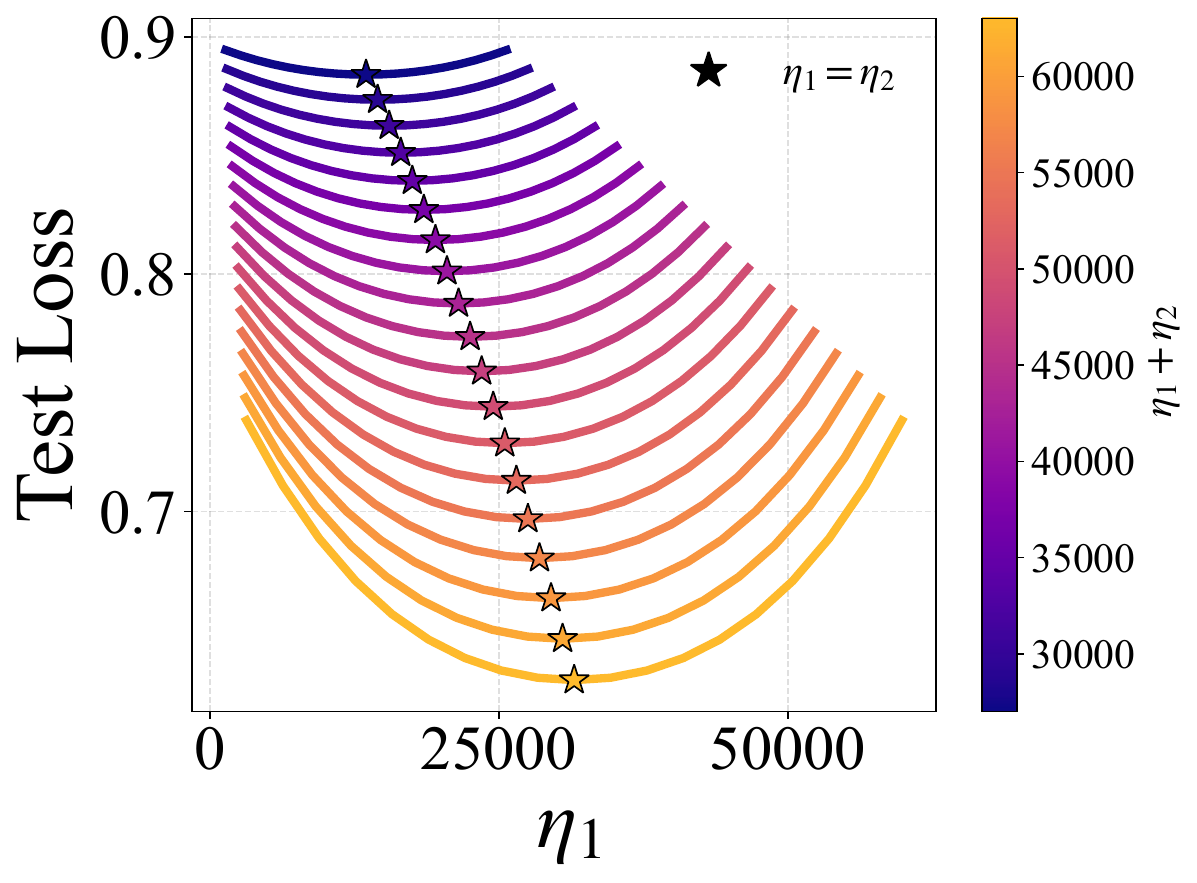}
        \caption{  2-step (experiment)  }
    \end{subfigure}
    \caption{ \textbf{ 2-layer NN under orthogonal initialization.} Here we set $\eta_1+\eta_2\leq O(h^{\frac{3}{2}})$  and  $h=1000$. We observe that the theoretical losses  closely track the empirical test losses measured after either one or two updates. Moreover, a clear qualitative shift emerges: after a single update, symmetric learning rates across layers are suboptimal, whereas after two updates they become locally optimal. We discuss this in Section~\ref{sec:exp}.\yaoqing{Did we discuss this figure?} \tianyu{In Experiment section.}\addressedyaoqing{In that case, it might be better to give a pointer or move the figure there.}
    \vigk{Lets follow the same plotting style as Fig 1 here as well.}\tianyu{Since Fig2 uses larger learning rates than Fig1, here both $\eta_1$ and $\eta_2$ are on the order of $O(h^{3/2})$. In this regime, our theory predicts that the two-step update incurs an $O(1/\sqrt{h})$ error in the loss (roughly $0.03$--$0.05$ when $h=1000$). I tried plotting the theory and experiments together; however, although  I try to adjust the $y$-axis, they still do not appear so closely aligned. Therefore, I plot them separately, but mark the point $\eta_1=\eta_2$ with a star. Fig3 follows the similar way.
    \vigk{If our theory incurs the error then it is better to show it and then say that it is expected. Otherwise, we can also add that small error term to our theory values and bring it closer to the experiment.}
}
}
\label{fig:2-NN-orthogonal}
\end{figure}

\paragraph{Two-layer networks: asymmetry vs. balance.} 
For the student network $f(\vx) = \frac{1}{h}\vx^\top \mW_1 \mW_2$ trained on labels $\vy = \mM^\top \vx$, we find:
\begin{itemize}
    \item \textit{Early asymmetry:} After a single gradient step, symmetric learning rates $\eta_1 = \eta_2$ do not minimize the test loss. Early training favors asymmetric updates, as the first layer primarily absorbs the label signal while the second layer transmits it. Enforcing symmetry too early limits the network's ability to exploit this distinction.
    \item \textit{Emergent balance:} After two steps, for sufficiently large width and an appropriate range of total learning-rate scale, symmetric layer-wise learning rates become locally optimal. At this stage, layers have coordinated sufficiently, making balanced updates advantageous for minimizing test loss.
\end{itemize}

\paragraph{Three-layer networks: extension and scaling.} 
For the student network $f^*(\vx) = \frac{1}{\sqrt{h}}\vx^\top \mW_1 \mW_2 \va$ with output vector $\va$, the qualitative phenomena is similar to the two-layer NNs. Early training favors asymmetric learning rates, reflecting distinct roles of hidden layers. After multiple steps, balanced learning rates emerge as optimal, but the admissible learning-rate regime is reduced to $O(h)$ due to differences in the initialization schemes of the two- and three-layer NNs, as well as the additional output layer. Two-step test loss also depends more strongly on higher-order products of $\eta_1 \eta_2$, highlighting deeper cross-layer interactions.

\paragraph{Key takeaway: balancing learning rates across layers.} 
Our results show that optimal layer-wise learning rates are dynamic. Early-stage training benefits from asymmetry to leverage layer-specific signal propagation, while deeper or later-stage updates promote balance, enabling coordinated alignment across layers. This perspective connects explicit gradient norms, test loss formulas, and width scaling to a principled understanding of when and why learning-rate balancing is beneficial in linear NNs. In Appendix~\ref{app:gaussian_initialization} and~\ref{app:more_experiments}, we extend the main results  under orthogonal initialization to the gaussian initialization setting, obtaining a theoretical loss expression for the one-step GD update and complementing it with simulation experiments for the multi-step case, we find that similar conclusions  hold.

\section{Main Results}

\subsection{Norm Analysis of Gradient Matrices 
}
\label{sec:orthog-norm}

Here we give the norm analysis of  update gradient matrices under random orthogonal  initialization. This analysis is an important step in simplifying the derivation of the theoretical test loss in the next Section (Section~\ref{sec:orthog-test_loss}). It also provides intuition about the range of learning rates that are beneficial for model training and offers a deeper understanding of the gradient matrices. Here, we examine the norm properties of the gradient matrices during one-step and two-step updates under both the two-layer and three-layer NN settings. We take two-layer NN case as a main example. The $t$-step update equations for the two-layer NN are as follows:
\begin{equation}
  \begin{aligned}
{\mW_1^{t}} = {\mW_1^{t-1}} - \eta_1 {\mG_1^{t-1}};\hspace{7pt}
{\mW_2^{t}} = {\mW_2^{t-1}} - \eta_2 {\mG_2^{t-1}},
\end{aligned}  
\end{equation}
where ${\mW_1^{t}}, {\mW_2^{t}}$ are two hidden layer weights after $t$-step update, $\eta_1$ and $\eta_2$ are the learning rate for the first  layer and second layer, respectively. ${\mW_1^{t}}$ and ${\mW_2^{t}}$ are the updated layer weights. ${\mG_1^{t-1}}$ and ${\mG_2^{t-1}}$ are the corresponding $t-$step exact gradient matrices, where:\addressedyaoqing{What is that prime $\prime$?} \tianyu{Removed}
\begin{align*}
{\mG_1^{t-1}} &= \frac{1}{h^2} {\mW_1^{t-1}}{\mW_2^{t-1}}{\mW_2^{t-1}}^{\top}-\frac{1}{h^2} \mX^{\top}\mY\mW_2^{{t-1}^{\top}},   \\
{\mG_2^{t-1}} &= \frac{1}{h^2}{\mW_1^{t-1}}^{\top} {\mW_1^{t-1}}{\mW_2^{t-1}}-\frac{1}{h^2}{\mW_1^{t-1}}^{\top}\mX^{\top}\mY.
\end{align*}


Since the gradients incorporate label information, we decompose each gradient matrix into two constituent components:
\begin{align}
    \mG_\ell^t = \mB_\ell^t - \mA_\ell^t, \quad \ell \in \{1,2\},
\end{align}
where the matrices $\mA_\ell^t$ correspond to data-aligned gradient components, while $\mB_\ell^t$ capture self-interaction effects arising from weight Gram matrices. Specifically for $\ell=1$:\addressedyaoqing{Maybe this is a stupid question. Is the update in (8) the same as the equation above it if we plug in (9) and (10)?} \tianyu{Yes.}
\begin{equation}
    \mA_1^t = \frac{1}{h}\mM \mW_2^{t\top}, \quad
    \mB_1^t = \frac{1}{h^2}\mW_1^t \mW_2^t \mW_2^{t\top},
\end{equation}
and for the second layer $\ell=2$,
\begin{equation}
    \mA_2^t = \frac{1}{h}\mW_1^{t\top}\mM, \quad
    \mB_2^t = \frac{1}{h^2}\mW_1^{t\top}\mW_1^t\mW_2^t.
\end{equation}
The $\mA_\ell^t$ terms describe how the network weights align with the target matrix $\mM$(~\eqref{eq:target_2_layer_nn}), and thus represent the primary learning signal. The $\mB_\ell^t$ terms arise from weight-weight interactions and act as an implicit regularization term whose magnitude grows with the norm of the weights.

\paragraph{One-Step Updates and Gradient Structure.} Using this decomposition, the one-step gradient descent updates can be written as:
\begin{equation}
    \mW_\ell^1 = \mW_\ell^0 + \eta_\ell \mA_\ell^0 - \eta_\ell \mB_\ell^0.
\end{equation}

Under orthogonal initialization, the norms of $\mA_\ell^0$ concentrate around deterministic quantities, while $\mB_\ell^0$ is initially small due to the orthogonality of $\mW_1^0$ and $\mW_2^0$. Thus, early-stage learning is dominated by the signal-aligned term $\mA_\ell^0$.

\paragraph{Signal-Only Reference Dynamics.} To isolate the contribution of the data-aligned terms, we consider a signal-only trajectory by removing the self-interaction components:
\begin{equation}
    \widetilde{\mW_\ell^1} = \mW_\ell^0 + \eta_\ell \mA_\ell^0.
\end{equation}

The corresponding signal-only gradient components at the next step are:
\begin{equation}
    \widetilde{\mA_1^1} = \frac{1}{h}\mM \widetilde{\mW_2^1}^{\top}, \quad
    \widetilde{\mA_2^1} = \frac{1}{h}\widetilde{\mW_1^1}^{\top}\mM.
\end{equation}

This fictitious trajectory captures pure signal propagation through the network and admits clean norm bounds that are independent of higher-order weight interactions. It serves as a reference point for comparing the true GD dynamics.
\paragraph{Two-Step Updates and Higher-Order Corrections.} The true two-step updates take the form:
\begin{align}
    \mW_\ell^2 = \mW_\ell^1 + \eta_\ell \mA_\ell^1 - \eta_\ell \mB_\ell^1.
\end{align}
To facilitate comparison with the signal-only trajectory, we define the corrected approximation
\begin{align}
        \overline{\mW_\ell^2} &= \widetilde{\mW_\ell^1} + \eta_\ell \widetilde{\mA_\ell^1} - \eta_\ell \widetilde{\mB_\ell^1}, \\
        \overline{\mG_\ell^1} &= \widetilde{\mB_\ell^1}-\widetilde{\mA_\ell^1} ,
\end{align}
where $\widetilde{\mB_\ell^1}$ denotes the self-interaction term evaluated along the signal-only path. This construction allows us to quantify the deviation between $\mW_\ell^2$ and $\widetilde{\mW_\ell^2}$ and to show that the difference is controlled by higher-order terms in $\eta_\ell$.

\subsection{Learning Rate Regimes and Gradient Dominance}
\label{sec:orthog-lr-norms}

We now formalize the effect of learning rate scaling on the relative magnitude of the signal and self-interaction components of the gradients. By analyzing the norms of the matrices
${\mA_1^0},{\mA_2^0}, {\mB_1^0}, {\mB_2^0}$, $\widetilde{\mA_1^1},\widetilde{\mA_2^1}, \widetilde{\mB_1^1}, \widetilde{\mB_2^1}$, we obtain the following characterization of the gradient structure under random orthogonal initialization.

\begin{proposition}
\label{sec:orthgo-proposition-2-layer}
(Two-layer NN under random orthogonal initialization.)
Under Assumption~\ref{sec:assumption-whiten}, if the learning rates satisfy
\(
\eta_1,\eta_2 \leq O(h\sqrt{h}),
\)
then the gradients are well-approximated by their signal-aligned components:
\begin{align*}
\begin{split}
        \norm{{\mG_1^0}-{\mA_1^0}} &\leq \frac{\norm{{\mG_1^0}}}{\sqrt{h}-1},\quad 
        \norm{{\mG_2^0}-{\mA_2^0}} \leq \frac{\norm{{\mG_2^0}}}{\sqrt{h}-1}, \\
        \norm{{\overline{\mG_1^1}}-{\widetilde{\mA_1^1}}} &\leq \frac{\norm{{\overline{\mG_1^1}}}}{\sqrt{h}-1}, \quad
        \norm{{\overline{\mG_2^1}}-{\widetilde{\mA_2^1}}} \leq \frac{\norm{{\overline{\mG_2^1}}}}{\sqrt{h}-1}.
\end{split}
\end{align*}
\end{proposition}

Proposition~\ref{sec:orthgo-proposition-2-layer} shows that, for sufficiently wide networks, the contribution of the self-interaction terms $\mB_\ell^t$ to the gradient norm is suppressed by a factor of $1/(\sqrt{h}-1)$. Consequently, both the one-step gradients $\mG_\ell^0$ and the corrected two-step gradients $\overline{\mG_\ell^1}$ are dominated by their signal-aligned components $\mA_\ell^0$ and $\widetilde{\mA_\ell^1}$, respectively. This justifies approximating the early-stage training dynamics using the signal-only trajectory introduced previously (see complete proof in Appendix~\ref{app:2-NN-orthgonal_one-step_prop} and~\ref{app:2-NN-orthgonal_two-step_prop}). In Figure~\ref{fig:spectral analysis} in Appendix~\ref{app:more_experiments} , we  perform spectral analysis of the $\mA_\ell^0$, $\widetilde{\mA_\ell^1}$, $\mB_\ell^0$, $\widetilde{\mB_\ell^1}$ , $\mG_\ell^0$ , $\overline{\mG_\ell^1}$ matrices and visualize the norm gap highlighted in Proposition~\ref{sec:orthgo-proposition-2-layer}, we further verify that $\mB_\ell^0$ and $\widetilde{\mB_\ell^1}$ are negligible, as they are dominated by $\mA_\ell^0$ and $\widetilde{\mA_\ell^1}$.

\paragraph{Large Learning Rate Regime.}
The proposition also identifies a critical scaling of the learning rate at which self-interaction effects become non-negligible for two-layer neural networks. In particular, when
$\eta_1 = \Theta(h\sqrt{h})$, we have:
\begin{align}
    \norm{{\mW_1^1}-{\mW_1^0}}_F &\asymp \norm{\mW_1^0}_F, \\
    \norm{{\overline{\mW_1^2}}-\widetilde{\mW_1^1}}_F &\asymp \norm{\widetilde{\mW_1^1}}_F,
\end{align}
and analogously for the second layer when $\eta_2 = \Theta(h\sqrt{h})$. This scaling marks a transition point where a single gradient step produces a weight update comparable in magnitude to the existing weights. 
Proposition~\ref{sec:orthgo-proposition-2-layer} shows that when the learning rates satisfy
\(
\eta_1,\eta_2 \leq O(h\sqrt{h}),
\)
the signal-aligned components dominate the gradients from a norm perspective. Specifically, the sets
\(
\{ {\mA_i^0} \}_{i=1}^2
\)
and
\(
\{ \widetilde{{\mA_i^1}} \}_{i=1}^2
\)
are very close to
\(
\{ {\mG_i^0} \}_{i=1}^2
\)
and
\(
\{ \overline{{\mG_i^1}} \}_{i=1}^2
\),
respectively. This implies that $\{ {\mA_i^0} \}_{i=1}^2$ and $\{ \widetilde{{\mA_i^1}} \}_{i=1}^2$ serve as the leading terms in the one-step gradients $\{ {\mG_i^0} \}_{i=1}^2$ and the corrected two-step gradients $\{ \overline{{\mG_i^1}} \}_{i=1}^2$. This approximation substantially simplifies the subsequent analysis of the theoretical test loss for the two-layer neural network. In particular, it allows us to replace the exact gradients $\{ {\mG_i^0} \}_{i=1}^2$ and $\{ \overline{{\mG_i^1}} \}_{i=1}^2$ with their leading-order counterparts $\{ {\mA_i^0} \}_{i=1}^2$ and $\{ \widetilde{{\mA_i^1}} \}_{i=1}^2$, respectively, as formally justified by Lemma~\ref{sec:orthog-approx-loss}. For the three-layer setting, we similarly replace the original gradients by their leading terms, as justified by Proposition~\ref{app:proposition-3-layer-orthogonal} and ~\ref{app:proposition-3-layer-orthogonal-two-steps}, in order to streamline the test loss analysis.

Finally, Proposition~\ref{sec:orthgo-proposition-2-layer} also identifies a critical learning-rate scaling. When the learning rates for the first and second layers are set to be on the order of
\(
\Theta(h\sqrt{h}),
\)
the gradient updates become comparable in magnitude to the initialized weight matrices, effectively overwhelming the initialization. This behavior mirrors the learning-rate scaling associated with the maximal update parameterization studied in~\citet{yang2021tensor,yang2022tensor}. Prior work~\citep{ba2022high,yang2021tensor,yang2022tensor,yang2023tensor} suggests that choosing learning rates within (or below) this large-learning-rate regime can be beneficial for training. While these studies primarily focus on two-layer networks with a fixed output layer, our analysis reaches a compatible conclusion without relying on this assumption.  A similar result can be obtained for the three-layer NN setting under random orthogonal initialization with $\eta_1,\eta_2 $ no more than $O(h)$ (See Proposition~\ref{app:proposition-3-layer-orthogonal} and ~\ref{app:proposition-3-layer-orthogonal-two-steps}). We provide the proof in Appendix~\ref{app:3-NN-orthgonal_one-step_prop} and~\ref{app:3-NN-orthgonal_two-step_prop}.

\subsection{Relationship between Test Loss and Layer-wise Learning Rates}
\label{sec:orthog-test_loss}



This section characterizes how the test loss depends on the learning rates of individual layers in the linear NNs trained under random orthogonal initialization. 


\subsubsection{ Two-layer Neural Networks}

Given test data $\tilde{\vx}_0 \!\sim\! \gN(\vzero, \mI_h)$, we consider the test loss
\begin{align*}
L_{\text{two-layer}} = \mathbb{E}_{\mW_1,\mW_2,\mM,\Tilde{\vx}_0,\mX} \left\| \frac{1}{h}\tilde{\vx}_0 \mW_1 \mW_2 - \tilde{\vx}_0 \mM \right\|^2.
\end{align*}
As the exact closed-form characterization of the test loss is nontrivial, our analysis proceeds by first simplifying the training dynamics using leading-order gradient approximations, and then translating these simplified dynamics into explicit expressions for the test loss.
\begin{lemma}
\label{sec:orthog-approx-loss}
Under Assumption ~\ref{sec:assumption-whiten} and ~\ref{sec:assumption}, for $\eta_1$,  $\eta_2 \leq O(h\sqrt{h})$, we have
\begin{align*}
    &\left |L_{\text{two-layer}}({\mW_1^1},{\mW_2^1})-L_{\text{two-layer}}({\widetilde{\mW_1^1}}, {\widetilde{\mW_2^1}}) \right|  \leq O(h^{-1}). \\ 
    &\left |L_{\text{two-layer}}({\mW_1^2},{\mW_2^2})-L_{\text{two-layer}}({\widetilde{\mW_1^2}}, {\widetilde{\mW_2^2}}) \right|  \leq O(h^{-\frac{1}{2}}). \\  
\end{align*}
\end{lemma}
Lemma \ref{sec:orthog-approx-loss} formalizes the idea that, for sufficiently large width and moderate learning rates, the test loss is insensitive to higher-order gradient corrections. Specifically, when $\eta_1,\eta_2 \le O(h\sqrt{h})$, replacing the true gradient with the signal-aligned approximations changes the test loss by at most $O(h^{-1})$ after one step and $O(h^{-1/2})$ after two steps.

Intuitively, this result builds on Proposition~\ref{sec:orthgo-proposition-2-layer}: since the signal components dominate the gradient norms, the parts of the update omitted in the approximation contribute only lower-order perturbations to the weights. As a consequence, the network’s input–output map after one or two steps is well-approximated by the signal-only dynamics, and the resulting test loss remains essentially unchanged at leading order. This lemma is crucial because it allows us to analyze the test loss using simplified weight trajectories that admit closed-form expressions, without sacrificing asymptotic accuracy.
The simplified analysis leads to the following result for two-layer networks.\yaoqing{Can we provide a simulation result to show that the gradient approximation makes sense? For example, we can show the loss as a function of $h$ for some particular choice of $\eta$'s and show the difference between true test loss and the approximated test loss?}\tianyu{I think Figure 1 maybe already show that the theoretical loss approximation matches the empirical results?. Do you think we still need an additional plot where the theory and experiments are overlaid in the same figure?}

\begin{theorem}
\label{theorem for orthog-2-layer}
    Given Assumption~\ref{sec:assumption-whiten},~\ref{sec:assumption} and in addition assume $\eta_1$ and $\eta_2$ are no more than $O({h\sqrt{h}})$, based on Proposition~\ref{sec:orthog-norm} and Lemma~\ref{sec:orthog-approx-loss}, consider the training procedure discussed in Section~\ref{sec:setup}, we obtain the following test loss after one-step and two-step GD update in a two-layer neural network under  random orthogonal initialization:
\begin{equation*}
\begin{aligned}
       & L_{\text{two-layer}}({\mW_1^1},{\mW_2^1})=\frac{\eta_1^2}{h^4}+\frac{\eta_2^2}{h^4}+\frac{2\eta_1\eta_2}{h^4}+\frac{\eta_1^2\eta_2^2}{h^7}\\ & \qquad \qquad \qquad \qquad  -\frac{2\eta_1}{h^2}-\frac{2\eta_2}{h^2} +\frac{1}{h}+\frac{2\eta_1\eta_2}{h^5}+1 \\
    &L_{\text{two-layer}}({\mW_1^2},{\mW_2^2})=\frac{1}{h}(1+\frac{\eta_1\eta_2}{h^3})^4 +\frac{16\eta_1^2\eta_2^2}{h^7} \\&+\left(\frac{2(\eta_1+\eta_2)(\eta_1\eta_2+h^3)}{h^5}-1\right)^2+(1+\frac{\eta_1\eta_2}{h^3})^2\frac{8\eta_1\eta_2}{h^5}
\end{aligned}    
\end{equation*}
\end{theorem}

We provide the proof in Appendix~\ref{app:orthog-2-NN-one-step} and~\ref{app:orthog-2-NN-two-steps}. Theorem \ref{theorem for orthog-2-layer} provides explicit formulas for the test loss after one-step and two-step gradient descent updates in a two-layer network. While the expressions themselves are algebraically involved, their structure reveals several key phenomena.

\paragraph{One-step test loss.} The one-step test loss decomposes into three types of terms: (i) Linear improvement terms (e.g., $-2\eta_1/h^2$, $-2\eta_2/h^2$), which reflect the reduction in error due to alignment with the target signal. (ii) Quadratic and interaction terms (e.g., $\eta_1^2/h^4$, $\eta_2^2/h^4$, $\eta_1\eta_2/h^4$), which capture over-updating and cross-layer coupling. (iii) Residual variance terms (e.g., $1/h + 1$), arising from the randomness of initialization and test inputs. These components make explicit how learning rates at different layers contribute asymmetrically and interactively to NN generalization.

\paragraph{Two-step test loss.} For the two-step update, the test loss exhibits higher-order dependence on the product $\eta_1\eta_2$. This reflects the fact that meaningful improvement in a two-layer linear network requires coordination between layers; updating only one layer is insufficient to substantially reduce the prediction error. The appearance of repeated factors of $(1+\eta_1\eta_2/h^3)$ highlights the multiplicative nature of representation learning across layers. Building on this insight, we obtain the following corollary:
\begin{corollary}
\label{cor:two-layer-NN}
Suppose $\eta_1+\eta_2 = 2h^{\alpha}$ and we consider $0<\alpha \le \tfrac{3}{2}$.
Then, for any $\alpha$ in this range, the point $\eta_1=\eta_2=h^{\alpha}$ is not a local minimum of the loss $L_{\text{two-layer}}({\mW_1^1},{\mW_2^1})$.
However,\addressedyaoqing{Do you mean "however"?} for $1<\alpha \le \tfrac{3}{2}$, if $h > \max \{{h^{*}}, 256\}$, then $\eta_1=\eta_2=h^{\alpha}$ is a local minimum of the loss $L_{\text{two-layer}}({\mW_1^2},{\mW_2^2})$, where ${h^{*}}$ is the root of the following equation:
\begin{equation*}
    (1+o(1))h^{1-\alpha} +16h^{\alpha-2}+2h^{-\alpha}+8h^{\alpha-3}+6h^{3\alpha-6}-2 =0
\end{equation*}
\end{corollary}

Corollary \ref{cor:two-layer-NN} studies the test loss landscape under a constrained learning-rate budget $\eta_1+\eta_2 = 2h^{\alpha}$ (Assumption~\ref{assumption:lr}), which isolates the effect of how learning rates are allocated across layers, rather than their total magnitude.

\paragraph{Asymmetric learning rates after one-step update.}
The first conclusion is that the symmetric choice $\eta_1=\eta_2$ is not a local minimum of the test loss after a single gradient descent step for any $0<\alpha\le \tfrac{3}{2}$. This result indicates that, in the initial stage of training, the test loss is optimized by an asymmetric allocation of learning rates across layers. After one update, the two layers contribute differently to the predictor: updates to the first layer primarily control the formation of internal representations, whereas updates to the second layer mainly affect the linear readout of these representations. Imposing equal learning rates at this stage restricts the network from exploiting this structural asymmetry, resulting in suboptimal test performance.

\paragraph{Symmetric learning rates after two-step update.}
In contrast, the second conclusion shows that for $1<\alpha\le \tfrac{3}{2}$ and sufficiently large network width, the symmetric choice $\eta_1=\eta_2$ becomes a local minimum of the test loss after two gradient descent steps. This behavior reflects a transition in the training dynamics: after multiple updates, the learning process becomes increasingly coupled across layers, and coordinated updates yield improved generalization. In this regime, balanced learning rates facilitate effective interaction between layers, leading to optimal performance. The lower bound on $\alpha$ ensures that the learning rates are sufficiently large to induce non-negligible cross-layer effects, while remaining within a stable training regime.

Overall, this corollary identifies a phase transition in the optimal allocation of layer-wise learning rates, governed by the interaction between network width, training depth, and the overall scale of the learning rates.





\subsubsection{Three-layer Neural Networks}
We now characterize the test loss of a three-layer neural network after one-step and two-step gradient descent updates under random orthogonal initialization. Compared to the two-layer setting, the presence of a vector-valued output layer $\va$ fundamentally alters both the learning-rate scaling and the structure of the resulting test loss. Given test data $\tilde{\vx}_0 \!\sim\! \gN(\vzero, \mI_h)$, we consider the test loss:
\begin{align*}
L_{\text{three-layer}} &= \mathbb{E}_{\substack{
\mW_1^0,\mW_2^0,\va, \\
\vbeta^*,\,\tilde{\vx}_0,\,\mX
}}
\left(
\frac{1}{\sqrt{h}}\, \tilde{\vx}_0 \mW_1 \mW_2 \va
- \tilde{\vx}_0 \vbeta^*
\right)^2
\end{align*}
\paragraph{Structural distinction from the two-layer network.}
In the three-layer network, the predictor takes the form $\tilde{\vx}_0 \mW_1 \mW_2 \va$, where the output weights $\va \in \mathbb{R}^{h}$ are fixed throughout training. Consequently, learning in the hidden layers affects the test loss only through their joint alignment with the target vector $\vbeta^*$. This additional linear mapping at the output introduces a bottleneck that attenuates the propagation of gradient updates, thereby reducing the scale at which layer-wise interactions become significant. As a result, the admissible learning-rate regime in the three-layer setting is $\eta_1,\eta_2 \le O(h)$, which is strictly smaller than the $O(h\sqrt{h})$ regime identified for the two-layer network.

\begin{theorem}
\label{theorem for orthog-3-layer}
    Given Assumption~\ref{sec:assumption-whiten},~\ref{sec:assumption} and in addition assume $\eta_1$ and $\eta_2$ are no more than $O({h})$ based on Proposition~\ref{app:proposition-3-layer-orthogonal} and ~\ref{app:proposition-3-layer-orthogonal-two-steps}, consider the training procedure discussed in Section~\ref{sec:setup}, we derive the test loss after one-step and two-step GD update in a three-layer neural network:
\begin{equation*}
\begin{aligned}
        L_{\text{three-layer}}({\mW_1^1},{\mW_2^1})&=\frac{\eta_1^2}{h^2}+\frac{\eta_2^2}{h^2} + \frac{2\eta_1\eta_2}{h^2}+ \frac{\eta_1^2\eta_2^2}{h^4}\\ &  -\frac{2\eta_1}{h}-\frac{2\eta_2}{h}+\frac{1}{h}+\frac{2\eta_1\eta_2}{h^3}+1 \\
    L_{\text{three-layer}}({\mW_1^2},{\mW_2^2})&=\left(\frac{2(\eta_1+\eta_2)(h+\eta_1\eta_2)}{h^2} -1 \right)^2 \\ &+\frac{1}{h}+ \frac{2\eta_1\eta_2}{h^2}+ \frac{10\eta_1\eta_2}{h^3}+\frac{\eta_1^2\eta_2^2}{h^3} \\ &+\frac{37\eta_1^2\eta_2^2}{h^4} + \frac{12\eta_1^3\eta_2^3}{h^5} + \frac{\eta_1^4\eta_2^4}{h^6}
\end{aligned}    
\end{equation*}
\end{theorem}

The first expression in Theorem~\ref{theorem for orthog-3-layer} gives the test loss after a single gradient descent step. We provide the proof in Appendix~\ref{app:orthog-3-NN-one-step}. Its structure mirrors that of the two-layer case, with appropriately rescaled terms.
The two-step test loss exhibits a substantially richer dependence on the learning rates, involving higher-order polynomial terms in the product $\eta_1\eta_2$. We provide the proof in Appendix~\ref{app:orthog-3-NN-two-steps}. This behavior reflects the fact that, in a three-layer network, a meaningful reduction in test loss requires coordinated updates across both hidden layers over multiple steps. Similar to Corollary~\ref{cor:two-layer-NN}, we have the following corollary for a three-layer neural network. 
\begin{figure}[tb]
    \centering
    \begin{subfigure}[t]{0.48\linewidth}
        \centering
        \includegraphics[width=\textwidth]{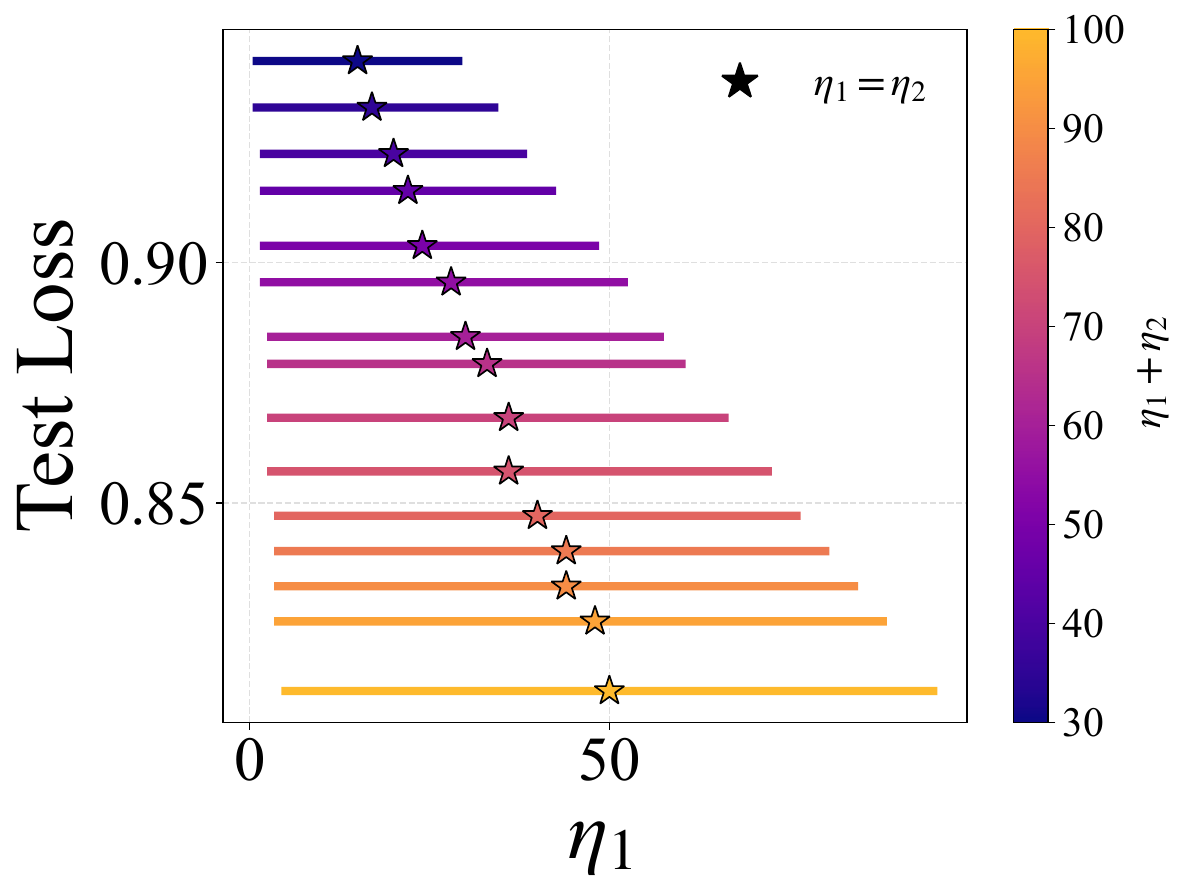}
        \caption{  1-step (theory) }
    \end{subfigure}
    \hfill
    \begin{subfigure}[t]{0.48\linewidth}
        \centering
        \includegraphics[width=\textwidth]{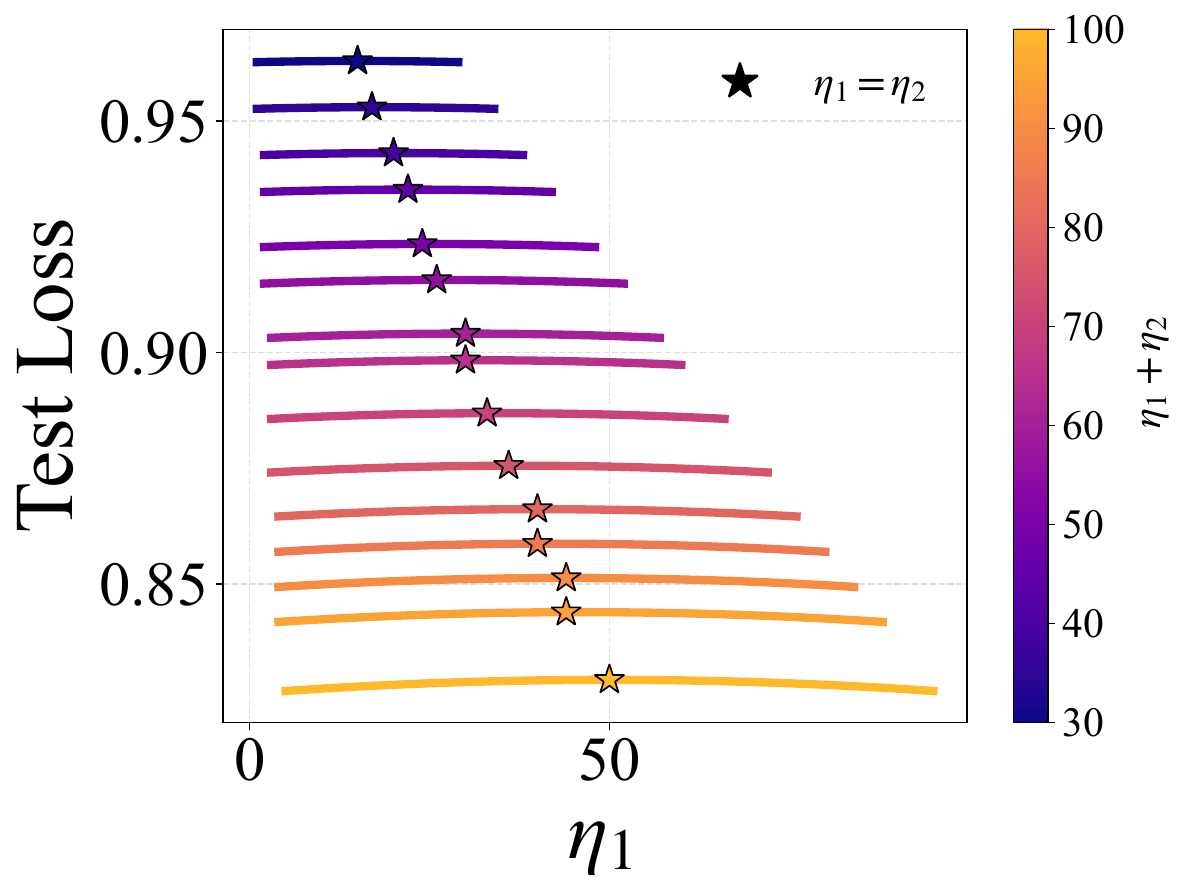}
        \caption{  1-step (experiment) }
    \end{subfigure}
    \hfill
    \begin{subfigure}[t]{0.48\linewidth}
        \centering
        \includegraphics[width=\textwidth]{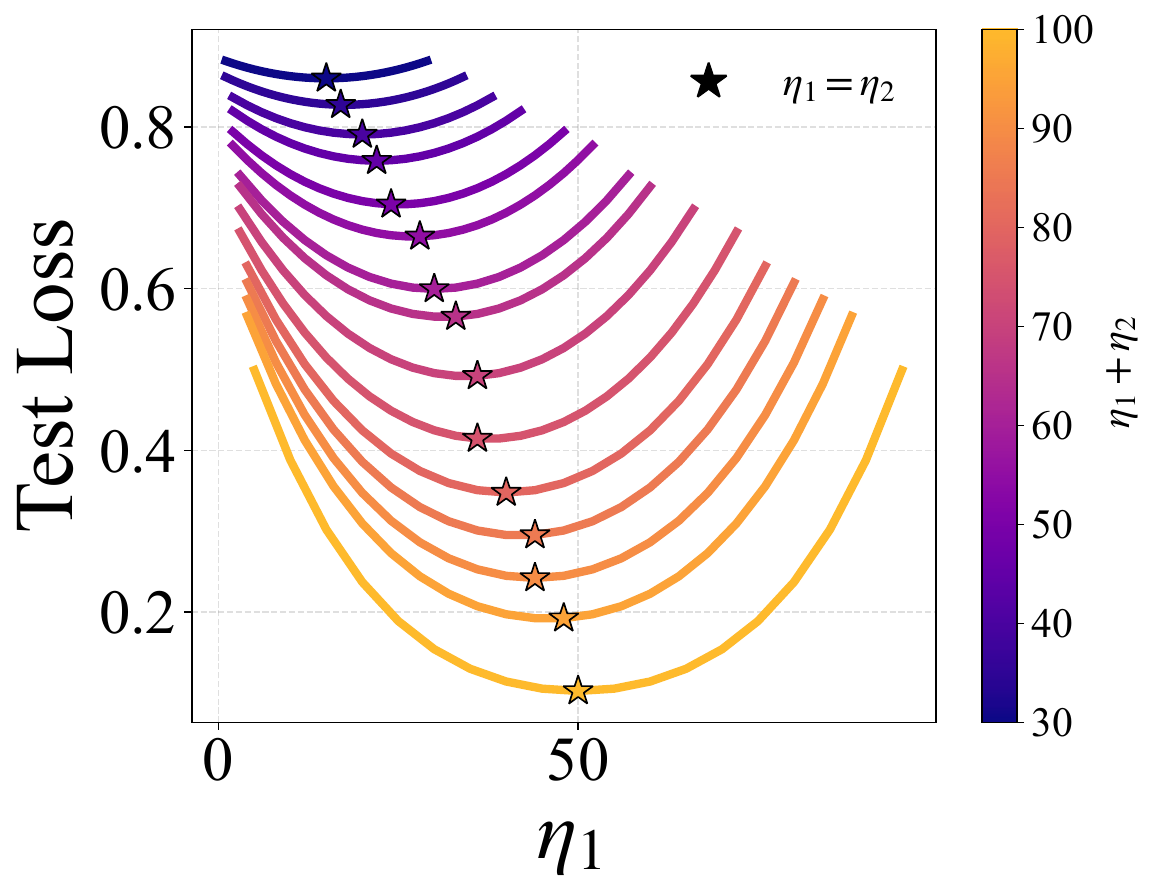}
        \caption{  2-step (theory) }
    \end{subfigure}
    \hfill
        \begin{subfigure}[t]{0.48\linewidth}
        \centering
        \includegraphics[width=\textwidth]{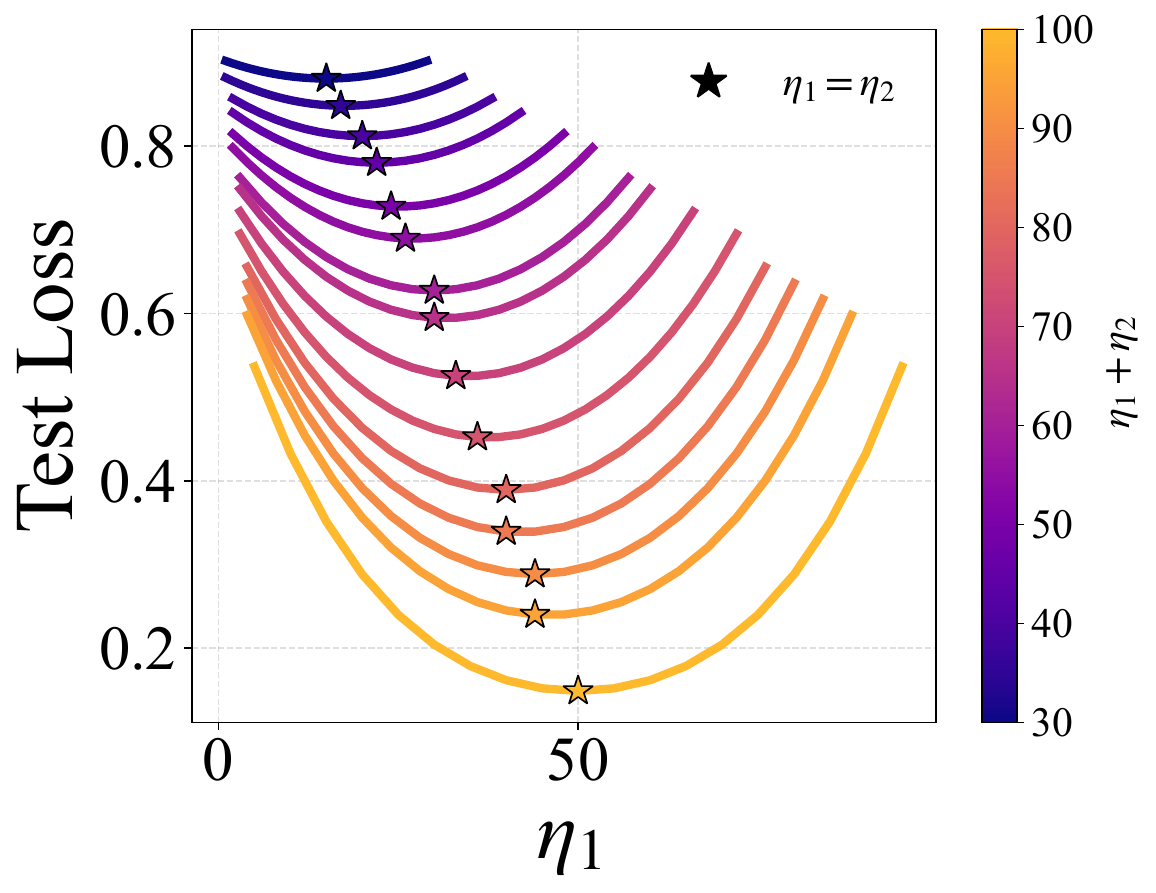}
        \caption{  2-step (experiment) }
    \end{subfigure}
\caption{ \textbf{ 3-layer NN under orthogonal initialization.} Here we set $\eta_1+\eta_2\leq O(h^{\frac{2}{3}})$  and  $h=1000$. We observe conclusions and results similar to those in Figure~\ref{fig:2-NN-orthogonal}. We discuss more in Section~\ref{sec:exp}.\vigk{Lets follow the same plotting style as Fig 1.}}
\vspace{-3mm}
\label{fig:3-NN-orthogonal}
\end{figure}

\begin{corollary}
\label{cor:three-layer-NN}
Suppose $\eta_1+\eta_2 = 2h^{\alpha}$ and we consider $0<\alpha < 1$.
Then, for any $\alpha$ in this range, the point $\eta_1=\eta_2=h^{\alpha}$ is not a local minimum of the loss $L_{\text{three-layer}}({\mW_1^1},{\mW_2^1})$.
However, for $0<\alpha \le \tfrac{2}{3}$, if $h > h^{*}$, then $\eta_1=\eta_2=h^{\alpha}$ is a local minimum of the loss $L_{\text{three-layer}}({\mW_1^2},{\mW_2^2})$, where ${h^{*}}$ is the root of the following equation:
\begin{align*}
\begin{split}
         &32h^{3\alpha-2}+ 33h^{\alpha-1}+74h^{\alpha-2}+2h^{-\alpha} \\+&10h^{-\alpha-1}+36h^{3\alpha-3}+4h^{5\alpha-4}-8=0
\end{split}
\end{align*}
\end{corollary}

\paragraph{Comparison with the two-layer case.}
While both two-layer and three-layer networks exhibit nontrivial dependence on layer-wise learning rates, the three-layer setting differs in two key aspects. First, the critical learning-rate scale is reduced from $O(h\sqrt{h})$ to $O(h)$ due to differences in the initialization schemes as well as the presence of the fixed output vector $\va$. Second, the two-step test loss has a stronger dependence on higher-order products of $\eta_1$ and $\eta_2$, reflecting enhanced cross-layer coupling.
Overall, Theorem~\ref{theorem for orthog-3-layer} demonstrates that, in three-layer networks with a vector-valued output layer, the test loss is governed by a delicate interaction between layer-wise learning rates, training depth, and network width. Although early-stage updates admit a decomposition similar to that of the two-layer case, deeper architectures amplify cross-layer interactions over successive steps, leading to a distinct learning-rate scaling regime and a richer dependence of test loss on the learning rates.

\section{Experiments}
\label{sec:exp}


\paragraph{Orthogonal initialization.} Here we numerically validate our theoretical results under orthogonal initialization. We set $h=n=d=1000$ and keep the model and data initialization the same as in Section~\ref{sec:setup}.  In Figure~\ref{fig:2-NN-orthogonal}, ~\ref{fig:3-NN-orthogonal} and ~\ref{fig:2-NN-orthogonal-large-lr}, we present both theoretical simulations and empirical experiments for two- and three-layer networks, comparing the one-step and two-step test loss as a function of the learning rates across a range of $\eta_1+\eta_2$ values below the critical threshold. Across all settings, the theoretical losses we derive closely match the observed test losses after either one or two updates. Moreover, when $\eta_1+\eta_2 = 2h^{\alpha}$ lies below the critical threshold, we can see a clear qualitative shift: after a single update, symmetric learning rates across layers are suboptimal, whereas after two updates they become locally optimal in sufficiently wide networks. This reveals a transition from asymmetric to balanced layer-wise learning-rate allocation. For different $h$, see Figure~\ref{fig:h=5000-2-NN} and~\ref{fig:h=5000-3-NN} in Appendix~\ref{app:more_experiments}). We also find that balanced layer-wise learning-rate allocation can  be locally optimal during early training over multiple steps (up to 512 update steps) under orthogonal initialization (see Figure~\ref{fig:2-NN-more-steps-orthgo},~\ref{fig:3-NN-more-steps-orthog} and~\ref{fig:3-NN-512-steps-orthog} in Appendix~\ref{app:more_experiments}).

\paragraph{Gaussian initialization.}  To demonstrate the generality of our results, we repeat the same set of experiments under gaussian initialization. We observe the same  behavior as in the orthogonal case: after the first update, symmetric layer-wise learning rates are suboptimal, whereas after two updates they become locally optimal in sufficiently wide networks. This phenomenon can also further extend to multiple training steps, please  see Figure~\ref{fig:2-NN-more-steps-gaussian} and~\ref{fig:3-NN-more-steps-gaussian} in Appendix~\ref{app:more_experiments}.
\vigk{(1) After we follow the same plotting style as Fig 1 for other plots, we will have space to bring the gaussian initialization based plots into the main paper as well. (2) Next, lets also improve the norm difference bar plot in Fig 4 and bring it to the main paper. Essentially, we want to convey more information with less space.}

\paragraph{Linear NN under noisy condition.} In Figure~\ref{fig:NN-noisy-orthgonal-initialization} in Appendix~\ref{app:more_experiments}, we consider adding label noise $\vxi \sim \mathcal{N}(0, \rho)$ to the teacher model in both two layer and three layer linear neural networks under orthogonal initialization. We still observe that  after the first update, symmetric layer-wise learning rates are suboptimal, whereas after two updates they become locally optimal.

\paragraph{Deep Linear NNs.} In Figure~\ref{fig:deep-linear-NN}, we consider 4-layer and 8-layer linear neural networks, which generalize the two layer and three layer settings. We observe that, for one and two update steps, the same transition from asymmetry to balance still appears.

\paragraph{Nonlinear NNs.} Here we consider a three layer nonlinear neural network, where the student model is $f(\vx_i)=\frac{1}{\sqrt{h}}\sigma(\sigma(\vx_i^{\top}\mW_1)\mW_2)\va$, and the teacher model is $y_i=\sigma({\vbeta^*}^{\top}\vx_i)$, with $\sigma$ being the ReLU activation. We use the same orthogonal initialization and training pipeline as in the two layer and three layer settings. In Figure~\ref{fig:nonlinear-NN}, we visualize the test loss as a function of $\eta_1$ after 1 and 8-step updates $\eta_1+\eta_2 < O(\sqrt{h})$. Although the curves are relatively less symmetric than in linear case, we still observe a similar asymmetry-to-balance transition, which generalizes the cases and results covered by our theoretical setup.

\section{Discussion}
\label{discussion}
\paragraph{Learning rate scheduler Design.} In previous sections, we revealed the asymmetry to balance transition in layer-wise learning rate allocation and offered theoretical support for layer-wise learning rate schedulers that aim to promote layer balance at later stages of training. Here, we provide a simple example to guide their practical design.
Consider a teacher model $\vy_i=\mM^\top \vx_i$ and a student model $f(\vx_i)=\vx_i\mW_1\mW_2$, where $\vx_i$ is the input and $\mW_1, \mW_2$ are the two trainable matrices. Since the Frobenius norm is a classic generalization metric, we can leverage it for $\mW_1$ and $\mW_2$ to design a learning rate scheduler.

First, we expect the layer with the larger Frobenius norm to be assigned a smaller learning rate, due to the property of the metric. More importantly, based on the theoretical insights in our paper, we expect the learning rates of the two layers to become balanced in the later stages of training, which motivates us to promote balance between the layer norms. As a result, at each step $t$,  we set the learning rates for ${\mW_1}^t$ and ${\mW_2}^t$ as
$\eta_{\mW_1}^{(t)}=\frac{2\norm{{\mW_2}^t}_F}{\norm{{\mW_1}^t}_F+\norm{{\mW_2}^t}_F}\mathrm{lr},
\qquad
\eta_{\mW_2}^{(t)}=\frac{2\norm{{\mW_1}^t}_F}{\norm{{\mW_1}^t}_F+\norm{{\mW_2}^t}_F}\mathrm{lr},$
where $\mathrm{lr}$ is a uniform base learning rate. As training enters the later stage, this balance-driven learning rate scheduler promotes $\big|\norm{{\mW_1}}_F-\norm{\mW_2}_F\big|\to 0$, and the learning rates also become balanced. It is worth noting that for this matrix-factorization type linear network, the curvature at convergence, measured by the largest Hessian eigenvalue, is related to $\big|\norm{\mW_1}_F-\norm{\mW_2}_F\big|$; in particular, smaller norm gap corresponds to a flatter solution~\citep{wang2021large}. Therefore, the transition of the learning rates from asymmetry to balance also corresponds to the process by which the model gradually converges to a flatter minima.

In Figure~\ref{fig:fro-balance}, we consider a setup where $\norm{\mW_2}_F=6$ and $\norm{\mW_1}_F=1$ at initialization, and compare this design with a uniform learning rate used throughout training. We find that this layer-wise schedule captures the asymmetry-to-balance transition observed in our paper, and achieves lower training loss and test loss than the fully uniform baseline. More specifically, we observe that $\big|\norm{\mW_1}_F-\norm{\mW_2}_F\big|$ approaches zero in the middle and late stages of training, which corresponds to increasingly balanced learning rates.

\paragraph{Step-dependent optimality of learning-rate symmetry.} Our results indicate that symmetric learning rates are suboptimal for the first step but optimal for two steps. This may not be the same as using asymmetric learning rates early and balancing them later during training.
This suggests that asymmetric learning rates may be preferable at the very beginning of training, and symmetric learning rates become optimal as cross-layer interactions develop over subsequent steps, even if the initial learning-rate allocation is not optimal. We believe this also points to a practical strategy that use asymmetric learning rates early in training and more symmetric ones later. we further clarify this question through examples involving a three-layer linear network and a CNN; please see Appendix~\ref{app:more_experiments}.

\section{Conclusion}

In this work,  we provide a finite-step characterization of how
layer-wise learning rates should be balanced during training
in linear neural networks. By analyzing gradient descent dynamics after one and two updates, we show that symmetric learning rates across layers are generally suboptimal at initialization, with early optimization favoring asymmetric allocations that reflect the distinct roles of different layers. As training progresses, a transition occurs where sufficiently large width and appropriate scaling of the total learning rate cause balanced learning rates to be locally optimal. Thus, signaling the emergence of coordinated layer-wise updates. This transition is architecture-dependent, with deeper networks exhibiting stricter conditions under which symmetry is optimal. Our results formalize balancing learning rates across layers as a dynamical phenomenon driven by optimization and scaling, rather than a static design choice, and elucidate how depth, width, and learning-rate magnitude jointly shape this behavior.

\section*{Acknowledgments}
This work is supported by the DARPA AIQ program, the U.S. Department of Energy under Award Number DE-SC0025584, the Allocation Year 2026 DOE Mission Science Award, Dartmouth College, and Lambda AI.

\section*{Impact Statement}

This paper presents research aimed at advancing  machine learning theory, particularly by providing an exact, finite-step characterization of how layer-wise learning-rate choices shape early training dynamics and generalization in multi-layer linear neural networks. Our analysis reveals a transition from initially asymmetric optimal learning rates to later balanced rates. While this work may have various potential societal implications, we do not find it necessary to highlight any specific ones here.

\bibliography{icml2026/references}

@article{howard2018universal,
  title={Universal language model fine-tuning for text classification},
  author={Howard, Jeremy and Ruder, Sebastian},
  journal={arXiv preprint arXiv:1801.06146},
  year={2018}
}

@inproceedings{long2015learning,
  title={Learning transferable features with deep adaptation networks},
  author={Long, Mingsheng and Cao, Yue and Wang, Jianmin and Jordan, Michael},
  booktitle={International conference on machine learning},
  pages={97--105},
  year={2015},
  organization={PMLR}
}

@article{kingma2014adam,
  title={Adam: A method for stochastic optimization},
  author={Kingma, Diederik P},
  journal={arXiv preprint arXiv:1412.6980},
  year={2014}
}

@article{liu2019variance,
  title={On the variance of the adaptive learning rate and beyond},
  author={Liu, Liyuan and Jiang, Haoming and He, Pengcheng and Chen, Weizhu and Liu, Xiaodong and Gao, Jianfeng and Han, Jiawei},
  journal={arXiv preprint arXiv:1908.03265},
  year={2019}
}

@article{you2017large,
  title={Large batch training of convolutional networks},
  author={You, Yang and Gitman, Igor and Ginsburg, Boris},
  journal={arXiv preprint arXiv:1708.03888},
  year={2017}
}

@inproceedings{you2018imagenet,
  title={Imagenet training in minutes},
  author={You, Yang and Zhang, Zhao and Hsieh, Cho-Jui and Demmel, James and Keutzer, Kurt},
  booktitle={Proceedings of the 47th international conference on parallel processing},
  pages={1--10},
  year={2018}
}

@inproceedings{yao2021adahessian,
  title={Adahessian: An adaptive second order optimizer for machine learning},
  author={Yao, Zhewei and Gholami, Amir and Shen, Sheng and Mustafa, Mustafa and Keutzer, Kurt and Mahoney, Michael},
  booktitle={proceedings of the AAAI conference on artificial intelligence},
  volume={35},
  number={12},
  pages={10665--10673},
  year={2021}
}

@article{zhou2023temperature,
  title={Temperature balancing, layer-wise weight analysis, and neural network training},
  author={Zhou, Yefan and Pang, Tianyu and Liu, Keqin and Mahoney, Michael W and Yang, Yaoqing and others},
  journal={Advances in Neural Information Processing Systems},
  volume={36},
  year={2023}
}

@article{lu2024alpha,
  title={AlphaPruning: Using Heavy-Tailed Self Regularization Theory for Improved Layer-wise Pruning of Large Language Models},
  author={Haiquan Lu and Yefan Zhou and Shiwei Liu and Zhangyang Wang and Michael W. Mahoney and Yaoqing Yang},
  journal={Advances in Neural Information Processing Systems},
  year={2024}
}

@article{liu2024model,
  title={Model Balancing Helps Low-data Training and Fine-tuning},
  author={Zihang Liu and Yuanzhe Hu and Tianyu Pang and Yefan Zhou and Yaoqing Yang},
  journal={Empirical Methods in Natural Language Processing},
  year={2024}
}

@inproceedings{dong2019hawq,
  title={Hawq: Hessian aware quantization of neural networks with mixed-precision},
  author={Dong, Zhen and Yao, Zhewei and Gholami, Amir and Mahoney, Michael W and Keutzer, Kurt},
  booktitle={Proceedings of the IEEE/CVF international conference on computer vision},
  pages={293--302},
  year={2019}
}

@inproceedings{shen2020q,
  title={Q-bert: Hessian based ultra low precision quantization of bert},
  author={Shen, Sheng and Dong, Zhen and Ye, Jiayu and Ma, Linjian and Yao, Zhewei and Gholami, Amir and Mahoney, Michael W and Keutzer, Kurt},
  booktitle={Proceedings of the AAAI Conference on Artificial Intelligence},
  volume={34},
  number={05},
  pages={8815--8821},
  year={2020}
}

@article{ba2022high,
  title={High-dimensional asymptotics of feature learning: How one gradient step improves the representation},
  author={Ba, Jimmy and Erdogdu, Murat A and Suzuki, Taiji and Wang, Zhichao and Wu, Denny and Yang, Greg},
  journal={Advances in Neural Information Processing Systems},
  volume={35},
  pages={37932--37946},
  year={2022}
}

@book{vershynin2018high,
  title={High-dimensional probability: An introduction with applications in data science},
  author={Vershynin, Roman},
  volume={47},
  year={2018},
  publisher={Cambridge university press}
}

@article{wang2023spectral,
  title={Spectral evolution and invariance in linear-width neural networks},
  author={Wang, Zhichao and Engel, Andrew and Sarwate, Anand D and Dumitriu, Ioana and Chiang, Tony},
  journal={Advances in neural information processing systems},
  volume={36},
  pages={20695--20728},
  year={2023}
}

@inproceedings{yang2021tensor,
  title={Tensor programs iv: Feature learning in infinite-width neural networks},
  author={Yang, Greg and Hu, Edward J},
  booktitle={International Conference on Machine Learning},
  pages={11727--11737},
  year={2021},
  organization={PMLR}
}

@article{yang2022tensor,
  title={Tensor programs v: Tuning large neural networks via zero-shot hyperparameter transfer},
  author={Yang, Greg and Hu, Edward J and Babuschkin, Igor and Sidor, Szymon and Liu, Xiaodong and Farhi, David and Ryder, Nick and Pachocki, Jakub and Chen, Weizhu and Gao, Jianfeng},
  journal={arXiv preprint arXiv:2203.03466},
  year={2022}
}

@article{yang2023tensor,
  title={Tensor programs vi: Feature learning in infinite-depth neural networks},
  author={Yang, Greg and Yu, Dingli and Zhu, Chen and Hayou, Soufiane},
  journal={arXiv preprint arXiv:2310.02244},
  year={2023}
}

@article{du2018algorithmic,
  title={Algorithmic regularization in learning deep homogeneous models: Layers are automatically balanced},
  author={Du, Simon S and Hu, Wei and Lee, Jason D},
  journal={Advances in neural information processing systems},
  volume={31},
  year={2018}
}

@article{zhang2024adam,
  title={Adam-mini: Use fewer learning rates to gain more},
  author={Zhang, Yushun and Chen, Congliang and Li, Ziniu and Ding, Tian and Wu, Chenwei and Kingma, Diederik P and Ye, Yinyu and Luo, Zhi-Quan and Sun, Ruoyu},
  journal={arXiv preprint arXiv:2406.16793},
  year={2024}
}

@article{vaswani2017attention,
  title={Attention is all you need},
  author={Vaswani, Ashish and Shazeer, Noam and Parmar, Niki and Uszkoreit, Jakob and Jones, Llion and Gomez, Aidan N and Kaiser, {\L}ukasz and Polosukhin, Illia},
  journal={Advances in neural information processing systems},
  volume={30},
  year={2017}
}

@inproceedings{ro2021autolr,
  title={Autolr: Layer-wise pruning and auto-tuning of learning rates in fine-tuning of deep networks},
  author={Ro, Youngmin and Choi, Jin Young},
  booktitle={Proceedings of the AAAI Conference on Artificial Intelligence},
  volume={35},
  number={3},
  pages={2486--2494},
  year={2021}
}

@article{yao2024layer,
  title={Layer-wise Importance Matters: Less Memory for Better Performance in Parameter-efficient Fine-tuning of Large Language Models},
  author={Yao, Kai and Gao, Penglei and Li, Lichun and Zhao, Yuan and Wang, Xiaofeng and Wang, Wei and Zhu, Jianke},
  journal={arXiv preprint arXiv:2410.11772},
  year={2024}
}

@article{hu2024minicpm,
  title={Minicpm: Unveiling the potential of small language models with scalable training strategies},
  author={Hu, Shengding and Tu, Yuge and Han, Xu and He, Chaoqun and Cui, Ganqu and Long, Xiang and Zheng, Zhi and Fang, Yewei and Huang, Yuxiang and Zhao, Weilin and others},
  journal={arXiv preprint arXiv:2404.06395},
  year={2024}
}

@article{loshchilov2016sgdr,
  title={Sgdr: Stochastic gradient descent with warm restarts},
  author={Loshchilov, Ilya and Hutter, Frank},
  journal={arXiv preprint arXiv:1608.03983},
  year={2016}
}

@article{yin2023outlier,
  title={Outlier weighed layerwise sparsity (owl): A missing secret sauce for pruning llms to high sparsity},
  author={Yin, Lu and Wu, You and Zhang, Zhenyu and Hsieh, Cheng-Yu and Wang, Yaqing and Jia, Yiling and Li, Gen and Jaiswal, Ajay and Pechenizkiy, Mykola and Liang, Yi and others},
  journal={arXiv preprint arXiv:2310.05175},
  year={2023}
}

@article{lin2020channel,
  title={Channel pruning via automatic structure search},
  author={Lin, Mingbao and Ji, Rongrong and Zhang, Yuxin and Zhang, Baochang and Wu, Yongjian and Tian, Yonghong},
  journal={arXiv preprint arXiv:2001.08565},
  year={2020}
}

@article{lee2020layer,
  title={Layer-adaptive sparsity for the magnitude-based pruning},
  author={Lee, Jaeho and Park, Sejun and Mo, Sangwoo and Ahn, Sungsoo and Shin, Jinwoo},
  journal={arXiv preprint arXiv:2010.07611},
  year={2020}
}

@inproceedings{ishii2017layer,
  title={Layer-wise weight decay for deep neural networks},
  author={Ishii, Masato and Sato, Atsushi},
  booktitle={Pacific-Rim Symposium on Image and Video Technology},
  pages={276--289},
  year={2017},
  organization={Springer}
}

@article{qing2024alphalora,
  title={AlphaLoRA: Assigning LoRA Experts Based on Layer Training Quality},
  author={Qing, Peijun and Gao, Chongyang and Zhou, Yefan and Diao, Xingjian and Yang, Yaoqing and Vosoughi, Soroush},
  journal={arXiv preprint arXiv:2410.10054},
  year={2024}
}

@inproceedings{gao2025mola,
  title={MoLA: MoE LoRA with Layer-wise Expert Allocation},
  author={Gao, Chongyang and Chen, Kezhen and Rao, Jinmeng and Liu, Ruibo and Sun, Baochen and Zhang, Yawen and Peng, Daiyi and Guo, Xiaoyuan and Subrahmanian, VS},
  booktitle={Findings of the Association for Computational Linguistics: NAACL 2025},
  pages={5097--5112},
  year={2025}
}

@article{kothapalli2025spikes,
  title={From spikes to heavy tails: Unveiling the spectral evolution of neural networks},
  author={Kothapalli, Vignesh and Pang, Tianyu and Deng, Shenyang and Liu, Zongmin and Yang, Yaoqing},
  journal={Transactions on Machine Learning Research},
  year={2025}
}

@article{he2025alphadecay,
  title={Alphadecay: Module-wise weight decay for heavy-tailed balancing in llms},
  author={He, Di and Tu, Songjun and Jaiswal, Ajay and Shen, Li and Yuan, Ganzhao and Liu, Shiwei and Yin, Lu},
  journal={arXiv preprint arXiv:2506.14562},
  year={2025}
}

@article{nakamura2019adaptive,
  title={Adaptive weight decay for deep neural networks},
  author={Nakamura, Kensuke and Hong, Byung-Woo},
  journal={IEEE Access},
  volume={7},
  pages={118857--118865},
  year={2019},
  publisher={IEEE}
}

@article{wang2025sharpness,
  title={The sharpness disparity principle in transformers for accelerating language model pre-training},
  author={Wang, Jinbo and Wang, Mingze and Zhou, Zhanpeng and Yan, Junchi and Wu, Lei and others},
  journal={arXiv preprint arXiv:2502.19002},
  year={2025}
}

@article{wang2021large,
  title={Large learning rate tames homogeneity: Convergence and balancing effect},
  author={Wang, Yuqing and Chen, Minshuo and Zhao, Tuo and Tao, Molei},
  journal={arXiv preprint arXiv:2110.03677},
  year={2021}
}

@article{ye2021global,
  title={Global convergence of gradient descent for asymmetric low-rank matrix factorization},
  author={Ye, Tian and Du, Simon S},
  journal={Advances in Neural Information Processing Systems},
  volume={34},
  pages={1429--1439},
  year={2021}
}

@article{kunin2024get,
  title={Get rich quick: exact solutions reveal how unbalanced initializations promote rapid feature learning},
  author={Kunin, Daniel and Ravent{\'o}s, Allan and Domin{\'e}, Cl{\'e}mentine and Chen, Feng and Klindt, David and Saxe, Andrew and Ganguli, Surya},
  journal={Advances in Neural Information Processing Systems},
  volume={37},
  pages={81157--81203},
  year={2024}
}

@article{Goldt_2020,
   title={Dynamics of stochastic gradient descent for two-layer neural networks in the teacher–student setup*},
   volume={2020},
   ISSN={1742-5468},
   url={http://dx.doi.org/10.1088/1742-5468/abc61e},
   DOI={10.1088/1742-5468/abc61e},
   number={12},
   journal={Journal of Statistical Mechanics: Theory and Experiment},
   publisher={IOP Publishing},
   author={Goldt, Sebastian and Advani, Madhu S and Saxe, Andrew M and Krzakala, Florent and Zdeborová, Lenka},
   year={2020}}

@article{saxe2019mathematical,
  title={A mathematical theory of semantic development in deep neural networks},
  author={Saxe, Andrew M and McClelland, James L and Ganguli, Surya},
  journal={Proceedings of the National Academy of Sciences},
  volume={116},
  number={23},
  pages={11537--11546},
  year={2019},
  publisher={National Academy of Sciences}
}

@misc{saxe2014exactsolutionsnonlineardynamics,
      title={Exact solutions to the nonlinear dynamics of learning in deep linear neural networks}, 
      author={Andrew M. Saxe and James L. McClelland and Surya Ganguli},
      year={2014},
      eprint={1312.6120},
      archivePrefix={arXiv},
      primaryClass={cs.NE},
      url={https://arxiv.org/abs/1312.6120}, 
}

@article{jacot2018neural,
  title={Neural tangent kernel: Convergence and generalization in neural networks},
  author={Jacot, Arthur and Gabriel, Franck and Hongler, Cl{\'e}ment},
  journal={Advances in neural information processing systems},
  volume={31},
  year={2018}
}

@article{hu2022universality,
  title={Universality laws for high-dimensional learning with random features},
  author={Hu, Hong and Lu, Yue M},
  journal={IEEE Transactions on Information Theory},
  volume={69},
  number={3},
  pages={1932--1964},
  year={2022},
  publisher={IEEE}
}

@article{Mei_2018,
   title={A mean field view of the landscape of two-layer neural networks},
   volume={115},
   ISSN={1091-6490},
   url={http://dx.doi.org/10.1073/pnas.1806579115},
   DOI={10.1073/pnas.1806579115},
   number={33},
   journal={Proceedings of the National Academy of Sciences},
   publisher={Proceedings of the National Academy of Sciences},
   author={Mei, Song and Montanari, Andrea and Nguyen, Phan-Minh},
   year={2018},
   month=jul }

@article{arora2018convergence,
  title={A convergence analysis of gradient descent for deep linear neural networks},
  author={Arora, Sanjeev and Cohen, Nadav and Golowich, Noah and Hu, Wei},
  journal={arXiv preprint arXiv:1810.02281},
  year={2018}
}

@article{arora2019implicit,
  title={Implicit regularization in deep matrix factorization},
  author={Arora, Sanjeev and Cohen, Nadav and Hu, Wei and Luo, Yuping},
  journal={Advances in neural information processing systems},
  volume={32},
  year={2019}
}

@misc{gidel2019implicitregularizationdiscretegradient,
      title={Implicit Regularization of Discrete Gradient Dynamics in Linear Neural Networks}, 
      author={Gauthier Gidel and Francis Bach and Simon Lacoste-Julien},
      year={2019},
      eprint={1904.13262},
      archivePrefix={arXiv},
      primaryClass={cs.LG},
      url={https://arxiv.org/abs/1904.13262}, 
}

@inproceedings{yang2024tensor,
  title={Tensor programs VI: Feature learning in infinite depth neural networks},
  author={Yang, Greg and Yu, Dingli and Zhu, Chen and Hayou, Soufiane},
  booktitle={International Conference on Learning Representations},
  volume={2024},
  pages={55099--55150},
  year={2024}
}

@inproceedings{
he2026one,
title={One {LR} Doesn{\textquoteright}t Fit All: Heavy-Tail Guided Layerwise Learning Rates for {LLM}s},
author={Di He and Songjun Tu and Keyu Wang and Lu Yin and Shiwei Liu},
booktitle={ICLR 2026 2nd Workshop on Deep Generative Model in Machine Learning: Theory, Principle and Efficacy},
year={2026},
url={https://openreview.net/forum?id=Aj3ZWgxYwt}
}
\bibliographystyle{icml2026/icml2026}

\newpage
\appendix
\onecolumn

\section*{Appendix}





\section{Limitations}
Our analysis is restricted to linear networks and focuses on the first few steps of gradient descent. However, unlike prior work that mainly focuses on asymptotic convergence, gradient flow, or kernel-based analyses, our work derives exact closed-form expressions for the gradients and test loss after one and two GD steps, enabling a precise characterization of early training dynamics. In particular, we directly link finite-step layer-wise learning-rate allocation to test loss. This is already nontrivial even for two-layer and three-layer linear networks: at each step, we decompose the gradient into dominant signal-aligned components and smaller residual terms, and rigorously bound the residual terms in operator norm, establishing conditions under which the approximate gradients accurately characterize the test loss dynamics. Extending these theoretical results to nonlinear activations, stochastic optimization, and adaptive learning-rate methods presents a natural and challenging direction. More broadly, understanding how and when balancing learning rates across layers emerges in realistic deep networks may provide new theoretical guidance for optimization strategies beyond the linearized or asymptotic regimes.

\section{More Related Work on Layer-wise Hyperparameter Tuning }
\label{app:more_realted_work}
Besides layer-wise learning rate tuning, strategies for assigning different layer-wise pruning ratios, for both unstructured and structured pruning, have been studied actively. ABCPruner~\citep{lin2020channel} first proposes layer-wise pruning strategies from a heuristic way and try to reduce the search space of possible layer sparsity combinations.
~\citet{lee2020layer} modify magnitude-based pruning by rescaling the importance scores in a layer by a factor dependent on the magnitude of surviving connections in that layer. ~\citet{yin2023outlier} allocate layer-wise pruning ratio proportional to the outlier ratio observed within each layer, thereby facilitating a more effective alignment between layer-wise weight sparsity and outlier ratios.
\texttt{Alphapruning}~\citep{lu2024alpha} assigns layer-wise pruning ratios based on the heavy-tailness across layers in large language models (LLMs), undertrained layers will be pruned more to ensure the post-pruning performance does not degrade aggressively across all layers, thereby achieving layer balancing.

Furthermore,~\citet{ishii2017layer, nakamura2019adaptive, he2025alphadecay} introduce layer-wise weight decay schedulers and works such as \texttt{HAWQ}~\citep{dong2019hawq} and \texttt{Q-BERT}~\citep{shen2020q} propose hessian-aware layer-wise quantization approaches. ~\citet{qing2024alphalora, gao2025mola} allocate layer-wise experts for MoE layers in transformers based on certain representation metrics.

\section{Norm Analysis of  Update Gradient Matrices}
\begin{lemma}
\label{app: Hanson-Wright Inequality}
    (Hanson-Wright Inequality \citep{vershynin2018high}). Let $\vx=(x_1, \cdots, x_n)\in \mathbb{R}^n$ be a random vector with independent, mean zero, sub-gaussian coordinates. Let $\mA$ be an $n \times n$ matrix. Then for every $t \geq 0$, we have $$ \mathbb{P} \left\{ \left| \vx^{\top}\mA\vx-\mathbb{E}\vx^{\top}\mA\vx \right| \geq t \right\} \leq 2\exp \left[-c\min(\frac{t^2}{K^4\norm{A}^2_F}, \frac{t}{K^2\norm{A}})\right],$$
    where $K=\max_i \norm{x_i}_{{\psi}_2}$.
\end{lemma}

\begin{lemma}
\label{app:Concentration of Lipschitz}
    (Concentration of Lipschitz function on the sphere \citep{vershynin2018high}). Consider a random vector $\vx \in \sqrt{n}\sS^{n-1}$. Given a Lipschitz function $f: \sqrt{n}\sS^{n-1} \rightarrow \mathbb{R}.$ Then 
     $$ \mathbb{P} \left\{ \left| f(\vx)-\mathbb{E} f(\vx) \right| \geq t \right\} \leq 2\exp(-\frac{ct^2}{\norm{f}^2_{Lip}}).$$
\end{lemma}

\begin{lemma}
\label{app:bernstein}
    (Bernstein Inequality \citep{vershynin2018high}). Let $x_1,\cdots,x_n$ be independent, mean zero, sub-exponential random variables. Then, for every $t\geq 0$, we have 
    $$\mathbb{P} \left\{ \left| \sum_{i=1}^n x_i \right| \geq t \right\} \leq 2\exp\left [-c\min (\frac{t^2}{\sum_{i=1}^n \norm{x_i}^2_{\psi_1}}, \frac{t}{\max_i\norm{x_i}_{\psi_1}})\right] $$
    where $c>0$ is an absolute constant.
\end{lemma}

\subsection{Norm Analysis of One-step Update Gradient Matrices}

\subsubsection{Three-layer Neural Network under Orthogonal Initialization } \label{app:3-NN-orthgonal_one-step_prop}

\begin{proposition}
\label{app:proposition-3-layer-orthogonal} 
 (Three-layer NN setting under  Orthogonal initialization.) Under Assumption~\ref{sec:assumption-whiten}, we have gradient approximation,
 \begin{equation}
 \begin{aligned}
          \norm{\mG_1^0-\mA_1^0} &\leq \frac{1}{\sqrt{h}-1}\norm{\mG_1^0}, \\
          \norm{\mG_2^0-\mA_2^0} &\leq  \frac{1}{\sqrt{h}-1}\norm{\mG_2^0}. \\
 \end{aligned}
 \end{equation}
And we have
 \\
 \begin{align}
     Small \ lr: \eta_1=\Theta(\sqrt{h}) \Rightarrow  &\norm{\mW_1^1-\mW_1^0} \asymp \norm{\mW_1^0} \\
     \eta_2=\Theta(\sqrt{h}) \Rightarrow  &\norm{\mW_2^1-\mW_2^0} \asymp \norm{\mW_2^0} \\
     Large \ lr: \eta_1=\Theta(h) \Rightarrow & \norm{\mW_1^1-\mW_1^0}_F\asymp\norm{\mW_1^0}_F\\
    \eta_2=\Theta(h) \Rightarrow & \norm{\mW_2^1-\mW_2^0} _F\asymp\norm{\mW_2^0}_F
 \end{align}
\end{proposition}

\paragraph{Proof of Proposition~\ref{app:proposition-3-layer-orthogonal}.} 
Note that $\mA_1^0=\frac{1}{\sqrt{h}}\vbeta^{*}\va^{\top}\mW_2^{0^{\top}}$, $\mA_2^0=\frac{1}{\sqrt{h}}\mW_1^{0^{\top}}\vbeta^{*}\va^{\top}$ are both rank-1 matrices. Here we consider the orthogonal  initialization where we  have $\mX^{\top}\mX=\mX\mX^{\top}=h\mI, \mW_1^{0^{\top}}\mW_1^0=\mW_1^0\mW_1^{0^{\top}}=\mI, \mW_2^{0^{\top}}\mW_2^0=\mW_2^0\mW_2^{0^{\top}}=\mI, \va^{\top}\va=1, \mathbb{E}[ \va \va^{\top}]=\frac{1}{h}\mI,   {\vbeta^*}^{{\top}}{\vbeta^*}=1, \mathbb{E}[{\vbeta^*}{\vbeta^*}^{{\top}} ]=\frac{1}{h}\mI$. Based on this,
\begin{align}
    \norm{\mA_1^0} &=\norm{\mA_1^0}_F= \sqrt{tr({\mA_1^0}^{\top}\mA_1^0)}=\frac{1}{\sqrt{h}} \\
    \norm{\mA_2^0} &= \norm{\mA_2^0}_F=\sqrt{tr({\mA_2^0}^{\top}\mA_2^0)}=\frac{1}{\sqrt{h}}. 
\end{align}
We also have $\mB_1^0=\frac{1}{h} \mW_1^0\mW_2^0\va\va^{\top}\mW_2^{0^{\top}}$, $\mB_2^0=\frac{1}{h}\mW_1^{0^{\top}}\mW_1^0\mW_2^0\va\va^{\top}$ are both rank-1 matrices. We have 
\begin{align}
    \norm{\mB_1^0} &=\norm{\mB_1^0}_F= \sqrt{tr({\mB_1^0}^{\top}\mB_1^0)}=\frac{1}{h} \\
    \norm{\mB_2^0} &= \norm{\mB_2^0}_F=\sqrt{tr({\mB_2^0}^{\top}\mB_2^0)}=\frac{1}{h}. 
\end{align}
Since $\mG_1^0= \mB_1^0-\mA_1^0$, $\mG_2^0= \mB_2^0-\mA_2^0$, we obtain that 
\begin{align*}
    \norm{\mG_1^0-\mA_1^0} & \leq \frac{1}{\sqrt{h}}\norm{\mA_1^0}\leq \frac{1}{\sqrt{h}}(\norm{\mG_1^0}+\norm{\mG_1^0-\mA_1^0}) \\
        \norm{\mG_2^0-\mA_2^0} & \leq \frac{1}{\sqrt{h}}\norm{\mA_2^0}\leq \frac{1}{\sqrt{h}}(\norm{\mG_1^0}+\norm{\mG_2^0-\mA_2^0}) .
\end{align*}
Thus, we get that 
 \begin{equation}
 \begin{aligned}
          \norm{\mG_1^0-\mA_1^0} &\leq \frac{1}{\sqrt{h}-1}\norm{\mG_1^0}, \\
          \norm{\mG_2^0-\mA_2^0} &\leq  \frac{1}{\sqrt{h}-1}\norm{\mG_2^0}. \\
 \end{aligned}
 \end{equation}

Based on this, we can get $\sqrt{h}\norm{\mG_1^0}= \Theta_{h, \mathbb{P}}(1), \sqrt{h}\norm{\mG_1^0}_F= \Theta_{h, \mathbb{P}}(1),\sqrt{h}\norm{\mG_2^0}= \Theta_{h, \mathbb{P}}(1),\sqrt{h}\norm{\mG_2^0}_F= \Theta_{h, \mathbb{P}}(1)$.

Since we have $\norm{\mW_1^0}=1, \norm{\mW_1^0}_F=\sqrt{h}, \norm{\mW_2^0}=1, \norm{\mW_2^0}_F=\sqrt{h}$, based on Assumption~\ref{sec:assumption-2}, we have 
 \begin{align*}
     Small \ lr: \eta_1=\Theta(\sqrt{h}) \Rightarrow  &\norm{\mW_1^1-\mW_1^0} \asymp \norm{\mW_1^0} \\
     \eta_2=\Theta(\sqrt{h}) \Rightarrow  &\norm{\mW_2^1-\mW_2^0} \asymp \norm{\mW_2^0} \\
     Large \ lr: \eta_1=\Theta(h) \Rightarrow & \norm{\mW_1^1-\mW_1^0}_F\asymp\norm{\mW_1^0}_F\\
    \eta_2=\Theta(h) \Rightarrow & \norm{\mW_2^1-\mW_2^0} _F\asymp\norm{\mW_2^0}_F
 \end{align*}
 \hfill $\square$

\subsubsection{Two-layer Neural Network under  Orthogonal Initialization }\label{app:2-NN-orthgonal_one-step_prop}

\begin{proposition}
\label{app:proposition-2-layer-orthogonal}
 (Two-layer NN setting under  Orthogonal initialization.) Under Assumption~\ref{sec:assumption-whiten}, we have gradient approximation,
 \begin{equation}
 \begin{aligned}
          \norm{{\mG_1^0}-{\mA_1^0}} &\leq \frac{1}{\sqrt{h}-1}\norm{{\mG_1^0}}, \\
          \norm{{\mG_2^0}-{\mA_2^0}} &\leq  \frac{1}{\sqrt{h}-1}\norm{{\mG_2^0}}. \\
 \end{aligned}
 \end{equation}
And we have
 \\
 \begin{align}
     Large \ lr: \eta_1=\Theta(h\sqrt{h}) \Rightarrow & \norm{{\mW_1^1}-{\mW_1^0}}_F\asymp\norm{\mW_1^0}_F\\
    \eta_2=\Theta(h\sqrt{h}) \Rightarrow & \norm{{\mW_2^1}-\mW_2^0} _F\asymp\norm{\mW_2^0}_F
 \end{align}
\end{proposition}

\paragraph{Proof of Proposition~\ref{app:proposition-2-layer-orthogonal}.} 
Note that ${\mA_1^0}=\frac{1}{h}\mM\mW_2^{0^{\top}}$, ${\mA_2^0}=\frac{1}{h}\mW_1^{0^{\top}}\mM$. Here we consider the  orthogonal  initialization where we  have $\mX^{\top}\mX=\mX\mX^{\top}=h\mI, \mW_1^{0^{\top}}\mW_1^0=\mW_1^0\mW_1^{0^{\top}}=\mI, \mW_2^{0^{\top}}\mW_2^0=\mW_2^0\mW_2^{0^{\top}}=\mI,    {\mM}^{{\top}} {\mM}= {\mM}\mM^{{\top}}=\frac{1}{h}\mI$. Based on this,
\begin{align}
    \norm{{\mA_1^0}}=\frac{1}{h\sqrt{h}} ,&\norm{{\mA_1^0}}_F= \sqrt{tr({{\mA_1^0}}^{\top}{\mA_1^0})}=\frac{1}{h} \\
    \norm{\mA_2^0}=\frac{1}{h\sqrt{h}} , & \norm{{{\mA_2^0}}}_F=\sqrt{tr({{\mA_2^0}}^{\top}{\mA_2^0})}=\frac{1}{h}. 
\end{align}
We also have ${\mB_1^0}=\frac{1}{h^2} \mW_1^0\mW_2^0\mW_2^{0^{\top}}$, ${\mB_2^0}=\frac{1}{h}\mW_1^{0^{\top}}\mW_1^0\mW_2^0$, so we can get that
\begin{align}
    \norm{{\mB_1^0}}=\frac{1}{h^2}, &\norm{{\mB_1^0}}_F= \sqrt{tr({{\mB_1^0}}^{\top}{\mB_1^0})}=\frac{1}{h\sqrt{h}} \\
    \norm{{\mB_2^0}}=\frac{1}{h^2}, &\norm{\mB_2^0}_F=\sqrt{tr({{\mB_2^0}}^{\top}{\mB_2^0})}=\frac{1}{h\sqrt{h}}. 
\end{align}
Since ${\mG_1^0}= {\mB_1^0}-{\mA_1^0}$, ${\mG_2^0}= {\mB_2^0}-{\mA_2^0}$, we obtain that 
\begin{align*}
    \norm{{\mG_1^0}-{\mA_1^0}} & \leq \frac{1}{\sqrt{h}}\norm{{\mA_1^0}}\leq \frac{1}{\sqrt{h}}(\norm{{\mG_1^0}}+\norm{{\mG_1^0}-{\mA_1^0}}) \\
        \norm{{\mG_2^0}-{\mA_2^0}} & \leq \frac{1}{\sqrt{h}}\norm{{\mA_2^0}}\leq \frac{1}{\sqrt{h}}(\norm{{\mG_1^0}}+\norm{{\mG_2^0}-{\mA_2^0}}) .
\end{align*}
Thus, we get that 
 \begin{equation}
 \begin{aligned}
          \norm{{\mG_1^0}-{\mA_1^0}} &\leq \frac{1}{\sqrt{h}-1}\norm{{\mG_1^0}}, \\
          \norm{{\mG_2^0}-{\mA_2^0}} &\leq  \frac{1}{\sqrt{h}-1}\norm{{\mG_2^0}}. \\
 \end{aligned}
 \end{equation}

Based on this, we can get $h\sqrt{h}\norm{{\mG_1^0}}= \Theta_{h, \mathbb{P}}(1), h\norm{{\mG_1^0}}_F= \Theta_{h, \mathbb{P}}(1),h\sqrt{h}\norm{{\mG_2^0}}= \Theta_{h, \mathbb{P}}(1),h\norm{{\mG_2^0}}_F= \Theta_{h, \mathbb{P}}(1)$.

Since we have $\norm{\mW_1^0}=1, \norm{\mW_1^0}_F=\sqrt{h}, \norm{\mW_2^0}=1, \norm{\mW_2^0}_F=\sqrt{h}$, based on Assumption~\ref{sec:assumption-2}, we have 
 \begin{align*}
     Large \ lr: \eta_1=\Theta(h\sqrt{h}) \Rightarrow & \norm{{\mW_1^1}-\mW_1^0}_F\asymp\norm{\mW_1^0}_F\\
    \eta_2=\Theta(h\sqrt{h}) \Rightarrow & \norm{{\mW_2^1}-\mW_2^0} _F\asymp\norm{\mW_2^0}_F
 \end{align*}
 \hfill $\square$

\subsubsection{Three-layer Neural Network under Gaussian Initialization}
Here, we present a one-step update gradient-norm analysis for three-layer neural networks under  Gaussian initialization. To ensure generality, here we incorporate noise with different $d,h,n$.

\label{app:norm property of 3-layer NN}
\begin{lemma}
\label{app:lemma_norm}
Consider that $\mW_1 \in \sR^{d \times h},\mW_2 \in \sR^{h \times h}, \va \in \sR^h$ are the first two hidden layers and  last layer weights, respectively, with entries sampled i.i.d as follows $ \sqrt{d}\left[\mW_1\right]_{i, j} \sim \mathcal{N}(0,1)$, $ \sqrt{h}\left[\mW_2\right]_{i, j} \sim \mathcal{N}(0,1)$, $\sqrt{h}\left[\va\right]_{i} \sim \mathcal{N}(0,1), \forall i \in [h], j \in [d]$. We define  \begin{align*} 
    \mA_1^0&=\frac{1}{n\sqrt{h}} \mX^{\top}\vy\va^{\top}\mW_2^{0^{\top}}\\
    \mB_1^0&=\frac{1}{nh} \mX^{\top}\mX\mW_1^0\mW_2^0\va\va^{\top}\mW_2^{0^{\top}}\\
    \mA_2^0&=\frac{1}{n\sqrt{h}}\mW_1^{0^{\top}}\mX^{\top}\vy\va^{\top}\\
    \mB_2^0&=\frac{1}{nh}\mW_1^{0^{\top}} \mX^{\top}\mX\mW_1^0\mW_2^0\va\va^{\top},
\end{align*}we have
\begin{enumerate}
\renewcommand{\labelenumi}{(\roman{enumi})}
    \item $\mathbb{E}\norm{\mA_1^0} \leq \mathbb{E}\norm{\mA_1^0}_{F}\leq C\left(\frac{1}{\sqrt{h}}+\frac{d}{n\sqrt{h}}+\sqrt{\frac{1}{h}+\frac{d}{nh}} \right)$
    
    \item $\mathbb{E}\norm{\mB_1^0} \leq \mathbb{E}\norm{\mB_1^0}_{F}\leq C\sqrt{1+\frac{h}{d}}\left(\frac{1}{d}+\frac{1}{n} \right)$

    \item $\mathbb{P}\left(\norm{\mA_1^0}_F\geq C \left(\frac{6}{\sqrt{h}}+\frac{d}{n\sqrt{h}}+\frac{5\sqrt{d}}{\sqrt{nh}} \right )\right )\leq 4e^{-ch}+6e^{-cn}  $ \label{ieq:A_1_F}

    \item $\mathbb{P}\left( \norm{\mA_1^0} \leq \frac{\rho_e\sqrt{2d}}{8\sqrt{nh}}\right)\leq 6\left( e^{-c\min \left\{ \frac{nd^2}{(n^2+d^2)},\frac{nd}{n+d} \right\}}+e^{-cn}+ e^{-ch}\right)$ \label{ieq:A_1}

    \item $\mathbb{P}\left( \norm{\mB_1^0}_F \geq C\left( \frac{8}{h}+\frac{2d}{nh}+\frac{9\sqrt{d}}{\sqrt{n}h}+\frac{4}{\sqrt{nh}}+\frac{4}{\sqrt{hd}}\right)\right)\leq 4e^{-cn}+8e^{-ch}+2e^{-cd}$

        \item $\mathbb{E}\norm{\mA_2^0} \leq \mathbb{E}\norm{\mA_2^0}_{F}\leq C\left( \frac{\sqrt{(h+d)(n+d)}(\sqrt{n+d}+\sqrt{n})}{n\sqrt{hd}} \right)$
    
    \item $\mathbb{E}\norm{\mB_2^0} \leq \mathbb{E}\norm{\mB_2^0}_{F}\leq C \left( \frac{1}{d}+\frac{1}{h}+\frac{1}{n}+\frac{d}{nh}\right)$

    \item $\mathbb{P}\left(\norm{\mA_2^0}_F\geq C \left(\frac{12}{\sqrt{h}}+\frac{2d}{n\sqrt{h}}+\frac{10\sqrt{d}}{\sqrt{nh}}+\frac{6\sqrt{d}}{d}+\frac{5\sqrt{n}}{n}+\frac{\sqrt{d}}{n} \right )\right )\leq 2e^{-cd}+2e^{-ch}+6e^{-cn}$ \label{ieq:A_2_F}

    \item $ \mathbb{P}\left( \norm{\mA_2^0} \leq \frac{\rho_e\sqrt{2d}}{8\sqrt{nh}}\right)\leq$ \\ $6\left( e^{-c\min \left\{ \frac{nd^2}{(n^2+d^2)},\frac{nd}{n+d}, \frac{nd^4}{(n^2+d^2)(h^2+d^2)}\right\}}+e^{-cn}+ e^{-ch}+e^{-cd}+nhd^2e^{-c\sqrt{d}}\right)$ \label{ieq:A_2}

    \item $\mathbb{P}\left( \norm{\mB_2^0}_F \geq C\left( \frac{4}{h}+\frac{32d}{nh}+\frac{16\sqrt{d}}{h\sqrt{n}}+\frac{4}{\sqrt{hd}}+\frac{16}{\sqrt{nh}}+\frac{4}{\sqrt{nd}} +\frac{4}{n}+\frac{16\sqrt{d}}{n\sqrt{h}} \right)\right)$\\ $\leq  8e^{-cd}+6e^{-ch}$
\end{enumerate}
\end{lemma}

\begin{remark} 

Note that $\mA_1^0$ is a \textit{rank-1} matrix, so the upper bound  and lower bound  of $\norm{\mA_1^0}_F$ and $\norm{\mA_1^0}$ is similar. The same is to $\mA_2^0$.

\end{remark}

\paragraph{Proof of Lemma~\ref{app:lemma_norm}.} We analyze these matrices of interest separately. 
\paragraph{ Part of (i).} Notice that
\begin{align}
\label{app:part_of_i}
    \norm{\mA_1^0}_F& \leq \frac{1}{n\sqrt{h}}\norm{\mX^{\top}\mX\vbeta^{*}\va^{\top}\mW_2^{0^{\top}}}_F+  \frac{1}{n\sqrt{h}}\norm{\mX^{\top}\vxi\va^{\top}\mW_2^{0^{\top}}}_F  \notag \\
          & \leq  \frac{1}{n\sqrt{h}}\norm{\mX^{\top}\mX\vbeta^{*}} \norm{\va^{\top}\mW_2^{0^{\top}}}+  \frac{1}{n\sqrt{h}}\norm{\mX^{\top}\vxi}\norm{\va^{\top}\mW_2^{0^{\top}}}  \notag \\
          & \leq \frac{1}{n\sqrt{h}}\norm{\mX}(\norm{\mX}+\norm{\vxi})\norm{\va}\norm{\mW_2^0}.
\end{align}
Based on basic probability theory, we know that Gaussian random matrices and vectors satisfy
\begin{equation}
\label{app: gauss_expect}
\begin{aligned}   
\mathbb{E}(\norm{\va})=1&, \quad \mathbb{E}(\norm{\vxi})=\rho_e\sqrt{n} \\
\mathbb{E}(\norm{\mW_2^0}^2)\leq C_0&, \quad \mathbb{E}(\norm{\mX}^2)\leq C_1(n+d),
\end{aligned}
\end{equation}
where $C_0$ and $C_1$ are consts.

Based on (~\ref{app: gauss_expect}), we obtain
$$ \mathbb{E}(\norm{\mA_1^0}) \leq  \mathbb{E}(\norm{\mA_1^0})_F \leq C\left(\frac{1}{\sqrt{h}}+\frac{d}{n\sqrt{h}}+\sqrt{\frac{1}{h}+\frac{d}{nh}} \right). $$ 


\paragraph{ Part of (ii).}  Notice that
\begin{align}
\label{app:part_of_ii}
    \norm{\mB_1^0}_F& \leq \frac{1}{nh}\norm{\mX^{\top}\mX\mW_1^0\mW_2^0\va}\norm{\va^{\top}\mW_2^{0^{\top}}}  \notag \\
          & \leq  \frac{1}{nh}\norm{\mX}^2\norm{\mW_1^0}\norm{\va}^2\norm{\mW_2^{0^{\top}}}^2
\end{align}

Besides (~\ref{app:bernstein}), based on basic probability theory, we know that Gaussian random matrices and vectors satisfy
\begin{equation}
\label{app:gauss_expect_2}
    \mathbb{E}(\norm{W_1^0}^2)\leq C_0(1+\frac{h}{d}).
\end{equation}

Based on (~\ref{app: gauss_expect}) and (~\ref{app:gauss_expect_2}), we obtain
$$ \mathbb{E}(\norm{\mB_1^0}) \leq  \mathbb{E}(\norm{\mB_1^0})_F \leq C\sqrt{1+\frac{h}{d}}\left(\frac{1}{d}+\frac{1}{n} \right). $$


\paragraph{ Part of (iii).}  Based on random vector and matrix concentration property of $\norm{\mX},\norm{\vxi}, \norm{\va}$ and $ \norm{\mW_2^0}$ (e.g. derived by Lemma~\ref{app:bernstein}: for any $t\geq 0$,

\begin{equation}
\label{app:norm_concentration}
\begin{aligned}
    \mathbb{P}(\left|\norm{\vxi}-\rho_e\sqrt{n}
    \right|\geq \frac{1}{2}\rho_e\sqrt{n} )\leq 2e^{-cn}&, \quad \mathbb{P}(\left|\norm{\va}-1
    \right|\geq \frac{1}{2} )\leq 2e^{-ch},  \\ 
     \mathbb{P}(\norm{\mX}\geq \sqrt{n}+\sqrt{d}+t)\leq 2e^{-ct^2}&, \quad \mathbb{P}(\norm{\mW_2^0}\geq 3)\leq 2e^{-ch}
\end{aligned}
\end{equation}
Hence, from (~\ref{app:part_of_i}) and (~\ref{app:norm_concentration}), we arrive at 
$$ \mathbb{P}\left(\norm{\mA_1^0}_F\geq C \left(\frac{2}{\sqrt{h}}+\frac{d}{n\sqrt{h}}+\frac{3\sqrt{d}}{\sqrt{nh}}+\frac{t^2}{n\sqrt{h}}+\frac{3t}{\sqrt{nh}}+\frac{2t\sqrt{d}}{n\sqrt{h}} \right )\right )\leq 2(e^{-cn}+2e^{-ch}+2e^{-ct^2})$$
Thus we can take $t=\sqrt{n}$ to obtain the result that:
$$ \mathbb{P}\left(\norm{\mA_1^0}_F\geq C \left(\frac{6}{\sqrt{h}}+\frac{d}{n\sqrt{h}}+\frac{5\sqrt{d}}{\sqrt{nh}} \right )\right )\leq 4e^{-ch}+6e^{-cn}.  $$

\paragraph{ Part of (iv).}
Here we try to give the lower bound of $\norm{\mA_1}$, since $\mA_1$ is rank-1 matrix, we have

\begin{align*}
    \norm{\mA_1^0}^2&=\frac{1}{n^2h}\norm{\mX^{\top}\vy\va^{\top}\mW_2^{0^\top}}^2 =\frac{1}{n^2h}\norm{\mX^{\top}\vy\va^{\top}\mW_2^{0^\top}}_F^2\\
    &=\frac{1}{n^2h}tr(\mW_2^{0}\va\vy^{\top}\mX\mX^{\top}\vy\va^{\top}\mW_2^{0^\top})\\
    &=\frac{1}{n^2h}tr(\vy^{\top}\mX\mX^{\top}\vy\va^{\top}\mW_2^{0^\top}\mW_2^{0}\va) \\
    &=\frac{1}{n^2h}\norm{\mW_2^{0}\va} ^2\norm{\mX^{\top}\vy}^2 \\
    & \geq \frac{1}{n^2h}\norm{\mW_2^{0}\va}^2\left( \vxi^{\top} \mX\mX^{\top}\vxi+2\vbeta^{*^{\top}}\mX^{\top}\mX\mX^{\top}\vxi \right).
\end{align*}

Following \citep{ba2022high}, we define events $\mE_1$, $\mE_2$ and $\mE_3$ by

$$ \mE_1:= \left \{ \left| tr(\mX\mX^{\top}) -nd\right| \leq  \frac{nd}{2}\right \}, \quad \mE_2:= \left \{ \norm{\mX}   \leq \sqrt{d}+2\sqrt{n}\right \}, \quad  \mE_3:= \left \{ \norm{\mW_2^0}  \leq 3 \right \}.$$  

Based on (~\ref{app:norm_concentration}) and Lemma~\ref{app:bernstein}, we know that $$
\mathbb{P}(\mE_1)\geq 1-2e^{-cn}, \quad \mathbb{P}(\mE_2)\geq 1-2e^{-cn}, \quad \mathbb{P}(\mE_3)\geq 1-2e^{-ch}.
$$

Thus condition on $\mE_1 \cap \mE_2$, it is easy to find that $\mathbb{E}\norm{\mX\mX^{\top}}_F^2 \leq C_2 n(n^2+d^2)$, also based on (~\ref{app: gauss_expect}),  by Lemma~\ref{app: Hanson-Wright Inequality}, we have $$ \mathbb{P}\left( \vxi^{\top} \mX\mX^{\top}\vxi \leq \frac{\rho_e^2}{2}nd-t | \mE_1 \cap \mE_2 \right  ) \leq 2e^{-c\min \left\{ \frac{t^2}{n(n^2+d^2)},\frac{t}{n+d} \right\}}.$$

Choosing $t=\frac{\rho_e^2}{4}nd$, we have 

\begin{equation}
\mathbb{P}\left( \vxi^{\top} \mX\mX^{\top}\vxi \leq \frac{\rho_e^2}{8}nd  \right  ) \leq 2e^{-c\min \left\{ \frac{nd^2}{(n^2+d^2)},\frac{nd}{n+d} \right\}}+4e^{-cn}.
\label{app:hanson-wright}
\end{equation}

In addition, by Lemma~\ref{app:Concentration of Lipschitz}, condition on $\mE_2$, we know that $$\norm{\vbeta^{*^{\top}}\mX^{\top}\mX\mX^{\top}\vxi}_{\psi_2}\leq C_3\sqrt{n(n^2+d^2)},$$ one can easily see that
$$ \mathbb{P}\left( \left| \vbeta^{*^{\top}}\mX^{\top}\mX\mX^{\top}\vxi \right|\geq t | \mE_2 \right)\leq 2e^{-\frac{ct^2}{n(n^2+d^2)}}.$$

Thus, let $t=\frac{\rho_e^2}{32}nd$, we obtain 
\begin{equation}
    \mathbb{P}\left( \left| \vbeta^{*^{\top}}\mX^{\top}\mX\mX^{\top}\vxi \right|\geq \frac{\rho_e^2}{32}nd \right)\leq 2e^{-\frac{cnd^2}{(n^2+d^2)}}+2e^{-cn}.
\label{app:lipchitz on xi}
\end{equation}
Similarly, by Lemma~\ref{app:Concentration of Lipschitz}, condition on $\mE_3$, we know that $\norm{\norm{\mW_2^0\va}-1}_{\psi_2}\leq \frac{3C_3}{\sqrt{h}}$, we obtain that $$ \mathbb{P}\left( \left| \norm{\mW_2^0\va} \right|\geq t | \mE_3 \right)\leq 2e^{-cht^2}. $$

Thus, let $t=\frac{1}{2}$, we obtain 
\begin{equation}
    \mathbb{P}\left(  \norm{\mW_2^0\va} \leq \frac{1}{2} \right)\leq 4e^{-ch}.
\label{app:lipchitz on a}
\end{equation}

Based on (~\ref{app:hanson-wright}), (~\ref{app:lipchitz on xi}) and (~\ref{app:lipchitz on a}), we arrive at 

$$ \mathbb{P}\left( \norm{\mA_1^0} \leq \frac{\rho_e\sqrt{2d}}{8\sqrt{nh}}\right)\leq 6\left( e^{-c\min \left\{ \frac{nd^2}{(n^2+d^2)},\frac{nd}{n+d} \right\}}+e^{-cn}+ e^{-ch}\right)$$
\paragraph{ Part of (v).} Based on random matrix concentration property of $\norm{\mW_1^0}$, we have for any $t\geq 0$
\begin{equation}
\label{app:norm_concentration_2}
    \mathbb{P}(\norm{\mW_1^0}\geq 2+\sqrt{\frac{h}{d}})\leq 2e^{-cd}
\end{equation}
Based on (~\ref{app:part_of_ii}), (~\ref{app:norm_concentration}) and (~\ref{app:norm_concentration_2}), similar to proof of (ii), by choosing $t=\sqrt{n}$, we arrive at

$$\mathbb{P}\left( \norm{\mB_1^0}_F \geq C\left( \frac{8}{h}+\frac{2d}{nh}+\frac{9\sqrt{d}}{\sqrt{n}h}+\frac{4}{\sqrt{nh}}+\frac{4}{\sqrt{hd}}\right)\right)\leq 4e^{-cn}+8e^{-ch}+2e^{-cd}.$$

\paragraph{ Part of (vi).} Notice that
\begin{align}
\label{app:part_of_vi}
    \norm{\mA_2^0}_F& \leq \frac{1}{n\sqrt{h}}\norm{\mW_1^{0^{\top}}\mX^{\top}\mX\vbeta^{*}\va^{\top}}_F+  \frac{1}{n\sqrt{h}}\norm{\mW_1^{0^{\top}}\mX^{\top}\vxi\va^{\top}}_F  \notag \\
          & \leq  \frac{1}{n\sqrt{h}}\norm{\mW_1^{0^{\top}}\mX^{\top}\mX\vbeta^{*}} \norm{\va^{\top}}+  \frac{1}{n\sqrt{h}}\norm{\mW_1^{0^{\top}}\mX^{\top}\vxi}\norm{\va^{\top}}  \notag \\
          & \leq \frac{1}{n\sqrt{h}}\norm{\mW_1^0}\norm{\mX}(\norm{\mX}+\norm{\vxi})\norm{\va}
\end{align}

Based on (~\ref{app: gauss_expect}) and (~\ref{app:gauss_expect_2}), we obtain
$$ \mathbb{E}(\norm{\mA_2^0}) \leq  \mathbb{E}(\norm{\mA_2^0})_F \leq C\left( \frac{\sqrt{(h+d)(n+d)}(\sqrt{n+d}+\sqrt{n})}{n\sqrt{hd}} \right). $$ 


\paragraph{ Part of (vii).}  Notice that
\begin{align}
\label{app:part_of_vii}
    \norm{\mB_2^0}_F& \leq \frac{1}{nh}\norm{\mW_1^{0^{\top}}\mX^{\top}\mX\mW_1^0\mW_2^0\va}\norm{\va^{\top}}  \notag \\
          & \leq  \frac{1}{nh}\norm{\mX}^2\norm{\mW_1^0}^2\norm{\va}^2\norm{\mW_2^{0^{\top}}}
\end{align}

Based on (~\ref{app: gauss_expect}) and (~\ref{app:gauss_expect_2}), we obtain
$$ \mathbb{E}(\norm{\mB_2^0}) \leq  \mathbb{E}(\norm{\mB_2^0})_F \leq C \left( \frac{1}{d}+\frac{1}{h}+\frac{1}{n}+\frac{d}{nh}\right). $$

\paragraph{Part of (viii).} 
Hence, from(~\ref{app:norm_concentration}) (~\ref{app:norm_concentration_2}) and  (~\ref{app:part_of_vi}), we arrive at 
\begin{align*}
&\mathbb{P}\left(\norm{\mA_2^0}_F\geq C 
\left( 2+\sqrt{\frac{h}{d}}\right)\left(\frac{2}{\sqrt{h}}+\frac{d}{n\sqrt{h}}+\frac{3\sqrt{d}}{\sqrt{nh}}+\frac{t^2}{n\sqrt{h}}+\frac{3t}{\sqrt{nh}}+\frac{2t\sqrt{d}}{n\sqrt{h}} \right )\right )\\&\leq 2(e^{-cn}+2e^{-ch}+2e^{-ct^2}). 
\end{align*}

Thus we can take $t=\sqrt{n}$ to obtain the result that:
$$ \mathbb{P}\left(\norm{\mA_2^0}_F\geq C \left(\frac{12}{\sqrt{h}}+\frac{2d}{n\sqrt{h}}+\frac{10\sqrt{d}}{\sqrt{nh}}+\frac{6\sqrt{d}}{d}+\frac{5\sqrt{n}}{n}+\frac{\sqrt{d}}{n} \right )\right )\leq 2e^{-cd}+2e^{-ch}+6e^{-cn}.  $$

\paragraph{ Part of (ix).}
Here we try to give the lower bound of $\norm{\mA_2}$, since $\mA_2$ is rank-1 matrix, we have

\begin{align*}
    \norm{\mA_2^0}^2&=\frac{1}{n^2h}\norm{\mW_1^{0^\top}\mX^{\top}\vy\va^{\top}}^2 =\frac{1}{n^2h}\norm{\mW_1^{0^\top}\mX^{\top}\vy\va^{\top}}_F^2\\
    &=\frac{1}{n^2h}tr(\va\vy^{\top}\mX\mW_1^{0}\mW_1^{0^\top}\mX^{\top}\vy\va^{\top})\\
    &=\frac{1}{n^2h}tr(\vy^{\top}\mX\mW_1^{0}\mW_1^{0^\top}\mX^{\top}\vy\va^{\top}\va) \\
    &=\frac{1}{n^2h}\norm{\va} ^2\norm{\mW_2^{0^{\top}}\mX^{\top}\vy}^2 \\
    & \geq \frac{1}{n^2h}\norm{\va}^2\left( \vxi^{\top} \mX\mW_1^{0}\mW_1^{0^\top}\mX^{\top}\vxi+2\vbeta^{*^{\top}}\mX^{\top}\mX\mW_1^{0}\mW_1^{0^\top}\mX^{\top}\vxi \right).
\end{align*}

Following \citep{ba2022high}, we define events $\mE_2$, $\mE_4$ and $\mE_5$ by

\begin{align*}
    &\mE_2:= \left \{ \norm{\mX}   \leq \sqrt{d}+2\sqrt{n}\right \}, \\  &\mE_4:= \left \{ \norm{\mW_1^0}  \leq 2+\frac{h}{d} \right \}, \\&\mE_5:= \left \{ \left| tr(\mX\mW_1^{0}\mW_1^{0^\top}\mX^{\top}) -nh\right| \leq  \frac{nh}{2}\right \}.
\end{align*}

Based on (~\ref{app:norm_concentration}) and  (~\ref{app:norm_concentration_2}), we know that $$
\mathbb{P}(\mE_2)\geq 1-2e^{-cn}, \quad \mathbb{P}(\mE_4)\geq 1-2e^{-cd}.
$$

We also know that by sub-gaussian and sub-exponential concentration inequality  
\begin{equation}
\label{app:norm_concentration_3}
   \mathbb{P}(\left | \mW_{1_{i,j}}^0 \right |\leq1)\geq 1-2e^{-cd}, \quad \mathbb{P}(\left | {\mW_{1_{i,j}}^{0}}^2-1 \right |\leq 1)\geq 1-2e^{-c\sqrt{d}}. 
\end{equation}

Based on Lemma~\ref{app:bernstein} and (~\ref{app:norm_concentration_3}), we have $$
\mathbb{P}(\mE_5)\geq 1-2e^{-cn}-2nhde^{-cd}-2nh(d^2-d)e^{-c\sqrt{d}}\geq 1-2e^{-cn}-4nhd^2e^{-c\sqrt{d}} $$.

Thus condition on $\mE_2 \cap \mE_4$, it is easy to find that $\mathbb{E}\norm{\mX\mW_1^{0}\mW_1^{0^\top}\mX^{\top}}_F^2 \leq C_4 n(n^2+h^2)$, also based on (~\ref{app: gauss_expect}),  by Lemma~\ref{app: Hanson-Wright Inequality}, we have $$ \mathbb{P}\left( \vxi^{\top} \mX\mX^{\top}\vxi \leq \frac{\rho_e^2}{2}nh-t | \mE_2 \cap \mE_5 \right  ) \leq 2e^{-c\min \left\{ \frac{t^2}{n(n^2+h^2)},\frac{t}{n+h} \right\}}.$$

Choosing $t=\frac{\rho_e^2}{4}nd$, we have 

\begin{equation}
\mathbb{P}\left( \vxi^{\top} \mX\mW_1^{0}\mW_1^{0^\top}\mX^{\top}\vxi \leq \frac{\rho_e^2}{8}nh  \right  ) \leq 2e^{-c\min \left\{ \frac{nh^2}{(n^2+h^2)},\frac{nh}{n+h} \right\}}+4e^{-cn}+4nhd^2e^{-c\sqrt{d}}.
\label{app:hanson-wright_2}
\end{equation}

In addition, by Lemma~\ref{app:Concentration of Lipschitz}, condition on $\mE_2$ and $\mE_4$, we know that $$\norm{\vbeta^{*^{\top}}\mX^{\top}\mX\mW_1^{0}\mW_1^{0^\top}\mX^{\top}\vxi}_{\psi_2}\leq C_5\frac{\sqrt{n(n^2+d^2)(h^2+d^2)}}{d},$$ one can easily see that
$$ \mathbb{P}\left( \left| \vbeta^{*^{\top}}\mX^{\top}\mX\mW_1^{0}\mW_1^{0^\top}\mX^{\top}\vxi \right|\geq t | \mE_2 \cap  \mE_4 \right)\leq 2e^{-\frac{cd^2t^2}{n(n^2+d^2)(h^2+d^2)}}.$$

Thus, let $t=\frac{\rho_e^2}{32}nd$, we obtain 
\begin{equation}
    \mathbb{P}\left( \left| \vbeta^{*^{\top}}\mX^{\top}\mX\mW_1^{0}\mW_1^{0^\top}\mX^{\top}\vxi \right|\geq \frac{\rho_e^2}{32}nd \right)\leq 2e^{-\frac{cnd^4}{(n^2+d^2)(h^2+d^2)}}+2e^{-cn}+2e^{-cd}.
\label{app:lipchitz on xi_2}
\end{equation}

Based on  (~\ref{app:norm_concentration}),(~\ref{app:hanson-wright_2})and(~\ref{app:lipchitz on xi_2})  , we arrive at 

\begin{align*}
    \mathbb{P}\left( \norm{\mA_2^0}  \leq  \frac{\rho_e\sqrt{2d}}{8\sqrt{nh}}\right)\leq &6\left( e^{-c\min \left\{ \frac{nd^2}{(n^2+d^2)},\frac{nd}{n+d}, \frac{nd^4}{(n^2+d^2)(h^2+d^2)}\right\}}+e^{-cn}+ e^{-ch}\right)\\ &+ 6\left(e^{-cd}+nhd^2e^{-c\sqrt{d}}\right)
\end{align*}


\paragraph{ Part of (x).} Based on (~\ref{app:norm_concentration}),(~\ref{app:norm_concentration_2}) and (~\ref{app:part_of_vii}), similar to proof of (vii), by choosing $t=\sqrt{d}$, we arrive at
\begin{align*}
&\mathbb{P}\left( \norm{\mB_2^0}_F \geq C\left( \frac{4}{h}+\frac{32d}{nh}+\frac{16\sqrt{d}}{h\sqrt{n}}+\frac{4}{\sqrt{hd}}+\frac{16}{\sqrt{nh}}+\frac{4}{\sqrt{nd}} +\frac{4}{n}+\frac{16\sqrt{d}}{n\sqrt{h}} \right)\right)\\&\leq 8e^{-cd}+6e^{-ch}.
\end{align*}

\begin{proposition}
 (Three-layer NN setting under Gaussian initialization.) Under Assumption~\ref{sec:assumption-2}, there exists some constant $c^* > 0$ such that for all large $n, h, d$ with probability at least $1- 32e^{-c^*n}-30n^4e^{-c^{*}\sqrt{n}}$, we have gradient approximation,
 \begin{equation}
 \label{eq: gradient_appro}
 \begin{aligned}
          \norm{\mG_1^0-\mA_1^0} &\leq \frac{1}{\sqrt{n}-1}\norm{\mG_1^0}, \\
          \norm{\mG_2^0-\mA_2^0} &\leq  \frac{1}{\sqrt{n}-1}\norm{\mG_2^0}. \\
 \end{aligned}
 \end{equation}
 We obtain the norm control of gradient matrices,
 \begin{equation}
     \begin{aligned}
         \sqrt{h}\norm{\mG_1^0}&= \Theta_{d, \mathbb{P}}(1),\hspace{10pt}
          \sqrt{h}\norm{\mG_1^0}_F = \Theta_{d, \mathbb{P}}(1),\\
          \sqrt{h}\norm{\mG_2^0}&= \Theta_{d, \mathbb{P}}(1),\hspace{10pt} \sqrt{h}\norm{\mG_2^0}_F = \Theta_{d, \mathbb{P}}(1).
     \end{aligned}
 \end{equation}

 Thus, we have
 \\
 \begin{align}
     Small \ lr: \eta_1=\Theta(\sqrt{h}) \Rightarrow  &\norm{\mW_1^1-\mW_1^0} \asymp \norm{\mW_1^0} \\
     \eta_2=\Theta(\sqrt{h}) \Rightarrow  &\norm{\mW_2^1-\mW_2^0} \asymp \norm{\mW_2^0} \\
     Large \ lr: \eta_1=\Theta(h) \Rightarrow & \norm{\mW_1^1-\mW_1^0}_F\asymp\norm{\mW_1^0}_F\\
    \eta_2=\Theta(h) \Rightarrow & \norm{\mW_2^1-\mW_2^0} _F\asymp\norm{\mW_2^0}_F
 \end{align}
\end{proposition}

\paragraph{Proof of Proposition~\ref{sec:proposition-3-layer}.} By Lemma~\ref{app:lemma_norm}, We know that in the proportional regime, there exist constants $C^*,c^* >0$ such that 
$$ \mathbb{P}\left( \norm{\mG_1^0-\mA_1^0} \leq C^*\frac{1}{n} \right )\geq 1-14e^{-c^*n}.$$

On the other hand, part(iv) in  Lemma~\ref{app:lemma_norm} implies that
$$ \mathbb{P}\left( \norm{\mA_1^0}\geq C^{*} \frac{1}{\sqrt{n}} \right)\geq 1-18e^{-c^{*}n}.$$
Conditioning on the two events stated above, we have $$ \norm{\mG_1^0-\mA_1^0} \leq \frac{1}{\sqrt{n}}\norm{\mA_1^0}\leq \frac{1}{\sqrt{n}}(\norm{\mG_1^0}+\norm{\mG_1^0-\mA_1^0}).$$
We finally obtain that $$\mathbb{P} \left( \norm{\mG_1^0-\mA_1^0} \leq \frac{1}{\sqrt{n}-1} \norm{\mG_1^0}\right )\geq 1- 34e^{-c^*n}$$

Similarly, We know that in the proportional regime, there exist constants $C^*,c^* >0$ such that 
$$ \mathbb{P}\left( \norm{\mG_2^0-\mA_2^0} \leq C^*\frac{1}{n} \right )\geq 1-14e^{-c^*n}.$$

On the other hand, part(iv) in  Lemma~\ref{app:lemma_norm} implies that
$$ \mathbb{P}\left( \norm{\mA_2^0}\geq C^{*} \frac{1}{\sqrt{n}} \right)\geq 1-30n^4e^{-c^{*}\sqrt{n}}.$$
Conditioning on the two events stated above, we have $$ \norm{\mG_2^0-\mA_2^0} \leq \frac{1}{\sqrt{n}}\norm{\mA_2^0}\leq \frac{1}{\sqrt{n}}(\norm{\mG_1^0}+\norm{\mG_2^0-\mA_2^0}).$$
We finally obtain that $$\mathbb{P} \left( \norm{\mG_2^0-\mA_2^0} \leq \frac{1}{\sqrt{n}-1} \norm{\mG_2^0}\right )\geq 1- 14e^{-c^*n}-30n^4e^{-c^{*}\sqrt{n}}.$$

Also we can get $\sqrt{h}\norm{\mG_1^0}= \Theta_{d, \mathbb{P}}(1), \sqrt{h}\norm{\mG_1^0}_F= \Theta_{d, \mathbb{P}}(1),\sqrt{h}\norm{\mG_2^0}= \Theta_{d, \mathbb{P}}(1),\sqrt{h}\norm{\mG_2^0}_F= \Theta_{d, \mathbb{P}}(1)$.

Since we have $\norm{\mW_1^0}=\Theta_{d, \mathbb{P}}(1), \norm{\mW_1^0}_F=\Theta_{d, \mathbb{P}}(\sqrt{h}), \norm{\mW_2^0}=\Theta_{d, \mathbb{P}}(1), \norm{\mW_2^0}_F=\Theta_{d, \mathbb{P}}(\sqrt{h})$, based on Assumption~\ref{sec:assumption-2}, we have 
 \begin{align*}
     Small \ lr: \eta_1=\Theta(\sqrt{h}) \Rightarrow  &\norm{\mW_1^1-\mW_1^0} \asymp \norm{\mW_1^0} \\
     \eta_2=\Theta(\sqrt{h}) \Rightarrow  &\norm{\mW_2^1-\mW_2^0} \asymp \norm{\mW_2^0} \\
     Large \ lr: \eta_1=\Theta(h) \Rightarrow & \norm{\mW_1^1-\mW_1^0}_F\asymp\norm{\mW_1^0}_F\\
    \eta_2=\Theta(h) \Rightarrow & \norm{\mW_2^1-\mW_2^0} _F\asymp\norm{\mW_2^0}_F
 \end{align*}
 \hfill $\square$


\subsubsection{Two-layer Neural Network Case Under Gaussian Initialization}
\label{app:norm property of 2-layer NN Case}
The one-step update equations for the two-layer neural network are as follows:
\begin{equation}
\label{eq:update_formula}
\begin{aligned}
\mW_1^{1} &= \mW_1^{0} - \eta_1 \mG_1^0 \\
\mW_2^{1} &= \mW_2^{0} - \eta_2 \mG_2^0
\end{aligned}
\end{equation}
where $W_1^{0},W_2^{0}$ are the initial hidden layer weights, $W_1^{1},W_2^{1}$ are the updated layer weights, $G_1^0$ and $G_2^0$ are the corresponding gradient matrix, where 
\begin{align}
\mG_1^0 &= \underbrace{\frac{1}{nh^2} \mX^{\top}\mX\mW_1^0\mW_2^0\mW_2^{0^{\top}}}_{\mB^{0}_1}-\underbrace{\frac{1}{nh} \mX^{\top}\mY\mW_2^{0^{\top}}}_{\mA^{0}_1}  \label{eq:G_1} \\
\mG_2^0 &= \underbrace{\frac{1}{nh^2}\mW_1^{0^{\top}} \mX^{\top}\mX\mW_1^0\mW_2^0}_{\mB^{0}_2}-\underbrace{\frac{1}{nh}\mW_1^{0^{\top}}\mX^{\top}\mY}_{\mA^{0}_2}\label{eq:G_2}
\end{align}

\begin{proposition}
\label{app:prop-two-layer}
  (Two-layer neural network under gaussian initialization.) There exists some constants $p^*, q^*>0$  such that for all large $n,h,d$ with high probability at least $1-p^{*}e^{-q^*n}$
  \begin{align}
      \norm{\mG_1^0-\mA^{0}_1}  &\leq \frac{1}{\sqrt{n}-1} \norm{\mG_1^0} \\
      \norm{\mG_2^0-\mA^{0}_2}  &\leq \frac{1}{\sqrt{n}-1} \norm{\mG_2^0}.
  \end{align}
  There exists some constants $p, q>0$  such that for all large $n, h, d$  with high probability at least $1-pe^{-qn}$ 
 \begin{align}
      Large \ lr: \eta_1=\Theta(h\sqrt{h}) \Rightarrow & \norm{\mW_1^1-\mW_1^0}_F\asymp\norm{\mW_1^0}_F\\
    \eta_2=\Theta(h\sqrt{h}) \Rightarrow & \norm{\mW_2^1-\mW_2^0} _F\asymp\norm{\mW_2^0}_F.
 \end{align}
\end{proposition}

\paragraph{Proof of Proposition~\ref{app:prop-two-layer}.} The proof is same to proofs of Proposition~\ref{sec:proposition-3-layer} and Lemma~\ref{app:lemma_norm}.

\subsection{Norm Analysis of Two-step Update Gradient Matrices}

\subsubsection{Two-layer Neural Network under  Orthogonal Initialization } \label{app:2-NN-orthgonal_two-step_prop}

\begin{proposition}
\label{app:proposition-2-layer-orthogonal-two-steps}
 (Two-layer NN setting under  Orthogonal initialization.) Under Assumption~\ref{sec:assumption-whiten}, if $\eta_1,\eta_2\leq O(h\sqrt{h}
 )$, we have the following gradient approximation,
 \begin{equation}
 \begin{aligned}
          \norm{{\overline{\mG_1^1}}-{\widetilde{\mA_1^1}}} &\leq \frac{1}{\sqrt{h}-1}\norm{{\overline{\mG_1^1}}}, \\
          \norm{{\overline{\mG_2^1}}-{\widetilde{\mA_2^1}}} &\leq  \frac{1}{\sqrt{h}-1}\norm{{\overline{\mG_2^1}}}. \\
 \end{aligned}
 \end{equation}
And we have
 \\
 \begin{align}
     Large \ lr: \eta_1=\Theta(h\sqrt{h}) \Rightarrow & \norm{{\overline{\mW_1^2}}-\widetilde{\mW_1^1}}_F\asymp\norm{{\widetilde{\mW_1^1}}}_F\\
    \eta_2=\Theta(h\sqrt{h}) \Rightarrow & \norm{{\overline{\mW_2^2}}-\widetilde{\mW_2^1}} _F\asymp\norm{\widetilde{\mW_2^1}}_F
 \end{align}
\end{proposition}

\paragraph{Proof of Proposition~\ref{app:proposition-2-layer-orthogonal-two-steps}.} 
Note that ${\widetilde{\mA_1^1}}=\frac{1}{h}\mM\widetilde{{{\mW_2^{1}}}}^{\top}$, ${\widetilde{\mA_2^1}}=\frac{1}{h}\widetilde{{{\mW_1^{1}}}}^{\top}\mM$. Based on Lemma~\ref{app:proposition-2-layer-orthogonal}, it is  easy to get 
\begin{align*}
    &\norm{\widetilde{{{\mW_1^{1}}}}-\mW_1^0}\asymp\norm{\mW_1^0},   \norm{\widetilde{{{\mW_2^{1}}}}-\mW_2^0}\asymp\norm{\mW_2^0} \\
       &\norm{\widetilde{{{\mW_1^{1}}}}-\mW_1^0}_F\asymp\norm{\mW_1^0}_F,   \norm{\widetilde{{{\mW_2^{1}}}}-\mW_2^0}_F\asymp\norm{\mW_2^0}_F
\end{align*}


We also have ${\widetilde{\mB_1^1}}=\frac{1}{h^2} \widetilde{{{\mW_1^{1}}}}\widetilde{{{\mW_2^{1}}}}\widetilde{{{\mW_2^{1}}}}^{\top}$, ${\widetilde{\mB_2^1}}=\frac{1}{h}\widetilde{{{\mW_1^{1}}}}^{\top}\widetilde{{{\mW_1^{1}}}}\widetilde{{{\mW_2^{1}}}}$ . Thus we can obtain that 

\begin{align*}
   &\norm{ {\widetilde{\mA_1^1}}} \asymp\norm{{\mA_1^0}},     \norm{ {\widetilde{\mA_1^1}}}_F \asymp\norm{{\mA_1^0}}_F ,   \norm{ {\widetilde{\mA_2^1}}} \asymp\norm{{\mA_2^0}},     \norm{ {\widetilde{\mA_2^1}}}_F \asymp\norm{{\mA_2^0}}_F \\
   &\norm{ {\widetilde{\mB_1^1}}} \asymp\norm{{\mB_1^0}},     \norm{ {\widetilde{\mB_1^1}}}_F \asymp\norm{{\mB_1^0}}_F ,   \norm{ {\widetilde{\mB_2^1}}} \asymp\norm{{\mB_2^0}},     \norm{ {\widetilde{\mB_2^1}}}_F \asymp\norm{{\mB_2^0}}_F \\
\end{align*}

Since ${\overline{\mG_1^1}}= {\widetilde{\mB_1^1}}-{\widetilde{\mA_1^1}}$, ${\overline{\mG_2^1}}= {\widetilde{\mB_2^1}}-{\widetilde{\mA_2^1}}$, we obtain that 
\begin{align*}
    \norm{{\overline{\mG_1^1}}-{\widetilde{\mA_1^1}}} & \leq \frac{1}{\sqrt{h}}\norm{{\widetilde{\mA_1^1}}}\leq \frac{1}{\sqrt{h}}({\overline{\mG_1^1}}+\norm{{\overline{\mG_1^1}}-{\widetilde{\mA_1^1}}}) \\
        \norm{{\overline{\mG_2^1}}-{\widetilde{\mA_2^1}}} & \leq \frac{1}{\sqrt{h}}\norm{{\widetilde{\mA_2^1}}}\leq \frac{1}{\sqrt{h}}({\overline{\mG_2^1}}+\norm{{\overline{\mG_2^1}}-{\widetilde{\mA_2^1}}}) .
\end{align*}
Thus, we get that 
 \begin{equation}
 \begin{aligned}
          \norm{{\overline{\mG_1^1}}-{\widetilde{\mA_1^1}}}  &\leq \frac{1}{\sqrt{h}-1}\norm{{\overline{\mG_1^1}}}, \\
          \norm{{\overline{\mG_2^1}}-{\widetilde{\mA_2^1}}}  &\leq  \frac{1}{\sqrt{h}-1}\norm{{\overline{\mG_2^1}}}. \\
 \end{aligned}
 \end{equation}

Based on this, we can get $\norm{{\overline{\mG_1^1}}}\asymp \norm{{\mA_1^0}}, \norm{{\overline{\mG_1^1}}}_F= \norm{{\mA_1^0}}_F ,\norm{{\overline{\mG_2^1}}}= \norm{{\mA_2^0}}, \norm{{\overline{\mG_2^1}}}_F= \norm{{\mA_2^0}}_F$.

Since we have $\norm{\mW_1^0}\asymp \norm{\widetilde{\mW_1^1}}, \norm{\mW_1^0}_F=\norm{\widetilde{\mW_1^1}}_F, \norm{\mW_2^0}=\norm{\widetilde{\mW_2^1}}, \norm{\mW_2^0}_F=\norm{\widetilde{\mW_2^1}}_F$, based on Assumption~\ref{sec:assumption-2}, we have 
 \begin{align}
     Large \ lr: \eta_1=\Theta(h\sqrt{h}) \Rightarrow & \norm{{\overline{\mW_1^2}}-\widetilde{\mW_1^1}}_F\asymp\norm{{\widetilde{\mW_1^1}}}_F\\
    \eta_2=\Theta(h\sqrt{h}) \Rightarrow & \norm{{\overline{\mW_2^2}}-\widetilde{\mW_2^1}} _F\asymp\norm{\widetilde{\mW_2^1}}_F
 \end{align}
 \hfill $\square$

\subsubsection{Three-layer Neural Network under  Orthogonal Initialization } \label{app:3-NN-orthgonal_two-step_prop}

\begin{proposition}
\label{app:proposition-3-layer-orthogonal-two-steps}
 (Three-layer NN setting under  Orthogonal initialization.) Under Assumption~\ref{sec:assumption-whiten}, if $\eta_1,\eta_2\leq O(h
 )$, we have the following gradient approximation,
 \begin{equation}
 \begin{aligned}
          \norm{{\overline{\mG_1^1}}-{\widetilde{\mA_1^1}}} &\leq \frac{1}{\sqrt{h}-1}\norm{{\overline{\mG_1^1}}}, \\
          \norm{{\overline{\mG_2^1}}-{\widetilde{\mA_2^1}}} &\leq  \frac{1}{\sqrt{h}-1}\norm{{\overline{\mG_2^1}}}. \\
 \end{aligned}
 \end{equation}
And we have
 \\
 \begin{align}
          Small \ lr: \eta_1=\Theta(\sqrt{h}) \Rightarrow & \norm{{\overline{\mW_1^2}}-\widetilde{\mW_1^1}}\asymp\norm{{\widetilde{\mW_1^1}}}\\
    \eta_2=\Theta(\sqrt{h}) \Rightarrow & \norm{{\overline{\mW_2^2}}-\widetilde{\mW_2^1}} \asymp\norm{\widetilde{\mW_2^1}} \\
     Large \ lr: \eta_1=\Theta(h) \Rightarrow & \norm{{\overline{\mW_1^2}}-\widetilde{\mW_1^1}}_F\asymp\norm{{\widetilde{\mW_1^1}}}_F\\
    \eta_2=\Theta(h) \Rightarrow & \norm{{\overline{\mW_2^2}}-\widetilde{\mW_2^1}} _F\asymp\norm{\widetilde{\mW_2^1}}_F
 \end{align}
\end{proposition}

\paragraph{Proof of Proposition~\ref{app:proposition-3-layer-orthogonal-two-steps}.} 
Note that ${\widetilde{\mA_1^1}}=\frac{1}{h}\vbeta^*\va^{\top}\widetilde{{{\mW_2^{1}}}}^{\top}$, ${\widetilde{\mA_2^1}}=\frac{1}{h}\widetilde{{{\mW_1^{1}}}}^{\top}\vbeta^*\va^{\top}$. Based on Lemma~\ref{app:proposition-2-layer-orthogonal}, it is  easy to get 
\begin{align*}
    &\norm{\widetilde{{{\mW_1^{1}}}}-\mW_1^0}\asymp\norm{\mW_1^0},   \norm{\widetilde{{{\mW_2^{1}}}}-\mW_2^0}\asymp\norm{\mW_2^0} \\
       &\norm{\widetilde{{{\mW_1^{1}}}}-\mW_1^0}_F\asymp\norm{\mW_1^0}_F,   \norm{\widetilde{{{\mW_2^{1}}}}-\mW_2^0}_F\asymp\norm{\mW_2^0}_F
\end{align*}


We also have ${\widetilde{\mB_1^1}}=\frac{1}{h^2} \widetilde{{{\mW_1^{1}}}}\widetilde{{{\mW_2^{1}}}}\va\va^{\top}\widetilde{{{\mW_2^{1}}}}^{\top}$, ${\widetilde{\mB_2^1}}=\frac{1}{h}\widetilde{{{\mW_1^{1}}}}^{\top}\widetilde{{{\mW_1^{1}}}}\widetilde{{{\mW_2^{1}}}}\va\va^{\top}$ . Thus we can obtain that 

\begin{align*}
   &\norm{ {\widetilde{\mA_1^1}}} \asymp\norm{{\mA_1^0}},     \norm{ {\widetilde{\mA_1^1}}}_F \asymp\norm{{\mA_1^0}}_F ,   \norm{ {\widetilde{\mA_2^1}}} \asymp\norm{{\mA_2^0}},     \norm{ {\widetilde{\mA_2^1}}}_F \asymp\norm{{\mA_2^0}}_F \\
   &\norm{ {\widetilde{\mB_1^1}}} \asymp\norm{{\mB_1^0}},     \norm{ {\widetilde{\mB_1^1}}}_F \asymp\norm{{\mB_1^0}}_F ,   \norm{ {\widetilde{\mB_2^1}}} \asymp\norm{{\mB_2^0}},     \norm{ {\widetilde{\mB_2^1}}}_F \asymp\norm{{\mB_2^0}}_F \\
\end{align*}

Since ${\overline{\mG_1^1}}= {\widetilde{\mB_1^1}}-{\widetilde{\mA_1^1}}$, ${\overline{\mG_2^1}}= {\widetilde{\mB_2^1}}-{\widetilde{\mA_2^1}}$, we obtain that 
\begin{align*}
    \norm{{\overline{\mG_1^1}}-{\widetilde{\mA_1^1}}} & \leq \frac{1}{\sqrt{h}}\norm{{\widetilde{\mA_1^1}}}\leq \frac{1}{\sqrt{h}}({\overline{\mG_1^1}}+\norm{{\overline{\mG_1^1}}-{\widetilde{\mA_1^1}}}) \\
        \norm{{\overline{\mG_2^1}}-{\widetilde{\mA_2^1}}} & \leq \frac{1}{\sqrt{h}}\norm{{\widetilde{\mA_2^1}}}\leq \frac{1}{\sqrt{h}}({\overline{\mG_2^1}}+\norm{{\overline{\mG_2^1}}-{\widetilde{\mA_2^1}}}) .
\end{align*}
Thus, we get that 
 \begin{equation}
 \begin{aligned}
          \norm{{\overline{\mG_1^1}}-{\widetilde{\mA_1^1}}}  &\leq \frac{1}{\sqrt{h}-1}\norm{{\overline{\mG_1^1}}}, \\
          \norm{{\overline{\mG_2^1}}-{\widetilde{\mA_2^1}}}  &\leq  \frac{1}{\sqrt{h}-1}\norm{{\overline{\mG_2^1}}}. \\
 \end{aligned}
 \end{equation}

Based on this, we can get $\norm{{\overline{\mG_1^1}}}\asymp \norm{{\mA_1^0}}, \norm{{\overline{\mG_1^1}}}_F= \norm{{\mA_1^0}}_F ,\norm{{\overline{\mG_2^1}}}= \norm{{\mA_2^0}}, \norm{{\overline{\mG_2^1}}}_F= \norm{{\mA_2^0}}_F$.

Since we have $\norm{\mW_1^0}\asymp \norm{\widetilde{\mW_1^1}}, \norm{\mW_1^0}_F=\norm{\widetilde{\mW_1^1}}_F, \norm{\mW_2^0}=\norm{\widetilde{\mW_2^1}}, \norm{\mW_2^0}_F=\norm{\widetilde{\mW_2^1}}_F$, based on Assumption~\ref{sec:assumption-2}, we have 
 \begin{align}
           Small \ lr: \eta_1=\Theta(\sqrt{h}) \Rightarrow & \norm{{\overline{\mW_1^2}}-\widetilde{\mW_1^1}}\asymp\norm{{\widetilde{\mW_1^1}}}\\
    \eta_2=\Theta(\sqrt{h}) \Rightarrow & \norm{{\overline{\mW_2^2}}-\widetilde{\mW_2^1}} \asymp\norm{\widetilde{\mW_2^1}} \\
     Large \ lr: \eta_1=\Theta(h) \Rightarrow & \norm{{\overline{\mW_1^2}}-\widetilde{\mW_1^1}}_F\asymp\norm{{\widetilde{\mW_1^1}}}_F\\
    \eta_2=\Theta(h) \Rightarrow & \norm{{\overline{\mW_2^2}}-\widetilde{\mW_2^1}} _F\asymp\norm{\widetilde{\mW_2^1}}_F
 \end{align}
 \hfill $\square$

\section{Orthogonal initialization}
Here we give one-step and two-step test loss under whiten initialization, also we give the gap bound between the exact test loss and the approximate test loss.

\begin{lemma}
\label{app:lemma_orth}
    Consider two stochastic random orthogonal matrices $\mO_1$, $\mO_2 \in \mathbb{R}^{h \times h}$ uniformly distributed on the orthogonal group  with respect to the Haar measure (i.e. $\mO_1\mO_1^{\top}=\mO_1^{\top}\mO_1 =\mI$, $\mO_2\mO_2^{\top}=\mO_2^{\top}\mO_2 =\mI$), we have $\mathbb{E}[\mO_1]=\mathbb{E}[\mO_2]=\vzero, \mathbb{E}[\mO_1^2]=\mathbb{E}[\mO_2^2]=\frac{1}{h}\mI, \mathbb{E}[\mO_1\mO_2]=\vzero.$
\end{lemma}

\paragraph{Proof of Lemma~\ref{app:lemma_orth}.} For orthogonal group, one key invariance property is for any fixed $\mR$ in orthogonal group, the distribution of $\mQ\mR$  and $\mR\mQ$ is same as $\mQ$. In particular, note that $-\mI$ is in orthogonal group, therefore, we have \begin{equation*}
    \mQ \stackrel{d}{=} (-\mI)\mQ = -\mQ,
\end{equation*}
so based on this, we  take a look at the expectation: $\mathbb{E} \left [ \mQ \right]=\mathbb{E} \left [ -\mQ \right]=-\mathbb{E} \left [ \mQ \right]$, which means we can get  $\mathbb{E} \left [ \mQ_1 \right]=\mathbb{E} \left [ \mQ_2 \right]=\vzero$. Also consider each row row (or column) of $\mQ$ is a random vector uniformly distributed on the unit sphere in $\mathbb{R}^h$. Hence by the definitions of  orthogonal group, we have
\begin{equation*}
    \sum_{a=1}^h \mQ_{ia}^2 = 1 \;\Rightarrow\; \mathbb{E}\big[\mQ_{ia}^2\big] = \frac{1}{h},
\end{equation*}
furthermore, if we consider flipping the sign of one row or one column like left-multiplying by $D= diag(-1,1,1,\cdots,1)$ in  orthogonal group, which flips the sign of every entry in the first row, but the distribution is unchanged. Thus, the expectation of any product involving an odd number of the entries from that row must be zero. Similarly for the flipping any column, so we can get unless $i=j $ and $a=b$,  $\mathbb{E}\left[\mQ_{ia}\mQ_{jb}\right]=0$, so it is easy to get $\mathbb{E}\left[\mQ_1^2\right]=\mathbb{E}\left[\mQ_2^2\right]=\frac{1}{h}\mI$. Based on above arguments we can deduce that $\mathbb{E}\left[\mQ_1\mQ_2\right]=\vzero$. \hfill $\square$

\subsection{Approximate one-step loss under orthogonal  initialization for two-layer NN}  \label{app:orthog-2-NN-one-step}

\begin{theorem}
    Given Assumption~\ref{sec:assumption},~\ref{sec:assumption-whiten}, and in addition assume $\eta_1$ and $\eta_2$ are no more than $O({h\sqrt{h}})$, based on Proposition~\ref{sec:orthog-norm} and Lemma~\ref{sec:orthog-approx-loss}, consider the training procedure discussed in Section~\ref{sec:setup}, we obtain the following test loss after one-step and two-step GD update in a two-layer neural network under  orthogonal initialization:
\begin{equation}
\begin{aligned}
       & L_{\text{two-layer}}({\mW_1^1},{\mW_2^1})=\frac{\eta_1^2}{h^4}+\frac{\eta_2^2}{h^4}+\frac{2\eta_1\eta_2}{h^4}+\frac{\eta_1^2\eta_2^2}{h^7}\\ & \qquad \qquad \qquad \qquad  -\frac{2\eta_1}{h^2}-\frac{2\eta_2}{h^2} +\frac{1}{h}+\frac{2\eta_1\eta_2}{h^5}+1 \\
    &L_{\text{two-layer}}({\mW_1^2},{\mW_2^2})=\frac{1}{h}(1+\frac{\eta_1\eta_2}{h^3})^4 +\frac{16\eta_1^2\eta_2^2}{h^7} \\&+\left(\frac{2(\eta_1+\eta_2)(\eta_1\eta_2+h^3)}{h^5}-1\right)^2+(1+\frac{\eta_1\eta_2}{h^3})^2\frac{8\eta_1\eta_2}{h^5}
\end{aligned}    
\end{equation}
\end{theorem}

We prove Theorem~\ref{theorem for orthog-2-layer} above by the following two subsection~\ref{app:orthog-2-NN-one-step} and~\ref{app:orthog-2-NN-two-steps}. 

For  orthogonal  initialization we follow  assumption ~\ref{sec:assumption-whiten} that $n=h=d$.

Here we consider the whiten  initialization which make the setting $\mX^{\top}\mX=\mX\mX^{\top}=h\mI, \mW_1^{0^{\top}}\mW_1^0=\mW_1^0\mW_1^{0^{\top}}=\mI, \mW_2^{0^{\top}}\mW_2^0=\mW_2^0\mW_2^{0^{\top}}=\mI, \mM^{{\top}}\mM=\mM\mM^{{\top}}=\frac{1}{h}\mI $.

We consider a  test data $\Tilde{\vx}_0$ under  two-layer setting, where $\frac{1}{\sqrt{h}}\Tilde{\vx}_0$ is  an random  orthogonal vector, we have
\begin{equation}
\begin{aligned}
    &L_{\text{two-layer}}(\mX, {\mW_1^1},{\mW_2^1},\Tilde{\vx}_0)\\=&\mathbb{E}_{\mW_1^0,\mW_2^0,\vxi,\Tilde{\vx}_0,\mX} \norm{ \frac{1}{h}\Tilde{\vx}_0\mW_1^1\mW_2^1-\Tilde{\vx}_0\mM}_{F}^2  \\
    =&tr\left( \mathbb{E}_{\mW_1^0,\mW_2^0,\vxi,\Tilde{\vx}_0,\mX} \left [\left( \frac{1}{h}{\mW_1^1}{\mW_2^1}-\mM\right)^{\top}{\Tilde{\vx}_0}^{\top}\Tilde{\vx}_0 \left( \frac{1}{h}{\mW_1^1}{\mW_2^1}-\mM\right)\right]\right) \\
    =& tr\left( \mathbb{E}_{\mW_1^0,\mW_2^0,\vxi,\Tilde{\vx}_0,\mX} \left [{\Tilde{\vx}_0}^{\top}\Tilde{\vx}_0 \left( \frac{1}{h}{\mW_1^1}{\mW_2^1}-\mM\right)\left( \frac{1}{h}{\mW_1^1}{\mW_2^1}-\mM\right)^{\top}\right] \right) \\
    =&tr\left( \mathbb{E}_{\mW_1^0,\mW_2^0,\vxi, \mX} \left [\left( \frac{1}{h}{\mW_1^1}{\mW_2^1}-\mM\right)\left( \frac{1}{h}\mW_1^1\mW_2^1-\mM\right)^{\top}\right] \right) \\
    =& tr\left( \mathbb{E}_{\mW_1^0,\mW_2^0,\vxi, \mX} \left [ \frac{1}{h^2}{\mW_1^1}{\mW_2^1}{{\mW_2^1}}^{\top}{{\mW_1^1}}^{\top}\right] \right) \\
    -&tr\left( \mathbb{E}_{\mW_1^0,\mW_2^0,\vxi, \mX} \left [ \frac{1}{h}\mM{{\mW_2^1}}^{\top}{{\mW_1^1}}^{\top}\right] \right) \\
    -&tr\left( \mathbb{E}_{\mW_1^0,\mW_2^0,\vxi, \mX}\left [ \frac{1}{h}{\mW_1^1}{\mW_2^1}{\mM}^{\top}\right] \right)\\
    +& tr\left( \mathbb{E} \left [{\mM} {\mM}^{\top}\right] \right).
\end{aligned}
\end{equation}
Here we define $L_1,L_2,L_3, L_4$, where
\begin{align*}
   L_1&=tr\left( \mathbb{E}_{\mW_1^0,\mW_2^0,\vxi, \mX} \left [ \frac{1}{h^2}{\mW_1^1}{\mW_2^1}{{\mW_2^1}}^{\top}{{\mW_1^1}}^{\top}\right] \right)\\
    L_2&=tr\left( \mathbb{E}_{\mW_1^0,\mW_2^0,\vxi, \mX} \left [ \frac{1}{h}\mM{{\mW_2^1}}^{\top}{{\mW_1^1}}^{\top}\right] \right) \\
    L_3 &=tr\left( \mathbb{E}_{\mW_1^0,\mW_2^0,\vxi, \mX}\left [ \frac{1}{h}{\mW_1^1}{\mW_2^1}{\mM}^{\top}\right] \right)\\
    L_4 &= tr\left( \mathbb{E} \left [{\mM} {\mM}^{\top}\right] \right)
\end{align*}
Thus $$L_{\text{two-layer}}=L_1-L_2-L_3+L_4$$

We have $L_1=\sum_{i=1}^{16} T_{i}$, where
\begin{align*}
T_1&=tr\left( \mathbb{E}_{\mW_1^0,\mW_2^0, \mX} \left [ \frac{1}{h^2}{\mW_2^0}^{\top}{\mW_1^0}^{\top}{\mW_1^0}{\mW_2^0}
\right] \right)=\frac{1}{h},\\
T_2&=tr\left( \mathbb{E}_{\mW_1^0,\mW_2^0, \mX} \left [ \frac{\eta_1}{h^4}{\mW_2^0}^{\top}{\mW_1^0}^{\top}\mX^{\top}\mY{\mW_2^0}^{\top}\mW_2^0
\right] \right)=0,\\
T_3&=tr\left( \mathbb{E}_{\mW_1^0,\mW_2^0,\mX} \left [ \frac{\eta_2}{h^4}{\mW_2^0}^{\top}{\mW_1^0}^{\top}\mW_1^0{\mW_1^0}^{\top}\mX^{\top}\mY
\right] \right)=0,\\
T_4&=tr\left( \mathbb{E}_{\mW_1^0,\mW_2^0, \mX} \left [ \frac{\eta_1\eta_2}{h^6}{\mW_2^0}^{\top}{\mW_1^0}^{\top}\mX^{\top}\mY{\mW_2^0}^{\top}{\mW_1^0}^{\top}\mX^{\top}\mY
\right] \right)=\frac{\eta_1\eta_2}{h^5},\\
T_5&=tr\left( \mathbb{E}_{\mW_1^0,\mW_2^0, \mX} \left [ \frac{\eta_1}{h^4}{\mW_2^0}^{\top}\mW_2^0 \mY^{\top}\mX\mW_1^0\mW_2^0
\right] \right)=0,\\
T_6&=tr\left( \mathbb{E}_{\mW_1^0,\mW_2^0, \mX} \left [ \frac{\eta_1^2}{h^6}{\mW_2^0}^{\top}\mW_2^0 \mY^{\top}\mX\mX^{\top}\mY{\mW_2^0}^{\top}\mW_2^0 
\right] \right)=\frac{\eta_1^2}{h^4},\\
T_7&=tr\left( \mathbb{E}_{\mW_1^0,\mW_2^0, \mX} \left [ \frac{\eta_1\eta_2}{h^6}{\mW_2^0}^{\top}\mW_2^0 \mY^{\top}\mX\mW_1^0{\mW_1^0}^{\top}\mX^{\top}\mY
\right] \right)=\frac{\eta_1\eta_2}{h^4},\\
T_8&=tr\left( \mathbb{E}_{\mW_1^0,\mW_2^0, \mX} \left [ \frac{\eta_1^2\eta_2}{h^8}{\mW_2^0}^{\top}\mW_2^0 \mY^{\top}\mX\mX^{\top}\mY{\mW_2^0}^{\top}{\mW_1^0}^{\top}\mX^{\top}\mY
\right] \right)=0,\\
T_9&=tr\left( \mathbb{E}_{\mW_1^0,\mW_2^0, \mX} \left [ \frac{\eta_2}{h^4}\mY^{\top}\mX\mW_1^0{\mW_1^0}^{\top}\mW_1^0\mW_2^0
\right] \right)=0,\\
T_{10}&=tr\left( \mathbb{E}_{\mW_1^0,\mW_2^0, \mX} \left [ \frac{\eta_1\eta_2}{h^6}\mY^{\top}\mX\mW_1^0{\mW_1^0}^{\top}\mX^{\top}\mY{\mW_2^0}^{\top}\mW_2^0 
\right] \right)=\frac{\eta_1\eta_2}{h^4},\\
T_{11}&=tr\left( \mathbb{E}_{\mW_1^0,\mW_2^0, \mX} \left [ \frac{\eta_2^2}{h^6}\mY^{\top}\mX\mW_1^0{\mW_1^0}^{\top}\mW_1^0{\mW_1^0}^{\top}\mX^{\top}\mY
\right] \right)=\frac{\eta_2^2}{h^4},\\
T_{12}&=tr\left( \mathbb{E}_{\mW_1^0,\mW_2^0, \mX} \left [ \frac{\eta_1\eta_2^2}{h^8}\mY^{\top}\mX\mW_1^0{\mW_1^0}^{\top}\mX^{\top}\mY{\mW_2^0}^{\top}{\mW_1^0}^{\top}\mX^{\top}\mY
\right] \right)=0,\\
T_{13}&=tr\left( \mathbb{E}_{\mW_1^0,\mW_2^0, \mX} \left [ \frac{\eta_1\eta_2}{h^6}\mY^{\top}\mX{\mW_1^0}{\mW_2^0}\mY^{\top}\mX\mW_1^0\mW_2^0
\right] \right)=\frac{\eta_1\eta_2}{h^5},\\
T_{14}&=tr\left( \mathbb{E}_{\mW_1^0,\mW_2^0, \mX} \left [ \frac{\eta_1^2\eta_2}{h^8}\mY^{\top}\mX{\mW_1^0}{\mW_2^0}\mY^{\top}\mX\mX^{\top}\mY{\mW_2^0}^{\top}\mW_2^0 
\right] \right)=0,\\
T_{15}&=tr\left( \mathbb{E}_{\mW_1^0,\mW_2^0, \mX} \left [ \frac{\eta_1\eta_2^2}{h^8}\mY^{\top}\mX{\mW_1^0}{\mW_2^0}\mY^{\top}\mX\mW_1^0{\mW_1^0}^{\top}\mX^{\top}\mY
\right] \right)=0,\\
T_{16}&=tr\left( \mathbb{E}_{\mW_1^0,\mW_2^0,\mX} \left [ \frac{\eta_1^2\eta_2^2}{h^{10}}\mY^{\top}\mX{\mW_1^0}{\mW_2^0}\mY^{\top}\mX\mX^{\top}\mY{\mW_2^0}^{\top}{\mW_1^0}^{\top}\mX^{\top}\mY
\right] \right)=\frac{\eta_1^2\eta_2^2}{h^7}.\\
\end{align*}
We have $L_2=\sum_{i=17}^{20} T_{i}$, where
\begin{align*}
T_{17}&=tr\left( \mathbb{E}_{\mW_1^0,\mW_2^0, \mX} \left [ \frac{1}{h}\mM{\mW_2^0}^{\top}{\mW_1^0}^{\top}
\right] \right)=0,\\
T_{18}&=tr\left( \mathbb{E}_{\mW_1^0,\mW_2^0, \mX} \left [ \frac{\eta_1}{h^3}\mM{\mW_2^0}^{\top}\mW_2^0 \mY^{\top}\mX
\right] \right)=\frac{\eta_1}{h^2},\\
T_{19}&=tr\left( \mathbb{E}_{\mW_1^0,\mW_2^0, \mX} \left [ \frac{\eta_2}{h^3}\mM\mY^{\top}\mX\mW_1^0{\mW_1^0}^{\top}
\right] \right)=\frac{\eta_2}{h^2},\\
T_{20}&=tr\left( \mathbb{E}_{\mW_1^0,\mW_2^0, \mX} \left [ \frac{\eta_1\eta_2}{h^5}\mM\mY^{\top}\mX{\mW_1^0}{\mW_2^0}\mY^{\top}\mX
\right] \right)=0.\\
\end{align*}

We have $L_3=\sum_{i=21}^{24} T_{i}$, where
\begin{align*}
    T_{21}&=tr\left( \mathbb{E}_{\mW_1^0,\mW_2^0, \mX} \left [ \frac{1}{h}\mW_1^0\mW_2^0\mM^{{\top}}
\right] \right)=0,\\
T_{22}&=tr\left( \mathbb{E}_{\mW_1^0,\mW_2^0, \mX} \left [ \frac{\eta_1}{h^3}\mX^{\top}\mY{\mW_2^0}^{\top}\mW_2^0\mM^{{\top}}
\right] \right)=\frac{\eta_1}{h^2},\\
T_{23}&=tr\left( \mathbb{E}_{\mW_1^0,\mW_2^0, \mX} \left [ \frac{\eta_2}{h^3}\mW_1^0{\mW_1^0}^{\top}\mX^{\top}\mY\mM^{{\top}}
\right] \right)=\frac{\eta_2}{h^2},\\
T_{24}&=tr\left( \mathbb{E}_{\mW_1^0,\mW_2^0, \mX} \left [ \frac{\eta_1\eta_2}{h^5}\mX^{\top}\mY{\mW_2^0}^{\top}{\mW_1^0}^{\top}\mX^{\top}\mY\mM^{{\top}}
\right] \right)=0,\\
\end{align*}

Based on the above computation, we see that for  orthogonal initialization, the one-step test loss for 2-layer NN is

\begin{equation}
    L_{\text{two-layer}}(\mX,{\mW_1^1},{\mW_2^1},\Tilde{\vx}_0)=\frac{\eta_1^2}{h^4}+\frac{\eta_2^2}{h^4}+\frac{2\eta_1\eta_2}{h^4}+\frac{\eta_1^2\eta_2^2}{h^7}-\frac{2\eta_1}{h^2}-\frac{2\eta_2}{h^2}+\frac{1}{h}+\frac{2\eta_1\eta_2}{h^5}+1
\end{equation}

Here, for the one-step updated  loss, we consider the following optimization problem, and we assume the following constraint   $\eta_1+\eta_2=2h^{\alpha}$,  our goal is to see whether $\eta_1=\eta_2=h^{\alpha}$ is local minima or local maxima.

\begin{equation}
    L_{\text{two-layer}}(\mX,{\mW_1^1},{\mW_2^1},\Tilde{\vx}_0)=\frac{\eta_1^2}{h^4}+\frac{\eta_2^2}{h^4}+\frac{2\eta_1\eta_2}{h^4}+\frac{\eta_1^2\eta_2^2}{h^7}-4h^{\alpha-2}+\frac{1}{h}+\frac{2\eta_1\eta_2}{h^5}+1
\end{equation}
It is easy to find that  $\eta_1=\eta_2=h^{\alpha}$ is a local  maxima.  \hfill $\square$

\subsection{Approximate two-step loss  for two-layer NN under orthogonal  initialization}  \label{app:orthog-2-NN-two-steps}

Here we consider the orthogonal  initialization which make the setting $\mX^{\top}\mX=\mX\mX^{\top}=h\mI, \mW_1^{0^{\top}}\mW_1^0=\mW_1^0\mW_1^{0^{\top}}=\mI, \mW_2^{0^{\top}}\mW_2^0=\mW_2^0\mW_2^{0^{\top}}=\mI, \mM^{{\top}}\mM=\mM\mM^{{\top}}=\frac{1}{h}\mI $.

For the simplification, we only consider replacing $\mG_1$ with $\mA_1$ and $\mG_2$ with $\mA_2$. 
\begin{align*}
{\mA_1^0} &= \frac{1}{h}\mM\mW_2^{0\top}
&\;
{\widetilde{\mA_1^1}} &= \frac{1}{h}\mM {\widetilde{\mW_2^{1}}}^{\top} \\
{\mB_1^0} &= \frac{1}{h^2}\mW_1^0
&\;
{\widetilde{\mB_1^1}} &= \frac{1}{h^2}{\widetilde{\mW_1^{1}}}{\widetilde{\mW_2^{1}}}{\widetilde{\mW_2^{1}}}^{\top} \\
{\mA_2^0} &= \frac{1}{h}\mW_1^{0\top}\mM
&\;
{\widetilde{\mA_2^1}} &= \frac{1}{h}{\widetilde{\mW_1^{1}}}^{\top}\mM \\
\mB_2^0 &= \frac{1}{h^2}\mW_2^0
&\;
{\widetilde{\mB_2^1}} &= \frac{1}{h^2}{\widetilde{\mW_1^{1}}}^{\top} {\widetilde{\mW_1^{1}}}{\widetilde{\mW_2^{1}}}
\end{align*}

Thus we have 
\begin{align*}
   {\widetilde{\mW_1^{1}}}&=\mW_1^0+\eta_1\mA_1^0 = \mW_1^0+\frac{\eta_1}{h} \mM\mW_2^{0^{\top}} \\
   {\widetilde{\mW_2^{1}}}& =\mW_2^0+\eta_2\mA_2^0 = \mW_2^0+\frac{\eta_2}{h} \mW_1^{0^{\top}}\mM\\
   {\widetilde{\mW_1^{2}}}&={\widetilde{\mW_1^{1}}}+\eta_1{\widetilde{\mA_1^1}} = {\widetilde{\mW_1^{1}}}+\frac{\eta_1}{h} \mM{\widetilde{\mW_2^{1}}}^{\top}  \\
    &=\mW_1^0+\frac{2\eta_1}{h} \mM\mW_2^{0^{\top}}+ \frac{\eta_1\eta_2}{h^3}\mW_1^0 \\
   {\widetilde{\mW_2^{2}}}& ={\widetilde{\mW_2^{1}}}+\eta_2{\widetilde{\mA_2^1}} = {\widetilde{\mW_2^{1}}}+\frac{\eta_2}{h} {\widetilde{\mW_1^{1}}}^{\top}\mM\\
   &=\mW_2^0+\frac{2\eta_2}{h}\mW_1^{0^\top}\mM+\frac{\eta_1\eta_2}{h^3}\mW_2^0
\end{align*}

we can derive that 
 
\begin{align*}
    {\widetilde{\mW_1^{2}}}{\widetilde{\mW_2^{2}}}&=(1+\frac{\eta_1\eta_2}{h^3})^2\mW_1^0\mW_2^0+ \frac{2(\eta_1+\eta_2)(\eta_1\eta_2+h^3)}{h^4}\mM +\frac{4\eta_1\eta_2}{h^2}\mM\mW_2^{0^{\top}}\mW_1^{0^{\top}}\mM
\end{align*}

Consider the following loss
\begin{equation*}
\begin{aligned}
    &L_{\text{two-layer}}(\mX, {\mW_1^2},{\mW_2^2},\Tilde{\vx}_0)\\ \approx & L_{\text{two-layer}}(\mX, {\widetilde{\mW_1^{2}}},{\widetilde{\mW_2^{2}}},\Tilde{\vx}_0)\\=&\mathbb{E}_{\mW_1^0,\mW_2^0,\vxi,\Tilde{\vx}_0,\mX} \norm{ \frac{1}{h}\Tilde{\vx}_0{\widetilde{\mW_1^{2}}}{\widetilde{\mW_2^{2}}}-\Tilde{\vx}_0\mM}_{F}^2  \\
    =& tr\left( \mathbb{E}_{\mW_1^0,\mW_2^0, \mX} \left [ \frac{1}{h^2}{\widetilde{\mW_1^{2}}}{\widetilde{\mW_2^{2}}}{\widetilde{\mW_2^{2}}}^{\top}{\widetilde{\mW_1^{2}}}^{\top}\right] \right) \\
    -&tr\left( \mathbb{E}_{\mW_1^0,\mW_2^0, \mX} \left [ \frac{1}{h}\mM{\widetilde{\mW_2^{2}}}^{\top}{\widetilde{\mW_1^{2}}}^{\top}\right] \right) \\
    -&tr\left( \mathbb{E}_{\mW_1^0,\mW_2^0, \mX}\left [\frac{1}{h}{\widetilde{\mW_1^{2}}}{\widetilde{\mW_2^{2}}}{\mM}^{\top}\right] \right)\\
    +& tr\left( \mathbb{E} \left [{\mM} {\mM}^{\top}\right] \right).
\end{aligned}
\end{equation*}
We first compute $tr\left( \mathbb{E}_{\mW_1^0,\mW_2^0,\mX} \left [ \frac{1}{h^2}{\widetilde{\mW_1^{2}}}{\widetilde{\mW_2^{2}}}{\widetilde{\mW_2^{2}}}^{\top}{\widetilde{\mW_1^{2}}}^{\top}\right] \right)$, we find that
\begin{align*}
    &\frac{1}{h^2}{\widetilde{\mW_1^{2}}}{\widetilde{\mW_2^{2}}}{\widetilde{\mW_2^{2}}}^{\top}{\widetilde{\mW_1^{2}}}^{\top}=\frac{1}{h^2}(1+\frac{\eta_1\eta_2}{h^3})^4\mI_h+ (1+\frac{\eta_1\eta_2}{h^3})^2\frac{2(\eta_1+\eta_2)(\eta_1\eta_2+h^3)}{h^6}\mW_1^0\mW_2^0\mM^{\top}\\
    &\quad + (1+\frac{\eta_1\eta_2}{h^3})^2\frac{4\eta_1\eta_2}{h^4}\mW_1^0\mW_2^0\mM^{\top}\mW_1^0\mW_2^0\mM^{\top} + (1+\frac{\eta_1\eta_2}{h^3})^2\frac{2(\eta_1+\eta_2)(\eta_1\eta_2+h^3)}{h^6}\mM{\mW_2^0}^{\top}{\mW_1^0}^{\top}   \\
     &\quad +  \frac{4(\eta_1+\eta_2)^2(\eta_1\eta_2+h^3)^2}{h^{11}} \mI_h +  \frac{8\eta_1\eta_2(\eta_1+\eta_2)(\eta_1\eta_2+h^3)}{h^7}\mW_1^0\mW_2^0\mM^{\top}  +  \frac{16\eta_1^2\eta_2^2}{h^8}\mI_h\\ &\quad + (1+\frac{\eta_1\eta_2}{h^3})^2\frac{4\eta_1\eta_2}{h^4}\mM\mW_2^{0^{\top}}\mW_1^{0^{\top}}\mM\mW_2^{0^{\top}}\mW_1^{0^{\top}}+\frac{8\eta_1\eta_2(\eta_1+\eta_2)(\eta_1\eta_2+h^3)}{h^9}\mM\mW_2^{0^{\top}}\mW_1^{0^{\top}}.
\end{align*}
Thus, we have
\begin{align*}
    tr\left( \mathbb{E}_{\mW_1^0,\mW_2^0,\vxi, \mX} \left [ \frac{1}{h^2}{\widetilde{\mW_1^{2}}}{\widetilde{\mW_2^{2}}}{\widetilde{\mW_2^{2}}}^{\top}{\widetilde{\mW_1^{2}}}^{\top}\right] \right) &= \frac{1}{h}(1+\frac{\eta_1\eta_2}{h^3})^4 +\frac{4(\eta_1+\eta_2)^2(\eta_1\eta_2+h^3)^2}{h^{10}}\\
    &\quad +\frac{16\eta_1^2\eta_2^2}{h^7}+ (1+\frac{\eta_1\eta_2}{h^3})^2\frac{8\eta_1\eta_2}{h^5}\\
    \end{align*}
Following the similar way, we get that 
\begin{align*}
        tr\left( \mathbb{E}_{\mW_1^0,\mW_2^0, \mX} \left [ \frac{1}{h}\mM{\widetilde{\mW_2^{2}}}^{\top}{\widetilde{\mW_1^{2}}}^{\top}\right] \right) &=\frac{2(\eta_1+\eta_2)(\eta_1\eta_2+h^3)}{h^5}, \\
    tr\left( \mathbb{E}_{\mW_1^0,\mW_2^0, \mX}\left [\frac{1}{h}{\widetilde{\mW_1^{2}}}{\widetilde{\mW_2^{2}}}{\mM}^{\top}\right] \right) &=\frac{2(\eta_1+\eta_2)(\eta_1\eta_2+h^3)}{h^5},\\
     tr\left( \mathbb{E} \left [{\mM} {\mM}^{\top}\right] \right)&=1 \\
\end{align*}
We have 
\begin{align*}
    L_{\text{two-layer}}(\mX,{\mW_1^2},{\mW_2^2},\Tilde{\vx}_0)&=\frac{1}{h}(1+\frac{\eta_1\eta_2}{h^3})^4 +\frac{4(\eta_1+\eta_2)^2(\eta_1\eta_2+h^3)^2}{h^{10}}+\frac{16\eta_1^2\eta_2^2}{h^7}\\
    & \quad +(1+\frac{\eta_1\eta_2}{h^3})^2\frac{8\eta_1\eta_2}{h^5}-\frac{4(\eta_1+\eta_2)(\eta_1\eta_2+h^3)}{h^5}+1 
\end{align*}

\begin{equation}
    L_{\text{two-layer}}(\mX,{\mW_1^2},{\mW_2^2},\Tilde{\vx}_0)=\frac{1}{h}(1+\frac{\eta_1\eta_2}{h^3})^4 +\frac{16\eta_1^2\eta_2^2}{h^7}+\left(\frac{2(\eta_1+\eta_2)(\eta_1\eta_2+h^3)}{h^5}-1\right)^2+(1+\frac{\eta_1\eta_2}{h^3})^2\frac{8\eta_1\eta_2}{h^5}
\end{equation}
\hfill $\square$

\begin{corollary}
Suppose $\eta_1+\eta_2 = 2h^{\alpha}$ and we consider $0<\alpha \le \tfrac{3}{2}$.
Then, for any $\alpha$ in this range, the point $\eta_1=\eta_2=h^{\alpha}$ is not a local minimum of the loss $L_{\text{two-layer}}({\mW_1^1},{\mW_2^1})$.
Moreover, for $1<\alpha \le \tfrac{3}{2}$, if $h > \max \{{h^{*}}, 256\}$, then $\eta_1=\eta_2=h^{\alpha}$ is a local minimum of the loss $L_{\text{two-layer}}({\mW_1^2},{\mW_2^2})$, where ${h^{*}}$ is the root of the following equation:
\begin{equation}
    (1+o(1))h^{1-\alpha} +16h^{\alpha-2}+2h^{-\alpha}+8h^{\alpha-3}+6h^{3\alpha-6}-2 =0
\end{equation}
\end{corollary}

\paragraph{Proof of Corollary~\ref{cor:two-layer-NN}.} Here, for the two-step updated  loss, we consider the following optimization problem, and we assume that   $\eta_1+\eta_2=2h^{\alpha}$, we want to find whether the local minima for  $L_{\text{two-layer}}$ is $\eta_1=\eta_2=h^{\alpha}$.

Since $ \eta_1+\eta_2=2h^{\alpha}$, we have 
\begin{align*}
        L_{\text{two-layer}}(\mX,{\mW_1^2},{\mW_2^2},,\Tilde{\vx}_0)&=\frac{1}{h}(1+\frac{\eta_1(2h^{\alpha}-\eta_1)}{h^3})^4 +\frac{16\eta_1^2(2h^{\alpha}-\eta_1)^2}{h^7} \\ &+\left(\frac{4(\eta_1(2h^{\alpha}-\eta_1)+h^3)}{h^{5-\alpha}}-1\right)^2+(1+\frac{\eta_1(2h^{\alpha}-\eta_1)}{h^3})^2\frac{8\eta_1(2h^{\alpha}-\eta_1)}{h^5}
\end{align*}

Taking the derivative, we have 
\begin{align*}
        L^{\prime}_{\text{two-layer}}(\mX,{\mW_1^2},{\mW_2^2},\Tilde{\vx}_0)&= 2(h^{\alpha}-\eta_1) \left[\frac{4}{h^4}(1+\frac{\eta_1(2h^{\alpha}-\eta_1)}{h^3})^3 +\frac{32\eta_1(2h^{\alpha}-\eta_1)}{h^7}\right] \\ &+2(h^{\alpha}-\eta_1) \left[\frac{8}{h^{5-\alpha}}\left(\frac{4(\eta_1(2h^{\alpha}-\eta_1)+h^3)}{h^{5-\alpha}}-1\right)\right] \\
        &+2(h^{\alpha}-\eta_1)  \left[ \frac{ 8}{h^5}+ \frac{32\eta_1(2h^{\alpha}-\eta_1)}{h^8}+\frac{24\eta_1^2(2h^{\alpha}-\eta_1)^2}{h^{11}}\right]
\end{align*}

If we let   $L_{\text{two-layer}}$ is $\eta_1=\eta_2=h^{\alpha}$ be local minima, we  must need 
\begin{itemize}
    \item  $4  > 5-\alpha \Rightarrow  \alpha >1$
    \item  $3(2\alpha-3)-4  < \alpha-5 \Rightarrow  \alpha < \frac{8}{5}$
    \item  $2(2\alpha-3)-4  < \alpha-5 \Rightarrow  \alpha < \frac{5}{3}$
    \item $2\alpha-7 < \alpha-5 \Rightarrow  \alpha < 2$
    \item  $4\alpha-10 < \alpha-5 \Rightarrow  \alpha < \frac{5}{3}$
    \item $2\alpha-8 < \alpha-5 \Rightarrow  \alpha < 3$
    \item $4\alpha-11 < \alpha-5 \Rightarrow  \alpha < 2$
\end{itemize}

Take the intersection, we have $1 < \alpha< \frac{8}{5}$. Given the fixed   $ 1<\alpha< \frac{8}{5}$, we will give how large $h$ is to ensure that $\eta_1=\eta_2=h^{\alpha}$ will are local minima,

\paragraph{Case 1.} If $ \alpha =\frac{3}{2}$,  we need $$8h^{\frac{1}{2}}-32-32-64-o(1) >0,$$ which means  $h>256+o(1)$.

\paragraph{Case 2.}  If $ 1 < \alpha <\frac{3}{2}$, we find $6\alpha-13<4\alpha-10<-4$ and   $1-\alpha >\alpha-2$, so we need 
$$ (4+o(1))h^{1-\alpha} +32h^{\alpha-2}+(32+o(1))h^{\alpha-2}+8h^{-\alpha}+32h^{\alpha-3}+24h^{3\alpha-6}-8 <0,$$
which means
$$ (1+o(1))h^{1-\alpha} +16h^{\alpha-2}+2h^{-\alpha}+8h^{\alpha-3}+6h^{3\alpha-6}-2 <0,$$

\paragraph{Case 3.}  If $  \frac{3}{2} < \alpha < \frac{8}{5}$, we find $6\alpha-13>4\alpha-10>-4$, so we need $$ (4+o(1))h^{3(2\alpha-3)-\alpha+1}+32h^{\alpha-2}+(32+o(1)) h^{3\alpha-5}+8h^{-\alpha}+32h^{\alpha-3}+24h^{3\alpha-6} -8<0, $$ 
Which means   $$ (1+o(1))h^{5\alpha-8}+8h^{\alpha-2}+8 h^{3\alpha-5}+2h^{-\alpha}+8h^{\alpha-3}+6h^{3\alpha-6} -2<0.$$ 
\hfill $\square$



\subsection{Bounded Loss Gap for Approximate two-step loss  for two-layer NN under  orthogonal  initialization} \label{app:orthog-bounded-loss}


Here we consider the orthogonal  initialization which make the setting $\mX^{\top}\mX=\mX\mX^{\top}=h\mI, \mW_1^{0^{\top}}\mW_1^0=\mW_1^0\mW_1^{0^{\top}}=\mI, \mW_2^{0^{\top}}\mW_2^0=\mW_2^0\mW_2^{0^{\top}}=\mI, \mM^{{\top}}\mM=\mM\mM^{{\top}}=\frac{1}{h}\mI $. 

We define that 
\begin{align*}
\mA_1^0 &= \frac{1}{h}\mM\mW_2^{0\top}
&\;
\mA_1^1 &= \frac{1}{h}\mM\mW_2^{1\top} 
&\; \widetilde{\mA_1^1} &= \frac{1}{h}\mM\widetilde{\mW_2^{1\top}} \\
\mB_1^0 &= \frac{1}{h^2}\mW_1^0
&\;
\mB_1^1 &= \frac{1}{h^2}\mW_1^1\mW_2^1\mW_2^{1\top} 
&\;  \widetilde{\mB_1^1} &= \frac{1}{h^2}\widetilde{\mW_1^1}\widetilde{\mW_2^1}\widetilde{\mW_2^{1\top}}\\
\mA_2^0 &= \frac{1}{h}\mW_1^{0\top}\mM
&\;
\mA_2^1 &= \frac{1}{h}\mW_1^{1\top}\mM 
&\; \widetilde{\mA_2^1} &= \frac{1}{h}\widetilde{\mW_1^{1\top}}\mM  \\
\mB_2^0 &= \frac{1}{h^2}\mW_2^0
&\;
\mB_2^1 &= \frac{1}{h^2}\mW_1^{1\top}\mW_1^1\mW_2^1
&\widetilde{\mB_2^1} &= \frac{1}{h^2}\widetilde{\mW_1^{1\top}}\widetilde{\mW_1^1}\widetilde{\mW_2^1}.
\end{align*}
And we denote that 
\begin{align*}
    \mW_1^1&=\mW_1^0+\eta_1 \mA_1^0-\eta_1 \mB_1^0, \\
    \widetilde{\mW_1^1}&=\mW_1^0+\eta_1 \mA_1^0, \\
    \mW_1^2&=\mW_1^1+\eta_1 \mA_1^1-\eta_1 \mB_1^1, \\
    \widetilde{\mW_1^2}&=\widetilde{\mW_1^1}+\eta_1 \widetilde{\mA_1^1}, \\
    \overline{\mW_1^2}&=\widetilde{\mW_1^1}+\eta_1 \widetilde{\mA_1^1}-\eta_1 \widetilde{\mB_1^1}, \\
    \mW_2^1&=\mW_2^0+\eta_2 \mA_2^0-\eta_2 \mB_2^0, \\
     \widetilde{\mW_2^1}&=\mW_2^0+\eta_2 \mA_2^0, \\
   \mW_2^2&=\mW_2^1+\eta_2 \mA_2^1-\eta_2 \mB_2^1, \\
    \widetilde{\mW_2^2}&=\widetilde{\mW_2^1}+\eta_2 \widetilde{\mA_2^1}, \\
    \overline{\mW_2^2}&=\widetilde{\mW_2^1}+\eta_2 \widetilde{\mA_2^1}-\eta_2 \widetilde{\mB_2^1}, \\
\end{align*}

\vigk{The equations above should go into the main text. If the gap between exact loss and approximated loss is small, then that means the $\mA$ terms dominate the norm of $\mW$'s, (which is the feature learning regime that we are interested in). So the main message should connect: 1. lr's required for feature learning $\to$ 2. $\mA$ terms dominating norm of $\mW$ $\to$ 3. true loss and approximated loss being close to each other. $\to$ 4. choice of lr1=lr2 resulting or not resulting in minimas.
}

\begin{lemma}
Under Assumption~\ref{sec:assumption} and~\ref{sec:assumption-whiten}, for $\eta_1$,  $\eta_2 \leq O(h\sqrt{h})$, we have
\begin{align*}
    &\left |L_{\text{two-layer}}({\mW_1^1},{\mW_2^1})-L_{\text{two-layer}}({\widetilde{\mW_1^1}}, {\widetilde{\mW_2^1}}) \right|  \leq O(\frac{1}{h}). \\ 
    &\left |L_{\text{two-layer}}({\mW_1^2},{\mW_2^2})-L_{\text{two-layer}}({\widetilde{\mW_1^2}}, {\widetilde{\mW_2^2}}) \right|  \leq O(\frac{1}{\sqrt{h}}). \\  
\end{align*}
\end{lemma}

\paragraph{Proof of Lemma~\ref{sec:orthog-approx-loss}.} We first  give bounded loss gap for approximate one-step loss under orthogonal  initialization for two-layer NN. For one step, we consider $\eta_1, \eta_2 \leq O(h^{\frac{3}{2}}).$
\begin{equation}
\label{bouned_gap_for_1-step_loss_orth}
\begin{aligned}
    &\left |\sqrt{L_{\text{two-layer}}(\mX,\mW_1^1,\mW_2^1,\Tilde{\vx}_0)} - \sqrt{L_{\text{two-layer}}(\mX,\widetilde{\mW_1^1}, \widetilde{\mW_2^1},\Tilde{\vx}_0)} \right|\\
    =&\left |\mathbb{E}_{\mW_1^0,\mW_2^0,\vxi,\Tilde{\vx}_0,\mX} \norm{ \frac{1}{h}\Tilde{\vx}_0\mW_1^1\mW_2^1-\Tilde{\vx}_0\mM}_{F}-\mathbb{E}_{\mW_1^0,\mW_2^0,\vxi,\Tilde{\vx}_0,\mX} \norm{ \frac{1}{h}\Tilde{\vx}_0\widetilde{\mW_1^1}\widetilde{\mW_2^1}-\Tilde{\vx}_0\mM}_{F}\right|\\
    \leq&\left |\mathbb{E}_{\mW_1^0,\mW_2^0,\vxi,\Tilde{\vx}_0,\mX} \left(\norm{ \frac{1}{h}\Tilde{\vx}_0\mW_1^1\mW_2^1-\frac{1}{h}\Tilde{\vx}_0\widetilde{\mW_1^1}\widetilde{\mW_2^1}}_F\right)   \right|\\
    \leq&\left |\mathbb{E}_{\mW_1^0,\mW_2^0,\vxi,\Tilde{\vx}_0,\mX} \left(\norm{\frac{1}{h}\Tilde{\vx}_0}_F\norm{\mW_1^1\mW_2^1-\widetilde{\mW_1^1}\widetilde{\mW_2^1}}\right)   \right|\\
    =&\frac{1}{\sqrt{h}}\left |\mathbb{E}_{\mW_1^0,\mW_2^0,\vxi,\Tilde{\vx}_0,\mX} \left(\norm{\mW_1^1\mW_2^1-\widetilde{\mW_1^1}\widetilde{\mW_2^1}}\right)   \right|\\
    =& \frac{1}{\sqrt{h}}\left |\mathbb{E}_{\mW_1^0,\mW_2^0,\vxi\Tilde{\vx}_0,\mX} \left(\norm{-\eta_1\mB^{0}_1\mW_2^0-\eta_1\eta_2\mB^{0}_1\mA^{0}_2-\eta_2\mW_1^0\mB^{0}_2-\eta_1\eta_2\mA^{0}_1\mB^{0}_2+\eta_1\eta_2\mB^{0}_1\mB^{0}_2}\right)   \right| \\
    \leq& \frac{1}{\sqrt{h}}\left |\mathbb{E} \left(\norm{\eta_1\mB^{0}_1\mW_2^0}+\norm{\eta_1\eta_2\mB^{0}_1\mA^{0}_2}+\norm{\eta_2\mW_1^0\mB^{0}_2}+\norm{\eta_1\eta_2\mA^{0}_1\mB^{0}_2}+\norm{\eta_1\eta_2\mB^{0}_1\mB^{0}_2}\right)   \right|
    \end{aligned}
    \end{equation}

Consider similar techniques in Lemma~\ref{app:lemma_norm}, we get that $$\mathbb{E}_{\mW_1^0,\mW_2^0,\vxi,\Tilde{\vx}_0,\mX}\norm{\eta_1\mB^{0}_1\mW_2^0}\leq \eta_1 \norm{\mB^{0}_1} \norm{\mW_2^0}\leq \frac{\eta_1}{h^2},$$
$$\mathbb{E}_{\mW_1^0,\mW_2^0,\vxi,\Tilde{\vx}_0,\mX}\norm{\eta_1\eta_2\mB^{0}_1\mA^{0}_2} \leq \eta_1\eta_2 \norm{\mB^{0}_1} \norm{\mA^{0}_2} \leq \frac{\eta_1\eta_2}{h^3    \sqrt{h}}, $$
$$\mathbb{E}_{\mW_1^0,\mW_2^0,\vxi,\Tilde{\vx}_0,\mX}\norm{\eta_2\mW_1^0}\mB^{0}_2\leq \eta_2 \norm{\mB^{0}_2} \norm{\mW_1^0}\leq \frac{\eta_2}{h^2},$$
$$\mathbb{E}_{\mW_1^0,\mW_2^0,\vxi,\Tilde{\vx}_0,\mX}\norm{\eta_1\eta_2\mA^{0}_1\mB^{0}_2} \leq \eta_1\eta_2 \norm{\mB^{0}_2} \norm{\mA^{0}_1} \leq \frac{\eta_1\eta_2}{h^3\sqrt{h}}, $$
$$\mathbb{E}_{\mW_1^0,\mW_2^0,\vxi,\Tilde{\vx}_0,\mX}\norm{\eta_1\eta_2\mB^{0}_1\mB^{0}_2} \leq \eta_1\eta_2 \norm{\mB^{0}_2} \norm{\mB^{0}_1} \leq \frac{\eta_1\eta_2}{h^4}, $$

taking these  inequalities into (\ref{app:approx-loss-2}), we have 
\begin{equation}
\label{app:approx-loss-one-step-2NN-orthg-1}
\begin{aligned}
    &\left |\sqrt{L_{\text{two-layer}}(\mX,\mW_1^1,\mW_2^1,\Tilde{\vx}_0)} - \sqrt{L_{\text{two-layer}}(\mX,\widetilde{\mW_1^1}, \widetilde{\mW_2^1},\Tilde{\vx}_0)} \right|\\
    \leq& \frac{1}{\sqrt{h}}\left |\mathbb{E} \left(\norm{\eta_1\mB_1\mW_2^0}+\norm{\eta_1\eta_2\mB_1\mA_2}+\norm{\eta_2\mW_1^0\mB_2}+\norm{\eta_1\eta_2\mA_1\mB_2}+\norm{\eta_1\eta_2\mB_1\mB_2}\right)   \right| \\
    \leq& \frac{2\eta_1\eta_2}{h^4}+\frac{\eta_1+\eta_2}{h^2\sqrt{h}}+\frac{\eta_1\eta_2}{h^4\sqrt{h}} \leq O(\frac{\eta_1+\eta_2}{h^2\sqrt{h}})
    \end{aligned}
    \end{equation}

Also, 
\begin{equation}
\label{app:approx-loss-one-step-2NN-orthg-2}
    \left |\sqrt{L_{\text{two-layer}}(\mX,\mW_1^1,\mW_2^1,\Tilde{\vx}_0)} +\sqrt{L_{\text{two-layer}}(\mX,\widetilde{\mW_1^1}, \widetilde{\mW_2^1},\Tilde{\vx}_0)} \right|\leq2 \sqrt{\max(\frac{\eta_1^2}{h^4},\frac{\eta_1}{h^2}, \frac{\eta_1^2\eta_2^2}{h^7},1)} . 
\end{equation}
We combine (\ref{app:approx-loss-one-step-2NN-orthg-1}), (\ref{app:approx-loss-one-step-2NN-orthg-2}) and Assumption~\ref{sec:assumption-2}, finally we get that 
\begin{align*}
    \left |L_{\text{two-layer}}(\mX,\mW_1^1,\mW_2^1,\Tilde{\vx}_0) - L_{\text{two-layer}}(\mX,\widetilde{\mW_1^1}, \widetilde{\mW_2^1},\Tilde{\vx}_0) \right| &\leq O\left(\frac{\eta_1+\eta_2}{h^2\sqrt{h}} \sqrt{\max(\frac{\eta_1^2}{h^4},\frac{\eta_1}{h^2}, \frac{\eta_1^2\eta_2^2}{h^7},1)}\right)\\
    &\leq O(\frac{1}{h})
\end{align*}
\hfill $\square$


We are here considering the 2-step loss under orthogonal initialization for two-layer NN.
We have 
\begin{equation}
\label{2-step-bounded-loss-decomp}
\begin{aligned}
    &\left |\sqrt{L_{\text{two-layer}}(\mX,\mW_1^2,\mW_2^2,\Tilde{\vx}_0)} - \sqrt{L_{\text{two-layer}}(\mX,\widetilde{\mW_1^2}, \widetilde{\mW_2^2},\Tilde{\vx}_0)} \right|\\
    \leq& \left |\sqrt{L_{\text{two-layer}}(\mX,\mW_1^2,\mW_2^2,\Tilde{\vx}_0)} - \sqrt{L_{\text{two-layer}}(\mX,\overline{\mW_1^2}, \overline{\mW_2^2},\Tilde{\vx}_0)} \right|\\
    + & \left |\sqrt{L_{\text{two-layer}}(\mX,\overline{\mW_1^2}, \overline{\mW_2^2},\Tilde{\vx}_0)} - \sqrt{L_{\text{two-layer}}(\mX,\widetilde{\mW_1^2}, \widetilde{\mW_2^2},\Tilde{\vx}_0)} \right| 
    \end{aligned}
    \end{equation}

We first  give bounded loss gap for $\left |\sqrt{L_{\text{two-layer}}(\mX,\overline{\mW_1^2}, \overline{\mW_2^2},\Tilde{\vx}_0)} - \sqrt{L_{\text{two-layer}}(\mX,\widetilde{\mW_1^2}, \widetilde{\mW_2^2},\Tilde{\vx}_0)} \right| $.  Similar to (\ref{bouned_gap_for_1-step_loss_orth}), we have 
\begin{equation}
\label{bouned_gap_for_2-step_loss_orth-1}
\begin{aligned}
    &\left |\sqrt{L_{\text{two-layer}}(\mX,\overline{\mW_1^2}, \overline{\mW_2^2},\Tilde{\vx}_0)} - \sqrt{L_{\text{two-layer}}(\mX,\widetilde{\mW_1^2}, \widetilde{\mW_2^2},\Tilde{\vx}_0)} \right|\\
    =&\left |\mathbb{E}_{\mW_1^0,\mW_2^0,\vxi,\Tilde{\vx}_0,\mX} \norm{ \frac{1}{h}\Tilde{\vx}_0\overline{\mW_1^2}\overline{\mW_2^2}-\Tilde{\vx}_0\mM}_{F}-\mathbb{E}_{\mW_1^0,\mW_2^0,\vxi,\Tilde{\vx}_0,\mX} \norm{ \frac{1}{h}\Tilde{\vx}_0\widetilde{\mW_1^2}\widetilde{\mW_2^2}-\Tilde{\vx}_0\mM}_{F}\right|\\
    \leq&\left |\mathbb{E}_{\mW_1^0,\mW_2^0,\vxi,\Tilde{\vx}_0,\mX} \left(\norm{ \frac{1}{h}\Tilde{\vx}_0\overline{\mW_1^2}\overline{\mW_2^2}-\frac{1}{h}\Tilde{\vx}_0\widetilde{\mW_1^2}\widetilde{\mW_2^2}}_F\right)   \right|\\
    \leq&\left |\mathbb{E}_{\mW_1^0,\mW_2^0,\vxi,\Tilde{\vx}_0,\mX} \left(\norm{\frac{1}{h}\Tilde{\vx}_0}_F\norm{\overline{\mW_1^2}\overline{\mW_2^2}-\widetilde{\mW_1^2}\widetilde{\mW_2^2}}\right)   \right|\\
    =&\frac{1}{\sqrt{h}}\left |\mathbb{E}_{\mW_1^0,\mW_2^0,\vxi,\Tilde{\vx}_0,\mX} \left(\norm{\overline{\mW_1^2}\overline{\mW_2^2}-\widetilde{\mW_1^2}\widetilde{\mW_2^2}}\right)   \right|\\
    =& \frac{1}{\sqrt{h}}\left |\mathbb{E}_{\mW_1^0,\mW_2^0,\vxi\Tilde{\vx}_0,\mX} \left(\norm{-\eta_1\widetilde{\mB^{1}_1}\widetilde{\mW_2^1}-\eta_1\eta_2\widetilde{\mB^{1}_1}\widetilde{\mA^{1}_2}-\eta_2\widetilde{\mW_1^1}\widetilde{\mB^{1}_2}-\eta_1\eta_2\widetilde{\mA^{1}_1}\widetilde{\mB^{1}_2}+\eta_1\eta_2\widetilde{\mB^{1}_1}\widetilde{\mB^{1}_2}}\right)   \right| \\
    \leq& \frac{1}{\sqrt{h}}\left |\mathbb{E} \left(\norm{\eta_1\widetilde{\mB^{1}_1}\widetilde{\mW_2^1}}+\norm{\eta_1\eta_2\widetilde{\mB^{1}_1}\widetilde{\mA^{1}_2}}+\norm{\eta_2\widetilde{\mW_1^1}\widetilde{\mB^{1}_2}}+\norm{\eta_1\eta_2\widetilde{\mA^{1}_1}\widetilde{\mB^{1}_2}}+\norm{\eta_1\eta_2\widetilde{\mB^{1}_1}\widetilde{\mB^{1}_2}}\right)   \right|
    \end{aligned}
    \end{equation}

We know that

\begin{align*}
   \widetilde{\mW_1^1}&=\mW_1^0+\eta_1 \mA_1^0
&\; \widetilde{\mW_2^1}&=\mW_2^0+\eta_2 \mA_2^0 \\
\mA_1^0 &= \frac{1}{h}\mM\mW_2^{0\top}
&\; \widetilde{\mA_1^1} &= \frac{1}{h}\mM\widetilde{\mW_2^{1}}^{\top} \\
\mB_1^0 &= \frac{1}{h^2}\mW_1^0
&\;  \widetilde{\mB_1^1} &= \frac{1}{h^2}\widetilde{\mW_1^1}\widetilde{\mW_2^1}\widetilde{\mW_2^{1}}^{\top}\\
\mA_2^0 &= \frac{1}{h}\mW_1^{0\top}\mM
&\; \widetilde{\mA_2^1} &= \frac{1}{h}\widetilde{\mW_1^{1}}^{\top}\mM  \\
\mB_2^0 &= \frac{1}{h^2}\mW_2^0
&\widetilde{\mB_2^1} &= \frac{1}{h^2}\widetilde{\mW_1^{1}}^{\top}\widetilde{\mW_1^1}\widetilde{\mW_2^1}.
\end{align*}

Thus, we have 
\begin{align*}
    \widetilde{\mA_1^1} &=\frac{1}{h}\mM{\mW_2^0}^{\top}+\frac{\eta_2}{h^3}\mW_1^0 \\
    \widetilde{\mA_2^1} &=\frac{1}{h}{\mW_1^0}^{\top}\mM+\frac{\eta_2}{h^3}\mW_2^0 \\
    \widetilde{\mB_1^1} &= \left( \frac{1}{h^2}+\frac{\eta_1\eta_2+\eta_2^2}{h^5}\right)\mW_1^0 +  \left( \frac{\eta_1+\eta_2}{h^3} +\frac{\eta_1\eta_2^2}{h^6}\right)\mM{\mW_2^0}^{\top} \\
     &+\frac{\eta_2}{h^3} \mW_1^0\mW_2^0\mM^{\top}\mW_1^0 +\frac{\eta_1\eta_2}{h^4}\mM{\mW_2^0}^{\top}{\mW_1^0}^{\top}\mM{\mW_2^0}^{\top} \\
    \widetilde{\mB_2^1} &= \left( \frac{1}{h^2}+\frac{\eta_1\eta_2+\eta_1^2}{h^5}\right)\mW_2^0 +  \left( \frac{\eta_1+\eta_2}{h^3} +\frac{\eta_1^2\eta_2}{h^6}\right){\mW_1^0}^{\top}\mM \\
     &+\frac{\eta_1}{h^3} \mW_2^0 \mM^{\top}\mW_1^0\mW_2^0 +\frac{\eta_1\eta_2}{h^4}{\mW_1^0}^{\top}\mM{\mW_2^0}^{\top}{\mW_1^0}^{\top}\mM \\
\end{align*}
We consider $\eta_1, \eta_2 \leq O(h^{\frac{3}{2}})$, it is easy to find that 
\begin{align*}
    \norm{\widetilde{\mW_1^1}} & \leq  \norm{\mW_1^0} +\frac{\eta_1}{h} \norm{\mM}\norm{\mW_2^0} \\
     &\leq O(1)+\frac{\eta_1}{h\sqrt{h}} =O(1)  \\
        \norm{\widetilde{\mW_2^1}} & \leq  \norm{\mW_2^0} +\frac{\eta_1}{h} \norm{\mM}\norm{\mW_2^0} \\
     &\leq O(1)+\frac{\eta_2}{h\sqrt{h}} =O(1)  \\
     \norm{\widetilde{\mA_1^1} } & \leq \frac{1}{h} \norm{\mM} \norm{\widetilde{\mW_2^1}}\leq O(\frac{1}{h\sqrt{h}}) \\
    \norm{\widetilde{\mA_2^1} } & \leq \frac{1}{h} \norm{\mM} \norm{\widetilde{\mW_1^1}}\leq O(\frac{1}{h\sqrt{h}}) \\
    \norm{\widetilde{\mB_1^1} }  & \leq \frac{1}{h^2} \norm{\widetilde{\mW_1^1}} \norm{\widetilde{\mW_2^1}}^2 \leq O(\frac{1}{h^2}) \\
       \norm{\widetilde{\mB_2^1} }  & \leq \frac{1}{h^2} \norm{\widetilde{\mW_2^1}} \norm{\widetilde{\mW_1^1}}^2 \leq O(\frac{1}{h^2}). \\
\end{align*}
Combining (\ref{bouned_gap_for_2-step_loss_orth-1}), we have

\begin{align*}
&\frac{1}{\sqrt{h}}\left |\mathbb{E} \left(\norm{\eta_1\widetilde{\mB^{1}_1}\widetilde{\mW_2^1}}+\norm{\eta_1\eta_2\widetilde{\mB^{1}_1}\widetilde{\mA^{1}_2}}+\norm{\eta_2\widetilde{\mW_1^1}\widetilde{\mB^{1}_2}}+\norm{\eta_1\eta_2\widetilde{\mA^{1}_1}\widetilde{\mB^{1}_2}}+\norm{\eta_1\eta_2\widetilde{\mB^{1}_1}\widetilde{\mB^{1}_2}}\right)   \right| \\
&\leq  \frac{2\eta_1\eta_2}{h^4}+\frac{\eta_1+\eta_2}{h^2\sqrt{h}}+\frac{\eta_1\eta_2}{h^4\sqrt{h}} \leq O(\frac{\eta_1+\eta_2}{h^2\sqrt{h}})
\end{align*}
Follow the same way to (\ref{app:approx-loss-one-step-2NN-orthg-2}), we can obtain that
\begin{equation*}
    \left |\sqrt{L_{\text{two-layer}}(\mX,\overline{\mW_1^2},\overline{\mW_2^2},\Tilde{\vx}_0)} +\sqrt{L_{\text{two-layer}}(\mX,\widetilde{\mW_1^2}, \widetilde{\mW_2^2},\Tilde{\vx}_0)} \right|\leq O(1) . 
\end{equation*}
Finally we get that 
\begin{align*}
    \left |L_{\text{two-layer}}(\mX,\overline{\mW_1^2},\overline{\mW_2^2},\Tilde{\vx}_0) - L_{\text{two-layer}}(\mX,\widetilde{\mW_1^2},\widetilde{\mW_2^2},\Tilde{\vx}_0) \right| &\leq O\left(\frac{\eta_1+\eta_2}{h^2\sqrt{h}} \right)\leq O(\frac{1}{h})
\end{align*}

We now give bounded loss gap for $ \left |\sqrt{L_{\text{two-layer}}(\mX,\mW_1^2,\mW_2^2,\Tilde{\vx}_0)} - \sqrt{L_{\text{two-layer}}(\mX,\overline{\mW_1^2}, \overline{\mW_2^2},\Tilde{\vx}_0)} \right|$.

We have
\begin{align*}
    \mW_1^2&=\mW_1^1+\eta_1 \mA_1^1-\eta_1 \mB_1^1=\mW_1^0+\eta_1 \mA_1^0-\eta_1 \mB_1^0+\eta_1 \mA_1^1-\eta_1 \mB_1^1, \\
    \overline{\mW_1^2}&=\widetilde{\mW_1^1}+\eta_1 \widetilde{\mA_1^1}-\eta_1 \widetilde{\mB_1^1}=\mW_1^0+\eta_1 \mA_1^0+\eta_1 \widetilde{\mA_1^1}-\eta_1 \widetilde{\mB_1^1}, \\
   \mW_2^2&=\mW_2^1+\eta_2 \mA_2^1-\eta_2 \mB_2^1=\mW_2^0+\eta_2 \mA_2^0-\eta_2 \mB_2^0+\eta_2 \mA_2^1-\eta_2 \mB_2^1, \\
    \overline{\mW_2^2}&=\widetilde{\mW_2^1}+\eta_2 \widetilde{\mA_2^1}-\eta_2 \widetilde{\mB_2^1}=\mW_2^0+\eta_2 \mA_2^0+\eta_2 \widetilde{\mA_2^1}-\eta_2 \widetilde{\mB_2^1}, \\
\end{align*}
For $\mW_1^2\mW_2^2$, we have
\begin{equation}
\label{w1w2}
\begin{aligned}
 \mW_1^2\mW_2^2&=  \mW_1^0\mW_2^0 + \eta_1\mA_1^0\mW_2^0-\eta_1 \mB_1^0\mW_2^0+\eta_1 \mA_1^1\mW_2^0-\eta_1 \mB_1^1\mW_2^0 \\
  &+\eta_2\mW_1^0 \mA_2^0+\eta_1\eta_2 \mA_1^0 \mA_2^0-\eta_1\eta_2 \mB_1^0 \mA_2^0+\eta_1\eta_2 \mA_1^1 \mA_2^0-\eta_1\eta_2 \mB_1^1 \mA_2^0 \\
  &-\eta_2\mW_1^0 \mB_2^0-\eta_1\eta_2 \mA_1^0 \mB_2^0+\eta_1\eta_2 \mB_1^0 \mB_2^0-\eta_1\eta_2 \mA_1^1 \mB_2^0+\eta_1\eta_2 \mB_1^1 \mB_2^0 \\
  &+\eta_2\mW_1^0 \mA_2^1+\eta_1\eta_2 \mA_1^0 \mA_2^1-\eta_1\eta_2 \mB_1^0 \mA_2^1+\eta_1\eta_2 \mA_1^1 \mA_2^1-\eta_1\eta_2 \mB_1^1 \mA_2^1 \\
    &-\eta_2\mW_1^0 \mB_2^1-\eta_1\eta_2 \mA_1^0 \mB_2^1+\eta_1\eta_2 \mB_1^0 \mB_2^1-\eta_1\eta_2 \mA_1^1 \mB_2^1+\eta_1\eta_2 \mB_1^1 \mB_2^1,
    \end{aligned}
    \end{equation}

For $\overline{\mW_1^2}\overline{\mW_2^2}$, we have 
\begin{equation}
\label{w1w2_overline}
\begin{aligned}
 \overline{\mW_1^2}\overline{\mW_2^2}&= \mW_1^0\mW_2^0+\eta_1 \mA_1^0\mW_2^0+\eta_1 \widetilde{\mA_1^1}\mW_2^0-\eta_1 \widetilde{\mB_1^1}\mW_2^0 \\
 &+ \eta_2\mW_1^0\mA_2^0+\eta_1\eta_2 \mA_1^0\mA_2^0+\eta_1\eta_2 \widetilde{\mA_1^1}\mA_2^0-\eta_1\eta_2 \widetilde{\mB_1^1}\mA_2^0 \\
 &+\eta_2\mW_1^0\widetilde{\mA_2^1}+\eta_1\eta_2 \mA_1^0\widetilde{\mA_2^1}+\eta_1\eta_2 \widetilde{\mA_1^1}\widetilde{\mA_2^1}-\eta_1\eta_2 \widetilde{\mB_1^1}\widetilde{\mA_2^1}\\
  &-\eta_2\mW_1^0\widetilde{\mB_2^1}-\eta_1\eta_2 \mA_1^0\widetilde{\mB_2^1}-\eta_1\eta_2 \widetilde{\mA_1^1}\widetilde{\mB_2^1}+\eta_1\eta_2 \widetilde{\mB_1^1}\widetilde{\mB_2^1}
 \end{aligned}
\end{equation}

Based on (\ref{w1w2}) and (\ref{w1w2_overline}), we have 
\begin{align*}
     \mW_1^2\mW_2^2-\overline{\mW_1^2}\overline{\mW_2^2}&=-\eta_1 \mB_1^0\mW_2^0+\eta_1 (\mA_1^1-\widetilde{\mA_1^1})\mW_2^0-\eta_1 (\mB_1^1-\widetilde{\mB_1^1})\mW_2^0 \\
     &-\eta_1\eta_2 \mB_1^0 \mA_2^0 +\eta_1\eta_2 (\mA_1^1-\widetilde{\mA_1^1}) \mA_2^0-\eta_1\eta_2 (\mB_1^1-\widetilde{\mB_1^1}) \mA_2^0 \\
     &+\eta_2\mW_1^0 (\mA_2^1-\widetilde{\mA_2^1})+\eta_1\eta_2 \mA_1^0 (\mA_2^1-\widetilde{\mA_2^1}) -\eta_1\eta_2 \mB_1^0 \mA_2^1 \\ &+\eta_1\eta_2 \mA_1^1 \mA_2^1-\eta_1\eta_2 \mB_1^1 \mA_2^1 -\eta_1\eta_2 \widetilde{\mA_1^1}\widetilde{\mA_2^1}+\eta_1\eta_2 \widetilde{\mB_1^1}\widetilde{\mA_2^1} \\
     &+\eta_2\mW_1^0 (\mB_2^1-\widetilde{\mB_2^1})+\eta_1\eta_2 \mA_1^0 (\mB_2^1-\widetilde{\mB_2^1})+\eta_1\eta_2 \mB_1^0 \mB_2^1\\
     &-\eta_1\eta_2 \mA_1^1 \mB_2^1+\eta_1\eta_2 \mB_1^1 \mB_2^1+\eta_1\eta_2 \widetilde{\mA_1^1}\widetilde{\mB_2^1}-\eta_1\eta_2 \widetilde{\mB_1^1}\widetilde{\mB_2^1} \\
      &-\eta_2\mW_1^0 \mB_2^0-\eta_1\eta_2 \mA_1^0 \mB_2^0+\eta_1\eta_2 \mB_1^0 \mB_2^0-\eta_1\eta_2 \mA_1^1 \mB_2^0+\eta_1\eta_2 \mB_1^1 \mB_2^0 \\
\end{align*}
We know that
\begin{align*}
    \norm{\mA_1^0}&\leq O(\frac{1}{h\sqrt{h}}), \norm{\mA_2^0}\leq O(\frac{1}{h\sqrt{h}}),  \norm{\mB_1^0}\leq O(\frac{1}{h^2}), \norm{\mB_2^0}\leq O(\frac{1}{h^2})\\
    \norm{\widetilde{\mW_1^1}} & \leq  \norm{\mW_1^0} +\frac{\eta_1}{h} \norm{\mM}\norm{\mW_2^0} \leq O(1)+\frac{\eta_1}{h\sqrt{h}} =O(1)  \\
        \norm{\widetilde{\mW_2^1}} & \leq  \norm{\mW_2^0} +\frac{\eta_1}{h} \norm{\mM}\norm{\mW_2^0}\leq O(1)+\frac{\eta_2}{h\sqrt{h}} =O(1)  \\
    \norm{\mW_1^1} & \leq  \norm{\mW_1^0} +\eta_1\norm{\mA_1^0}+\eta_1\norm{\mB_1^0} \leq O(1)  \\
        \norm{\mW_2^1} & \leq  \norm{\mW_2^0} +\eta_2\norm{\mA_2^0}+\eta_2\norm{\mB_2^0} \leq O(1)  \\
     \norm{\widetilde{\mA_1^1} } & \leq \frac{1}{h} \norm{\mM} \norm{\widetilde{\mW_2^1}}\leq O(\frac{1}{h\sqrt{h}}),     \norm{\mA_1^1 } \leq \frac{1}{h} \norm{\mM} \norm{\mW_2^1}\leq O(\frac{1}{h\sqrt{h}}) \\
    \norm{\widetilde{\mA_2^1} } & \leq \frac{1}{h} \norm{\mM} \norm{\widetilde{\mW_1^1}}\leq O(\frac{1}{h\sqrt{h}}),    \norm{\mA_2^1 } \leq \frac{1}{h} \norm{\mM} \norm{\mW_1^1}\leq O(\frac{1}{h\sqrt{h}})  \\
    \norm{\widetilde{\mB_1^1} }  & \leq \frac{1}{h^2} \norm{\widetilde{\mW_1^1}} \norm{\widetilde{\mW_2^1}}^2 \leq O(\frac{1}{h^2}), \norm{\mB_1^1 }   \leq \frac{1}{h^2} \norm{\mW_1^1} \norm{\mW_2^1}^2 \leq O(\frac{1}{h^2}) \\
       \norm{\widetilde{\mB_2^1} }  & \leq \frac{1}{h^2} \norm{\widetilde{\mW_2^1}} \norm{\widetilde{\mW_1^1}}^2 \leq O(\frac{1}{h^2}), \norm{\mB_2^1 }   \leq \frac{1}{h^2} \norm{\mW_2^1} \norm{\mW_1^1}^2 \leq O(\frac{1}{h^2}) . \\
\end{align*}

We find that 
\begin{equation}
    \norm{\mW_1^2\mW_2^2-\overline{\mW_1^2}\overline{\mW_2^2}} \leq 3\frac{\eta_1}{h^2} + 2\frac{\eta_1}{h\sqrt{h}} +12\frac{\eta_1\eta_2}{h^3\sqrt{h}} +6\frac{\eta_1\eta_2}{h^3}+5\frac{\eta_1\eta_2}{h^4} +2\frac{\eta_2}{h\sqrt{h}}+3\frac{\eta_2}{h^2}\leq O(\frac{\eta_1+\eta_2}{h\sqrt{h}})
\end{equation}

Similar to (\ref{bouned_gap_for_2-step_loss_orth-1})
\begin{equation}
\label{bouned_gap_for_2-step_loss_orth-2}
\begin{aligned}
    &\left |\sqrt{L_{\text{two-layer}}(\mX,\mW_1^2, \mW_2^2,\Tilde{\vx}_0)} - \sqrt{L_{\text{two-layer}}(\mX,\overline{\mW_1^2}, \overline{\mW_2^2},\Tilde{\vx}_0)} \right|\\
    =&\left |\mathbb{E}_{\mW_1^0,\mW_2^0,\vxi,\Tilde{\vx}_0,\mX} \norm{ \frac{1}{h}\Tilde{\vx}_0\overline{\mW_1^2}\overline{\mW_2^2}-\Tilde{\vx}_0\mM}_{F}-\mathbb{E}_{\mW_1^0,\mW_2^0,\vxi,\Tilde{\vx}_0,\mX} \norm{ \frac{1}{h}\Tilde{\vx}_0\widetilde{\mW_1^2}\widetilde{\mW_2^2}-\Tilde{\vx}_0\mM}_{F}\right|\\
    \leq&\left |\mathbb{E}_{\mW_1^0,\mW_2^0,\vxi,\Tilde{\vx}_0,\mX} \left(\norm{ \frac{1}{h}\Tilde{\vx}_0\overline{\mW_1^2}\overline{\mW_2^2}-\frac{1}{h}\Tilde{\vx}_0\widetilde{\mW_1^2}\widetilde{\mW_2^2}}_F\right)   \right|\\
    \leq&\left |\mathbb{E}_{\mW_1^0,\mW_2^0,\vxi,\Tilde{\vx}_0,\mX} \left(\norm{\frac{1}{h}\Tilde{\vx}_0}_F\norm{\overline{\mW_1^2}\overline{\mW_2^2}-\widetilde{\mW_1^2}\widetilde{\mW_2^2}}\right)   \right|\\
    =&\frac{1}{\sqrt{h}}\left |\mathbb{E}_{\mW_1^0,\mW_2^0,\vxi,\Tilde{\vx}_0,\mX} \left(\norm{\overline{\mW_1^2}\overline{\mW_2^2}-\widetilde{\mW_1^2}\widetilde{\mW_2^2}}\right)   \right|\\
    \leq&  O(\frac{\eta_1+\eta_2}{h^2})
    \end{aligned}
    \end{equation}

Follow the same way to (\ref{app:approx-loss-one-step-2NN-orthg-2}), we can obtain that
\begin{equation*}
  \left |\sqrt{L_{\text{two-layer}}(\mX,\mW_1^2, \mW_2^2,\Tilde{\vx}_0)} +\sqrt{L_{\text{two-layer}}(\mX,\overline{\mW_1^2}, \overline{\mW_2^2},\Tilde{\vx}_0)} \right|\leq O(1) . 
\end{equation*}
Finally we get that 
\begin{align*}
    \left |L_{\text{two-layer}}(\mX,\mW_1^2,\mW_2^2,\Tilde{\vx}_0) - L_{\text{two-layer}}(\mX,\overline{\mW_1^2},\overline{\mW_2^2},\Tilde{\vx}_0) \right| &\leq O\left(\frac{\eta_1+\eta_2}{h^2} \right)\leq O(\frac{1}{\sqrt{h}})
\end{align*}

Thus, due to (\ref{2-step-bounded-loss-decomp}), we have 

\begin{equation}
\begin{aligned}
    &\left |\sqrt{L_{\text{two-layer}}(\mX,\mW_1^2,\mW_2^2,\Tilde{\vx}_0)} - \sqrt{L_{\text{two-layer}}(\mX,\widetilde{\mW_1^2}, \widetilde{\mW_2^2},\Tilde{\vx}_0)} \right|\\
    \leq& \left |\sqrt{L_{\text{two-layer}}(\mX,\mW_1^2,\mW_2^2,\Tilde{\vx}_0)} - \sqrt{L_{\text{two-layer}}(\mX,\overline{\mW_1^2}, \overline{\mW_2^2},\Tilde{\vx}_0)} \right|\\
    + & \left |\sqrt{L_{\text{two-layer}}(\mX,\overline{\mW_1^2}, \overline{\mW_2^2},\Tilde{\vx}_0)} - \sqrt{L_{\text{two-layer}}(\mX,\widetilde{\mW_1^2}, \widetilde{\mW_2^2},\Tilde{\vx}_0)} \right|  \\
    \leq & O(\frac{1}{\sqrt{h}})+ O(\frac{1}{h}) = O(\frac{1}{\sqrt{h}}).
    \end{aligned}
    \end{equation}
\hfill $\square$

\subsection{Approximate one-step loss under orthogonal initialization for three-layer NN}  \label{app:orthog-3-NN-one-step}

\begin{theorem}
    Given Assumption~\ref{sec:assumption},~\ref{sec:assumption-whiten},  and in addition assume $\eta_1$ and $\eta_2$ are no more than $O({h})$ based on Proposition~\ref{app:proposition-3-layer-orthogonal} and ~\ref{app:proposition-3-layer-orthogonal-two-steps}, consider the training procedure discussed in Section~\ref{sec:setup}, we derive the test loss after one-step and two-step GD update in a three-layer neural network:
\begin{equation}
\begin{aligned}
        L_{\text{two-layer}}({\mW_1^1},{\mW_2^1})&=\frac{\eta_1^2}{h^2}+\frac{\eta_2^2}{h^2} + \frac{2\eta_1\eta_2}{h^2}+ \frac{\eta_1^2\eta_2^2}{h^4}\\ &  -\frac{2\eta_1}{h}-\frac{2\eta_2}{h}+\frac{1}{h}+\frac{2\eta_1\eta_2}{h^3}+1 \\
    L_{\text{three-layer}}({\mW_1^2},{\mW_2^2})&=\left(\frac{2(\eta_1+\eta_2)(h+\eta_1\eta_2)}{h^2} -1 \right)^2 \\ &+\frac{1}{h}+ \frac{2\eta_1\eta_2}{h^2}+ \frac{10\eta_1\eta_2}{h^3}+\frac{\eta_1^2\eta_2^2}{h^3} \\ &+\frac{37\eta_1^2\eta_2^2}{h^4} + \frac{12\eta_1^3\eta_2^3}{h^5} + \frac{\eta_1^4\eta_2^4}{h^6}
\end{aligned}    
\end{equation}
\end{theorem}

We prove Theorem~\ref{theorem for orthog-3-layer} above by the following two subsection~\ref{app:orthog-3-NN-one-step} and~\ref{app:orthog-3-NN-two-steps}.

For  orthogonal  initialization we assume $n=h=d$.

Here we consider the  orthogonal  initialization where we  make the setting $\mX^{\top}\mX=\mX\mX^{\top}=h\mI, \mW_1^{0^{\top}}\mW_1^0=\mW_1^0\mW_1^{0^{\top}}=\mI, \mW_2^{0^{\top}}\mW_2^0=\mW_2^0\mW_2^{0^{\top}}=\mI, \va^{\top}\va=1, \mathbb{E}[ \va \va^{\top}]=\frac{1}{h}\mI,   {\vbeta^*}^{{\top}}{\vbeta^*}=1, \mathbb{E}[{\vbeta^*}{\vbeta^*}^{{\top}} ]=\frac{1}{h}\mI$.

We consider a  test data $\Tilde{\vx}_0$ under  three-layer setting, where $\frac{1}{\sqrt{h}}\Tilde{\vx}_0$ is  an random  orthogonal vector, we have
\begin{equation}
\begin{aligned}
    &L(\mX,\mW_1^1,\mW_2^1,\va,\Tilde{\vx}_0)\\=&\mathbb{E}_{\mW_1^0,\mW_2^0,\va,\vxi,\Tilde{\vx}_0,\mX} \left( \frac{1}{\sqrt{h}}\Tilde{\vx}_0\mW_1^1\mW_2^1\va-\Tilde{\vx}_0\vbeta^*\right)^2  \\
    =&\mathbb{E}_{\mW_1^0,\mW_2^0,\va,\vxi,\Tilde{\vx}_0,\mX} \left [\left( \frac{1}{\sqrt{h}}\mW_1^1\mW_2^1\va-\vbeta^*\right)^{\top}{\Tilde{\vx}_0}^{\top}\Tilde{\vx}_0 \left( \frac{1}{\sqrt{h}}\mW_1^1\mW_2^1\va-\vbeta^*\right)\right] \\
    =& tr\left( \mathbb{E}_{\mW_1^0,\mW_2^0,\va,\vxi,\Tilde{\vx}_0,\mX} \left [{\Tilde{\vx}_0}^{\top}\Tilde{\vx}_0 \left( \frac{1}{\sqrt{h}}\mW_1^1\mW_2^1\va-\vbeta^*\right)\left( \frac{1}{\sqrt{h}}\mW_1^1\mW_2^1\va-\vbeta^*\right)^{\top}\right] \right) \\
    =&tr\left( \mathbb{E}_{\mW_1^0,\mW_2^0,\va,\vxi, \mX} \left [\left( \frac{1}{\sqrt{h}}\mW_1^1\mW_2^1\va-\vbeta^*\right)\left( \frac{1}{\sqrt{h}}\mW_1^1\mW_2^1\va-\vbeta^*\right)^{\top}\right] \right) \\
    =& tr\left( \mathbb{E}_{\mW_1^0,\mW_2^0,\va,\vxi, \mX} \left [ \frac{1}{h}\mW_1^1\mW_2^1\va\va^{\top}{\mW_2^1}^{\top}{\mW_1^1}^{\top}\right] \right) \\
    -&tr\left( \mathbb{E}_{\mW_1^0,\mW_2^0,\va,\vxi, \mX} \left [ \frac{1}{\sqrt{h}}\vbeta^*\va^{\top}{\mW_2^1}^{\top}{\mW_1^1}^{\top}\right] \right) \\
    -&tr\left( \mathbb{E}_{\mW_1^0,\mW_2^0,\va,\vxi, \mX}\left [ \frac{1}{\sqrt{h}}{\mW_1^1}{\mW_2^1}\va{\vbeta^*}^{\top}\right] \right) + tr\left( \mathbb{E} \left [{\vbeta^*} {\vbeta^*}^{\top}\right] \right).
\end{aligned}
\end{equation}
Here we define $L_1,L_2,L_3, L_4$, where
\begin{align*}
   L_1&=tr\left( \mathbb{E}_{\mW_1^0,\mW_2^0,\va,\vxi, \mX} \left [ \frac{1}{h}\mW_1^1\mW_2^1\va\va^{\top}{\mW_2^1}^{\top}{\mW_1^1}^{\top}\right] \right)\\
    L_2&=tr\left( \mathbb{E}_{\mW_1^0,\mW_2^0,\va,\vxi, \mX} \left [ \frac{1}{\sqrt{h}}\vbeta^*\va^{\top}{\mW_2^1}^{\top}{\mW_1^1}^{\top}\right] \right) \\
    L_3 &=tr\left( \mathbb{E}_{\mW_1^0,\mW_2^0,\va,\vxi, \mX} \left [ \frac{1}{\sqrt{h}}{\mW_1^1}{\mW_2^1}\va{\vbeta^*}^{\top}\right] \right)\\
    L_4 &= tr\left( \mathbb{E} \left [{\vbeta^*} {\vbeta^*}^{\top}\right] \right)
\end{align*}
Thus $$L_{\text{three-layer} }=L_1-L_2-L_3+L_4$$

We have $L_1=\sum_{i=1}^{16}T_i$, where
\begin{align*}
T_1&=tr\left( \mathbb{E}_{\mW_1^0,\mW_2^0,\va, \mX} \left [ \frac{1}{h}\va\va^{\top}{\mW_2^0}^{\top}{\mW_1^0}^{\top}{\mW_1^0}{\mW_2^0}
\right] \right)=\frac{1}{h},\\
T_2&=tr\left( \mathbb{E}_{\mW_1^0,\mW_2^0,\va, \mX} \left [ \frac{\eta_1}{h^2\sqrt{h}}\va\va^{\top}{\mW_2^0}^{\top}{\mW_1^0}^{\top}\mX^{\top}\vy\va^{\top}{\mW_2^0}^{\top}\mW_2^0
\right] \right)=0,\\
T_3&=tr\left( \mathbb{E}_{\mW_1^0,\mW_2^0,\va, \mX} \left [ \frac{\eta_2}{h^2\sqrt{h}}\va\va^{\top}{\mW_2^0}^{\top}{\mW_1^0}^{\top}\mW_1^0{\mW_1^0}^{\top}\mX^{\top}\vy\va^{\top}
\right] \right)=0,\\
T_4&=tr\left( \mathbb{E}_{\mW_1^0,\mW_2^0,\va, \mX} \left [ \frac{\eta_1\eta_2}{h^4}\va\va^{\top}{\mW_2^0}^{\top}{\mW_1^0}^{\top}\mX^{\top}\vy\va^{\top}{\mW_2^0}^{\top}{\mW_1^0}^{\top}\mX^{\top}\vy\va^{\top}
\right] \right)=\frac{\eta_1\eta_2}{h^3},\\
T_5&=tr\left( \mathbb{E}_{\mW_1^0,\mW_2^0,\va, \mX} \left [ \frac{\eta_1}{h^2\sqrt{h}}\va\va^{\top}{\mW_2^0}^{\top}\mW_2^0 \va\vy^{\top}\mX\mW_1^0\mW_2^0
\right] \right)=0,\\
T_6&=tr\left( \mathbb{E}_{\mW_1^0,\mW_2^0,\va, \mX} \left [ \frac{\eta_1^2}{h^4}\va\va^{\top}{\mW_2^0}^{\top}\mW_2^0 \va\vy^{\top}\mX\mX^{\top}\vy\va^{\top}{\mW_2^0}^{\top}\mW_2^0 
\right] \right)=\frac{\eta_1^2}{h^2},\\
T_7&=tr\left( \mathbb{E}_{\mW_1^0,\mW_2^0,\va, \mX} \left [ \frac{\eta_1\eta_2}{h^4}\va\va^{\top}{\mW_2^0}^{\top}\mW_2^0 \va\vy^{\top}\mX\mW_1^0{\mW_1^0}^{\top}\mX^{\top}\vy\va^{\top}
\right] \right)=\frac{\eta_1\eta_2}{h^2},\\
T_8&=tr\left( \mathbb{E}_{\mW_1^0,\mW_2^0,\va, \mX} \left [ \frac{\eta_1^2\eta_2}{h^5\sqrt{h}}\va\va^{\top}{\mW_2^0}^{\top}\mW_2^0 \va\vy^{\top}\mX\mX^{\top}\vy\va^{\top}{\mW_2^0}^{\top}{\mW_1^0}^{\top}\mX^{\top}\vy\va^{\top}
\right] \right)=0,\\
T_9&=tr\left( \mathbb{E}_{\mW_1^0,\mW_2^0,\va, \mX} \left [ \frac{\eta_2}{h^2\sqrt{h}}\va\va^{\top}\va\vy^{\top}\mX\mW_1^0{\mW_1^0}^{\top}\mW_1^0\mW_2^0
\right] \right)=0,\\
T_{10}&=tr\left( \mathbb{E}_{\mW_1^0,\mW_2^0,\va, \mX} \left [ \frac{\eta_1\eta_2}{h^4}\va\va^{\top}\va\vy^{\top}\mX\mW_1^0{\mW_1^0}^{\top}\mX^{\top}\vy\va^{\top}{\mW_2^0}^{\top}\mW_2^0 
\right] \right)=\frac{\eta_1\eta_2}{h^2},\\
T_{11}&=tr\left( \mathbb{E}_{\mW_1^0,\mW_2^0,\va, \mX} \left [ \frac{\eta_2^2}{h^4}\va\va^{\top}\va\vy^{\top}\mX\mW_1^0{\mW_1^0}^{\top}\mW_1^0{\mW_1^0}^{\top}\mX^{\top}\vy\va^{\top}
\right] \right)=\frac{\eta_2^2}{h^2},\\
T_{12}&=tr\left( \mathbb{E}_{\mW_1^0,\mW_2^0,\va, \mX} \left [ \frac{\eta_1\eta_2^2}{h^5\sqrt{h}}\va\va^{\top}\va\vy^{\top}\mX\mW_1^0{\mW_1^0}^{\top}\mX^{\top}\vy\va^{\top}{\mW_2^0}^{\top}{\mW_1^0}^{\top}\mX^{\top}\vy\va^{\top}
\right] \right)=0,\\
T_{13}&=tr\left( \mathbb{E}_{\mW_1^0,\mW_2^0,\va, \mX} \left [ \frac{\eta_1\eta_2}{h^4}\va\va^{\top}\va\vy^{\top}\mX{\mW_1^0}{\mW_2^0}\va\vy^{\top}\mX\mW_1^0\mW_2^0
\right] \right)=\frac{\eta_1\eta_2}{h^3},\\
T_{14}&=tr\left( \mathbb{E}_{\mW_1^0,\mW_2^0,\va, \mX} \left [ \frac{\eta_1^2\eta_2}{h^5\sqrt{h}}\va\va^{\top}\va\vy^{\top}\mX{\mW_1^0}{\mW_2^0}\va\vy^{\top}\mX\mX^{\top}\vy\va^{\top}{\mW_2^0}^{\top}\mW_2^0 
\right] \right)=0,\\
T_{15}&=tr\left( \mathbb{E}_{\mW_1^0,\mW_2^0,\va, \mX} \left [ \frac{\eta_1\eta_2^2}{h^5\sqrt{h}}\va\va^{\top}\va\vy^{\top}\mX{\mW_1^0}{\mW_2^0}\va\vy^{\top}\mX\mW_1^0{\mW_1^0}^{\top}\mX^{\top}\vy\va^{\top}
\right] \right)=0,\\
T_{16}&=tr\left( \mathbb{E}_{\mW_1^0,\mW_2^0,\va, \mX} \left [ \frac{\eta_1^2\eta_2^2}{h^7}\va\va^{\top}\va\vy^{\top}\mX{\mW_1^0}{\mW_2^0}\va\vy^{\top}\mX\mX^{\top}\vy\va^{\top}{\mW_2^0}^{\top}{\mW_1^0}^{\top}\mX^{\top}\vy\va^{\top}
\right] \right)=\frac{\eta_1^2\eta_2^2}{h^4}.\\
\end{align*}
We have $L_2=\sum_{i=17}^{20} T_{i}$, where
\begin{align*}
T_{17}&=tr\left( \mathbb{E}_{\mW_1^0,\mW_2^0,\va, \mX} \left [ \frac{1}{\sqrt{h}}\vbeta^{*}\va^{\top}{\mW_2^0}^{\top}{\mW_1^0}^{\top}
\right] \right)=0,\\
T_{18}&=tr\left( \mathbb{E}_{\mW_1^0,\mW_2^0,\va, \mX} \left [ \frac{\eta_1}{h^2}\vbeta^{*}\va^{\top}{\mW_2^0}^{\top}\mW_2^0 \va\vy^{\top}\mX
\right] \right)=\frac{\eta_1}{h},\\
T_{19}&=tr\left( \mathbb{E}_{\mW_1^0,\mW_2^0,\va, \mX} \left [ \frac{\eta_2}{h^2}\vbeta^{*}\va^{\top}\va\vy^{\top}\mX\mW_1^0{\mW_1^0}^{\top}
\right] \right)=\frac{\eta_2}{h},\\
T_{20}&=tr\left( \mathbb{E}_{\mW_1^0,\mW_2^0,\va, \mX} \left [ \frac{\eta_1\eta_2}{h^3\sqrt{h}}\vbeta^{*}\va^{\top}\va\vy^{\top}\mX{\mW_1^0}{\mW_2^0}\va\vy^{\top}\mX
\right] \right)=0.\\
\end{align*}

We have $L_3=\sum_{i=21}^{24} T_{i}$, where
\begin{align*}
    T_{21}&=tr\left( \mathbb{E}_{\mW_1^0,\mW_2^0,\va, \mX} \left [ \frac{1}{\sqrt{h}}\mW_1^0\mW_2^0\va\vbeta^{*^{\top}}
\right] \right)=0,\\
T_{22}&=tr\left( \mathbb{E}_{\mW_1^0,\mW_2^0,\va, \mX} \left [ \frac{\eta_1}{h^2}\mX^{\top}\vy\va^{\top}{\mW_2^0}^{\top}\mW_2^0\va\vbeta^{*^{\top}}
\right] \right)=\frac{\eta_1}{h},\\
T_{23}&=tr\left( \mathbb{E}_{\mW_1^0,\mW_2^0,\va, \mX} \left [ \frac{\eta_2}{h^2}\mW_1^0{\mW_1^0}^{\top}\mX^{\top}\vy\va^{\top}\va\vbeta^{*^{\top}}
\right] \right)=\frac{\eta_2}{h},\\
T_{24}&=tr\left( \mathbb{E}_{\mW_1^0,\mW_2^0,\va, \mX} \left [ \frac{\eta_1\eta_2}{h^3\sqrt{h}}\mX^{\top}\vy\va^{\top}{\mW_2^0}^{\top}{\mW_1^0}^{\top}\mX^{\top}\vy\va^{\top}\va\vbeta^{*^{\top}}
\right] \right)=0,\\
\end{align*}

Based on the above computation, we see that for  orthogonal initialization, the one-step test loss for 3-layer NN is

\begin{equation}
    L_{\text{three-layer}}(\mX,\mW_1^1,\mW_2^1, \va^1,\Tilde{\vx}_0)=\frac{\eta_1^2}{h^2}+\frac{\eta_2^2}{h^2} + \frac{2\eta_1\eta_2}{h^2}+ \frac{\eta_1^2\eta_2^2}{h^4}-\frac{2\eta_1}{h}-\frac{2\eta_2}{h}+\frac{1}{h}+\frac{2\eta_1\eta_2}{h^3}+1
\end{equation}

Here, for the one-step updated  loss, we consider the following optimization problem, and we assume the following constraint   $\eta_1+\eta_2=2h^{\alpha}$,  our goal is to see whether $\eta_1=\eta_2=h^{\alpha}$ is local minima or local maxima.

\begin{equation}
    L_{\text{three-layer}}(\mX,\mW_1^1,\mW_2^1, \va^1,\Tilde{\vx}_0)=\frac{\eta_1^2}{h^2}+\frac{\eta_2^2}{h^2} + \frac{2\eta_1\eta_2}{h^2}+ \frac{\eta_1^2\eta_2^2}{h^4}-4h^{\alpha-1}+\frac{1}{h}+1+O(\frac{\eta_1\eta_2}{h^3})
\end{equation}
It is easy to find that  $\eta_1=\eta_2=h^{\alpha}$ is a local  maxima.  \hfill $\square$

\subsection{Approximate two-step loss  for three-layer NN under orthogonal  initialization}  \label{app:orthog-3-NN-two-steps}
For  orthogonal  initialization we assume $n=h=d$.

Here we consider the  orthogonal  initialization where we  make the setting $\mX^{\top}\mX=\mX\mX^{\top}=h\mI, \mW_1^{0^{\top}}\mW_1^0=\mW_1^0\mW_1^{0^{\top}}=\mI, \mW_2^{0^{\top}}\mW_2^0=\mW_2^0\mW_2^{0^{\top}}=\mI, \va^{\top}\va=1, \mathbb{E}[ \va \va^{\top}]=\frac{1}{h}\mI,   {\vbeta^*}^{{\top}}{\vbeta^*}=1, \mathbb{E}[{\vbeta^*}{\vbeta^*}^{{\top}} ]=\frac{1}{h}\mI$.

For the simplification, we only consider replacing $\mG_1$ with $\mA_1$ and $\mG_2$ with $\mA_2$. We consider $s=\frac{1}{h}$
\begin{align*}
\mA_1^0 &= \frac{1}{\sqrt{h}}\vbeta^*\va^{\top}\mW_2^{0\top}
&\;
\widetilde{\mA_1^1} &= \frac{1}{\sqrt{h}}\vbeta^*\va^{\top}\widetilde{\mW_2^{1}}^{\top} \\
\mB_1^0 &= \frac{1}{h}\mW_1^0\mW_2^0\va\va^{\top}{\mW_2^0}^{\top}
&\;
\widetilde{\mB_1^1} &= \frac{1}{h}\widetilde{\mW_1^{1}}\widetilde{\mW_2^{1}}\va\va^{\top}\widetilde{\mW_2^{1}}^{\top} \\
\mA_2^0 &= \frac{1}{\sqrt{h}}\mW_1^{0\top}\vbeta^*\va^{\top}
&\;
\widetilde{\mA_2^1} &= \frac{1}{\sqrt{h}}\widetilde{\mW_1^{1}}^{\top}\vbeta^*\va^{\top} \\
\mB_2^0 &= \frac{1}{h}\mW_2^0\va\va^{\top}
&\;
\widetilde{\mB_2^1} &= \frac{1}{h}\widetilde{\mW_1^{1}}^{\top}\widetilde{\mW_1^{1}}\widetilde{\mW_2^{1}}\va\va^{\top}
\end{align*}

Thus we have 
\begin{align*}
   \widetilde{\mW_1^1}&=\mW_1^0+\eta_1\mA_1^0 = \mW_1^0+\frac{\eta_1}{\sqrt{h}} \vbeta^*\va^{\top}\mW_2^{0\top} \\
   \widetilde{\mW_2^1}& =\mW_2^0+\eta_2\mA_2^0 = \mW_2^0+\frac{\eta_2}{\sqrt{h}} \mW_1^{0\top}\vbeta^*\va^{\top}\\
    \widetilde{\mW_1^2}&=\widetilde{\mW_1^1}+\eta_1\widetilde{\mA_1^1} = \widetilde{\mW_1^1}+\frac{\eta_1}{\sqrt{h}} \vbeta^*\va^{\top} \widetilde{\mW_2^{1}}^{\top}  \\
    &= \mW_1^0+\frac{2\eta_1}{\sqrt{h}} \vbeta^*\va^{\top}\mW_2^{0\top} + \frac{\eta_1\eta_2}{h}\vbeta^* {\vbeta^*}^{\top}\mW_1^0 \\
   \widetilde{\mW_2^2}& =\widetilde{\mW_2^1}+\eta_2\widetilde{\mA_2^1} = \widetilde{\mW_2^1}+\frac{\eta_2}{\sqrt{h}} \widetilde{\mW_1^{1}}^{\top}\vbeta^*\va^{\top}\\
   &=\mW_2^0+\frac{2\eta_2}{\sqrt{h}} \mW_1^{0\top}\vbeta^*\va^{\top} + \frac{\eta_1\eta_2}{h}\mW_2^0\va\va^{\top}
\end{align*}

we can derive that
\begin{align*}
    \widetilde{\mW_1^2} \widetilde{\mW_2^2} &=\mW_1^0\mW_2^0 +\frac{2\eta_1}{\sqrt{h}} \vbeta^*\va^{\top} + \frac{\eta_1\eta_2}{h}\vbeta^* {\vbeta^*}^{\top}\mW_1^0\mW_2^0   \\ 
     &+\frac{2\eta_2}{\sqrt{h}} \vbeta^*\va^{\top}+\frac{4\eta_1\eta_2}{h} \vbeta^*\va^{\top}\mW_2^{0\top}\mW_1^{0\top}\vbeta^*\va^{\top}+\frac{2\eta_1\eta_2^2}{h\sqrt{h}}\vbeta^*\va^{\top}\\
     &+\frac{\eta_1\eta_2}{h}\mW_1^0\mW_2^0\va\va^{\top}+\frac{2\eta_1^2\eta_2}{h\sqrt{h}}\vbeta^*\va^{\top}+ \frac{\eta_1^2\eta_2^2}{h^2}\vbeta^* {\vbeta^*}^{\top}\mW_1^0\mW_2^0\va\va^{\top} \\
     &= \mW_1^0\mW_2^0 +(\frac{2\eta_1}{\sqrt{h}}+\frac{2\eta_2}{\sqrt{h}}+\frac{2\eta_1\eta_2^2}{h\sqrt{h}}+\frac{2\eta_1^2\eta_2}{h\sqrt{h}}) \vbeta^*\va^{\top} + \frac{\eta_1\eta_2}{h}\vbeta^* {\vbeta^*}^{\top}\mW_1^0\mW_2^0  \\
     & + \frac{4\eta_1\eta_2}{h} \vbeta^*\va^{\top}\mW_2^{0\top}\mW_1^{0\top}\vbeta^*\va^{\top}+\frac{\eta_1\eta_2}{h}\mW_1^0\mW_2^0\va\va^{\top}+ \frac{\eta_1^2\eta_2^2}{h^2}\vbeta^* {\vbeta^*}^{\top}\mW_1^0\mW_2^0\va\va^{\top}\\
    \widetilde{\mW_2^{2}}^{\top} \widetilde{\mW_1^{2}}^{\top}&= {\mW_2^0}^{\top}  {\mW_1^0}^{\top} +(\frac{2\eta_1}{\sqrt{h}}+\frac{2\eta_2}{\sqrt{h}}+\frac{2\eta_1\eta_2^2}{h\sqrt{h}}+\frac{2\eta_1^2\eta_2}{h\sqrt{h}}) \va{\vbeta^*}^{\top} + \frac{\eta_1\eta_2}{h}{\mW_2^0}^{\top} {\mW_1^0}^{\top}\vbeta^* {\vbeta^*}^{\top} \\
     & + \frac{4\eta_1\eta_2}{h} \va{\vbeta^*}^{\top}\mW_1^{0}\mW_2^{0}\va{\vbeta^*}^{\top}+\frac{\eta_1\eta_2}{h}\va\va^{\top}{\mW_2^0}^{\top} {\mW_1^0}^{\top}+ \frac{\eta_1^2\eta_2^2}{h^2}\va\va^{\top}{\mW_2^0}^{\top}  {\mW_1^0}^{\top}\vbeta^* {\vbeta^*}^{\top}\\
\end{align*}

Thus, we have 
\begin{align*}
    tr\left(\mathbb{E}_{\mW_1^0,\mW_2^0,\va,\vbeta^*, \mX} \left [ \frac{1}{h}\widetilde{\mW_1^2} \widetilde{\mW_2^2}\va\va^{\top}\widetilde{\mW_2^{2}}^{\top} \widetilde{\mW_1^{2}}^{\top}\right]\right) &=     tr\left(\mathbb{E}_{\mW_1^0,\mW_2^0,\va,\vbeta^*, \mX} \left [ \frac{1}{h}\va\va^{\top}\widetilde{\mW_2^{2}}^{\top} \widetilde{\mW_1^{2}}^{\top}\widetilde{\mW_1^2} \widetilde{\mW_2^2}\right]\right)  \\&= \frac{1}{h}+ \frac{2\eta_1\eta_2}{h^2}+ \frac{2\eta_1\eta_2}{h^3}+ \frac{21\eta_1^2\eta_2^2}{h^4} \\
      &+ \frac{4\eta_1\eta_2}{h^2} tr \left(\mathbb{E}  \left [\va\va^{\top}\va{\vbeta^*}^{\top}\mW_1^{0}\mW_2^{0}\va{\vbeta^*}^{\top}\mW_1^{0}\mW_2^{0}\right]\right) \\ 
      &+\frac{4((\eta_1+\eta_2)^2(h+\eta_1\eta_2)^2)}{h^4} + \frac{4\eta_1^3\eta_2^3}{h^5} + \frac{\eta_1^4\eta_2^4}{h^6}\\
      &+ \frac{4\eta_1^2\eta_2^2}{h^3} tr \left(\mathbb{E}  \left [\va\va^{\top}\va{\vbeta^*}^{\top}\mW_1^{0}\mW_2^{0}\va{\vbeta^*}^{\top}\mW_1^{0}\mW_2^{0}\right]\right)\\
       &+ \frac{4\eta_1\eta_2}{h^2} tr\left(\mathbb{E}\left [ \va\va^{\top}{\mW_2^0}^{\top} {\mW_1^0}^{\top}  \vbeta^*\va^{\top}\mW_2^{0\top}\mW_1^{0\top}\vbeta^*\va^{\top}  \right]\right )  \\
       &+ \frac{4\eta_1^2\eta_2^2}{h^3} tr \left(\mathbb{E}  \left [\va\va^{\top}{\mW_2^0}^{\top} {\mW_1^0}^{\top}\vbeta^* \va^{\top}\mW_2^{0\top}\mW_1^{0\top}\vbeta^*\va^{\top} \right]\right)\\
        &+ \frac{4\eta_1^2\eta_2^2}{h^3} tr \left(\mathbb{E}  \left [\va\va^{\top}{\mW_2^0}^{\top} {\mW_1^0}^{\top}\vbeta^* \va^{\top}\mW_2^{0\top}\mW_1^{0\top}\vbeta^*\va^{\top} \right]\right)\\
            &+ \frac{4\eta_1^3\eta_2^3}{h^4} tr \left(\mathbb{E}  \left [\va\va^{\top}{\mW_2^0}^{\top} {\mW_1^0}^{\top}\vbeta^* \va^{\top}\mW_2^{0\top}\mW_1^{0\top}\vbeta^*\va^{\top} \right]\right)\\
  &+ \frac{4\eta_1^2\eta_2^2}{h^3} tr \left(\mathbb{E}  \left [\va\va^{\top}\va{\vbeta^*}^{\top}\mW_1^{0}\mW_2^{0}\va{\vbeta^*}^{\top}\mW_1^{0}\mW_2^{0}\right]\right)\\
              &+ \frac{4\eta_1^3\eta_2^3}{h^4} tr \left(\mathbb{E}  \left [\va\va^{\top}{\mW_2^0}^{\top} {\mW_1^0}^{\top}\vbeta^* \va^{\top}\mW_2^{0\top}\mW_1^{0\top}\vbeta^*\va^{\top} \right]\right)\\
                &= \frac{1}{h}+ \frac{2\eta_1\eta_2}{h^2}+ \frac{10\eta_1\eta_2}{h^3}+ \frac{\eta_1^2\eta_2^2}{h^3}+\frac{37\eta_1^2\eta_2^2}{h^4}\\ 
      &+\frac{4((\eta_1+\eta_2)^2(h+\eta_1\eta_2)^2}{h^4} + \frac{12\eta_1^3\eta_2^3}{h^5} + \frac{\eta_1^4\eta_2^4}{h^6}\\
 tr\left(\mathbb{E}_{\mW_1^0,\mW_2^0,\va,\vbeta^*, \mX} \left [ \frac{1}{\sqrt{h}}\vbeta^*\va^{\top}\widetilde{\mW_2^{2}}^{\top} \widetilde{\mW_1^{2}}^{\top}\right]\right) &= \frac{2(\eta_1+\eta_2)(h+\eta_1\eta_2)}{h^2}  \\
  tr\left(\mathbb{E}_{\mW_1^0,\mW_2^0,\va,\vbeta^*, \mX} \left [ \frac{1}{\sqrt{h}}\widetilde{\mW_1^2} \widetilde{\mW_2^2}\va{\vbeta^*}^{\top}\right]\right) &= \frac{2(\eta_1+\eta_2)(h+\eta_1\eta_2)}{h^2}  \\
    tr\left(\mathbb{E}_{\mW_1^0,\mW_2^0,\va,\vbeta^*, \mX} \left [ {\vbeta^*}{\vbeta^*}^{\top}\right]\right) &= 1
\end{align*}

Thus, we have 
\begin{equation}
    L_{\text{three-layer}}(\mX,\mW_1^2,\mW_2^2,\va,\Tilde{\vx}_0)=\left(\frac{2(\eta_1+\eta_2)(h+\eta_1\eta_2)}{h^2} -1 \right)^2+\frac{1}{h}+ \frac{2\eta_1\eta_2}{h^2}+ \frac{10\eta_1\eta_2}{h^3}+\frac{\eta_1^2\eta_2^2}{h^3}+\frac{37\eta_1^2\eta_2^2}{h^4} + \frac{12\eta_1^3\eta_2^3}{h^5} + \frac{\eta_1^4\eta_2^4}{h^6}
\end{equation} \hfill $\square$

\begin{corollary}
\label{cor:three-layer-NN}
Suppose $\eta_1+\eta_2 = 2h^{\alpha}$ and we consider $0<\alpha < 1$.
Then, for any $\alpha$ in this range, the point $\eta_1=\eta_2=h^{\alpha}$ is not a local minimum of the loss $L_{\text{three-layer}}({\mW_1^1},{\mW_2^1})$.
Moreover, for $0<\alpha \le \tfrac{2}{3}$, if $h > h^{*}$, then $\eta_1=\eta_2=h^{\alpha}$ is a local minimum of the loss $L_{\text{three-layer}}({\mW_1^2},{\mW_2^2})$, where ${h^{*}}$ is the root of the following equation:
\begin{equation}
\begin{aligned}
         &32h^{3\alpha-2}+ 33h^{\alpha-1}+74h^{\alpha-2}+2h^{-\alpha} \\+&10h^{-\alpha-1}+36h^{3\alpha-3}+4h^{5\alpha-4}-8=0
\end{aligned}
\end{equation}
\end{corollary}

\paragraph{Proof of Corollary~\ref{cor:three-layer-NN}.} Here, for the two-step updated  loss, we consider the following optimization problem, and we assume that   $\eta_1+\eta_2=2h^{\alpha}$, we want to find whether the local minima for  $L_{\text{three-layer}}$ is $\eta_1=\eta_2=h^{\alpha}$.

Since $ \eta_1+\eta_2=2h^{\alpha}$, we have 
\begin{align*}
        L_{\text{three-layer}}(\mX,\mW_1^2,\mW_2^2,\va,\Tilde{\vx}_0)&=\left(\frac{4(h+\eta_1(2h^{\alpha}-\eta_1))}{h^{2-\alpha}} -1 \right)^2+\frac{1}{h}+ \frac{2\eta_1(2h^{\alpha}-\eta_1)}{h^2}+ \frac{10\eta_1(2h^{\alpha}-\eta_1)}{h^3} \\& + \frac{\eta_1^2(2h^{\alpha}-\eta_1)^2}{h^3}+\frac{37\eta_1^2(2h^{\alpha}-\eta_1)^2}{h^4} + \frac{12\eta_1^3(2h^{\alpha}-\eta_1)^3}{h^5} + \frac{\eta_1^4(2h^{\alpha}-\eta_1)^4}{h^6}
\end{align*}

Taking the derivative, we have 
\begin{align*}
        L^{\prime}_{\text{three-layer}}(\mX,\mW_1^2,\mW_2^2,\va,\Tilde{\vx}_0)&= 2(h^{\alpha}-\eta_1) \left[\frac{8}{h^{2-\alpha}}\left(\frac{4(\eta_1(2h^{\alpha}-\eta_1)+h)}{h^{2-\alpha}}-1\right)\right] \\
        &+2(h^{\alpha}-\eta_1)  \left[ \frac{ 2}{h^2}+ \frac{ 10}{h^3}+ \frac{(h+74)\eta_1(2h^{\alpha}-\eta_1)}{h^4}+\frac{36\eta_1^2(2h^{\alpha}-\eta_1)^2}{h^{5}} +\frac{4\eta_1^3(2h^{\alpha}-\eta_1)^3}{h^{6}}\right]
\end{align*}

If we let   $L_{\text{two-layer}}$ is $\eta_1=\eta_2=h^{\alpha}$ be local minima, we  must need 
\begin{itemize}
    \item  $2  > 2-\alpha \Rightarrow  \alpha >0$
      \item  $2\alpha-3  < \alpha-2 \Rightarrow  \alpha < 1$
    \item  $2\alpha-4  < \alpha-2 \Rightarrow  \alpha < 2$
    \item  $4\alpha-5 < \alpha-2 \Rightarrow  \alpha < 1$
    \item $6\alpha-6< \alpha-2 \Rightarrow  \alpha < \frac{4}{5}$
    \item $2\alpha-3 < \alpha-2 \Rightarrow  \alpha < 1$
    \item $4\alpha-4 < \alpha-2 \Rightarrow  \alpha < \frac{2}{3}$
\end{itemize}

Take the intersection, we have $0 < \alpha< \frac{2}{3}$. Given the fixed   $ 0<\alpha< \frac{2}{3}$, we will give how large $h$ is to ensure that $\eta_1=\eta_2=h^{\alpha}$ will be the local minima,

We need $$ 32h^{3\alpha-2}+ 33h^{\alpha-1}+74h^{\alpha-2}+2h^{-\alpha}+10h^{-\alpha-1}+36h^{3\alpha-3}+4h^{5\alpha-4}-8<0.$$ \hfill $\square$

\section{Gaussian Initialization}
\label{app:gaussian_initialization}

\yaoqing{If I understand correctly, this section and the previous one almost have the same structure. This is why I had the question of two different settings for datasets, etc. I feel we should merge these two sections and have a single chain of assumptions, datasets, models, gradient approximations, etc. Perhaps we can think about it. If this is not going to be too much of a hassle, we should do it. But if we want to prioritize other things, such as the introduction and the discussion of results, we can.}

In this section, to obtain more general and practical  results, we extend the one-step loss analysis to gaussian initialization \addressedyaoqing{What do you mean by Gaussian and practical initialization?} while also accounting for label noise.

\begin{assumption}
\label{sec:assumption-2}
For gaussian initialization,  we consider more general case with $d=h$,  we also  assume $ c_1 n\leq h\leq C_1n$, where $C_1,  c_1$ are finite constants.
\end{assumption}

\paragraph{Dataset.} 
Here we use linear \textit{teacher} models to generate the training data of both two-layer and three-layer \textit{student} networks under gaussian initialization.
We sample $n$ data points $\{\vx_1, \cdots, \vx_n\}$ from the isotropic Gaussian $\vx_i \sim \gN(\vzero_h, \mI_h), \forall i \in [n]$ as our input data. 

\yaoqing{Why do we have two different teacher model settings? Do we have the same teacher but only different random matrix ensembles, or do we actually need different teachers?}
\begin{itemize}
    \item \textbf{Two-layer NN Case. } For a given $\vx_i \in \sR^h$, we use a linear \textit{teacher} model $F:\sR^h \to \sR^h$ to generate the corresponding label $\vy_i \in \sR^h$ \citep{du2018algorithmic} as follows
\begin{equation}
    \vy_i = F(\vx_i) + \vxi^{\prime}_i = \mM^{\top} \vx_i + \vxi^{\prime}_i.
\end{equation}
Here, $\mM \in \sR^{h \times h}$ with entries sampled i.i.d as follows $h\left[\mM\right]_{i,j} \sim \mathcal{N}(0,1) $  is the \textit{target matrix}, and $\vxi^{\prime}_i  \in \sR^{h} $ with entries sampled i.i.d as follows $\sqrt{h}\left[\vxi^{\prime}_i\right]_j \sim \gN(0, \rho_e^2)$ is the independent additive label noise. We represent $\mX \in \sR^{n \times h}, \mY \in \sR^{n\times h}$
as the input matrix and the label matrix, respectively.

    \item \textbf{Three-layer NN Case. } For a given $\vx_i \in \sR^h$, we use a linear teacher model $F^*:\sR^h \to \sR$ to generate the corresponding scalar label $y_i \in \sR$ as follows:
\begin{equation}
    y_i = F^*(\vx_i) + \xi_i = \vbeta^{*\top} \vx_i + \vxi_i.
\end{equation}
Here $\vbeta^* \in \sR^h$ with $\sqrt{h}\vbeta^* \sim \mathcal{N}(0,1) $  is the \textit{target direction}, and $\vxi_i \sim \gN(0, \rho_e^2)$ is the independent additive label noise.
We represent $\mX \in \sR^{n \times h}, \vy \in \sR^n$ as the input matrix and the label vector, respectively.
\end{itemize}

\paragraph{Model.} For two-layer and three-layer NNs, we consider the entries sampled i.i.d  follows $ \sqrt{h}\left[\mW_1^0\right]_{i, j} \sim \mathcal{N}(0,1)$, $ \sqrt{h}\left[\mW_2^0\right]_{i, j} \sim \mathcal{N}(0,1)$, $\sqrt{h}\left[\va\right]_{i} \sim \mathcal{N}(0,1), \forall i \in [h], j \in [h]$.

\subsection{Norm Analysis of One-step Update Gradient Matrices Under Gaussian Initialization}
\label{sec:norm}

We first give the norm analysis of one-step update gradient matrices under gaussian initialization. This analysis is an important step in simplifying the derivation of the theoretical test loss in the next Section (Section~\ref{sec:test_loss}). It also provides intuition about the range of learning rates that are beneficial for model training and offers a deeper understanding of the gradient matrices. Here, we follow the work of~\citet{ba2022high} and examine the norm properties of the hidden layers’ gradient matrices during a one-step update under both the three-layer and two-layer NN settings.
We give the norm analysis of  three-layer NN setting as an example, for the norm analysis of  two-layer NN setting, please see Appendix~\ref{app:norm property of 2-layer NN Case}.

The one-step update equations for the three-layer NN are as follows:
\begin{equation}
\label{eq:update_formula}
\begin{aligned}
\mW_1^{1} &= \mW_1^{0} - \eta_1 \mG_1^0, \hspace{10pt}
\mW_2^{1} = \mW_2^{0} - \eta_2 \mG_2^0,
\end{aligned}
\end{equation}
where $\mW_1^{0}, \mW_2^{0}$ are the initial hidden layer weights, $\eta_1$ and $\eta_2$ are the learning rate for the first  layer and second layer, respectively. $\mW_1^{1}$ and $\mW_2^{1}$ are the updated layer weights. $\mG_1$ and $\mG_2$ are the corresponding gradient matrices, where 
\begin{align}
\mG_1^0 &= \underbrace{\frac{1}{nh} \mX^{\top}\mX\mW_1^0\mW_2^0\va\va^{\top}\mW_2^{0^{\top}}}_{\mB_1^0}-\underbrace{\frac{1}{n\sqrt{h}} \mX^{\top}\vy\va^{\top}\mW_2^{0^{\top}}}_{\mA_1^0},  \label{eq:G_1} \\
\mG_2^0 &= \underbrace{\frac{1}{nh}\mW_1^{0^{\top}} \mX^{\top}\mX\mW_1^0\mW_2^0\va\va^{\top}}_{\mB_2^0}-\underbrace{\frac{1}{n\sqrt{h}}\mW_1^{0^{\top}}\mX^{\top}\vy\va^{\top}}_{\mA_2^0}.\label{eq:G_2}
\end{align}

By analyzing the norm of $\mA_1^0, \mA_2^0, \mB_1^0, \mB_2^0$, we have the following proposition: 
\begin{proposition}
\label{sec:proposition-3-layer}
 (Three-layer NN setting under gaussian initialization.) Under Assumption~\ref{sec:assumption-2}, there exists some constant $c^* > 0$ such that for all large $n, h$ with probability at least $1- 32e^{-c^*n}-30n^4e^{-c^{*}\sqrt{n}}$, we have gradient approximation,
 \begin{equation}
 \label{eq: gradient_appro}
 \begin{aligned}
          \norm{\mG_1^0-\mA_1^0} &\leq \frac{1}{\sqrt{n}-1}\norm{\mG_1^0}, \\
          \norm{\mG_2^0-\mA_2^0} &\leq  \frac{1}{\sqrt{n}-1}\norm{\mG_2^0}. \\
 \end{aligned}
 \end{equation}
 We obtain the norm control of gradient matrices,
 \begin{equation}
     \begin{aligned}
         \sqrt{h}\norm{\mG_1^0}&= \Theta_{h, \mathbb{P}}(1),\hspace{10pt}
          \sqrt{h}\norm{\mG_1^0}_F = \Theta_{h, \mathbb{P}}(1),\\
          \sqrt{h}\norm{\mG_2^0}&= \Theta_{h, \mathbb{P}}(1),\hspace{10pt} \sqrt{h}\norm{\mG_2^0}_F = \Theta_{h, \mathbb{P}}(1).
     \end{aligned}
 \end{equation}
 Thus, we have
 \\
 \begin{align}
     Small \ lr: \eta_1=\Theta(\sqrt{h}) \Rightarrow  &\norm{\mW_1^1-\mW_1^0} \asymp \norm{\mW_1^0} \\
     \eta_2=\Theta(\sqrt{h}) \Rightarrow  &\norm{\mW_2^1-\mW_2^0} \asymp \norm{\mW_2^0} \\
     Large \ lr: \eta_1=\Theta(h) \Rightarrow & \norm{\mW_1^1-\mW_1^0}_F\asymp\norm{\mW_1^0}_F\\
    \eta_2=\Theta(h) \Rightarrow & \norm{\mW_2^1-\mW_2^0} _F\asymp\norm{\mW_2^0}_F
 \end{align}
\end{proposition}


We provide the complete proof in the Appendix~\ref{app:norm property of 3-layer NN}. A similar result can be obtained for the two-layer NN setting. See Proposition~\ref{app:prop-two-layer}, we provide the proof in Appendix~\ref{app:norm property of 2-layer NN Case}.

Proposition~\ref{sec:proposition-3-layer}  shows that in terms of norm, $\{ \mA_i^0 \}_{i=1}^2$ is very close to $\{ \mG_i^0 \}_{i=1}^2$, which means $\{ \mA_i^0 \}_{i=1}^2$ serves as the leading term in $\{ \mG_i^0 \}_{i=1}^2$.  This approximation  can significantly simplify the subsequent analysis of the theoretical test loss for the three-layer NN when we replace the gradients $\{ \mG_i^0 \}_{i=1}^2$ with their approximated version $\{ \mA_i^0 \}_{i=1}^2$.
For the two-layer NN setting, we similarly replace the original gradient with its leading term, as justified by Proposition~\ref{app:prop-two-layer} and Lemma~\ref{sec: approx-loss}, to simplify the test loss analysis.


\subsection{Relationship between Test Loss and Layer-wise Learning Rates}
\label{sec:test_loss}

In this section, we first derive the theoretical test loss after a one-step update for both two-layer and three-layer neural networks under our setup. Based on this theoretical test loss, we vary the learning rates for each layer in these two networks.
We aim to determine whether using the same learning rates across layers leads to minimal test loss for networks trained with a one-step GD update when $\eta_1+\eta_2= 2h^{\alpha}$, where $h^{\alpha}\leq$ \textit{Large lr}.
\subsubsection{ Two-layer Neural Networks}

Given test data $\tilde{\vx}_0 \!\sim\! \gN(\vzero, \mI_h)$, we consider the test loss
\[
L_{\text{two-layer}} = \mathbb{E}_{\mW_1^0,\mW_2^0,\vxi^{\prime},\mM,\Tilde{\vx}_0,\mX} \left\| \frac{1}{\sqrt{h}} \tilde{\vx}_0 \mW_1 \mW_2 - \tilde{\vx}_0 \mM \right\|^2.
\]
The key lemma in this subsection uses the approximate gradient to replace the true gradient updates, thereby simplifying the analysis of the test loss.

\begin{lemma}
\label{sec: approx-loss}
We define the following matrices: 
\begin{align*}
    {\mA_1^0}^{\prime}&=\frac{1}{nh} \mX^{\top}\mY\mW_2^{0^{\top}},
    {\mB_1^0}^{\prime} =\frac{1}{nh^2} \mX^{\top}\mX\mW_1^0\mW_2^0\mW_2^{0^{\top}},\\
    {\mA_2^0}^{\prime}&=\frac{1}{nh}\mW_1^{0^{\top}}\mX^{\top}\mY,
    {\mB_2^0}^{\prime} =\frac{1}{nh^2}\mW_1^{0^{\top}} \mX^{\top}\mX\mW_1^0\mW_2^0.
    \end{align*}
We also define ${\widetilde{\mW_1^1}}^{\prime} = \mW_1^0+\eta_1{\mA_1^0}^{\prime}$ and  ${\widetilde{\mW_2^1}}^{\prime} = \mW_2^0+\eta_2{\mA_2^0}^{\prime}$.
Then, under Assumption~\ref{sec:assumption} and~\ref{sec:assumption-2}, for $\eta_1$,  $\eta_2$  no more than $O(h\sqrt{h})$, we have
\begin{align*}
    &\left|L_{\text{two-layer}}({\mW_1^1}^{\prime},{\mW_2^1}^{\prime})-L_{\text{two-layer}}({\widetilde{\mW_1^1}}^{\prime}, {\widetilde{\mW_2^1}}^{\prime}) \right|  \leq O(\frac{1}{h}).
\end{align*}
\end{lemma}

The simplified analysis leads to the following result for two-layer networks.

\begin{theorem}
\label{theorem for 2-layer}
    Given Assumption~\ref{sec:assumption},~\ref{sec:assumption-2}, and in addition assume $\eta_1$ and $\eta_2$ are no more than $O({h\sqrt{h}})$, based on Proposition~\ref{app:prop-two-layer} and Lemma~\ref{sec: approx-loss}, consider the training procedure discussed in Section~\ref{sec:setup}, we obtain the following test loss after one-step GD update in a two-layer neural network under gaussian initialization:
    \begin{equation}
        \begin{aligned}
            &L_{\text{two-layer}}=\frac{2\eta_1^2}{h^4}+\frac{2\eta_1^2(1+\rho_e^2)}{nh^3}+ \frac{2\eta_2^2}{h^4}+\frac{2\eta_2^2(1+\rho_e^2)}{nh^3} \\&-2\frac{\eta_1}{h^2}-2\frac{\eta_2}{h^2}+\frac{\eta_1^2\eta_2^2}{n^2h^5}+  \frac{2\eta_1\eta_2}{h^4}+\frac{2\eta_1\eta_2}{nh^3}+\frac{2\eta_1\eta_2\rho_e^2}{nh^3}\\
        &+\frac{\eta_1^2\eta_2^2(\rho_e^2+1)^2}{n^2h^5}+\frac{2\eta_1^2\eta_2^2}{h^7}+\frac{2\eta_1^2\eta_2^2\rho_e^2}{n^3h^4}+1+\frac{1}{h}\\
        &+O(\frac{\eta_1^2}{h^5})+O(\frac{\eta_2^5}{h^2})+O(\frac{\eta_1\eta_2}{h^5})+O(\frac{\eta_1^2\eta_2^2}{h^8}).
        \end{aligned}
    \end{equation}
\end{theorem}

The complete proof is provided in Appendix~\ref{app:2-layer Test Loss}.

\paragraph{Analysis of Special Cases.}
Here, we consider a special case. Specifically, we take $h = n = d$ and $\rho_e = 0$, under which the loss simplifies to:
\begin{equation}
\label{eq:two-layer-special-case}
        \begin{aligned}
                            &L_{\text{two-layer}}=\frac{4\eta_1^2}{h^4}+ \frac{4\eta_2^2}{h^4} -2\frac{\eta_1}{h^2}-2\frac{\eta_2}{h^2} +  \frac{4\eta_1\eta_2}{h^4}+\frac{4\eta_1^2\eta_2^2}{h^7}\\&+1+\frac{1}{h}+O(\frac{\eta_1^2}{h^5})+O(\frac{\eta_2^5}{h^2})+O(\frac{\eta_1\eta_2}{h^5})+O(\frac{\eta_1^2\eta_2^2}{h^8}).
        \end{aligned}
    \end{equation}

Taking special case (~\ref{eq:two-layer-special-case}) as an example~\ref{eq:two-layer-special-case}, we obtain the following corollary for two-layer neural network under gaussian initialization.
\begin{corollary}
\label{cor:2-NN-gaussian}
Suppose $\eta_1+\eta_2 = 2h^{\alpha}$ and we consider $0<\alpha \le \tfrac{3}{2}$.
Then, for any $\alpha$ in this range, the point $\eta_1=\eta_2=h^{\alpha}$ is not a local minimum of the loss $L_{\text{two-layer}}({\mW_1^1}^{\prime},{\mW_2^1}^{\prime})$.
\end{corollary}

We do simulations in Figure~\ref{fig:2-NN-more-steps-gaussian} in Appendix~\ref{app:more_experiments} to support  Corollary~\ref{cor:2-NN-gaussian}.

\subsubsection{Three-layer Neural Networks}

Given test data $\tilde{\vx}_0 \!\sim\! \gN(\vzero, \mI_d)$, we consider the test loss
\[
L_{\text{three-layer}} = \mathbb{E}_{\mW_1^0,\mW_2^0,\va,\vxi,\Tilde{\vx}_0,\mX} \left( \frac{1}{\sqrt{h}} \tilde{\vx}_0 \mW_1 \mW_2 
\va- \tilde{\vx}_0 \vbeta^* \right)^2
\]



\begin{theorem}
\label{theorem for 3-layer}
    Given Assumption~\ref{sec:assumption},~\ref{sec:assumption-2},  and in addition assume $\eta_1$ and $\eta_2$ are no more than $O({h})$ based on Proposition~\ref{sec:proposition-3-layer}, consider the training procedure discussed in Section~\ref{sec:setup}, we derive the test loss after one-step GD update in a three-layer neural network:
    \begin{equation}
        \begin{aligned}
&L_{\text{three-layer}}=\frac{\eta_1^2}{h^2}+\frac{\eta_1^2(1+\rho_e^2)}{hn}-2\frac{\eta_1}{h} +\frac{2\eta_2^2}{h^2}\\
     &+\frac{2\eta_2^2(1+\rho_e^2)}{nh}-2\frac{\eta_2}{h}+\frac{2\eta_1\eta_2}{h^2}+\frac{2\eta_1\eta_2(1+\rho_e^2)}{nh}\\
     & +\frac{\eta_1^2\eta_2^2\rho_e^2}{nh^3}+\frac{\eta_1^2\eta_2^2\rho_e^2}{n^2h^2}+\frac{4\eta_1^2\eta_2^2}{n^2h^2}+1  \\ &+O\left(\frac{\eta_1^2}{h^3}\right)  +O\left(\frac{\eta_2^2}{h^3}\right)+O\left(\frac{\eta_1\eta_2}{h^3}\right)+O\left(\frac{\eta_1^2\eta_2^2}{h^5}\right).
        \end{aligned}
    \end{equation}
\end{theorem}
The complete proof is provided in Appendix~\ref{app:3-layer NN test loss}.

\paragraph{Analysis of Special Cases.}

Here we consider a special case. Specifically, we take $h = n = d$ and $\rho_e = 0$, under which the loss becomes:
    \begin{equation}
    \label{eq:three-layer-special-case}
        \begin{aligned}
&L_{\text{three-layer}}=\frac{2\eta_1^2}{h^2}-\frac{2\eta_1}{h}+\frac{4\eta_2^2}{h^2}-\frac{2\eta_2}{h}+\frac{4\eta_1\eta_2}{h^2}+\frac{4\eta_1^2\eta_2^2h}{n^2h^3}\\
     & +1 +O\left(\frac{\eta_1^2}{h^3}\right)+O\left(\frac{\eta_2^2}{h^3}\right)+O\left(\frac{\eta_1\eta_2}{h^3}\right)+O\left(\frac{\eta_1^2\eta_2^2}{h^5}\right).
        \end{aligned}
    \end{equation}

Taking special case (~\ref{eq:three-layer-special-case}) as an example, we obtain the following corollary for three-layer neural network under gaussian initialization.
\begin{corollary}
\label{cor:3-NN-gaussian}
Suppose $\eta_1+\eta_2 = 2h^{\alpha}$ and we consider $0<\alpha < 1$.
Then, for any $\alpha$ in this range, the point $\eta_1=\eta_2=h^{\alpha}$ is not a local minimum of the loss $L_{\text{three-layer}}({\mW_1^1},{\mW_2^1})$.
\end{corollary}

We do simulations in Figure~\ref{fig:3-NN-more-steps-gaussian} in Appendix~\ref{app:more_experiments} to support  Corollary~\ref{cor:3-NN-gaussian}.

\subsection{Two-layer NN Test Loss under Gaussian initialization}
\label{app:2-layer Test Loss}

\begin{lemma}
\label{app: approx-loss}
Consider that \begin{align*}
    {\mA_1^0}&=\frac{1}{nh} \mX^{\top}\mY\mW_2^{0^{\top}}\\
    {\mB_1^0}&=\frac{1}{nh^2} \mX^{\top}\mX\mW_1^0\mW_2^0\mW_2^{0^{\top}}\\
   {\mA_2^0}&=\frac{1}{nh}\mW_1^{0^{\top}}\mX^{\top}\mY\\
    {\mB_2^0}&=\frac{1}{nh^2}\mW_1^{0^{\top}} \mX^{\top}\mX\mW_1^0\mW_2^0,
    \end{align*}
    under Assumption~\ref{sec:assumption},~\ref{sec:assumption-2}, consider ${\widetilde{\mW_1^1}} = \mW_1^0+\eta_1{\mA_1^0}$ and  ${\widetilde{\mW_2^1}} = \mW_2^0+\eta_2{\mA_2^0}$, then for $\eta_1$,  $\eta_2 \sim O(h\sqrt{h})$, we have $$\left |L_{\text{two-layer}}(\mX,{\mW_1^1},{\mW_2^1},\Tilde{\vx}_0) - L_{\text{two-layer}}(\mX,{\widetilde{\mW_1^1}}, {\widetilde{\mW_2^1}},\Tilde{\vx}_0) \right| \leq O(\frac{1}{h})$$ 
\end{lemma}

\paragraph{Proof of Lemma~\ref{app: approx-loss}.}
\begin{equation}
\label{app:approx-loss-2}
\begin{aligned}
    &\left |\sqrt{L_{\text{two-layer}}(\mX,{\mW_1^1}, {\mW_2^1},\Tilde{\vx}_0)} - \sqrt{L_{\text{two-layer}}(\mX,{\widetilde{\mW_1^1}}, {\widetilde{\mW_2^1}},\Tilde{\vx}_0)} \right|\\
    =&\left |\mathbb{E}_{\mW_1^0,\mW_2^0,\vxi,\Tilde{\vx}_0,\mX} \norm{ \frac{1}{h}\Tilde{\vx}_0 {\mW_1^1}{\mW_2^1}-\Tilde{\vx}_0\mM}_{F}-\mathbb{E}_{\mW_1^0,\mW_2^0,\vxi,\Tilde{\vx}_0,\mX} \norm{ \frac{1}{h}\Tilde{\vx}_0{\widetilde{\mW_1^1}}{\widetilde{\mW_2^1}}-\Tilde{\vx}_0\mM}_{F}\right|\\
    \leq&\left |\mathbb{E}_{\mW_1^0,\mW_2^0,\vxi,\Tilde{\vx}_0,\mX} \left(\norm{ \frac{1}{h}\Tilde{\vx}_0{\mW_1^1}{\mW_2^1}-\frac{1}{h}\Tilde{\vx}_0{\widetilde{\mW_1^1}}{\widetilde{\mW_2^1}}}_F\right)   \right|\\
    \leq&\left |\mathbb{E}_{\mW_1^0,\mW_2^0,\vxi,\Tilde{\vx}_0,\mX} \left(\norm{ \frac{1}{h}\Tilde{\vx}_0}_F\norm{{\mW_1^1}{\mW_2^1}-{\widetilde{\mW_1^1}}{\widetilde{\mW_2^1}}}\right)   \right|\\
    =&\sqrt{\frac{1}{h}}\left |\mathbb{E}_{\mW_1^0,\mW_2^0,\vxi,\Tilde{\vx}_0,\mX} \left(\norm{{\mW_1^1}{\mW_2^1}-{\widetilde{\mW_1^1}}{\widetilde{\mW_2^1}}}\right)   \right|\\
    =& \sqrt{\frac{1}{h}}\left |\mathbb{E}_{\mW_1^0,\mW_2^0,\vxi\Tilde{\vx}_0,\mX} \left(\norm{-\eta_1{\mB_1^0}\mW_2^0-\eta_1\eta_2{\mB_1^0}{\mA_2^0}-\eta_2\mW_1^0{\mB_2^0}-\eta_1\eta_2{\mA_1^0}{\mB_2^0}+\eta_1\eta_2{\mB_1^0}{\mB_2^0}}\right)   \right| \\
    \leq& \sqrt{\frac{1}{h}}\left |\mathbb{E} \left(\norm{\eta_1{\mB_1^0}\mW_2^0}+\norm{\eta_1\eta_2{\mB_1^0}{\mA_2^0}}+\norm{\eta_2\mW_1^0{\mB_2^0}}+\norm{\eta_1\eta_2{\mA_1^0}{\mB_2^0}}+\norm{\eta_1\eta_2{\mB_1^0}{\mB_2^0}}\right)   \right|
    \end{aligned}
    \end{equation}

Consider similar techniques in Lemma~\ref{app:lemma_norm}, we get that $$\mathbb{E}_{\mW_1^0,\mW_2^0,\vxi,\Tilde{\vx}_0,\mX}\norm{\eta_1{\mB_1^0}\mW_2^0}\leq \eta_1 \norm{{\mB_1^0}} \norm{\mW_2^0}\leq O(\frac{\eta_1}{h^2}),$$
$$\mathbb{E}_{\mW_1^0,\mW_2^0,\vxi,\Tilde{\vx}_0,\mX}\norm{\eta_1\eta_2{\mB_1^0}{\mA_2^0}} \leq \eta_1\eta_2 \norm{{\mB_1^0}} \norm{{\mA_2^0}} \leq O(\frac{\eta_1\eta_2}{h^3\sqrt{h}}), $$
$$\mathbb{E}_{\mW_1^0,\mW_2^0,\vxi,\Tilde{\vx}_0,\mX}\norm{\eta_2\mW_1^0{\mB_2^0}}\leq \eta_2 \norm{{\mB_2^0}} \norm{\mW_1^0}\leq O(\frac{\eta_2}{h^2}),$$
$$\mathbb{E}_{\mW_1^0,\mW_2^0,\vxi,\Tilde{\vx}_0,\mX}\norm{\eta_1\eta_2{\mA_1^0}{\mB_2^0}} \leq \eta_1\eta_2 \norm{{\mB_2^0}} \norm{{\mA_1^0}} \leq O(\frac{\eta_1\eta_2}{h^3\sqrt{h}}), $$
$$\mathbb{E}_{\mW_1^0,\mW_2^0,\vxi,\Tilde{\vx}_0,\mX}\norm{\eta_1\eta_2{\mB_1^0}{\mB_2^0}} \leq \eta_1\eta_2 \norm{{\mB_2^0}} \norm{{\mB_1^0}} \leq O(\frac{\eta_1\eta_2}{h^4}), $$

taking these  inequalities into (\ref{app:approx-loss-2}), we have 
\begin{equation}
\label{app:approx-loss-3}
\begin{aligned}
    &\left |\sqrt{L_{\text{two-layer}}(\mX,{\mW_1^1},{\mW_2^1},\Tilde{\vx}_0)} - \sqrt{L_{\text{two-layer}}(\mX,{\widetilde{\mW_1^1}}, {\widetilde{\mW_2^1}},\Tilde{\vx}_0)} \right|\\
    \leq& \sqrt{\frac{1}{h}}\left |\mathbb{E} \left(\norm{\eta_1{\mB_1^0}\mW_2^0}+\norm{\eta_1\eta_2{\mB_1^0}{\mA_2^0}}+\norm{\eta_2\mW_1^0{\mB_2^0}}+\norm{\eta_1\eta_2{\mA_1^0}{\mB_2^0}}+\norm{\eta_1\eta_2{\mB_1^0}{\mB_2^0}}\right)   \right| \\
    \leq& O(\frac{\eta_1+\eta_2}{h^2\sqrt{h}}) +O(\frac{\eta_1\eta_2}{h^4})
    \end{aligned}
    \end{equation}

Also, based on the Assumption~\ref{sec:assumption}  and  theorem~\ref{app:two-layer-loss-one-step}, we have
\begin{equation}
\label{app:approx-loss-4}
    \left |\sqrt{L_{\text{two-layer}}(\mX,{\mW_1^1},{\mW_2^1},\Tilde{\vx}_0)} +\sqrt{L_{\text{two-layer}}(\mX,{\widetilde{\mW_1^1}}, {\widetilde{\mW_2^1}},\Tilde{\vx}_0)} \right|\leq O(1). 
\end{equation}
We combine (\ref{app:approx-loss-3}), (\ref{app:approx-loss-4}) and Assumption~\ref{sec:assumption-2},  and assume $\eta_1,\eta_2 \sim O(h\sqrt{h})$, finally we get that $$\left |L_{\text{two-layer}}(\mX,{\mW_1^1},{\mW_2^1},\Tilde{\vx}_0) - L_{\text{two-layer}}(\mX,{\widetilde{\mW_1^1}}, {\widetilde{\mW_2^1}},\Tilde{\vx}_0) \right| \leq  O(\frac{\eta_1+\eta_2}{h^2\sqrt{h}}) \leq O(\frac{1}{h})$$ 
\hfill $\square$

\begin{theorem}
\label{app:two-layer-loss-one-step}
    Given Assumption~\ref{sec:assumption},~\ref{sec:assumption-2},  and in addition assume $\eta_1$ and $\eta_2$ are no more than $O({h\sqrt{h}})$ based on Proposition~\ref{app:prop-two-layer}, consider training procedure discussed in section~\ref{sec:setup},  we derive the test loss after  one-step GD update in a two-layer neural network under guassian initialization:
    \begin{equation}
        \begin{aligned}
                L_{\text{two-layer}}&=\frac{2\eta_1^2}{h^4}+\frac{2\eta_1^2(1+\rho_e^2)}{nh^3}-2\frac{\eta_1}{h^2} \\&+ \frac{2\eta_2^2}{h^4}+\frac{2\eta_2^2(1+\rho_e^2)}{nh^3}-2\frac{\eta_2}{h^2}+\frac{\eta_1^2\eta_2^2}{n^2h^5}\\
        &+  \frac{2\eta_1\eta_2}{h^4}+\frac{2\eta_1\eta_2}{nh^3}+\frac{2\eta_1\eta_2\rho_e^2}{nh^3}+\frac{\eta_1^2\eta_2^2(\rho_e^2+1)^2}{n^2h^5}+\frac{2\eta_1^2\eta_2^2}{h^7}+\frac{2\eta_1^2\eta_2^2\rho_e^2}{n^3h^4}\\
        &+1+\frac{1}{h}+O(\frac{\eta_1^2}{h^5})+O(\frac{\eta_2^5}{h^2})+O(\frac{\eta_1\eta_2}{h^5})+O(\frac{\eta_1^2\eta_2^2}{h^8})
        \end{aligned}
    \end{equation}
\end{theorem}

\paragraph{Proof of Theorem~\ref{theorem for 2-layer}.} We consider a test data $\Tilde{\vx}_0 \sim  \gN(\vzero, \mI_d)\in \sR ^{1\times d}$ under  two-layer setting , we have
\begin{equation}
\begin{aligned}
    &L_{\text{two-layer}}(\mX, {\mW_1^1},{\mW_2^1},\Tilde{\vx}_0)\\=&\mathbb{E}_{\mW_1^0,\mW_2^0,\vxi,\Tilde{\vx}_0,\mX} \norm{ \frac{1}{h}\Tilde{\vx}_0\mW_1^1\mW_2^1-\Tilde{\vx}_0\mM}_{F}^2  \\
    =&tr\left( \mathbb{E}_{\mW_1^0,\mW_2^0,\vxi,\Tilde{\vx}_0,\mX} \left [\left( \frac{1}{h}{\mW_1^1}{\mW_2^1}-\mM\right)^{\top}{\Tilde{\vx}_0}^{\top}\Tilde{\vx}_0 \left( \frac{1}{h}{\mW_1^1}{\mW_2^1}-\mM\right)\right]\right) \\
    =& tr\left( \mathbb{E}_{\mW_1^0,\mW_2^0,\vxi,\Tilde{\vx}_0,\mX} \left [{\Tilde{\vx}_0}^{\top}\Tilde{\vx}_0 \left( \frac{1}{h}{\mW_1^1}{\mW_2^1}-\mM\right)\left( \frac{1}{h}{\mW_1^1}{\mW_2^1}-\mM\right)^{\top}\right] \right) \\
    =&tr\left( \mathbb{E}_{\mW_1^0,\mW_2^0,\vxi, \mX} \left [\left( \frac{1}{h}{\mW_1^1}{\mW_2^1}-\mM\right)\left( \frac{1}{h}\mW_1^1\mW_2^1-\mM\right)^{\top}\right] \right) \\
    =& tr\left( \mathbb{E}_{\mW_1^0,\mW_2^0,\vxi, \mX} \left [ \frac{1}{h^2}{\mW_1^1}{\mW_2^1}{{\mW_2^1}}^{\top}{{\mW_1^1}}^{\top}\right] \right) \\
    -&tr\left( \mathbb{E}_{\mW_1^0,\mW_2^0,\vxi, \mX} \left [ \frac{1}{h}\mM{{\mW_2^1}}^{\top}{{\mW_1^1}}^{\top}\right] \right) \\
    -&tr\left( \mathbb{E}_{\mW_1^0,\mW_2^0,\vxi, \mX}\left [ \frac{1}{h}{\mW_1^1}{\mW_2^1}{\mM}^{\top}\right] \right)\\
    +& tr\left( \mathbb{E} \left [{\mM} {\mM}^{\top}\right] \right).
\end{aligned}
\end{equation}
Here we define $L_1,L_2,L_3, L_4$, where
\begin{align*}
   L_1&=tr\left( \mathbb{E}_{\mW_1^0,\mW_2^0,\vxi, \mX} \left [ \frac{1}{h^2}{\mW_1^1}{\mW_2^1}{{\mW_2^1}}^{\top}{{\mW_1^1}}^{\top}\right] \right)\\
    L_2&=tr\left( \mathbb{E}_{\mW_1^0,\mW_2^0,\vxi, \mX} \left [ \frac{1}{h}\mM{{\mW_2^1}}^{\top}{{\mW_1^1}}^{\top}\right] \right) \\
    L_3 &=tr\left( \mathbb{E}_{\mW_1^0,\mW_2^0,\vxi, \mX}\left [ \frac{1}{h}{\mW_1^1}{\mW_2^1}{\mM}^{\top}\right] \right)\\
    L_4 &= tr\left( \mathbb{E} \left [{\mM} {\mM}^{\top}\right] \right)
\end{align*}
Thus $$L_{\text{two-layer}}=L_1-L_2-L_3+L_4$$

Consider the exact gradient update,
\begin{align*}
        {\mW_1^1} &= \mW_1^0+\eta_1{\mA_1^0}-\eta_1{\mB_1^0} \\
           {\mW_2^1} &= \mW_2^0+\eta_2{\mA_2^0}-\eta_2{\mB_2^0}.
\end{align*}
We have 
\begin{align*}
    {\mW_1^1} {\mW_2^1}=&\mW_1^0\mW_2^0+\eta_1{\mA_1^0}\mW_2^0-\eta_1{\mB_1^0}\mW_2^0+\eta_2\mW_1^0{\mA_2^0}+\eta_1\eta_2{\mA_1^0}{\mA_2^0}\\&-\eta_1\eta_2{\mB_1^0}{\mA_2^0}-\eta_2\mW_1^0{\mB_2^0}-\eta_1\eta_2{\mA_1^0}{\mB_2^0}+\eta_1\eta_2{\mB_1^0}{\mB_2^0},
\end{align*}

 where
\begin{equation*}
\begin{aligned}
    {\mA_1^0}{\mB_2^0}&=\frac{1}{n^2h^3} \mX^{\top}\mY\mW_2^{0^{\top}}\mW_1^{0^{\top}} \mX^{\top}\mX\mW_1^0\mW_2^0\\
    {\mB_1^0}{\mA_2^0}&= \frac{1}{n^2h^3} \mX^{\top}\mX\mW_1^0\mW_2^0\mW_2^{0^{\top}}\mW_1^{0^{\top}}\mX^{\top}\mY\\
    {\mB_1^0}{\mB_2^0}&= \frac{1}{n^2h^4} \mX^{\top}\mX\mW_1^0\mW_2^0\mW_2^{0^{\top}}\mW_1^{0^{\top}} \mX^{\top}\mX\mW_1^0\mW_2^0.
\end{aligned}
\end{equation*}
Based on Lemma~\ref{app: approx-loss}, we consider replacing ${\mW_2^1}, {\mW_2^1}$ with ${\widetilde{\mW_1^1}},{\widetilde{\mW_2^1}}$.

Thus we have
\begin{align*}
{\mW_1^1} {\mW_2^1}\approx {\widetilde{\mW_1^1}}{\widetilde{\mW_2^1}}&= \mW_1^0\mW_2^0+\frac{\eta_1}{nh}\mX^{\top}\mY{\mW_2^0}^{\top}\mW_2^0 \\
&+\frac{\eta_2}{nh} \mW_1^0{\mW_1^0}^{\top}\mX^{\top}\mY\\
& +\frac{\eta_1\eta_2}{n^2h^2}\mX^{\top}\mY{\mW_2^0}^{\top}{\mW_1^0}^{\top}\mX^{\top}\mY\\
{{\mW_2^1}}^{\top}{{\mW_1^1}}^{\top}\approx \left({\widetilde{\mW_1^1}}{{\widetilde{\mW_2^1}}}\right)^{\top}&=  {\mW_2^0}^{\top}{\mW_1^0}^{\top}+\frac{\eta_1}{nh}{\mW_2^0}^{\top}\mW_2^0 \mY^{\top}\mX \\
&+\frac{\eta_2}{nh} \mY^{\top}\mX\mW_1^0{\mW_1^0}^{\top}\\
& +\frac{\eta_1\eta_2}{n^2h^2}\mY^{\top}\mX{\mW_1^0}{\mW_2^0}\mY^{\top}\mX
\end{align*}

We have $L_1=\sum_{i=1}^{16}T_i$, where
\begin{align*}
T_1&=tr\left( \mathbb{E}_{\mW_1^0,\mW_2^0,\vxi, \mX} \left [ \frac{1}{h^2}{\mW_2^0}^{\top}{\mW_1^0}^{\top}{\mW_1^0}{\mW_2^0}
\right] \right),\\
T_2&=tr\left( \mathbb{E}_{\mW_1^0,\mW_2^0,\vxi, \mX} \left [ \frac{\eta_1}{nh^3}{\mW_2^0}^{\top}{\mW_1^0}^{\top}\mX^{\top}\mY{\mW_2^0}^{\top}\mW_2^0
\right] \right),\\
T_3&=tr\left( \mathbb{E}_{\mW_1^0,\mW_2^0,\vxi, \mX} \left [ \frac{\eta_2}{nh^2}{\mW_2^0}^{\top}{\mW_1^0}^{\top}\mW_1^0{\mW_1^0}^{\top}\mX^{\top}\mY
\right] \right),\\
T_4&=tr\left( \mathbb{E}_{\mW_1^0,\mW_2^0,\vxi, \mX} \left [ \frac{\eta_1\eta_2}{n^2h^4}{\mW_2^0}^{\top}{\mW_1^0}^{\top}\mX^{\top}\mY{\mW_2^0}^{\top}{\mW_1^0}^{\top}\mX^{\top}\mY
\right] \right),\\
T_5&=tr\left( \mathbb{E}_{\mW_1^0,\mW_2^0,\vxi, \mX} \left [ \frac{\eta_1}{nh^3}{\mW_2^0}^{\top}\mW_2^0 \mY^{\top}\mX\mW_1^0\mW_2^0
\right] \right),\\
T_6&=tr\left( \mathbb{E}_{\mW_1^0,\mW_2^0,\vxi, \mX} \left [ \frac{\eta_1^2}{n^2h^4}{\mW_2^0}^{\top}\mW_2^0 \mY^{\top}\mX\mX^{\top}\mY{\mW_2^0}^{\top}\mW_2^0 
\right] \right),\\
T_7&=tr\left( \mathbb{E}_{\mW_1^0,\mW_2^0,\vxi, \mX} \left [ \frac{\eta_1\eta_2}{n^2h^4}{\mW_2^0}^{\top}\mW_2^0 \mY^{\top}\mX\mW_1^0{\mW_1^0}^{\top}\mX^{\top}\mY
\right] \right),\\
T_8&=tr\left( \mathbb{E}_{\mW_1^0,\mW_2^0,\vxi, \mX} \left [ \frac{\eta_1^2\eta_2}{n^3h^5}{\mW_2^0}^{\top}\mW_2^0 \mY^{\top}\mX\mX^{\top}\mY{\mW_2^0}^{\top}{\mW_1^0}^{\top}\mX^{\top}\mY
\right] \right),\\
T_9&=tr\left( \mathbb{E}_{\mW_1^0,\mW_2^0,\vxi, \mX} \left [ \frac{\eta_2}{nh^3}\mY^{\top}\mX\mW_1^0{\mW_1^0}^{\top}\mW_1^0\mW_2^0
\right] \right),\\
T_{10}&=tr\left( \mathbb{E}_{\mW_1^0,\mW_2^0,\vxi, \mX} \left [ \frac{\eta_1\eta_2}{n^2h^4}\mY^{\top}\mX\mW_1^0{\mW_1^0}^{\top}\mX^{\top}\mY{\mW_2^0}^{\top}\mW_2^0 
\right] \right),\\
T_{11}&=tr\left( \mathbb{E}_{\mW_1^0,\mW_2^0,\vxi, \mX} \left [ \frac{\eta_2^2}{n^2h^4}\mY^{\top}\mX\mW_1^0{\mW_1^0}^{\top}\mW_1^0{\mW_1^0}^{\top}\mX^{\top}\mY
\right] \right),\\
T_{12}&=tr\left( \mathbb{E}_{\mW_1^0,\mW_2^0,\vxi, \mX} \left [ \frac{\eta_1\eta_2^2}{n^3h^5\sqrt{h}}\mY^{\top}\mX\mW_1^0{\mW_1^0}^{\top}\mX^{\top}\mY{\mW_2^0}^{\top}{\mW_1^0}^{\top}\mX^{\top}\mY
\right] \right),\\
T_{13}&=tr\left( \mathbb{E}_{\mW_1^0,\mW_2^0,\vxi, \mX} \left [ \frac{\eta_1\eta_2}{n^2h^4}\mY^{\top}\mX{\mW_1^0}{\mW_2^0}\mY^{\top}\mX\mW_1^0\mW_2^0
\right] \right),\\
T_{14}&=tr\left( \mathbb{E}_{\mW_1^0,\mW_2^0,\vxi, \mX} \left [ \frac{\eta_1^2\eta_2}{n^3h^5\sqrt{h}}\mY^{\top}\mX{\mW_1^0}{\mW_2^0}\mY^{\top}\mX\mX^{\top}\mY{\mW_2^0}^{\top}\mW_2^0 
\right] \right),\\
T_{15}&=tr\left( \mathbb{E}_{\mW_1^0,\mW_2^0,\vxi, \mX} \left [ \frac{\eta_1\eta_2^2}{n^3h^5\sqrt{h}}\mY^{\top}\mX{\mW_1^0}{\mW_2^0}\mY^{\top}\mX\mW_1^0{\mW_1^0}^{\top}\mX^{\top}\mY
\right] \right),\\
T_{16}&=tr\left( \mathbb{E}_{\mW_1^0,\mW_2^0,\vxi, \mX} \left [ \frac{\eta_1^2\eta_2^2}{n^4h^6}\mY^{\top}\mX{\mW_1^0}{\mW_2^0}\mY^{\top}\mX\mX^{\top}\mY{\mW_2^0}^{\top}{\mW_1^0}^{\top}\mX^{\top}\mY
\right] \right).\\
\end{align*}
We have $L_2=\sum_{i=17}^{20} T_{i}$, where
\begin{align*}
T_{17}&=tr\left( \mathbb{E}_{\mW_1^0,\mW_2^0,\vxi, \mX} \left [ \frac{1}{h}\mM{\mW_2^0}^{\top}{\mW_1^0}^{\top}
\right] \right),\\
T_{18}&=tr\left( \mathbb{E}_{\mW_1^0,\mW_2^0,\vxi, \mX} \left [ \frac{\eta_1}{nh^2}\mM{\mW_2^0}^{\top}\mW_2^0 \mY^{\top}\mX
\right] \right),\\
T_{19}&=tr\left( \mathbb{E}_{\mW_1^0,\mW_2^0,\vxi, \mX} \left [ \frac{\eta_2}{nh^2}\mM\mY^{\top}\mX\mW_1^0{\mW_1^0}^{\top}
\right] \right),\\
T_{20}&=tr\left( \mathbb{E}_{\mW_1^0,\mW_2^0,\vxi, \mX} \left [ \frac{\eta_1\eta_2}{n^2h^3}\mM\mY^{\top}\mX{\mW_1^0}{\mW_2^0}\mY^{\top}\mX
\right] \right).\\
\end{align*}

We have $L_3=\sum_{i=21}^{24} T_{i}$, where
\begin{align*}
    T_{21}&=tr\left( \mathbb{E}_{\mW_1^0,\mW_2^0,\vxi, \mX} \left [ \frac{1}{h}\mW_1^0\mW_2^0\mM^{{\top}}
\right] \right),\\
T_{22}&=tr\left( \mathbb{E}_{\mW_1^0,\mW_2^0,\vxi, \mX} \left [ \frac{\eta_1}{nh^2}\mX^{\top}\mY{\mW_2^0}^{\top}\mW_2^0\mM^{{\top}}
\right] \right),\\
T_{23}&=tr\left( \mathbb{E}_{\mW_1^0,\mW_2^0,\vxi, \mX} \left [ \frac{\eta_2}{nh^2}\mW_1^0{\mW_1^0}^{\top}\mX^{\top}\mY\mM^{{\top}}
\right] \right),\\
T_{24}&=tr\left( \mathbb{E}_{\mW_1^0,\mW_2^0,\vxi, \mX} \left [ \frac{\eta_1\eta_2}{n^2h^3}\mX^{\top}\mY{\mW_2^0}^{\top}{\mW_1^0}^{\top}\mX^{\top}\mY\mM^{{\top}}
\right] \right),\\
\end{align*}

Thus, we obtain that $$L_{\text{two-layer}}=\sum_{i=1}^{16} T_{i}-\sum_{i=17}^{20} T_{i}-\sum_{i=21}^{24} T_{i}+L_4$$

\paragraph{Analysis of $T_1$.}

\begin{equation}
    T_1=tr\left( \mathbb{E}_{\mW_1^0,\mW_2^0,\vxi, \mX} \left [ \frac{1}{h^2}{\mW_2^0}^{\top}{\mW_1^0}^{\top}{\mW_1^0}{\mW_2^0}
\right] \right)=\frac{1}{h}
\end{equation}
\hfill $\square$

\paragraph{ Analysis of $T_{4}$ and $T_{13}$.}

\begin{equation}
\begin{aligned}
    T_4&=tr\left( \mathbb{E}_{\mW_1^0,\mW_2^0,\vxi, \mX} \left [ \frac{\eta_1\eta_2}{n^2h^4}{\mW_2^0}^{\top}{\mW_1^0}^{\top}\mX^{\top}\mY{\mW_2^0}^{\top}{\mW_1^0}^{\top}\mX^{\top}\mY
\right] \right)\\
&=\frac{\eta_1\eta_2}{n^2h^4} \mathbb{E}\sum_{i=1}^h\sum_{k=1}^h\sum_{m=1}^d\sum_{p=1}^n\sum_{q=1}^d\sum_{t=1}^h\sum_{k=1}^h\sum_{m=1}^d\sum_{p=1}^n\sum_{q=1}^d\\
&{\mW_2^0}_{ki}{\mW_1^0}_{mk}{\mX}_{pm}{\mX}_{pq}{\mM}_{qt}{\mW_2^0}_{kt}{\mW_1^0}_{mk}{\mX}_{pm}{\mX}_{pq}{\mM}_{qi}\\
&+\frac{\eta_1\eta_2}{n^2h^4} \mathbb{E}\sum_{i=1}^h\sum_{k=1}^h\sum_{m=1}^d\sum_{p=1}^n\sum_{t=1}^h\sum_{k=1}^h\sum_{m=1}^d\sum_{p=1}^n\\
&{\mW_2^0}_{ki}{\mW_1^0}_{mk}{\mX}_{pm}{\vxi}_{pt}{\mW_2^0}_{kt}{\mW_1^0}_{mk}{\mX}_{pm}{\vxi}_{pi}\\
\end{aligned}  
\end{equation}

For $Term_1=\frac{\eta_1\eta_2}{n^2h^4} \mathbb{E}\sum_{i=1}^h\sum_{k=1}^h\sum_{m=1}^d\sum_{p=1}^n\sum_{q=1}^d\sum_{t=1}^h\sum_{k=1}^h\sum_{m=1}^d\sum_{p=1}^n\sum_{q=1}^d $ \\$ {\mW_2^0}_{ki}{\mW_1^0}_{mk}{\mX}_{pm}{\mX}_{pq}{\mM}_{qt}{\mW_2^0}_{kt}{\mW_1^0}_{mk}{\mX}_{pm}{\mX}_{pq}{\mM}_{qi} $, we consider the following cases:

\textbf{Case 1. $k=k, i=t, m=m,q=q, p=p, m\neq q$.} 
$$Term_1^1=\frac{\eta_1\eta_2}{n^2h^5}\times h \times h \times (d^2-d)\times n \times \frac{1}{d} \times \frac{1}{d}\times \frac{1}{h}=\frac{\eta_1\eta_2}{nh^4}+O(\frac{\eta_1\eta_2}{nh^5}).$$

\textbf{Case 2. $k=k, i=t, m=m,q=q, p\neq p, m=q$.}

$$Term_1^2=\frac{\eta_1\eta_2}{n^2h^5}\times h \times h \times d\times (n^2-n) \times \frac{1}{d} \times \frac{1}{d}\times \frac{1}{h}=\frac{\eta_1\eta_2}{h^5}+O(\frac{\eta_1\eta_2}{nh^5}).$$

\textbf{Case 3. $k=k, i=t, m=m,q=q, p= p, m=q$.}

$$Term_1^3=\frac{\eta_1\eta_2}{n^2h^5}\times h \times h \times d\times n \times \frac{1}{d} \times \frac{1}{d}\times \frac{1}{h}\times3=O(\frac{\eta_1\eta_2}{nh^5}).$$

Thus we have $$Term_1=\frac{\eta_1\eta_2}{nh^4}+\frac{\eta_1\eta_2}{h^5}+O(\frac{\eta_1\eta_2}{nh^5})$$

For $Term_2=\frac{\eta_1\eta_2}{n^2h^4} \mathbb{E}\sum_{i=1}^h\sum_{k=1}^h\sum_{m=1}^d\sum_{p=1}^n\sum_{t=1}^h\sum_{k=1}^h\sum_{m=1}^d\sum_{p=1}^n$\\
${\mW_2^0}_{ki}{\mW_1^0}_{mk}{\mX}_{pm}{\vxi}_{pt}{\mW_2^0}_{kt}{\mW_1^0}_{mk}{\mX}_{pm}{\vxi}_{pi}$, we consider the following case:

\textbf{Case 1. $k=k, i=t, m=m,p=p$.} 
$$Term_2=\frac{\eta_1\eta_2}{n^2h^5}\times h \times h \times d\times n \times \frac{1}{d} \times \frac{1}{h}\times \rho_e^2=\frac{\eta_1\eta_2\rho_e^2}{nh^4}.$$

Combine $Term_1$ and $Term_2$

We finally get that
\begin{equation}
    T_4=O(\frac{\eta_1\eta_2}{h^5})
\end{equation}
\hfill $\square$

Since it is easy to see that $T_4=T_{13}$, we have 
\begin{equation}
    T_{13}=O(\frac{\eta_1\eta_2}{h^5})
\end{equation}
\hfill $\square$

\paragraph{ Analysis of $T_{6}$ and $T_{11}$.}

\begin{align*}
    T_6&=tr\left( \mathbb{E}_{\mW_1^0,\mW_2^0,\vxi, \mX} \left [ \frac{\eta_1^2}{n^2h^4}{\mW_2^0}^{\top}\mW_2^0 \mY^{\top}\mX\mX^{\top}\mY{\mW_2^0}^{\top}\mW_2^0 
\right] \right)\\
&=tr\left( \mathbb{E}_{\mW_1^0,\mW_2^0,\vxi, \mX} \left [ \frac{\eta_1^2}{n^2h^4}{\mW_2^0}^{\top}\mW_2^0 {\mW_2^0}^{\top}\mW_2^0 \mY^{\top}\mX\mX^{\top}\mY
\right] \right)
\end{align*}

Similar to ~\ref{app:four square of x}, we have 
$$\mathbb{E}({\mW_2^0}^{\top}\mW_2^0 {\mW_2^0}^{\top}\mW_2^0)=\left(2+\frac{1}{h}\right)\mI_h.$$
Taking $\mathbb{E}({\mW_2^0}^{\top}\mW_2^0 {\mW_2^0}^{\top}\mW_2^0)$ into $T_6$, we have 
\begin{equation}
   \begin{aligned}
T_6&=\frac{\eta_1^2}{n^2h^4}\left(2+\frac{1}{h}\right)tr\left( \mathbb{E}_{\mW_1^0,\mW_2^0,\vxi, \mX} \left [ \mY^{\top}\mX\mX^{\top}\mY
\right] \right)\\
&=\frac{\eta_1^2}{n^2h^4}\left(2+\frac{1}{h}\right)tr\left( \mathbb{E}_{\mW_1^0,\mW_2^0,\vxi, \mX} \left [ \mM^{\top}\mX^{\top}\mX\mX^{\top}\mX\mM+{\vxi}^{\top}\mX\mX^{\top}{\vxi}
\right] \right) \\
&=\frac{\eta_1^2}{n^2h^4}\left(2+\frac{1}{h}\right)tr\left( \mathbb{E}_{\mW_1^0,\mW_2^0,\vxi, \mX} \left [ \mM^{\top}\mX^{\top}\mX\mX^{\top}\mX\mM\right] \right)\\
&+\frac{\eta_1^2}{n^2h^4}\left(2+\frac{1}{h}\right)tr\left( \mathbb{E}_{\mW_1^0,\mW_2^0,\vxi, \mX} \left [{\vxi}^{\top}\mX\mX^{\top}{\vxi}
\right] \right) \\
&=\frac{\eta_1^2}{n^2h^4d}\left(2+\frac{1}{h}\right)\left( n^2d+nd^2+nd\right)+\frac{\eta_1^2}{n^2h^5}\left(2+\frac{1}{h}\right)\left(nhd\rho_e^2\right)\\
&=\frac{2\eta_1^2}{h^4}+\frac{2\eta_1^2d}{nh^4}+\frac{2\eta_1^2d\rho_e^2}{nh^4}+O(\frac{\eta_1^2}{h^5})
\end{aligned} 
\end{equation}
\hfill $\square$

Similar to $T_{6}$, For $T_{11}$
we have 
\begin{equation}
  \begin{aligned}
    T_{11}&=tr\left( \mathbb{E}_{\mW_1^0,\mW_2^0,\vxi, \mX} \left [ \frac{\eta_2^2}{n^2h^4}\mY^{\top}\mX\mW_1^0{\mW_1^0}^{\top}\mW_1^0{\mW_1^0}^{\top}\mX^{\top}\mY
\right] \right)\\
&=tr\left( \mathbb{E}_{\mW_1^0,\mW_2^0,\vxi, \mX} \left [ \frac{\eta_2^2}{n^2h^4}\mX^{\top}\mY\mY^{\top}\mX\mW_1^0{\mW_1^0}^{\top}\mW_1^0{\mW_1^0}^{\top}
\right] \right)\\
&=\frac{\eta_2^2}{n^2h^4}\left(\frac{h^2+hd+h}{d^2}\right)tr\left( \mathbb{E}_{\mW_1^0,\mW_2^0,\vxi, \mX} \left [ \mX^{\top}\mY\mY^{\top}\mX
\right] \right)\\
&=\frac{\eta_2^2}{n^2h^4}\left(\frac{h^2+hd+h}{d^2}\right)tr\left( \mathbb{E}_{\mW_1^0,\mW_2^0,\vxi, \mX} \left [ \mX^{\top}\mX\mM\mM^{\top}\mX^{\top}\mX+\mX^{\top}{\vxi}{\vxi}^{\top}\mX
\right] \right) \\
&=\frac{\eta_2^2}{n^2h^4d}\left(\frac{h^2+hd+h}{d^2}\right)\left(n^2d+nd^2+nd \right)+\frac{\eta_2^2}{n^2h^5}\left(\frac{h^2+hd+h}{d^2}\right)\left(ndh\rho_e^2\right)\\
&=\frac{\eta_2^2}{d^2h^2}+\frac{\eta_2^2}{dh^3}+\frac{\eta_2^2}{ndh^2}+\frac{\eta_2^2}{nh^3}+\frac{\eta_2^2\rho_e^2}{ndh^2}+\frac{\eta_2^2\rho_e^2}{nh^3}+O(\frac{\eta_2^2}{h^5})
\end{aligned}  
\end{equation}
\hfill $\square$

\paragraph{ Analysis of $T_{7}$ and $T_{10}$.}
\begin{equation}
 \begin{aligned}
    T_7&=tr\left( \mathbb{E}_{\mW_1^0,\mW_2^0,\vxi, \mX} \left [ \frac{\eta_1\eta_2}{n^2h^4}{\mW_2^0}^{\top}\mW_2^0 \mY^{\top}\mX\mW_1^0{\mW_1^0}^{\top}\mX^{\top}\mY
\right] \right)\\
&=\frac{\eta_1\eta_2}{n^2h^4}tr\left( \mathbb{E}_{\mW_1^0,\mW_2^0,\vxi, \mX} \left [ \mX^{\top}\mY{\mW_2^0}^{\top}\mW_2^0 \mY^{\top}\mX\mW_1^0{\mW_1^0}^{\top}
\right] \right)\\
&=\frac{\eta_1\eta_2}{n^2h^3d}tr\left( \mathbb{E}_{\mW_1^0,\mW_2^0,\vxi, \mX} \left [ \mX^{\top}\mY{\mW_2^0}^{\top}\mW_2^0 \mY^{\top}\mX
\right] \right)\\
&=\frac{\eta_1\eta_2}{n^2h^3d}tr\left( \mathbb{E}_{\mW_1^0,\mW_2^0,\vxi, \mX} \left [ \mY^{\top}\mX\mX^{\top}\mY{\mW_2^0}^{\top}\mW_2^0 
\right] \right)\\
&=\frac{\eta_1\eta_2}{n^2h^3d}tr\left( \mathbb{E}_{\mW_1^0,\mW_2^0,\vxi, \mX} \left [ \mY^{\top}\mX\mX^{\top}\mY 
\right] \right)\\
&=\frac{\eta_1\eta_2}{n^2d^2h^3}\left(n^2d+nd^2+nd\right)+\frac{\eta_1\eta_2}{n^2h^4d}(nhd\rho_e^2)\\
&=\frac{\eta_1\eta_2}{dh^3}+\frac{\eta_1\eta_2}{nh^3}+\frac{\eta_1\eta_2\rho_e^2}{nh^3}+O(\frac{\eta_1\eta_2}{h^5})
\end{aligned}   
\end{equation}
\hfill $\square$

It is easy to find 
\begin{equation}
T_{10}=T_{7}=\frac{\eta_1\eta_2}{dh^3}+\frac{\eta_1\eta_2}{nh^3}+\frac{\eta_1\eta_2\rho_e^2}{nh^3}+O(\frac{\eta_1\eta_2}{h^5})    
\end{equation}
\hfill $\square$

\paragraph{Analysis of $T_{16}$}
\begin{equation}
    \begin{aligned}
    T_{16}&=tr\left( \mathbb{E}_{\mW_1^0,\mW_2^0,\vxi, \mX} \left [ \frac{\eta_1^2\eta_2^2}{n^4h^6}\mY^{\top}\mX{\mW_1^0}{\mW_2^0}\mY^{\top}\mX\mX^{\top}\mY{\mW_2^0}^{\top}{\mW_1^0}^{\top}\mX^{\top}\mY
\right] \right)\\
&=\frac{\eta_1^2\eta_2^2}{n^4h^6}tr\left( \mathbb{E} \left [\mM^{\top}\mX^{\top}\mX{\mW_1^0}{\mW_2^0}\mM^{\top}\mX^{\top}\mX\mX^{\top}\mX\mM{\mW_2^0}^{\top}{\mW_1^0}^{\top}\mX^{\top}\mX\mM
\right] \right)\\
&+\frac{\eta_1^2\eta_2^2}{n^4h^6}tr\left( \mathbb{E} \left [{\vxi}^{\top}\mX{\mW_1^0}{\mW_2^0}{\vxi}^{\top}\mX\mX^{\top}{\vxi}{\mW_2^0}^{\top}{\mW_1^0}^{\top}\mX^{\top}{\vxi}
\right] \right)\\
&+ \frac{2\eta_1^2\eta_2^2}{n^4h^6}tr\left( \mathbb{E} \left [{\vxi}^{\top}\mX{\mW_1^0}{\mW_2^0}{\vxi}^{\top}\mX\mX^{\top}\mX\mM{\mW_2^0}^{\top}{\mW_1^0}^{\top}\mX^{\top}\mX\mM
\right] \right)\\
&+\frac{\eta_1^2\eta_2^2}{n^4h^6}tr\left( \mathbb{E} \left [\mM^{\top}\mX^{\top}\mX{\mW_1^0}{\mW_2^0}{\vxi}^{\top}\mX\mX^{\top}{\vxi}{\mW_2^0}^{\top}{\mW_1^0}^{\top}\mX^{\top}\mX\mM
\right] \right)\\
&+ \frac{\eta_1^2\eta_2^2}{n^4h^6}tr\left( \mathbb{E} \left [{\vxi}^{\top}\mX{\mW_1^0}{\mW_2^0}\mM^{\top}\mX^{\top}\mX\mX^{\top}\mX\mM{\mW_2^0}^{\top}{\mW_1^0}^{\top}\mX^{\top}{\vxi}
\right] \right)\\
\end{aligned}
\end{equation}

For $\frac{\eta_1^2\eta_2^2}{n^4h^6}tr\left( \mathbb{E} \left [\mM^{\top}\mX^{\top}\mX{\mW_1^0}{\mW_2^0}\mM^{\top}\mX^{\top}\mX\mX^{\top}\mX\mM{\mW_2^0}^{\top}{\mW_1^0}^{\top}\mX^{\top}\mX\mM
\right] \right)$, we have 
\begin{align*}
    &\frac{\eta_1^2\eta_2^2}{n^4h^6}tr\left( \mathbb{E} \left [\mM^{\top}\mX^{\top}\mX{\mW_1^0}{\mW_2^0}\mM^{\top}\mX^{\top}\mX\mX^{\top}\mX\mM{\mW_2^0}^{\top}{\mW_1^0}^{\top}\mX^{\top}\mX\mM
\right] \right)\\
=&\frac{\eta_1^2\eta_2^2}{n^4h^6}\mathbb{E}\sum_{i=1}^h\sum_{k=1}^d\sum_{m=1}^n\sum_{q=1}^d\sum_{s=1}^h\sum_{t=1}^h\sum_{b=1}^d\sum_{p=1}^n\sum_{c=1}^d \sum_{p=1}^n\sum_{b=1}^d\sum_{t=1}^h\sum_{s=1}^h\sum_{q=1}^d\sum_{m=1}^n\sum_{k=1}^d \\
& \mM_{ki}\mX_{mk}\mX_{mq}{\mW_1^0}_{qs}{\mW_2^0}_{st}\mM_{bt}\mX_{pb}\mX_{pc}\mX_{pc}\mX_{pb}\mM_{bt}{\mW_2^0}_{st}{\mW_1^0}_{qs}\mX_{mq}\mX_{mk}\mM_{ki}\\
=&\frac{\eta_1^2\eta_2^2}{n^4h^6}\mathbb{E}\sum_{i=1}^h\sum_{k=1}^d\sum_{m=1}^n\sum_{q=1}^d\sum_{s=1}^h\sum_{t=1}^h\sum_{b=1}^d\sum_{p=1}^n\sum_{c=1}^d \sum_{p=1}^n\sum_{b=1}^d\sum_{m=1}^n\sum_{k=1}^d \\
& \mM_{ki}\mX_{mk}\mX_{mq}{{\mW_1^0}_{qs}}^2{{\mW_2^0}_{st}}^2\mM_{bt}\mX_{pb}\mX_{pc}\mX_{pc}\mX_{pb}\mM_{bt}\mX_{mq}\mX_{mk}\mM_{ki}\\
\end{align*}
\textbf{We focus only on the case dominated by the leading term.} Since other cases will  be $O(\frac{\eta_1^2\eta_2^2}{h^8})$.

\textbf{Case 1. $b=b, k=k, i\neq t, m=m,p=p,m\neq p, k\neq q, b\neq c  $.}
\begin{align*}
    =&\frac{\eta_1^2\eta_2^2}{n^4h^6}\mathbb{E}\sum_{i=1}^h\sum_{k=1}^d\sum_{m=1}^n\sum_{q=1}^d\sum_{s=1}^h\sum_{t=1}^h\sum_{b=1}^d\sum_{p=1}^n\sum_{c=1}^d \sum_{p=1}^n\sum_{m=1}^n \\
& {\mM_{ki}}^2{{\mW_1^0}_{qs}}^2{{\mW_2^0}_{st}}^2{\mM_{bt}}^2\mX_{mk}\mX_{mq}\mX_{pb}\mX_{pc}\mX_{pc}\mX_{pb}\mX_{mq}\mX_{mk}\\
    =&\frac{\eta_1^2\eta_2^2}{n^4h^6}\mathbb{E}\sum_{i=1}^h\sum_{k=1}^d\sum_{m=1}^n\sum_{q=1}^d\sum_{s=1}^h\sum_{t=1}^h\sum_{b=1}^d\sum_{p=1}^n\sum_{c=1}^d\\
& {\mM_{ki}}^2{{\mW_1^0}_{qs}}^2{{\mW_2^0}_{st}}^2{\mM_{bt}}^2\mX_{mk}^2\mX_{mq}^2\mX_{pb}^2\mX_{pc}^2\\
=&\frac{\eta_1^2\eta_2^2}{n^4h^8} \times  \frac{1}{d} \times  \frac{1}{d} \times  \frac{1}{d} \times \frac{1}{h} \times h \times (d^2-d) \times (n^2-n)\times (d^2-d)  \times (h^2-h) \\
=& \frac{\eta_1^2\eta_2^2d}{n^2h^5}+O(\frac{\eta_1^2\eta_2^2}{h^8}).
\end{align*}

\textbf{Case 2. $b=b, k=k, i\neq t,  k= q, b= c, m\neq m,  p \neq p , k\neq b,  $.}

Similar to Case 1 , we have 
\begin{align*}
    =&\frac{\eta_1^2\eta_2^2}{n^4h^6}\mathbb{E}\sum_{i=1}^h\sum_{k=1}^d\sum_{m=1}^n\sum_{q=1}^d\sum_{s=1}^h\sum_{t=1}^h\sum_{b=1}^d\sum_{p=1}^n\sum_{c=1}^d \sum_{p=1}^n\sum_{m=1}^n \\
& {\mM_{ki}}^2{{\mW_1^0}_{qs}}^2{{\mW_2^0}_{st}}^2{\mM_{bt}}^2\mX_{mk}\mX_{mq}\mX_{pb}\mX_{pc}\mX_{pc}\mX_{pb}\mX_{mq}\mX_{mk}\\
  =&\frac{\eta_1^2\eta_2^2}{n^4h^6}\mathbb{E}\sum_{i=1}^h\sum_{k=1}^d\sum_{m=1}^n\sum_{s=1}^h\sum_{t=1}^h\sum_{b=1}^d\sum_{p=1}^n\sum_{m=1}^n\sum_{p=1}^n\\
& {\mM_{ki}}^2{{\mW_1^0}_{qs}}^2{{\mW_2^0}_{st}}^2{\mM_{bt}}^2\mX_{mk}^2\mX_{mk}^2\mX_{pb}^2\mX_{pb}^2\\
  =&\frac{\eta_1^2\eta_2^2}{n^4h^8} \times  \frac{1}{d} \times  \frac{1}{d} \times  \frac{1}{d} \times \frac{1}{h} \times h \times (n^2-n) \times (n^2-n)\times (d^2-d)  \times (h^2-h) \\
=& \frac{\eta_1^2\eta_2^2}{dh^6}+O(\frac{\eta_1^2\eta_2^2}{h^8}).
\end{align*}
\hfill $\square$

\textbf{Case 3. $b=b, k=k, i\neq t, k=q,p=p,m\neq p, m\neq m, b\neq c  $.}
\begin{align*}
    =&\frac{\eta_1^2\eta_2^2}{n^4h^6}\mathbb{E}\sum_{i=1}^h\sum_{k=1}^d\sum_{m=1}^n\sum_{q=1}^d\sum_{s=1}^h\sum_{t=1}^h\sum_{b=1}^d\sum_{p=1}^n\sum_{c=1}^d \sum_{p=1}^n\sum_{m=1}^n \\
& {\mM_{ki}}^2{{\mW_1^0}_{qs}}^2{{\mW_2^0}_{st}}^2{\mM_{bt}}^2\mX_{mk}\mX_{mq}\mX_{pb}\mX_{pc}\mX_{pc}\mX_{pb}\mX_{mq}\mX_{mk}\\
    =&\frac{\eta_1^2\eta_2^2}{n^4h^6}\mathbb{E}\sum_{i=1}^h\sum_{k=1}^d\sum_{m=1}^n\sum_{m=1}^n\sum_{s=1}^h\sum_{t=1}^h\sum_{b=1}^d\sum_{p=1}^n\sum_{c=1}^d\\
& {\mM_{ki}}^2{{\mW_1^0}_{qs}}^2{{\mW_2^0}_{st}}^2{\mM_{bt}}^2\mX_{mk}^2\mX_{mq}^2\mX_{pb}^2\mX_{pc}^2\\
 =&\frac{\eta_1^2\eta_2^2}{n^4h^8} \times  \frac{1}{d} \times  \frac{1}{d} \times  \frac{1}{d} \times \frac{1}{h} \times h \times (n^2-n) \times (n^2-n)\times (d^2-d)  \times (h^2-h) \\
=& \frac{\eta_1^2\eta_2^2}{dh^6}+O(\frac{\eta_1^2\eta_2^2}{h^8}).
\end{align*}

\textbf{Case 4. $b=b, k= k, m=m, i\neq t, , b= c, k\neq q,  p \neq p , k\neq b,  $.}

Similar to Case 1 , we have 
\begin{align*}
    =&\frac{\eta_1^2\eta_2^2}{n^4h^6}\mathbb{E}\sum_{i=1}^h\sum_{k=1}^d\sum_{m=1}^n\sum_{q=1}^d\sum_{s=1}^h\sum_{t=1}^h\sum_{b=1}^d\sum_{p=1}^n\sum_{c=1}^d \sum_{p=1}^n\sum_{m=1}^n \\
& {\mM_{ki}}^2{{\mW_1^0}_{qs}}^2{{\mW_2^0}_{st}}^2{\mM_{bt}}^2\mX_{mk}\mX_{mq}\mX_{pb}\mX_{pc}\mX_{pc}\mX_{pb}\mX_{mq}\mX_{mk}\\
  =&\frac{\eta_1^2\eta_2^2}{n^4h^6}\mathbb{E}\sum_{i=1}^h\sum_{k=1}^d\sum_{m=1}^n\sum_{s=1}^h\sum_{t=1}^h\sum_{b=1}^d\sum_{p=1}^n\sum_{q=1}^d\sum_{p=1}^n\\
& {\mM_{ki}}^2{{\mW_1^0}_{qs}}^2{{\mW_2^0}_{st}}^2{\mM_{bt}}^2\mX_{mk}^2\mX_{mq}^2\mX_{pb}^2\mX_{pb}^2\\
 =&\frac{\eta_1^2\eta_2^2}{n^4h^8} \times  \frac{1}{d} \times  \frac{1}{d} \times  \frac{1}{d} \times \frac{1}{h} \times h \times (d^2-d) \times (n^2-n)\times (d^2-d)  \times (h^2-h) \\
=& \frac{\eta_1^2\eta_2^2d}{n^2h^5}+O(\frac{\eta_1^2\eta_2^2}{h^8}).
\end{align*}
\hfill $\square$

For $\frac{\eta_1^2\eta_2^2}{n^4h^6}tr\left( \mathbb{E} \left [{\vxi}^{\top}\mX{\mW_1^0}{\mW_2^0}{\vxi}^{\top}\mX\mX^{\top}{\vxi}{\mW_2^0}^{\top}{\mW_1^0}^{\top}\mX^{\top}{\vxi}
\right] \right)$, we have 
\begin{align*}
    &\frac{\eta_1^2\eta_2^2}{n^4h^6}tr\left( \mathbb{E} \left [{\vxi}^{\top}\mX{\mW_1^0}{\mW_2^0}{\vxi}^{\top}\mX\mX^{\top}{\vxi}{\mW_2^0}^{\top}{\mW_1^0}^{\top}\mX^{\top}{\vxi}
\right] \right)\\
=&\frac{\eta_1^2\eta_2^2}{n^4h^3}\mathbb{E}\sum_{i=1}^h\sum_{k=1}^n\sum_{q=1}^d\sum_{s=1}^h\sum_{t=1}^h\sum_{b=1}^n\sum_{c=1}^d \sum_{b=1}^n\sum_{t=1}^h\sum_{s=1}^h\sum_{q=1}^d\sum_{k=1}^n \\
&{\vxi_{ki}}\mX_{kq}{\mW_1^0}_{qs}{\mW_2^0}_{st}{\vxi_{bt}}\mX_{bc}\mX_{bc}{\vxi_{bt}}{\mW_2^0}_{st}{\mW_1^0}_{qs}\mX_{kq}{\vxi_{ki}}\\
=&\frac{\eta_1^2\eta_2^2}{n^4h^6}\mathbb{E}\sum_{i=1}^h\sum_{k=1}^n\sum_{q=1}^d\sum_{s=1}^h\sum_{t=1}^h\sum_{b=1}^n\sum_{c=1}^d \sum_{b=1}^n\sum_{k=1}^n{\vxi_{ki}}\mX_{kq}{\mW_1^0}_{qs}^2{\mW_2^0}_{st}^2{\vxi_{bt}}\mX_{bc}\mX_{bc}{\vxi_{bt}}\mX_{kq}{\vxi_{ki}}\\
\end{align*}

\textbf{We focus only on the case dominated by the leading term.} Since other cases will  be $O(\frac{\eta_1^2\eta_2^2}{h^8})$.

\textbf{Case 1. $b=b, k=k, i\neq t, k\neq b $.}
\begin{align*}
    =&\frac{\eta_1^2\eta_2^2}{n^4h^6}\mathbb{E}\sum_{i=1}^h\sum_{k=1}^n\sum_{q=1}^d\sum_{s=1}^h\sum_{t=1}^h\sum_{b=1}^n\sum_{c=1}^d {\vxi_{ki}}^2\mX_{kq}^2{\mW_1^0}_{qs}^2{\mW_2^0}_{st}^2{\vxi_{bt}}^2\mX_{bc}^2\\
    =&\frac{\eta_1^2\eta_2^2}{n^4h^8}\times \rho_e^2\times \rho_e^2\times \frac{1}{d}\times \frac{1}{h} \times (h^2-h) \times (n^2-n) \times d \times d \times h \\
    =&\frac{\eta_1^2\eta_2^2d\rho_e^4}{n^2h^6}+O(\frac{\eta_1^2\eta_2^2}{h^8}).
\end{align*}
\hfill $\square$

For $\frac{2\eta_1^2\eta_2^2}{n^4h^6}tr\left( \mathbb{E} \left [{\vxi}^{\top}\mX{\mW_1^0}{\mW_2^0}{\vxi}^{\top}\mX\mX^{\top}\mX\mM{\mW_2^0}^{\top}{\mW_1^0}^{\top}\mX^{\top}\mX\mM
\right] \right)$, we have 
\begin{align*}
    &\frac{2\eta_1^2\eta_2^2}{n^4h^6}tr\left( \mathbb{E} \left [{\vxi}^{\top}\mX{\mW_1^0}{\mW_2^0}{\vxi}^{\top}\mX\mX^{\top}\mX\mM{\mW_2^0}^{\top}{\mW_1^0}^{\top}\mX^{\top}\mX\mM
\right] \right) \\
=&\frac{\eta_1^2\eta_2^2}{n^4h^6}\mathbb{E}\sum_{i=1}^h\sum_{k=1}^n\sum_{q=1}^d\sum_{s=1}^h\sum_{t=1}^h\sum_{b=1}^n\sum_{c=1}^d \sum_{b=1}^d\sum_{t=1}^h\sum_{s=1}^h\sum_{q=1}^d\sum_{k=1}^d\sum_{p=1}^n \sum_{m=1}^n\\
&{\vxi_{ki}}\mX_{kq}{\mW_1^0}_{qs}{\mW_2^0}_{st}{\vxi_{bt}}\mX_{bc}\mX_{pc}\mX_{pb}\mM_{bt}{\mW_2^0}_{st}{\mW_1^0}_{qs}\mX_{mq}\mX_{mk}\mM_{ki}\\
=&\frac{\eta_1^2\eta_2^2}{n^4h^6}\mathbb{E}\sum_{i=1}^h\sum_{k=1}^n\sum_{q=1}^d\sum_{s=1}^h\sum_{b=1}^n\sum_{c=1}^d \sum_{p=1}^n \sum_{m=1}^n\sum_{k=1}^d \\&{\vxi_{ki}}^2\mX_{kq}{\mW_1^0}_{qs}^2{\mW_2^0}_{si}^2\mX_{kc}\mX_{pc}\mX_{pk}\mM_{ki}^2\mX_{mq}\mX_{mk}
\end{align*}

\textbf{We focus only on the case dominated by the leading term.}

\textbf{Case 1. $q=c=k$, $k\neq p \neq m. $ }
\begin{align*}
    &\frac{\eta_1^2\eta_2^2}{n^4h^6}\mathbb{E}\sum_{i=1}^h\sum_{k=1}^n\sum_{q=1}^d\sum_{s=1}^h\sum_{b=1}^n\sum_{p=1}^n \sum_{m=1}^n{\vxi_{ki}}^2\mX_{kq}^2{\mW_1^0}_{qs}^2{\mW_2^0}_{si}^2\mX_{pq}^2\mM_{ki}^2\mX_{mq}^2\\
    &=\frac{\eta_1^2\eta_2^2}{n^4h^8} \times \rho_e^2 \times \frac{1}{d} \times \frac{1}{h} \times \frac{1}{d} \times h \times h \times d \times (n^3-n) \\
    &=O(\frac{\eta_1^2\eta_2^2}{h^8}).
\end{align*}
\hfill $\square$

For $\frac{\eta_1^2\eta_2^2}{n^4h^6}tr\left( \mathbb{E} \left [\mM^{\top}\mX^{\top}\mX{\mW_1^0}{\mW_2^0}{\vxi}^{\top}\mX\mX^{\top}{\vxi}{\mW_2^0}^{\top}{\mW_1^0}^{\top}\mX^{\top}\mX\mM
\right] \right)$, we have 
\begin{align*}
    &\frac{\eta_1^2\eta_2^2}{n^4h^6}tr\left( \mathbb{E} \left [\mM^{\top}\mX^{\top}\mX{\mW_1^0}{\mW_2^0}{\vxi}^{\top}\mX\mX^{\top}{\vxi}{\mW_2^0}^{\top}{\mW_1^0}^{\top}\mX^{\top}\mX\mM
\right] \right) \\
&=\frac{\eta_1^2\eta_2^2}{n^4h^5d}tr\left( \mathbb{E} \left [\mX^{\top}\mX{\mW_1^0}{\mW_2^0}{\vxi}^{\top}\mX\mX^{\top}{\vxi}{\mW_2^0}^{\top}{\mW_1^0}^{\top}\mX^{\top}\mX
\right] \right) \\
&=\frac{\eta_1^2\eta_2^2}{n^4h^5d} \mathbb{E} \sum_{i=1}^d \sum_{m=1}^n  \sum_{q=1}^d \sum_{s=1}^h  \sum_{t=1}^h  \sum_{b=1}^d\sum_{c=1}^d \sum_{b=1}^d\sum_{t=1}^h \sum_{q=1}^d\sum_{s=1}^h  \sum_{m=1}^n   \\
& \mX_{mi}\mX_{mq}{\mW_1^0}_{qs}{\mW_2^0}_{st}{\vxi}_{bt}\mX_{bc}\mX_{bc}{\vxi}_{bt}{\mW_2^0}_{st}{\mW_1^0}_{qs}\mX_{mq}\mX_{mi}\\
&=\frac{\eta_1^2\eta_2^2}{n^4h^5d} \mathbb{E} \sum_{i=1}^d \sum_{m=1}^n  \sum_{q=1}^d \sum_{s=1}^h  \sum_{t=1}^h  \sum_{b=1}^d\sum_{c=1}^d \  \sum_{m=1}^n  \mX_{mi}\mX_{mq}{\mW_1^0}_{qs}^2{\mW_2^0}_{st}^2{\vxi}_{bt}^2 \mX_{bc}^2\mX_{mq}\mX_{mi}
\end{align*}

\textbf{We focus only on the case dominated by the leading term.}

\textbf{Case 1. $ q=i, m\neq m $. }

\begin{align*}
    &\frac{\eta_1^2\eta_2^2}{n^4h^5d} \mathbb{E} \sum_{i=1}^d \sum_{m=1}^n  \sum_{q=1}^d \sum_{s=1}^h  \sum_{t=1}^h  \sum_{b=1}^d\sum_{c=1}^d \  \sum_{m=1}^n  \mX_{mi}\mX_{mq}{\mW_1^0}_{qs}^2{\mW_2^0}_{st}^2{\vxi}_{bt}^2 \mX_{bc}^2\mX_{mq}\mX_{mi}\\
    &=\frac{\eta_1^2\eta_2^2}{n^4h^5d} \mathbb{E} \sum_{i=1}^d \sum_{m=1}^n   \sum_{s=1}^h  \sum_{t=1}^h  \sum_{b=1}^d\sum_{c=1}^d \  \sum_{m=1}^n  \mX_{mi}^2{\mW_1^0}_{qs}^2{\mW_2^0}_{st}^2{\vxi}_{bt}^2 \mX_{bc}^2\mX_{mi}^2\\
    &=\frac{\eta_1^2\eta_2^2}{n^4h^7d}\times \rho_e^2  \times \frac{1}{d} \times \frac{1}{h} \times (n^2-n)\times d\times h\times h \times d\times d \\
    &=\frac{\eta_1^2\eta_2^2d\rho_e^2}{n^2h^6}+O(\frac{\eta_1^2\eta_2^2}{h^8}).
\end{align*}

\textbf{Case 2. $  m= m, q\neq i $. }

\begin{align*}
    &\frac{\eta_1^2\eta_2^2}{n^4h^5d} \mathbb{E} \sum_{i=1}^d \sum_{m=1}^n  \sum_{q=1}^d \sum_{s=1}^h  \sum_{t=1}^h  \sum_{b=1}^d\sum_{c=1}^d \  \sum_{m=1}^n  \mX_{mi}\mX_{mq}{\mW_1^0}_{qs}^2{\mW_2^0}_{st}^2{\vxi}_{bt}^2 \mX_{bc}^2\mX_{mq}\mX_{mi}\\
    &=\frac{\eta_1^2\eta_2^2}{n^4h^7d} \mathbb{E} \sum_{i=1}^d \sum_{m=1}^n   \sum_{s=1}^h  \sum_{t=1}^h  \sum_{b=1}^d\sum_{c=1}^d \  \sum_{q=1}^d \mX_{mi}^2{\mW_1^0}_{qs}^2{\mW_2^0}_{st}^2{\vxi}_{bt}^2 \mX_{bc}^2\mX_{mq}^2\\
    &=\frac{\eta_1^2\eta_2^2}{n^4h^2d}\times \rho_e^2  \times \frac{1}{d} \times \frac{1}{h} \times (d^2-d)\times n\times h\times h \times d\times d \\
    &=\frac{\eta_1^2\eta_2^2d^2\rho_e^2}{n^3h^6}+O(\frac{\eta_1^2\eta_2^2}{h^8})
\end{align*}

It is easy to see 
\begin{align*}
    &\frac{\eta_1^2\eta_2^2}{n^4h^6}tr\left( \mathbb{E} \left [\mM^{\top}\mX^{\top}\mX{\mW_1^0}{\mW_2^0}{\vxi}^{\top}\mX\mX^{\top}{\vxi}{\mW_2^0}^{\top}{\mW_1^0}^{\top}\mX^{\top}\mX\mM
\right] \right)\\
=&\frac{\eta_1^2\eta_2^2}{n^4h^6}tr\left( \mathbb{E} \left [{\vxi}^{\top}\mX{\mW_1^0}{\mW_2^0}\mM^{\top}\mX^{\top}\mX\mX^{\top}\mX\mM{\mW_2^0}^{\top}{\mW_1^0}^{\top}\mX^{\top}{\vxi}\right]\right)\\
=&\frac{\eta_1^2\eta_2^2d\rho_e^2}{n^2h^6}+\frac{\eta_1^2\eta_2^2d^2\rho_e^2}{n^3h^6}+O(\frac{\eta_1^2\eta_2^2}{h^8})
\end{align*}

Finally we get 
\begin{equation}
    T_{16}=\frac{\eta_1^2\eta_2^2d}{n^2h^6}+\frac{\eta_1^2\eta_2^2}{dh^6}+\frac{\eta_1^2\eta_2^2d\rho_e^4}{n^2h^6}+\frac{2\eta_1^2\eta_2^2d\rho_e^2}{n^2h^6}+\frac{2\eta_1^2\eta_2^2d^2\rho_e^2}{n^3h^6}+O(\frac{\eta_1^2\eta_2^2}{h^8})
\end{equation}
\hfill $\square$

\paragraph{Analysis of $T_2$, $T_3$, $T_5$, $T_8$, $T_9$, $T_{12}$, $T_{14}$, $T_{15}$, $T_{17}$, $T_{20}$, $T_{21}$ and $T_{24}$.}
All terms involve the product of an odd number of identical random matrices with zero mean, and due to their independence from other random matrices, these terms are all $0$. \hfill $\square$

\paragraph{Analysis of $T_{18}$.}
\begin{equation}
    \begin{aligned}
        T_{18}&=tr\left( \mathbb{E}_{\mW_1^0,\mW_2^0,\vxi, \mX} \left [ \frac{\eta_1}{nh^2}\mM{\mW_2^0}^{\top}\mW_2^0 \mY^{\top}\mX 
\right] \right)\\
&=\frac{\eta_1}{nh^2}tr\left( \mathbb{E}_{\mW_1^0,\mW_2^0,\vxi, \mX} \left [ {\mW_2^0}^{\top}\mW_2^0 \mY^{\top}\mX \mM
\right] \right)\\
&=\frac{\eta_1}{nh^2}tr\left( \mathbb{E}_{\mW_1^0,\mW_2^0,\vxi, \mX} \left [  \mM^{\top}\mX^{\top}\mX \mM
\right] \right) \\
&=\frac{\eta_1}{h^2}.
    \end{aligned}
\end{equation}
\hfill $\square$

Similar to $T_{18}$, we can get 
\begin{align}
     T_{19}&=\frac{\eta_2}{dh} \\
     T_{22}&=\frac{\eta_1}{h^2} \\
     T_{23}&=\frac{\eta_2}{dh}.
\end{align}
\hfill $\square$
   
Finally, we obtain the exact loss 
  \begin{equation}
        \begin{aligned}
            L_{\text{two-layer}}&=\frac{2\eta_1^2}{h^4}+\frac{2\eta_1^2d(1+\rho_e^2)}{nh^4}-2\frac{\eta_1}{h^2} \\&+ \frac{\eta_2^2}{d^2h^2}+\frac{\eta_2^2}{dh^3}+\frac{\eta_2^2(1+\rho_e^2)}{ndh^2}+\frac{\eta_2^2(1+\rho_e^2)}{nh^3} -2\frac{\eta_2}{dh} +\frac{\eta_1^2\eta_2^2d}{n^2h^6}\\
        &+  \frac{2\eta_1\eta_2}{dh^3}+\frac{2\eta_1\eta_2}{nh^3}+\frac{2\eta_1\eta_2\rho_e^2}{nh^3}+\frac{\eta_1^2\eta_2^2d(\rho_e^2+1)^2}{n^2h^6}+\frac{2\eta_1^2\eta_2^2}{dh^6}+\frac{2\eta_1^2\eta_2^2d^2\rho_e^2}{n^3h^6}\\
        &+1+\frac{1}{h}+O(\frac{\eta_1^2}{h^5})+O(\frac{\eta_2^5}{h^5})+O(\frac{\eta_1\eta_2}{h^2})+O(\frac{\eta_1^2\eta_2^2}{h^8})
        \end{aligned}
    \end{equation}

Under Assumption~\ref{sec:assumption-2}, we have 
    \begin{equation}
    \label{eq:L_two_layer}
        \begin{aligned}
            L_{\text{two-layer}}&=\frac{2\eta_1^2}{h^4}+\frac{2\eta_1^2(1+\rho_e^2)}{nh^3}-2\frac{\eta_1}{h^2} \\&+ \frac{2\eta_2^2}{h^4}+\frac{2\eta_2^2(1+\rho_e^2)}{nh^3}-2\frac{\eta_2}{h^2}+\frac{\eta_1^2\eta_2^2}{n^2h^5}\\
        &+  \frac{2\eta_1\eta_2}{h^4}+\frac{2\eta_1\eta_2}{nh^3}+\frac{2\eta_1\eta_2\rho_e^2}{nh^3}+\frac{\eta_1^2\eta_2^2(\rho_e^2+1)^2}{n^2h^5}+\frac{2\eta_1^2\eta_2^2}{h^7}+\frac{2\eta_1^2\eta_2^2\rho_e^2}{n^3h^4}\\
        &+1+\frac{1}{h}+O(\frac{\eta_1^2}{h^5})+O(\frac{\eta_2^5}{h^2})+O(\frac{\eta_1\eta_2}{h^5})+O(\frac{\eta_1^2\eta_2^2}{h^8})
        \end{aligned}
    \end{equation}
\hfill $\square$

\subsection{Three-layer NN Test Loss under Gaussian initialization}
\label{app:3-layer NN test loss}

\begin{theorem}
    Given Assumption~\ref{sec:assumption},~\ref{sec:assumption-2},  and in addition assume $\eta_1$ and $\eta_2$ are no more than $O({h})$ based on Proposition~\ref{sec:proposition-3-layer}, consider training procedure discussed in section~\ref{sec:setup},  we derive the test loss after  one-step GD update in a three-layer neural network:
 \begin{equation}
        \begin{aligned}
L_{\text{three-layer}}&=\frac{\eta_1^2}{h^2}+\frac{\eta_1^2(1+\rho_e^2)}{hn}-2\frac{\eta_1}{h} +\frac{2\eta_2^2}{h^2}+\frac{2\eta_2^2(1+\rho_e^2)}{nh}-2\frac{\eta_2}{h}\\
     &+\frac{2\eta_1\eta_2}{h^2}+\frac{2\eta_1\eta_2(1+\rho_e^2)}{nh}+\frac{\eta_1^2\eta_2^2\rho_e^2}{nh^3}+\frac{\eta_1^2\eta_2^2d\rho_e^2}{n^2h^3}+\frac{4\eta_1^2\eta_2^2}{n^2h^2}\\
     & +1 +O(\frac{\eta_1^2}{h^3})+O(\frac{\eta_2^2}{h^3})+O(\frac{\eta_1\eta_2}{h^3})+O\left(\frac{\eta_1^2\eta_2^2}{h^5}\right)
        \end{aligned}
    \end{equation}
\end{theorem}

\paragraph{Proof of Theorem~\ref{theorem for 3-layer}.} Similar to appendix ~\ref{app:2-layer Test Loss}, we consider a test data $\Tilde{\vx}_0 \sim \gN(\vzero, \mI_d)\in \sR ^{1\times d}$.
\begin{equation}
\begin{aligned}
    &L(\mX,\mW_1^1,\mW_2^1,\va,\Tilde{\vx}_0)\\=&\mathbb{E}_{\mW_1^0,\mW_2^0,\va,\vxi,\Tilde{\vx}_0,\mX} \left( \frac{1}{\sqrt{h}}\Tilde{\vx}_0\mW_1^1\mW_2^1\va-\Tilde{\vx}_0\vbeta^*\right)^2  \\
    =&\mathbb{E}_{\mW_1^0,\mW_2^0,\va,\vxi,\Tilde{\vx}_0,\mX} \left [\left( \frac{1}{\sqrt{h}}\mW_1^1\mW_2^1\va-\vbeta^*\right)^{\top}{\Tilde{\vx}_0}^{\top}\Tilde{\vx}_0 \left( \frac{1}{\sqrt{h}}\mW_1^1\mW_2^1\va-\vbeta^*\right)\right] \\
    =& tr\left( \mathbb{E}_{\mW_1^0,\mW_2^0,\va,\vxi,\Tilde{\vx}_0,\mX} \left [{\Tilde{\vx}_0}^{\top}\Tilde{\vx}_0 \left( \frac{1}{\sqrt{h}}\mW_1^1\mW_2^1\va-\vbeta^*\right)\left( \frac{1}{\sqrt{h}}\mW_1^1\mW_2^1\va-\vbeta^*\right)^{\top}\right] \right) \\
    =&tr\left( \mathbb{E}_{\mW_1^0,\mW_2^0,\va,\vxi, \mX} \left [\left( \frac{1}{\sqrt{h}}\mW_1^1\mW_2^1\va-\vbeta^*\right)\left( \frac{1}{\sqrt{h}}\mW_1^1\mW_2^1\va-\vbeta^*\right)^{\top}\right] \right) \\
    =& tr\left( \mathbb{E}_{\mW_1^0,\mW_2^0,\va,\vxi, \mX} \left [ \frac{1}{h}\mW_1^1\mW_2^1\va\va^{\top}{\mW_2^1}^{\top}{\mW_1^1}^{\top}\right] \right) \\
    -&tr\left( \mathbb{E}_{\mW_1^0,\mW_2^0,\va,\vxi, \mX} \left [ \frac{1}{\sqrt{h}}\vbeta^*\va^{\top}{\mW_2^1}^{\top}{\mW_1^1}^{\top}\right] \right) \\
    -&tr\left( \mathbb{E}_{\mW_1^0,\mW_2^0,\va,\vxi, \mX}\left [ \frac{1}{\sqrt{h}}{\mW_1^1}{\mW_2^1}\va{\vbeta^*}^{\top}\right] \right) + tr\left( \mathbb{E} \left [{\vbeta^*} {\vbeta^*}^{\top}\right] \right).
\end{aligned}
\end{equation}
Here we define $L_1,L_2,L_3, L_4$, where
\begin{align*}
   L_1&=tr\left( \mathbb{E}_{\mW_1^0,\mW_2^0,\va,\vxi, \mX} \left [ \frac{1}{h}\mW_1^1\mW_2^1\va\va^{\top}{\mW_2^1}^{\top}{\mW_1^1}^{\top}\right] \right)\\
    L_2&=tr\left( \mathbb{E}_{\mW_1^0,\mW_2^0,\va,\vxi, \mX} \left [ \frac{1}{\sqrt{h}}\vbeta^*\va^{\top}{\mW_2^1}^{\top}{\mW_1^1}^{\top}\right] \right) \\
    L_3 &=tr\left( \mathbb{E}_{\mW_1^0,\mW_2^0,\va,\vxi, \mX} \left [ \frac{1}{\sqrt{h}}{\mW_1^1}{\mW_2^1}\va{\vbeta^*}^{\top}\right] \right)\\
    L_4 &= tr\left( \mathbb{E} \left [{\vbeta^*} {\vbeta^*}^{\top}\right] \right)
\end{align*}
Thus $$L_{\text{three-layer} }=L_1-L_2-L_3+L_4$$

Due to Proposition~\ref{sec:proposition-3-layer} and Appendix~\ref{app:norm property of 3-layer NN}, we know that the norm of $G_1$ is dominated by $A_1$, the norm of $G_2$ is dominated by $A_2$, to simplify the gradient, we consider the following approximation.

\begin{equation}
\label{app:approx_weight_1}
    \mW_1^1\approx \mW_1^0+\frac{\eta_1}{n\sqrt{h}}\mX^{\top}\vy\va^{\top}{\mW_2^0}^{\top}.
\end{equation}

\begin{equation}
\label{app:approx_weight_2}
    \mW_2^1\approx \mW_2^0+\frac{\eta_2}{n\sqrt{h}}{\mW_1^0}^{\top}\mX^{\top}\vy\va^{\top}.
\end{equation}

Thus we have
\begin{equation}
\begin{aligned}
\mW_1^1\mW_2^1\approx & \mW_1^0\mW_2^0+\frac{\eta_1}{n\sqrt{h}}\mX^{\top}\vy\va^{\top}{\mW_2^0}^{\top}\mW_2^0 \\
&+\frac{\eta_2}{n\sqrt{h}} \mW_1^0{\mW_1^0}^{\top}\mX^{\top}\vy\va^{\top}\\
& +\frac{\eta_1\eta_2}{n^2h}\mX^{\top}\vy\va^{\top}{\mW_2^0}^{\top}{\mW_1^0}^{\top}\mX^{\top}\vy\va^{\top},
\end{aligned}
\end{equation}

\begin{equation}
\begin{aligned}
{\mW_2^1}^{\top}{\mW_1^1}^{\top}\approx & {\mW_2^0}^{\top}{\mW_1^0}^{\top}+\frac{\eta_1}{n\sqrt{h}}{\mW_2^0}^{\top}\mW_2^0 \va\vy^{\top}\mX \\
&+\frac{\eta_2}{n\sqrt{h}} \va\vy^{\top}\mX\mW_1^0{\mW_1^0}^{\top}\\
& +\frac{\eta_1\eta_2}{n^2h}\va\vy^{\top}\mX{\mW_1^0}{\mW_2^0}\va\vy^{\top}\mX,
\end{aligned}
\end{equation}
we take (~\ref{app:approx_weight_1}) and (~\ref{app:approx_weight_2}) into $L_1,L_2,L_3$.

We have $L_1=\sum_{i=1}^{16}T_i$, where
\begin{align*}
T_1&=tr\left( \mathbb{E}_{\mW_1^0,\mW_2^0,\va,\vxi, \mX} \left [ \frac{1}{h}\va\va^{\top}{\mW_2^0}^{\top}{\mW_1^0}^{\top}{\mW_1^0}{\mW_2^0}
\right] \right),\\
T_2&=tr\left( \mathbb{E}_{\mW_1^0,\mW_2^0,\va,\vxi, \mX} \left [ \frac{\eta_1}{nh\sqrt{h}}\va\va^{\top}{\mW_2^0}^{\top}{\mW_1^0}^{\top}\mX^{\top}\vy\va^{\top}{\mW_2^0}^{\top}\mW_2^0
\right] \right),\\
T_3&=tr\left( \mathbb{E}_{\mW_1^0,\mW_2^0,\va,\vxi, \mX} \left [ \frac{\eta_2}{nh\sqrt{h}}\va\va^{\top}{\mW_2^0}^{\top}{\mW_1^0}^{\top}\mW_1^0{\mW_1^0}^{\top}\mX^{\top}\vy\va^{\top}
\right] \right),\\
T_4&=tr\left( \mathbb{E}_{\mW_1^0,\mW_2^0,\va,\vxi, \mX} \left [ \frac{\eta_1\eta_2}{n^2h^2}\va\va^{\top}{\mW_2^0}^{\top}{\mW_1^0}^{\top}\mX^{\top}\vy\va^{\top}{\mW_2^0}^{\top}{\mW_1^0}^{\top}\mX^{\top}\vy\va^{\top}
\right] \right),\\
T_5&=tr\left( \mathbb{E}_{\mW_1^0,\mW_2^0,\va,\vxi, \mX} \left [ \frac{\eta_1}{nh\sqrt{h}}\va\va^{\top}{\mW_2^0}^{\top}\mW_2^0 \va\vy^{\top}\mX\mW_1^0\mW_2^0
\right] \right),\\
T_6&=tr\left( \mathbb{E}_{\mW_1^0,\mW_2^0,\va,\vxi, \mX} \left [ \frac{\eta_1^2}{n^2h^2}\va\va^{\top}{\mW_2^0}^{\top}\mW_2^0 \va\vy^{\top}\mX\mX^{\top}\vy\va^{\top}{\mW_2^0}^{\top}\mW_2^0 
\right] \right),\\
T_7&=tr\left( \mathbb{E}_{\mW_1^0,\mW_2^0,\va,\vxi, \mX} \left [ \frac{\eta_1\eta_2}{n^2h^2}\va\va^{\top}{\mW_2^0}^{\top}\mW_2^0 \va\vy^{\top}\mX\mW_1^0{\mW_1^0}^{\top}\mX^{\top}\vy\va^{\top}
\right] \right),\\
T_8&=tr\left( \mathbb{E}_{\mW_1^0,\mW_2^0,\va,\vxi, \mX} \left [ \frac{\eta_1^2\eta_2}{n^3h^2\sqrt{h}}\va\va^{\top}{\mW_2^0}^{\top}\mW_2^0 \va\vy^{\top}\mX\mX^{\top}\vy\va^{\top}{\mW_2^0}^{\top}{\mW_1^0}^{\top}\mX^{\top}\vy\va^{\top}
\right] \right),\\
T_9&=tr\left( \mathbb{E}_{\mW_1^0,\mW_2^0,\va,\vxi, \mX} \left [ \frac{\eta_2}{nh\sqrt{h}}\va\va^{\top}\va\vy^{\top}\mX\mW_1^0{\mW_1^0}^{\top}\mW_1^0\mW_2^0
\right] \right),\\
T_{10}&=tr\left( \mathbb{E}_{\mW_1^0,\mW_2^0,\va,\vxi, \mX} \left [ \frac{\eta_1\eta_2}{n^2h^2}\va\va^{\top}\va\vy^{\top}\mX\mW_1^0{\mW_1^0}^{\top}\mX^{\top}\vy\va^{\top}{\mW_2^0}^{\top}\mW_2^0 
\right] \right),\\
T_{11}&=tr\left( \mathbb{E}_{\mW_1^0,\mW_2^0,\va,\vxi, \mX} \left [ \frac{\eta_2^2}{n^2h^2}\va\va^{\top}\va\vy^{\top}\mX\mW_1^0{\mW_1^0}^{\top}\mW_1^0{\mW_1^0}^{\top}\mX^{\top}\vy\va^{\top}
\right] \right),\\
T_{12}&=tr\left( \mathbb{E}_{\mW_1^0,\mW_2^0,\va,\vxi, \mX} \left [ \frac{\eta_1\eta_2^2}{n^3h^2\sqrt{h}}\va\va^{\top}\va\vy^{\top}\mX\mW_1^0{\mW_1^0}^{\top}\mX^{\top}\vy\va^{\top}{\mW_2^0}^{\top}{\mW_1^0}^{\top}\mX^{\top}\vy\va^{\top}
\right] \right),\\
T_{13}&=tr\left( \mathbb{E}_{\mW_1^0,\mW_2^0,\va,\vxi, \mX} \left [ \frac{\eta_1\eta_2}{n^2h^2}\va\va^{\top}\va\vy^{\top}\mX{\mW_1^0}{\mW_2^0}\va\vy^{\top}\mX\mW_1^0\mW_2^0
\right] \right),\\
T_{14}&=tr\left( \mathbb{E}_{\mW_1^0,\mW_2^0,\va,\vxi, \mX} \left [ \frac{\eta_1^2\eta_2}{n^3h^2\sqrt{h}}\va\va^{\top}\va\vy^{\top}\mX{\mW_1^0}{\mW_2^0}\va\vy^{\top}\mX\mX^{\top}\vy\va^{\top}{\mW_2^0}^{\top}\mW_2^0 
\right] \right),\\
T_{15}&=tr\left( \mathbb{E}_{\mW_1^0,\mW_2^0,\va,\vxi, \mX} \left [ \frac{\eta_1\eta_2^2}{n^3h^2\sqrt{h}}\va\va^{\top}\va\vy^{\top}\mX{\mW_1^0}{\mW_2^0}\va\vy^{\top}\mX\mW_1^0{\mW_1^0}^{\top}\mX^{\top}\vy\va^{\top}
\right] \right),\\
T_{16}&=tr\left( \mathbb{E}_{\mW_1^0,\mW_2^0,\va,\vxi, \mX} \left [ \frac{\eta_1^2\eta_2^2}{n^4h^3}\va\va^{\top}\va\vy^{\top}\mX{\mW_1^0}{\mW_2^0}\va\vy^{\top}\mX\mX^{\top}\vy\va^{\top}{\mW_2^0}^{\top}{\mW_1^0}^{\top}\mX^{\top}\vy\va^{\top}
\right] \right).\\
\end{align*}
We have $L_2=\sum_{i=17}^{20} T_{i}$, where
\begin{align*}
T_{17}&=tr\left( \mathbb{E}_{\mW_1^0,\mW_2^0,\va,\vxi, \mX} \left [ \frac{1}{\sqrt{h}}\vbeta^{*}\va^{\top}{\mW_2^0}^{\top}{\mW_1^0}^{\top}
\right] \right),\\
T_{18}&=tr\left( \mathbb{E}_{\mW_1^0,\mW_2^0,\va,\vxi, \mX} \left [ \frac{\eta_1}{nh}\vbeta^{*}\va^{\top}{\mW_2^0}^{\top}\mW_2^0 \va\vy^{\top}\mX
\right] \right),\\
T_{19}&=tr\left( \mathbb{E}_{\mW_1^0,\mW_2^0,\va,\vxi, \mX} \left [ \frac{\eta_2}{nh}\vbeta^{*}\va^{\top}\va\vy^{\top}\mX\mW_1^0{\mW_1^0}^{\top}
\right] \right),\\
T_{20}&=tr\left( \mathbb{E}_{\mW_1^0,\mW_2^0,\va,\vxi, \mX} \left [ \frac{\eta_1\eta_2}{n^2h\sqrt{h}}\vbeta^{*}\va^{\top}\va\vy^{\top}\mX{\mW_1^0}{\mW_2^0}\va\vy^{\top}\mX
\right] \right).\\
\end{align*}

We have $L_3=\sum_{i=21}^{24} T_{i}$, where
\begin{align*}
    T_{21}&=tr\left( \mathbb{E}_{\mW_1^0,\mW_2^0,\va,\vxi, \mX} \left [ \frac{1}{\sqrt{h}}\mW_1^0\mW_2^0\va\vbeta^{*^{\top}}
\right] \right),\\
T_{22}&=tr\left( \mathbb{E}_{\mW_1^0,\mW_2^0,\va,\vxi, \mX} \left [ \frac{\eta_1}{nh}\mX^{\top}\vy\va^{\top}{\mW_2^0}^{\top}\mW_2^0\va\vbeta^{*^{\top}}
\right] \right),\\
T_{23}&=tr\left( \mathbb{E}_{\mW_1^0,\mW_2^0,\va,\vxi, \mX} \left [ \frac{\eta_2}{nh}\mW_1^0{\mW_1^0}^{\top}\mX^{\top}\vy\va^{\top}\va\vbeta^{*^{\top}}
\right] \right),\\
T_{24}&=tr\left( \mathbb{E}_{\mW_1^0,\mW_2^0,\va,\vxi, \mX} \left [ \frac{\eta_1\eta_2}{n^2h\sqrt{h}}\mX^{\top}\vy\va^{\top}{\mW_2^0}^{\top}{\mW_1^0}^{\top}\mX^{\top}\vy\va^{\top}\va\vbeta^{*^{\top}}
\right] \right),\\
\end{align*}

Thus, we obtain that $$L=\sum_{i=1}^{16} T_{i}-\sum_{i=17}^{20} T_{i}-\sum_{i=21}^{24} T_{i}+L_4$$

\paragraph{Analysis of  $T_1$.}
\begin{equation}
    \begin{aligned}
        T_1&=tr\left( \mathbb{E}_{\mW_1^0,\mW_2^0,\va,\vxi, \mX} \left [ \frac{1}{h}\va\va^{\top}{\mW_2^0}^{\top}{\mW_1^0}^{\top}{\mW_1^0}{\mW_2^0}\right)\right]=\frac{1}{h}
    \end{aligned}
\label{app:T1}
\end{equation}
\hfill $\square$

\paragraph{Analysis of  $T_4$ and $T_{13}$.} 
For $T_4$, we have 
\begin{equation}
    \begin{aligned}
 T_4&=tr\left( \mathbb{E}_{\mW_1^0,\mW_2^0,\va,\vxi, \mX} \left [ \frac{\eta_1\eta_2}{n^2h^2}\va\va^{\top}{\mW_2^0}^{\top}{\mW_1^0}^{\top}\mX^{\top}\vy\va^{\top}{\mW_2^0}^{\top}{\mW_1^0}^{\top}\mX^{\top}\vy\va^{\top}
\right] \right)\\
&=\frac{\eta_1\eta_2}{n^2h^2}\mathbb{E}tr\left(  \left [ \va\va^{\top}{\mW_2^0}^{\top}{\mW_1^0}^{\top}\mX^{\top}\vy\vy^{\top}\mX{\mW_1^0}{\mW_2^0}\va\va^{\top}
\right] \right)\\
&=\frac{\eta_1\eta_2}{n^2h^2}tr\left( \mathbb{E} \left [ \va\va^{\top}\va\va^{\top}\right]\mathbb{E}\left [{\mW_2^0}^{\top}{\mW_1^0}^{\top}\mX^{\top}\vy\vy^{\top}\mX{\mW_1^0}{\mW_2^0}
\right] \right)\\
\end{aligned}
\end{equation}
For $\mathbb{E} \left [ \va\va^{\top}\va\va^{\top}\right]$, similar to (~\ref{app:four square of x}), we have
\begin{equation}
    \begin{aligned}
\mathbb{E}\left(\va\va^{\top}\va\va^{\top}\right)&= \left(\frac{1}{h}+\frac{2}{h^2}\right)\mI_h\\
\end{aligned}
\label{app:four square of a_transpose}
\end{equation}
By taking (~\ref{app:four square of a_transpose}) in to $T_4$, we have 
\begin{equation}
    \begin{aligned}
        T_4&=\frac{\eta_1\eta_2}{n^2h^2}\left(\frac{1}{h}+\frac{2}{h^2}\right)tr\left( \mathbb{E}\left [{\mW_2^0}^{\top}{\mW_1^0}^{\top}\mX^{\top}\vy\vy^{\top}\mX{\mW_1^0}{\mW_2^0}
\right] \right)\\
&=\frac{\eta_1\eta_2}{n^2h^2}\left(\frac{1}{h}+\frac{2}{h^2}\right)tr\left( \mathbb{E}\left [{\mW_2^0}{\mW_2^0}^{\top}{\mW_1^0}^{\top}\mX^{\top}\vy\vy^{\top}\mX{\mW_1^0}\right]\right)\\
&=\frac{\eta_1\eta_2}{n^2h^2}\left(\frac{1}{h}+\frac{2}{h^2}\right)tr\left( \mathbb{E}\left [{\mW_1^0}^{\top}\mX^{\top}\vy\vy^{\top}\mX{\mW_1^0}\right]\right)\\
&=\frac{\eta_1\eta_2}{n^2h^2}\left(\frac{1}{h}+\frac{2}{h^2}\right)tr\left( \mathbb{E}\left [{\mW_1^0}{\mW_1^0}^{\top}\mX^{\top}\vy\vy^{\top}\mX\right]\right)\\
&=\frac{\eta_1\eta_2}{n^2hd}\left(\frac{1}{h}+\frac{2}{h^2}\right)tr\left( \mathbb{E}\left [\mX^{\top}\vy\vy^{\top}\mX\right]\right)\\
&=\frac{\eta_1\eta_2}{n^2hd}\left(\frac{1}{h}+\frac{2}{h^2}\right)\left(n^2+nd(1+\rho_e^2)+n\right)\\
&=O\left(\frac{\eta_1\eta_2}{h^3}\right).
\end{aligned}
\label{app:T_4}
\end{equation}
  \hfill $\square$

It is easy to find that $T_4=T_{13}$, so we have 
\begin{equation}
    T_{13}=O\left(\frac{\eta_1\eta_2}{h^3}\right).
\label{app:T_13}
\end{equation}
  \hfill $\square$

\paragraph{Analysis of  $T_6$ and $T_{11}$.} For $T_6$, we have
\begin{align*}
    T_6&=\frac{\eta_1^2}{n^2h^2} tr\left( \mathbb{E}_{\mW_1^0,\mW_2^0,\va,\vxi, \mX} \left [ \va\va^{\top}{\mW_2^0}^{\top}\mW_2^0 \va\vy^{\top}\mX\mX^{\top}\vy\va^{\top}{\mW_2^0}^{\top}\mW_2^0 
\right] \right)\\
    &=\frac{\eta_1^2}{n^2h^2} tr\left(\mathbb{E}_{\mW_1^0,\mW_2^0,\va,\vxi, \mX} \left [ \va^{\top}{\mW_2^0}^{\top}\mW_2^0\va\va^{\top}{\mW_2^0}^{\top}\mW_2^0 \va\vy^{\top}\mX\mX^{\top}\vy\right] \right)\\
    &= \frac{\eta_1^2}{n^2h^2} \mathbb{E}_{\mW_1^0,\mW_2^0,\va,\vxi, \mX} \left [ \va^{\top}{\mW_2^0}^{\top}\mW_2^0\va\va^{\top}{\mW_2^0}^{\top}\mW_2^0 \va\vy^{\top}\mX\mX^{\top}\vy\right] 
\end{align*}
let $\mS_2={\mW_2^0}^{\top}\mW_2^{0}$. By replacing them into $T_6$,
\begin{align*}
    T_6&=\frac{\eta_1^2}{n^2h^2} \mathbb{E}_{\mW_1^0,\mW_2^0,\va,\vxi, \mX} \left [ \va^{\top}{\mW_2^0}^{\top}\mW_2^0\va\va^{\top}{\mW_2^0}^{\top}\mW_2^0 \va\vy^{\top}\mX\mX^{\top}\vy\right] \\
    &=\frac{\eta_1^2}{n^2h^2} \mathbb{E}\left [ \va^{\top}\mS_{2}\va\va^{\top}\mS_{2} \va\right]  \mathbb{E}\left[\vy^{\top}\mX\mX^{\top}\vy\right]. 
\end{align*}
Here we first analyze $ \mathbb{E}\left [ \va^{\top}\mS_{2}\va\va^{\top}\mS_{2} \va\right]=\mathbb{E}\left [ \left(\sum_{i,j} a_{i}S_{2_{ij}}a_{j}\right) \left(\sum_{p,q} a_{p}S_{2_{pq}}a_{q}\right)\right] $, This can be reduced to the following three cases, since the expectations in all other cases are zero.

\textbf{Case 1 $(i,j)=(p,q), i \neq j$.}
\begin{equation}
\begin{aligned}
    \mathbb{E}\left [ \va^{\top}\mS_{2}\va\va^{\top}\mS_{2} \va\right]_1&=\mathbb{E}\left [ \left(\sum_{\substack{i,j \\ i \neq j}} a_{i}S_{2_{ij}}a_{j}\right)^2\right]\\
    &= \sum_{\substack{i,j \\ i \neq j}} \mathbb{E}a_i^2 \mathbb{E}S_{2_{ij}}^2 \mathbb{E}a_j^2\\
    &=(h^2-h)\times\frac{1}{h}\times\frac{1}{h}\times\frac{1}{h}\times h\times\frac{1}{h}\\
    &=\frac{1}{h}-\frac{1}{h^2}  
\end{aligned} 
\label{app:T6_case1}
\end{equation}

\textbf{Case 2 $(i,j)=(q,p), i \neq j$.}  Same to Case 1,
\begin{equation}
    \mathbb{E}\left [ \va^{\top}\mS_{2}\va\va^{\top}\mS_{2} \va\right]_2=\frac{1}{h}-\frac{1}{h^2}  
\label{app:T6_case2}
\end{equation}

\textbf{Case 3 $i=j, p=q$.} 
\begin{equation}
\begin{aligned}
     \mathbb{E}\left [ \va^{\top}\mS_{2}\va\va^{\top}\mS_{2}\va\right]_3&=\mathbb{E}\left[\left(\sum_{i}S_{2_{ii}}a_i^2\right)\left(\sum_{p}S_{2_{pp}}a_p^2\right)\right]\\
     &=\sum_{i,p}\left[\mathbb{E}\left(S_{2_{ii}}S_{2_{pp}}\right) \mathbb{E}\left(a_i^2a_p^2\right)\right]\\
&=\sum_{i=p}\left[\mathbb{E}\left(S_{2_{ii}}^2\right) \mathbb{E}\left(a_i^4\right)\right]+\sum_{i\neq p}\left[\mathbb{E}\left(S_{2_{ii}}S_{2_{pp}}\right) \mathbb{E}\left(a_i^2a_p^2\right)\right].
\end{aligned}
\end{equation}
For $\mathbb{E}S_{2_{ii}}^2$, we have 
\begin{equation}
\begin{aligned}
    \mathbb{E}S_{2_{ii}}^2&=\mathbb{E}S_{2_{11}}^2\\
    &=\mathbb{E}\left(W_{2_{11}}^2+W_{2_{21}}^2+\cdots+W_{2_{h1}}^2\right)^2\\
    &=h\times\frac{3}{h^2}+(h^2-h)\times\frac{1}{h}\times\frac{1}{h}\\
    &=1+\frac{2}{h}.
\end{aligned}
\label{app:square of S_2_ii}
\end{equation}
It is east to see 
\begin{equation}
    \mathbb{E}a_i^4=\frac{3}{h^2}
\label{app:4-moment of a}
\end{equation}
\begin{equation}
    \mathbb{E}_{i\neq p}\left(S_{2_{ii}}S_{2_{pp}}\right)=\mathbb{E}S_{2_{ii}}\mathbb{E}S_{2_{pp}}=1
\label{app:SiiSpp}
\end{equation}
   \begin{equation}
    \mathbb{E}_{i\neq p}\left(a_{ii}^2a_{pp}^2\right)=\mathbb{E}a_{{ii}}^2\mathbb{E}a_{{pp}}^2=\frac{1}{h^2}
\label{app:square of aiap}
\end{equation}

by combining (~\ref{app:square of S_2_ii}) to (~\ref{app:square of aiap}), we have  
\begin{equation}
\begin{aligned}
        \mathbb{E}\left [ \va^{\top}\mS_{2}\va\va^{\top}\mS_{2}\va\right]_3&=h\times(1+\frac{1}{h^2})\times \frac{3}{h^2}+(h^2-h)\times \frac{1}{h^2}\\
        &=1+\frac{2}{h}+\frac{6}{h^2}.
\end{aligned}
\label{app:T6_case3}
\end{equation}
By combining (~\ref{app:T6_case1}),(~\ref{app:T6_case2})and(~\ref{app:T6_case3}), we have \begin{equation}
    \mathbb{E}\left [ \va^{\top}\mS_{2}\va\va^{\top}\mS_{2}\va\right]=1+\frac{4}{h}+\frac{4}{h^2}.
\label{app:part of T6}
\end{equation}
 \hfill $\square$

 We then analyze $\mathbb{E}\left[\vy^{\top}\mX\mX^{\top}\vy\right]$. note that $\vy=\mX\vbeta^{*}+\vxi$, we have 
 \begin{equation}
     \begin{aligned}
         \mathbb{E}\left[\vy^{\top}\mX\mX^{\top}\vy\right]&=\mathbb{E}\left[\left(\vbeta^{*^{\top}}\mX^{\top}\mX+\vxi^{\top}\mX\right)\left(\mX^{\top}\mX\vbeta^{*}+\mX^{\top}\vxi\right)\right]\\ 
&=\mathbb{E}\left(\vbeta^{*^{\top}}\mX^{\top}\mX\mX^{\top}\mX\vbeta^{*}\right)+\mathbb{E}\left(\vxi^{\top}\mX\mX^{\top}\vxi\right) \\
&=tr\left[\mathbb{E}\left(\mX^{\top}\mX\mX^{\top}\mX\right)\mathbb{E}\left(\vbeta^{*}\vbeta^{*^{\top}}\right)\right]+ tr\left[\mathbb{E}\left(\mX\mX^{\top}\right)\mathbb{E}\left(\vxi\vxi^{\top}\right)\right].
\end{aligned}
\label{app: part_of_T6_decom}
 \end{equation}
For $\mathbb{E}\left(\mX^{\top}\mX\mX^{\top}\mX\right)$, we show that \begin{equation}
    \begin{aligned}
    \mathbb{E}\left(\mX^{\top}\mX\right)^2_{ii}&=\mathbb{E}\left(\mX^{\top}\mX\right)^2_{11}\\
    &=\mathbb{E}\left(\sum_{m=1}^nX^2_{m1}\right)^2+\mathbb{E}\left[\sum_{j=2}^{d}\left(\sum_{k=1}^n X_{k1}X_{kj}\right)^2\right]\\
    &=3n+(n^2-n)+(d-1)n\\
    &=n^2+nd+n\\
    \mathbb{E}\left(\mX^{\top}\mX\right)^2_{ij}&=\mathbb{E}\left[\left(\sum_{m=1}^n X_{m1}^2\right)\left(\sum_{k=1}^n X_{k1}X_{k2}\right)\right]\\
    &+\mathbb{E}\left[\left(\sum_{m=1}^n X_{m2}^2\right)\left(\sum_{k=1}^n X_{k2}X_{k1}\right)\right]\\
    &+ \sum_{j=3}^n\mathbb{E}\left[\left(\sum_{k=1}^n X_{k1}X_{kj}\right)\left(\sum_{p=1}^n X_{pj}X_{p2}\right)\right]\\
    &=0.
    \end{aligned}
\label{app:four square of x}
\end{equation}
Which means $\mathbb{E}\left(\mX^{\top}\mX\mX^{\top}\mX\right)= (n^2+nd+n)\mI_d$.

It is easy to see $\mathbb{E}\left(\vbeta^{*}\vbeta^{*^{\top}}\right)=\frac{1}{d}\mI_d$, $\mathbb{E}\left(\mX\mX^{\top}\right)=n\mI_n$, $\mathbb{E}\left(\vxi\vxi^{\top}\right)=\rho_{e}^2\mI_n$. Thus, consider (~\ref{app: part_of_T6_decom}) we have 
\begin{equation}
\mathbb{E}\left[\vy^{\top}\mX\mX^{\top}\vy\right]=n^2+nd(1+\rho_e^2)+n
\label{app:part of T6_2}.
\end{equation}
 \hfill $\square$

 By (~\ref{app:part of T6}) and (~\ref{app:part of T6_2}), we finally get 
 \begin{equation}
     \begin{aligned}
         T_6&=\frac{\eta_1^2\left(n^2+nd(1+\rho_e^2)+n\right)\left(1+\frac{4
         }{h}+\frac{4}{h^4}\right)}{n^2h^2}\\
         &= \frac{\eta_1^2}{h^2}+\frac{\eta_1^2d(1+\rho_e^2)}{h^2n}+O(\frac{\eta_1^2}{h^3})
     \end{aligned}
\label{app:T_6}
 \end{equation}
  \hfill $\square$

For $T_{11}$, similar to $T_6$, let $\mW_1^0{\mW_1^0}^{\top}=\mS_1$, we have
\begin{align*}
    T_{11}&=tr\left( \mathbb{E}_{\mW_1^0,\mW_2^0,\va,\vxi, \mX} \left [ \frac{\eta_2^2}{n^2h^2}\va\va^{\top}\va\vy^{\top}\mX\mW_1^0{\mW_1^0}^{\top}\mW_1^0{\mW_1^0}^{\top}\mX^{\top}\vy\va^{\top}
\right] \right)\\
&=tr\left( \mathbb{E}_{\mW_1^0,\mW_2^0,\va,\vxi, \mX} \left [ \frac{\eta_2^2}{n^2h^2}\va^{\top}\va\va^{\top}\va\vy^{\top}\mX\mW_1^0{\mW_1^0}^{\top}\mW_1^0{\mW_1^0}^{\top}\mX^{\top}\vy
\right] \right)\\
&=\frac{\eta_2^2}{n^2h^2}tr\left(\mathbb{E}\left(\va^{\top}\va\va^{\top}\va \right)\mathbb{E}\left( \vy^{\top}\mX\mW_1^0{\mW_1^0}^{\top}\mW_1^0{\mW_1^0}^{\top}\mX^{\top}\vy\right)\right)\\
&=\frac{\eta_2^2}{n^2h^2}tr\left(\mathbb{E}\left(\va^{\top}\va\va^{\top}\va \right)\mathbb{E}\left( \vy^{\top}\mX\mS_1^2\mX^{\top}\vy\right)\right)
\end{align*}

For $\mathbb{E}\left(\va^{\top}\va\va^{\top}\va \right)$, we have 

\begin{equation}
\begin{aligned}
\mathbb{E}\left(\va^{\top}\va\va^{\top}\va \right)&=   \mathbb{E}\left(\sum_{i=1}^{h}a_i^2 \right)^2\\
&=h\mathbb{E}\left(a_1^4\right)+(h^2-h)\mathbb{E}\left(a_1^2a_2^2\right)\\
&=h\times \frac{3}{h^2}+(h^2-h)\times\frac{1}{h}\times\frac{1}{h}\\
&=1+\frac{2}{h}
\end{aligned}
\label{app:four square of a}
\end{equation}

Take (~\ref{app:four square of a}) into $T_6$, we have \begin{equation}
\begin{aligned}
      T_{11}&=\frac{\eta_2^2}{n^2h^2}\left(1+\frac{2}{h} \right)tr\mathbb{E}\left[\left( \vy^{\top}\mX\mS_1^2\mX^{\top}\vy\right)\right]\\
      &= \frac{\eta_2^2}{n^2h^2}\left(1+\frac{2}{h} \right)tr\mathbb{E}\left[\left( \mX^{\top}\vy\vy^{\top}\mX\mS_1^2\right)\right]\\
      & =\frac{\eta_2^2}{n^2h^2}\left(1+\frac{2}{h} \right)tr\mathbb{E}\left[\left( \mX^{\top}\vy\vy^{\top}\mX\right)\mathbb{E}\left( \mS_1^2\right)\right]\\.
\end{aligned}
\end{equation}
Similar to compute $\mathbb{E}\left(\mX^{\top}\mX\mX^{\top}\mX\right)$, we have 
\begin{equation}
    \mathbb{E}\left( \mS_1^2\right)=\frac{h^2+h+hd}{d^2}\mI_d.
\label{app:four square of W_1}
\end{equation}
By taking (~\ref{app:four square of W_1}) into $T_{11}$
\begin{equation}
    \begin{aligned}
         T_{11}&=\frac{\eta_2^2}{n^2h^2}\left(1+\frac{2}{h}\right)\frac{h^2+h+hd}{d^2}tr\mathbb{E}\left[\left( \mX^{\top}\vy\vy^{\top}\mX\right)\right].
    \end{aligned}
\end{equation}
By (~\ref{app:part of T6_2}), we have 
\begin{equation}
\begin{aligned}
     T_{11}&=\frac{\eta_2^2}{n^2h^2}\left(1+\frac{2}{h}\right)\left(\frac{h^2+h+hd}{d^2}\right)\left(n^2+nd(1+\rho_e^2)+n\right)\\
     &=\frac{\eta_2^2}{d^2}+\frac{\eta_2^2}{hd}+\frac{\eta_2^2(1+\rho_e^2)}{nh}+\frac{\eta_2^2(1+\rho_e^2)}{nd}+O(\frac{\eta_2^2}{h^3}).
\end{aligned}
\label{app:T_11}
\end{equation}
  \hfill $\square$

\paragraph{Analysis of  $T_7$ and $T_{10}$.} We have

\begin{equation}
    \begin{aligned}
      T_7&=tr\left( \mathbb{E}_{\mW_1^0,\mW_2^0,\va,\vxi, \mX} \left [ \frac{\eta_1\eta_2}{n^2h^2}\va\va^{\top}{\mW_2^0}^{\top}\mW_2^0 \va\vy^{\top}\mX\mW_1^0{\mW_1^0}^{\top}\mX^{\top}\vy\va^{\top}
\right] \right)\\
&=\frac{\eta_1\eta_2}{n^2h^2}\mathbb{E}tr\left( \left [ {\mW_2^0}^{\top}\mW_2^0 \va\vy^{\top}\mX\mW_1^0{\mW_1^0}^{\top}\mX^{\top}\vy\va^{\top}\va\va^{\top}\right] \right)\\
&=\frac{\eta_1\eta_2}{n^2h^2}\mathbb{E}tr\left( \left [  \va\vy^{\top}\mX\mW_1^0{\mW_1^0}^{\top}\mX^{\top}\vy\va^{\top}\va\va^{\top}\right] \right)\\
&=\frac{\eta_1\eta_2}{n^2h^2}\mathbb{E}tr\left( \left [  \mW_1^0{\mW_1^0}^{\top}\mX^{\top}\vy\va^{\top}\va\va^{\top}\va\vy^{\top}\mX\right] \right)\\
&=\frac{\eta_1\eta_2}{n^2hd}\mathbb{E}tr\left( \left [ \mX^{\top}\vy\va^{\top}\va\va^{\top}\va\vy^{\top}\mX\right] \right)\\
&=\frac{\eta_1\eta_2}{n^2hd}\mathbb{E}tr\left( \left [ \vy^{\top}\mX\mX^{\top}\vy\va^{\top}\va\va^{\top}\va\right] \right)\\
&=\frac{\eta_1\eta_2}{n^2hd} \mathbb{E}\left [ \vy^{\top}\mX\mX^{\top}\vy\right]\mathbb{E}\left [ \va^{\top}\va\va^{\top}\va\right].\\
\end{aligned}
\end{equation}

By (~\ref{app:part of T6_2}) and (~\ref{app:four square of a}), we show 
\begin{equation}
\begin{aligned}
        T_{7}&=\frac{\eta_1\eta_2}{n^2hd}\left(n^2+nd(1+\rho_e^2)+n \right)\left(1+\frac{2}{h}\right)\\
        &=\frac{\eta_1\eta_2}{hd}+\frac{\eta_1\eta_2(1+\rho_e^2)}{nh}+O(\frac{\eta_1\eta_2}{h^3}).
\end{aligned}
\label{app:T_7}
\end{equation}
  \hfill $\square$

It is easy to find that $T_7=T_{10}$, so we have 
\begin{equation}
    T_{10}=\frac{\eta_1\eta_2}{nh}+\frac{\eta_1\eta_2(1+\rho_e^2)}{nh}+O(\frac{\eta_1\eta_2}{h^3}).
\label{app:T_10}
\end{equation}
  \hfill $\square$
\paragraph{Analysis of  $T_{16}$.} 
\begin{equation}
    \begin{aligned}
       T_{16}&=\frac{\eta_1^2\eta_2^2}{n^4h^3}tr\left( \mathbb{E}_{\mW_1^0,\mW_2^0,\va,\vxi, \mX} \left [ \va\va^{\top}\va\vy^{\top}\mX{\mW_1^0}{\mW_2^0}\va\vy^{\top}\mX\mX^{\top}\vy\va^{\top}{\mW_2^0}^{\top}{\mW_1^0}^{\top}\mX^{\top}\vy\va^{\top}
\right] \right) \\
&=\frac{\eta_1^2\eta_2^2}{n^4h^3}tr\left( \mathbb{E}\left [ \va\va^{\top}\va \left[\vy^{\top}\mX{\mW_1^0}{\mW_2^0}\va\right]^{\top}\vy^{\top}\mX\mX^{\top}\vy\left[\va^{\top}{\mW_2^0}^{\top}{\mW_1^0}^{\top}\mX^{\top}\vy\right]^{\top}\va^{\top}
\right] \right)\\
&=\frac{\eta_1^2\eta_2^2}{n^4h^3}tr\left( \mathbb{E}\left [ \va\va^{\top}\va \va^{\top}{\mW_2^0}^{\top}{\mW_1^0}^{\top}\mX^{\top}\vy\vy^{\top}\mX\mX^{\top}\vy\vy^{\top}\mX{\mW_1^0}{\mW_2^0}\va\va^{\top}
\right] \right)\\
&=\frac{\eta_1^2\eta_2^2}{n^4h^3}tr\left( \mathbb{E}\left [ \va\va^{\top}\va \va^{\top}\va\va^{\top}{\mW_2^0}^{\top}{\mW_1^0}^{\top}\mX^{\top}\vy\vy^{\top}\mX\mX^{\top}\vy\vy^{\top}\mX{\mW_1^0}{\mW_2^0}
\right] \right)\\
&=\frac{\eta_1^2\eta_2^2}{n^4h^3}tr\left( \mathbb{E}\left [ \va\va^{\top}\va \va^{\top}\va\va^{\top}\right]\mathbb{E}\left [  {\mW_2^0}^{\top}{\mW_1^0}^{\top}\mX^{\top}\vy\vy^{\top}\mX\mX^{\top}\vy\vy^{\top}\mX{\mW_1^0}{\mW_2^0}
\right] \right)
    \end{aligned}
\end{equation}

For $\mathbb{E}\left [ \va\va^{\top}\va \va^{\top}\va\va^{\top}\right]$, we have 
\begin{equation}
    \begin{aligned}
       \mathbb{E}\left [ \va\va^{\top}\va \va^{\top}\va\va^{\top}\right]_{11}&= \mathbb{E}\left [\left(\sum_{i=1}^ha_i^2\right)^2a_{1}^2\right] =\frac{1}{h}+O(\frac{1}{h^2}),\\
\mathbb{E}\left [ \va\va^{\top}\va \va^{\top}\va\va^{\top}\right]_{11}&= \mathbb{E}\left [\left(\sum_{i=1}^ha_i^2\right)^2a_{1}a_2\right] =0.
\end{aligned}
\label{app:six square of a}
\end{equation}
Which means $\mathbb{E}\left [ \va\va^{\top}\va \va^{\top}\va\va^{\top}\right]=\left(\frac{1}{h}+O(\frac{1}{h^2})\right)\mI_h$, taking it into $T_{16}$
\begin{equation}
    \begin{aligned}
        T_{16}&=\frac{\eta_1^2\eta_2^2}{n^4h^3}\left(\frac{1}{h}+O(\frac{1}{h^2})\right)tr\mathbb{E}\left [  {\mW_2^0}^{\top}{\mW_1^0}^{\top}\mX^{\top}\vy\vy^{\top}\mX\mX^{\top}\vy\vy^{\top}\mX{\mW_1^0}{\mW_2^0}\right]\\
        &=\frac{\eta_1^2\eta_2^2}{n^4h^3}\left(\frac{1}{h}+O(\frac{1}{h^2})\right)tr\mathbb{E}\left [  {\mW_2^0}{\mW_2^0}^{\top}{\mW_1^0}^{\top}\mX^{\top}\vy\vy^{\top}\mX\mX^{\top}\vy\vy^{\top}\mX{\mW_1^0}\right]\\
         &=\frac{\eta_1^2\eta_2^2}{n^4h^3}\left(\frac{1}{h}+O(\frac{1}{h^2})\right)tr\mathbb{E}\left [ {\mW_1^0}^{\top}\mX^{\top}\vy\vy^{\top}\mX\mX^{\top}\vy\vy^{\top}\mX{\mW_1^0}\right]\\
         &=\frac{\eta_1^2\eta_2^2}{n^4h^3}\left(\frac{1}{h}+O(\frac{1}{h^2})\right)tr\mathbb{E}\left [{\mW_1^0} {\mW_1^0}^{\top}\mX^{\top}\vy\vy^{\top}\mX\mX^{\top}\vy\vy^{\top}\mX\right]\\
        &=\frac{\eta_1^2\eta_2^2}{n^4h^2d}\left(\frac{1}{h}+O(\frac{1}{h^2})\right)tr\mathbb{E}\left [\mX^{\top}\vy\vy^{\top}\mX\mX^{\top}\vy\vy^{\top}\mX\right]\\
    \end{aligned}
\end{equation}
For $tr\mathbb{E}\left [\mX^{\top}\vy\vy^{\top}\mX\mX^{\top}\vy\vy^{\top}\mX\right]$, we have 
\begin{align*}
    tr\mathbb{E}\left [\mX^{\top}\vy\vy^{\top}\mX\mX^{\top}\vy\vy^{\top}\mX\right]&=tr\mathbb{E}\left [\mX^{\top}\mX\vbeta^{*}\vbeta^{*^{\top}}\mX^{\top}\mX \mX^{\top}\vxi\vxi^{\top}\mX\right] \\ 
    &+tr\mathbb{E}\left [\mX^{\top}\vxi\vxi^{\top}\mX\mX^{\top}\mX\vbeta^{*}\vbeta^{*^{\top}}\mX^{\top}\mX \right] \\
    &+tr\mathbb{E}\left [\mX^{\top}\mX\vbeta^{*}\vxi^{\top}\mX\mX^{\top}\mX\vbeta^{*}\vxi^{\top}\mX\right]\\
    &+tr\mathbb{E}\left [\mX^{\top}\vxi\vbeta^{*^{\top}}\mX^{\top}\mX\mX^{\top}\vxi\vbeta^{*^{\top}}\mX^{\top}\mX\right]\\
    &+tr\mathbb{E}\left [\mX^{\top}\vxi\vxi^{\top}\mX\mX^{\top}\vxi\vxi^{\top}\mX\right]\\
    &+tr\mathbb{E}\left [\mX^{\top}\mX\vbeta^{*}\vxi^{\top}\mX\mX^{\top}\vxi\vbeta^{*^{\top}}\mX^{\top}\mX\right]\\
    &+tr\mathbb{E}\left [\mX^{\top}\vxi\vbeta^{*^{\top}}\mX^{\top}\mX\mX^{\top}\mX\vbeta^{*}\vxi^{\top}\mX\right]\\
    &+tr\mathbb{E}\left [\mX^{\top}\mX\vbeta^{*}\vbeta^{*^{\top}}\mX^{\top}\mX\mX^{\top}\mX\vbeta^{*}\vbeta^{*^{\top}}\mX^{\top}\mX\right]\\
\end{align*}

It is easy to see 
\begin{align*}
    tr\mathbb{E}\left [\mX^{\top}\mX\vbeta^{*}\vbeta^{*^{\top}}\mX^{\top}\mX \mX^{\top}\vxi\vxi^{\top}\mX\right]&=tr\mathbb{E}\left [\mX^{\top}\vxi\vxi^{\top}\mX\mX^{\top}\mX\vbeta^{*}\vbeta^{*^{\top}}\mX^{\top}\mX \right] \\
    &=tr\mathbb{E}\left [\mX^{\top}\mX\vbeta^{*}\vxi^{\top}\mX\mX^{\top}\mX\vbeta^{*}\vxi^{\top}\mX\right]\\
    &=tr\mathbb{E}\left [\mX^{\top}\vxi\vbeta^{*^{\top}}\mX^{\top}\mX\mX^{\top}\vxi\vbeta^{*^{\top}}\mX^{\top}\mX\right]\\
    &=\frac{\rho_e^2}{d}tr\mathbb{E}\left [\mX\mX^{\top}\mX\mX\mX^{\top}\mX\right]\\
    &=\frac{\rho_e^2}{d} \mathbb{E}\norm{\mX\mX^{\top}\mX}_F^2\\
    &\leq\frac{\rho_e^2}{d}\mathbb{E}\norm{\mX}^2\norm{\mX}^2\norm{\mX}_F^2\\
    &=O(n^3)
\end{align*}
It is also easy to see $tr\mathbb{E}\left [\mX^{\top}\vxi\vxi^{\top}\mX\mX^{\top}\vxi\vxi^{\top}\mX\right]=\mathbb{E}\norm{\mX\mX^{\top}\vxi\vxi^{\top}}_F^2\leq O(n^3)$.

Here we focus on computing $$tr\mathbb{E}\left [\mX^{\top}\mX\vbeta^{*}\vxi^{\top}\mX\mX^{\top}\vxi\vbeta^{*^{\top}}\mX^{\top}\mX\right], tr\mathbb{E}\left [\mX^{\top}\mX\vbeta^{*}\vbeta^{*^{\top}}\mX^{\top}\mX\mX^{\top}\mX\vbeta^{*}\vbeta^{*^{\top}}\mX^{\top}\mX\right]$$  since we have $$tr\mathbb{E}\left [\mX^{\top}\mX\vbeta^{*}\vxi^{\top}\mX\mX^{\top}\vxi\vbeta^{*^{\top}}\mX^{\top}\mX\right]=tr\mathbb{E}\left [\mX^{\top}\vxi\vbeta^{*^{\top}}\mX^{\top}\mX\mX^{\top}\mX\vbeta^{*}\vxi^{\top}\mX\right]$$

For $tr\mathbb{E}\left [\mX^{\top}\mX\vbeta^{*}\vxi^{\top}\mX\mX^{\top}\vxi\vbeta^{*^{\top}}\mX^{\top}\mX\right]$, we have 
\begin{equation}
    \begin{aligned}
        &tr\mathbb{E}\left [\mX^{\top}\mX\vbeta^{*}\vxi^{\top}\mX\mX^{\top}\vxi\vbeta^{*^{\top}}\mX^{\top}\mX\right]\\=& tr\mathbb{E}\left [\vxi^{\top}\mX\mX^{\top}\vxi\mX^{\top}\mX\vbeta^{*}\vbeta^{*^{\top}}\mX^{\top}\mX\right]\\
        =&tr\mathbb{E}\left [\vbeta^{*}\vbeta^{*^{\top}}\mX^{\top}\mX\vxi^{\top}\mX\mX^{\top}\vxi\mX^{\top}\mX\right]\\
        =&\frac{\rho_e^2}{d}\mathbb{E}\left[tr\left(\mX^{\top}\mX\right)tr\left(\mX^{\top}\mX\mX^{\top}\mX\right)\right]\\
=&\frac{\rho_e^2}{d}\mathbb{E}\left[\norm{\mX^{\top}\mX}_F^2\norm{\mX}_F^2\right]\\
=&\frac{\rho_e^2}{d}\mathbb{E}\left[ \left( \sum_{i,j}X_{ij}^2\right) \left(\sum_{p,q}\left(X^{\top}X\right)_{pq}^2\right)\right]\\
=&\frac{\rho_e^2}{d}\mathbb{E}\left[\left( \sum_{i,j}X_{ij}^2\right)\left(\sum_{p=1}^{d}\left(\sum_{m=1}^nX_{mp}^2\right)^2+\sum_{\substack{p,q \\ p \neq q}}\left(\sum_{k=1}^nX_{kp}X_{kq}\right)^2\right)\right]\\
=&n\rho_e^2\mathbb{E}\left[X_{11}^2\left(\sum_{p=1}^{d}\left(\sum_{m=1}^nX_{mp}^2\right)^2+\sum_{\substack{p,q \\ p \neq q}}\left(\sum_{k=1}^nX_{kp}X_{kq}\right)^2\right)\right]\\
    \end{aligned}
\end{equation}
\begin{equation}
    \begin{aligned}
&n\rho_e^2\mathbb{E}\left[X_{11}^2\left(\sum_{p=1}^{d}\left(\sum_{m=1}^nX_{mp}^2\right)^2\right)\right]\\=&nd\times\left[(d-1)(3n+n^2-n)+15 +2\times(n-1)\times3+n^2-2\times(n-1)-1\right]\\
=&n^3d\rho_e^2+O(n^2d)
    \end{aligned}
\label{app:part of T_16_2_1}
\end{equation}
\begin{equation}
    \begin{aligned}
        n\rho_e^2\mathbb{E}\left[X_{11}^2\left(\sum_{\substack{p,q \\ p \neq q}}\left(\sum_{k=1}^nX_{kp}X_{kq}\right)^2\right)\right]&=n\rho_e^2\times\left[(d^2-d-2)n+2\times3n\right]\\
        &=n^2d^2\rho_e^2+O(n^2d)
    \end{aligned}
    \label{app:part of T_16_2_2}
\end{equation}
By (~\ref{app:part of T_16_2_1}) and  (~\ref{app:part of T_16_2_2})
we have \begin{equation}
     tr\mathbb{E}\left [\mX^{\top}\mX\vbeta^{*}\vxi^{\top}\mX\mX^{\top}\vxi\vbeta^{*^{\top}}\mX^{\top}\mX\right]=n^3d\rho_e^2+n^2d^2\rho_e^2+O(n^2d)
    \label{app:part of T_16_2}
\end{equation}

For $tr\mathbb{E}\left [\mX^{\top}\mX\vbeta^{*}\vbeta^{*^{\top}}\mX^{\top}\mX\mX^{\top}\mX\vbeta^{*}\vbeta^{*^{\top}}\mX^{\top}\mX\right]$, we have 
\begin{equation}
    \begin{aligned}
       &tr\mathbb{E}\left [\mX^{\top}\mX\vbeta^{*}\vbeta^{*^{\top}}\mX^{\top}\mX\mX^{\top}\mX\vbeta^{*}\vbeta^{*^{\top}}\mX^{\top}\mX\right]\\=&\mathbb{E}\norm{\mX^{\top}\mX\vbeta^{*}\vbeta^{*^{\top}}\mX^{\top}\mX}_F^2 \\
=&\sum_{i,j}\left(\sum_{k=1}^{n}\sum_{m=1}^{d}\sum_{s=1}^{d}\sum_{q=1}^{n}X_{ki}X_{km}\beta^{*}_{m}\beta^{*^{\top}}_{s}X_{qs}X_{qj}\right)^2.
    \end{aligned}
\end{equation}
\textbf{Case 1} $i\neq j, m=s$. It is easy to see in this case, the main term holds when $3\leq m \leq d$ and $k\neq q $, since other conditions will only have up to $O(n^2d)$. Combining with the condition  $3\leq m \leq d$, we have 
$$tr\mathbb{E}\left [\mX^{\top}\mX\vbeta^{*}\vbeta^{*^{\top}}\mX^{\top}\mX\mX^{\top}\mX\vbeta^{*}\vbeta^{*^{\top}}\mX^{\top}\mX\right]_1=O(n^2d).$$

\textbf{Case 2} $i\neq j, m\neq s.$ It is easy to see in this case, the main term holds when $s\geq 2, m\geq 2$, and other conditions will only have up to $O(n^2d)$. Combining with the condition $s\geq 2, m\geq 2$, we have  $$tr\mathbb{E}\left [\mX^{\top}\mX\vbeta^{*}\vbeta^{*^{\top}}\mX^{\top}\mX\mX^{\top}\mX\vbeta^{*}\vbeta^{*^{\top}}\mX^{\top}\mX\right]_2=4n^2d^2+O(n^2d).$$

\textbf{Case 3} $i=j, m=s.$ In this case, the main term still holds when $3\leq m \leq d$ and $k\neq q $. We have 

$$tr\mathbb{E}\left [\mX^{\top}\mX\vbeta^{*}\vbeta^{*^{\top}}\mX^{\top}\mX\mX^{\top}\mX\vbeta^{*}\vbeta^{*^{\top}}\mX^{\top}\mX\right]_3=O(n^2d).$$

\textbf{Case 4} $i= j, m\neq s.$ In this case, the main term holds when $s\geq 2, m\geq 2$,  we have  $$tr\mathbb{E}\left [\mX^{\top}\mX\vbeta^{*}\vbeta^{*^{\top}}\mX^{\top}\mX\mX^{\top}\mX\vbeta^{*}\vbeta^{*^{\top}}\mX^{\top}\mX\right]_4=O(n^2d).$$

By Case 1 to Case 4, We finally get 
\begin{equation}
tr\mathbb{E}\left [\mX^{\top}\mX\vbeta^{*}\vbeta^{*^{\top}}\mX^{\top}\mX\mX^{\top}\mX\vbeta^{*}\vbeta^{*^{\top}}\mX^{\top}\mX\right]=4n^2d^2+ O(n^2d)
\label{app: part of T_16_3}
\end{equation}

Here, we take (~\ref{app:part of T_16_2}) and(~\ref{app: part of T_16_3}), we finally get 
\begin{equation}
\begin{aligned}
      T_{16}&=\frac{\eta_1^2\eta_2^2}{n^4h^2d}\left(\frac{1}{h}+O(\frac{1}{h^2})\right)\left[(2n^3d+2n^2d^2)\rho_e^2+4n^2d^2+O(n^2d)\right]\\
      &=\frac{\eta_1^2\eta_2^2\rho_e^2}{nh^e}+\frac{\eta_1^2\eta_2^2d\rho_e^2}{n^2h^3}+\frac{4\eta_1^2\eta_2^2d}{n^2h^3}+O\left(\frac{\eta_1^2\eta_2^2}{h^5}\right)
  \end{aligned}
\label{app:T_{16}}
\end{equation}
\hfill $\square$

\paragraph{Analysis of $T_2$, $T_3$, $T_5$, $T_8$, $T_9$, $T_{12}$, $T_{14}$, $T_{15}$, $T_{17}$, $T_{20}$, $T_{21}$ and $T_{24}$.}
All terms involve the product of an odd number of identical random matrices with zero mean, and due to their independence from other random matrices, these terms are all $0$. \hfill $\square$

\paragraph{Analysis of  $T_{18}$, $T_{19}$, $T_{22}$ and $T_{23}$.} We take $T_{18}$ as an example.

\begin{equation}
    \begin{aligned}
        T_{18}&=tr\left( \mathbb{E}_{\mW_1^0,\mW_2^0,\va,\vxi, \mX} \left [ \frac{\eta_1}{nh}\vbeta^{*}\va^{\top}{\mW_2^0}^{\top}\mW_2^0 \va\vy^{\top}\mX
\right] \right)\\
&=\frac{\eta_1}{nh}\mathbb{E}tr\left[ {\mW_2^0}^{\top}\mW_2^0 \va\vy^{\top}\mX\vbeta^{*}\va^{\top}
\right]\\
&=\frac{\eta_1}{nh}\mathbb{E}tr\left[  \va^{\top}\va\vy^{\top}\mX\vbeta^{*}
\right]\\
&=\frac{\eta_1}{nh}\mathbb{E}tr\left[  \vy^{\top}\mX\vbeta^{*}
\right]\\
&=\frac{\eta_1}{nh}\mathbb{E}tr\left[  \vy^{\top}\mX\vbeta^{*}
\right]\\
&= \frac{\eta_1}{nh}\mathbb{E}tr\left[\vbeta^{*^{\top}}  \mX^{\top}\mX\vbeta^{*}
\right]\\
&= \frac{\eta_1}{nh}\mathbb{E}tr\left[\mX^{\top}\mX\vbeta^{*}\vbeta^{*^{\top}}  
\right]\\
&=\frac{\eta_1}{h}
    \end{aligned}
\label{app:T_18}
\end{equation}
Similar to $T_{18}$, we can get 
\begin{align}
    T_{19}&=\frac{\eta_2}{d} \label{app:T_19}\\
     T_{22}&=\frac{\eta_1}{h} \label{app:T_22}\\
      T_{23}&=\frac{\eta_2}{d} \label{app:T_23}
\end{align}
\hfill $\square$

Finally, we get the exact loss\begin{equation}
        \begin{aligned}
L_{\text{three-layer}}&=\frac{\eta_1^2}{h^2}+\frac{\eta_1^2d(1+\rho_e^2)}{h^2n}-2\frac{\eta_1}{h}\\&+\frac{\eta_2^2}{d^2}+\frac{\eta_2^2}{hd}+\frac{\eta_2^2(1+\rho_e^2)}{nh}+\frac{\eta_2^2(1+\rho_e^2)}{nd}-2\frac{\eta_2}{d}\\
     &+\frac{2\eta_1\eta_2}{hd}+\frac{2\eta_1\eta_2(1+\rho_e^2)}{nh}+\frac{\eta_1^2\eta_2^2\rho_e^2}{nh^3}+\frac{\eta_1^2\eta_2^2d\rho_e^2}{n^2h^3}+\frac{4\eta_1^2\eta_2^2d}{n^2h^3}\\
     & +1 +O(\frac{\eta_1^2}{h^3})+O(\frac{\eta_2^2}{h^3})+O(\frac{\eta_1\eta_2}{h^3})+O\left(\frac{\eta_1^2\eta_2^2}{h^5}\right)
        \end{aligned}
\label{app:exact_loss_3_NN}
    \end{equation}

Under Assumption~\ref{sec:assumption-2}, we have
\begin{equation}
        \begin{aligned}
L_{\text{three-layer}}&=\frac{\eta_1^2}{h^2}+\frac{\eta_1^2(1+\rho_e^2)}{hn}-2\frac{\eta_1}{h}\\&+\frac{2\eta_2^2}{h^2}+\frac{2\eta_2^2(1+\rho_e^2)}{nh}-2\frac{\eta_2}{h}\\
     &+\frac{2\eta_1\eta_2}{h^2}+\frac{2\eta_1\eta_2(1+\rho_e^2)}{nh}+\frac{\eta_1^2\eta_2^2\rho_e^2}{nh^3}+\frac{\eta_1^2\eta_2^2\rho_e^2}{n^2h^2}+\frac{4\eta_1^2\eta_2^2}{n^2h^2}\\
     & +1 +O(\frac{\eta_1^2}{h^3})+O(\frac{\eta_2^2}{h^3})+O(\frac{\eta_1\eta_2}{h^3})+O\left(\frac{\eta_1^2\eta_2^2}{h^5}\right)
        \end{aligned}
    \end{equation}
\hfill $\square$

\section{Additional Experiments} \label{app:more_experiments}

\subsection{Spectral Analysis }
\begin{figure*}[!htb]
    \centering
    \begin{subfigure}[t]{0.24\linewidth}
        \centering
        \includegraphics[width=\textwidth]{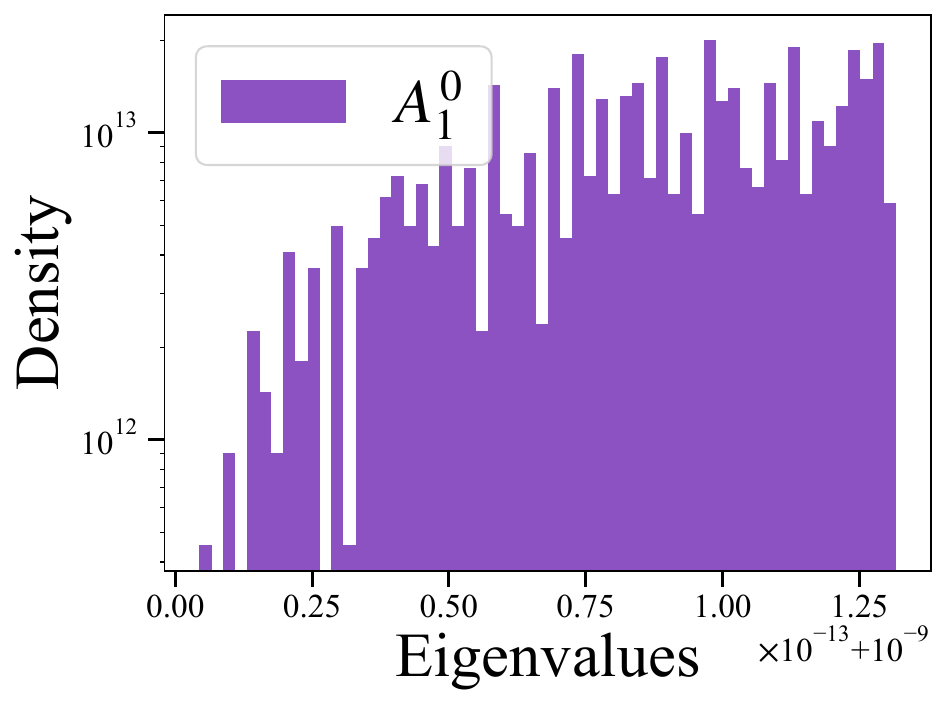}
        \caption{   $\mA_1^0$ }
    \end{subfigure}
    \hfill
    \begin{subfigure}[t]{0.24\linewidth}
        \centering
        \includegraphics[width=\textwidth]{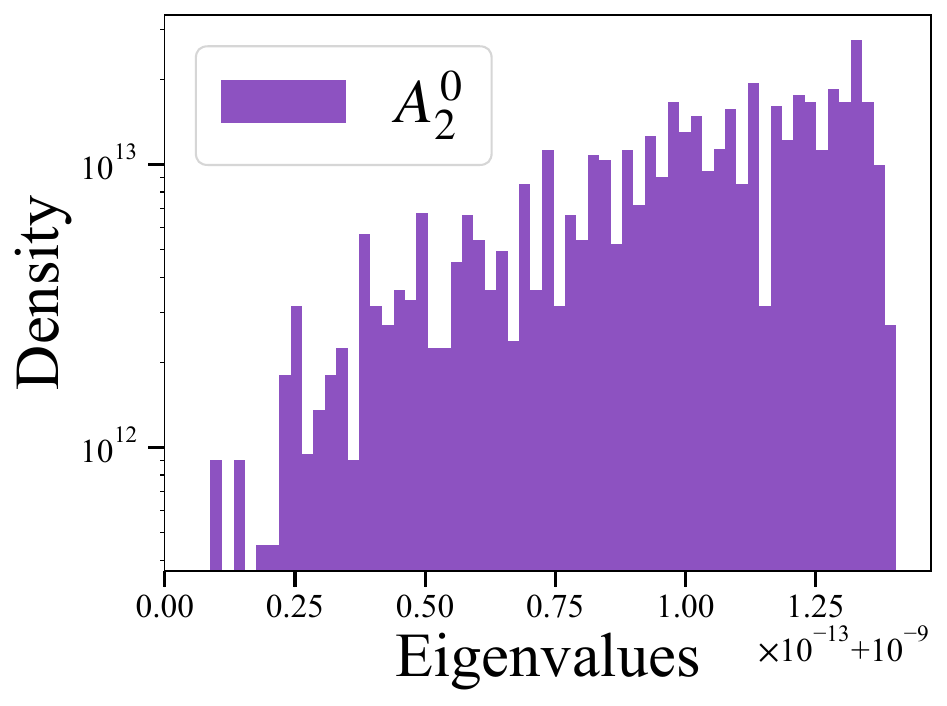}
        \caption{ $\mA_2^0$  }
    \end{subfigure}
    \hfill
        \begin{subfigure}[t]{0.24\linewidth}
        \centering
        \includegraphics[width=\textwidth]{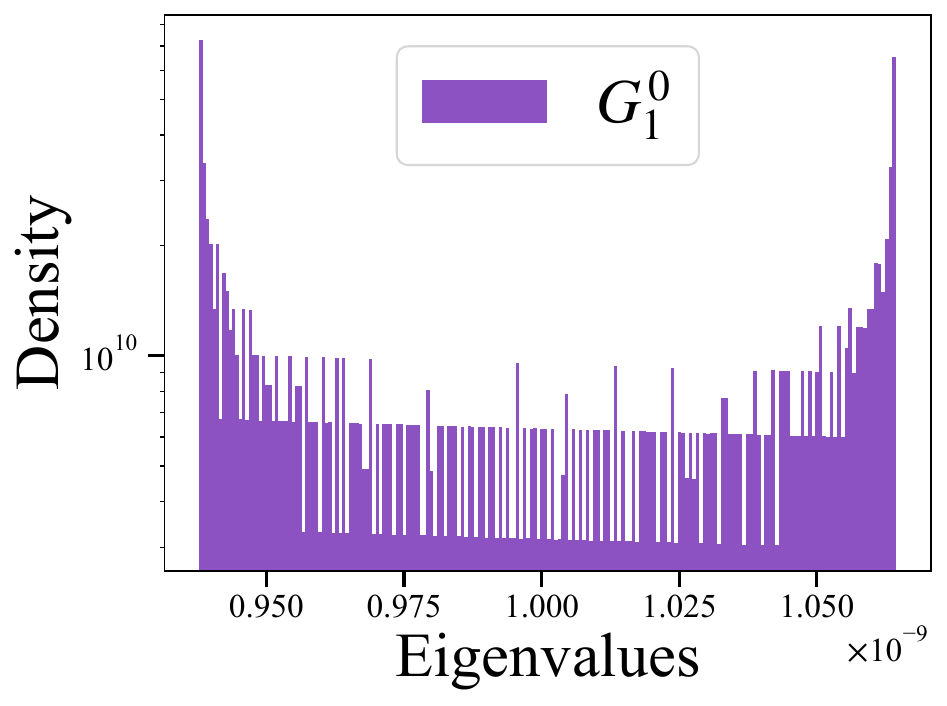}
        \caption{ $\mG_1^0$}
    \end{subfigure}
    \hfill
        \begin{subfigure}[t]{0.24\linewidth}
        \centering
        \includegraphics[width=\textwidth]{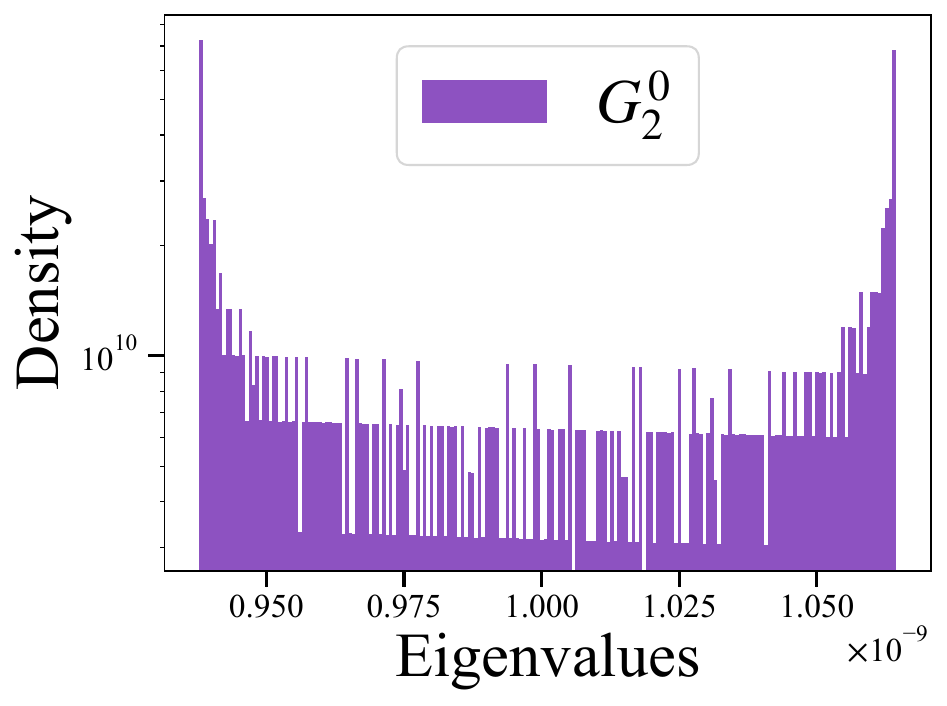}
        \caption{ $\mG_2^0$ }
    \end{subfigure}
    \hfill

    \begin{subfigure}[t]{0.24\linewidth}
        \centering
        \includegraphics[width=\textwidth]{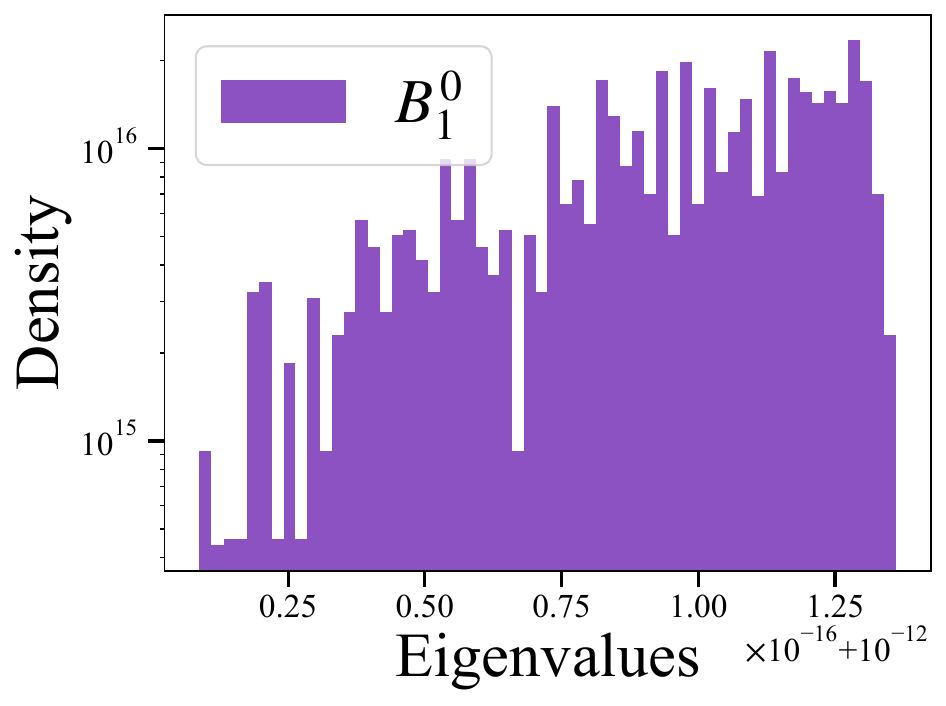}
        \caption{ $\mB_1^0$ }
    \end{subfigure}
    \hfill
    \begin{subfigure}[t]{0.24\linewidth}
        \centering
        \includegraphics[width=\textwidth]{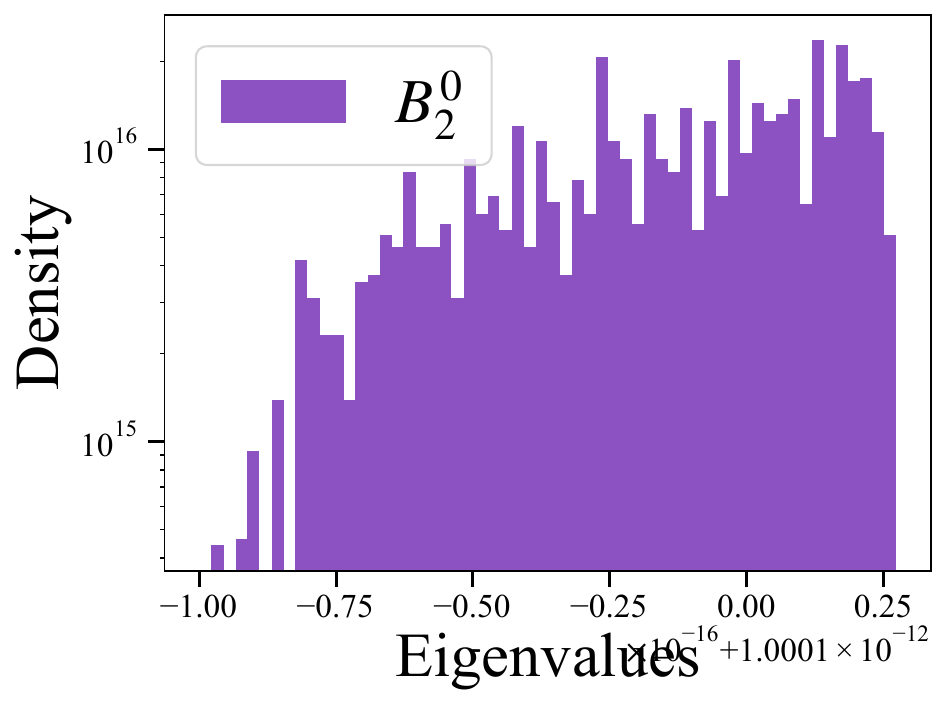}
        \caption{   $\mB_2^0$ }
    \end{subfigure}
    \hfill
        \begin{subfigure}[t]{0.24\linewidth}
        \centering
        \includegraphics[width=\textwidth]{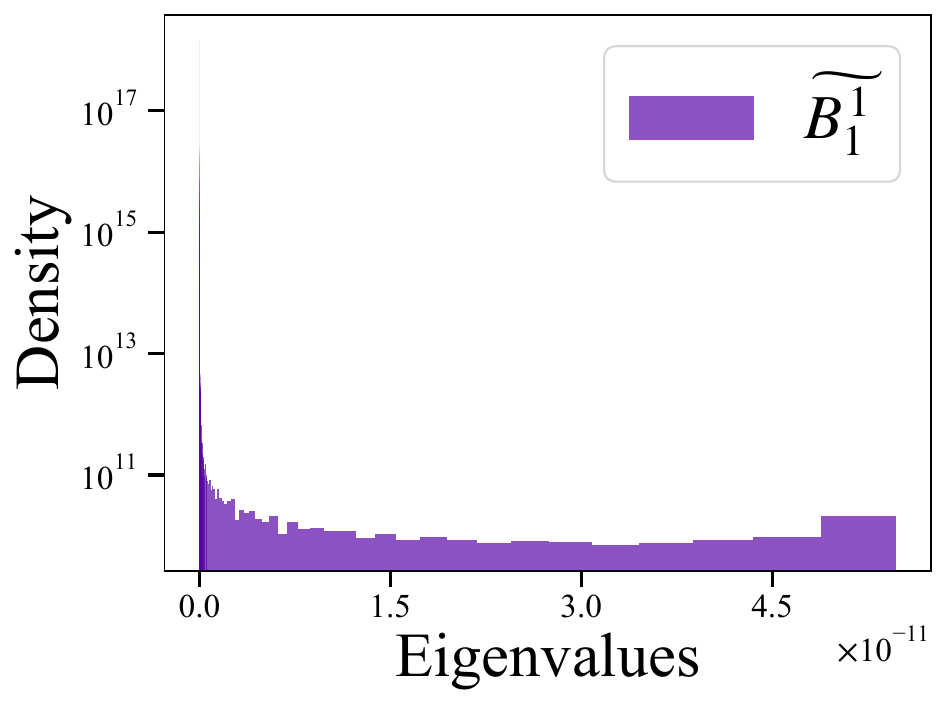}
        \caption{ $\widetilde{\mB_1^1}$ }
    \end{subfigure}
    \hfill
        \begin{subfigure}[t]{0.24\linewidth}
        \centering
        \includegraphics[width=\textwidth]{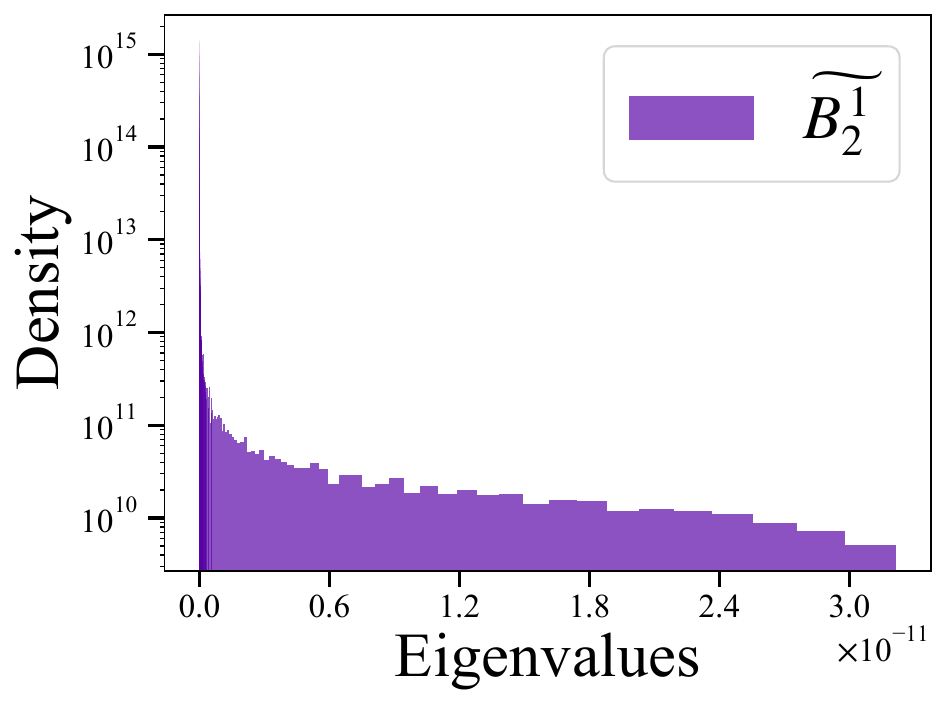}
        \caption{ $\widetilde{\mB_2^1}$ }
    \end{subfigure}
    \hfill

    \begin{subfigure}[t]{0.24\linewidth}
        \centering
        \includegraphics[width=\textwidth]{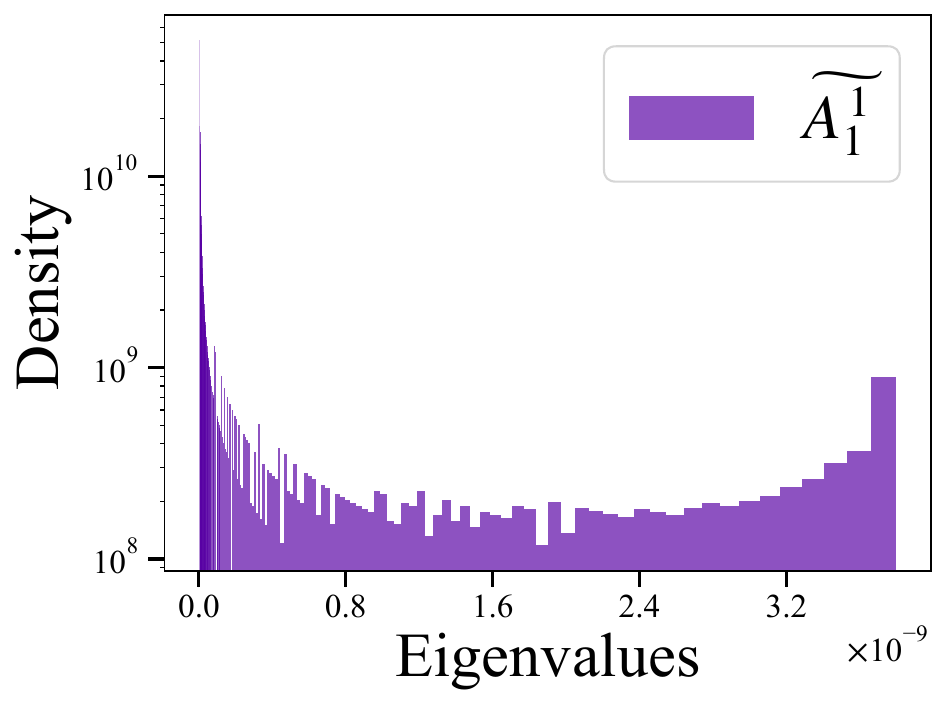}
        \caption{ $\widetilde{\mA_1^1}$  }
    \end{subfigure}
    \hfill
    \begin{subfigure}[t]{0.24\linewidth}
        \centering
        \includegraphics[width=\textwidth]{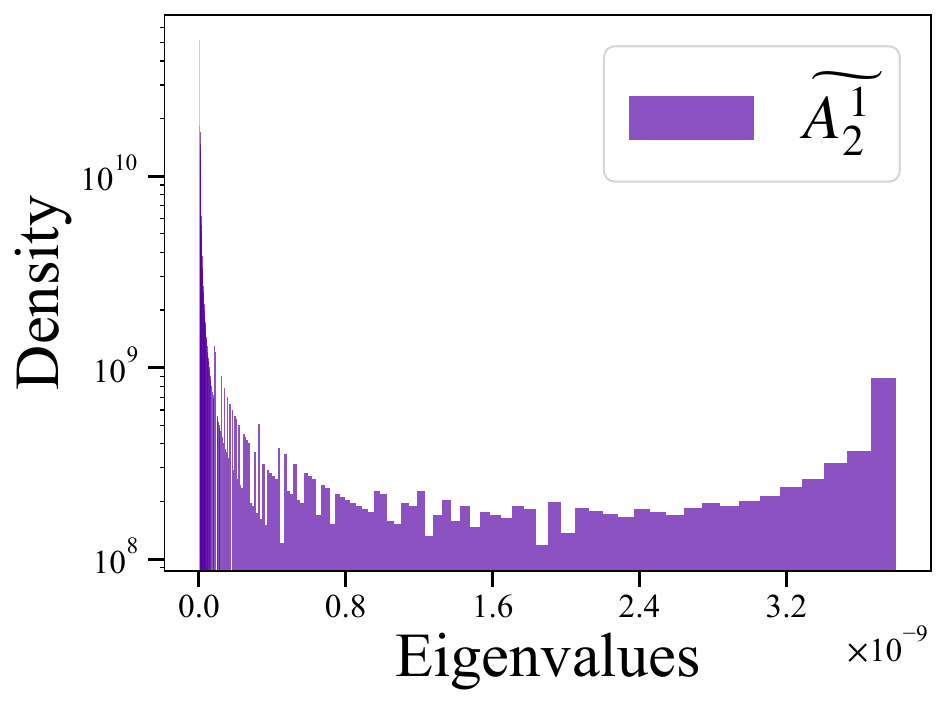}
        \caption{  $\widetilde{\mA_2^1}$  }
    \end{subfigure}
    \hfill
        \begin{subfigure}[t]{0.24\linewidth}
        \centering
        \includegraphics[width=\textwidth]{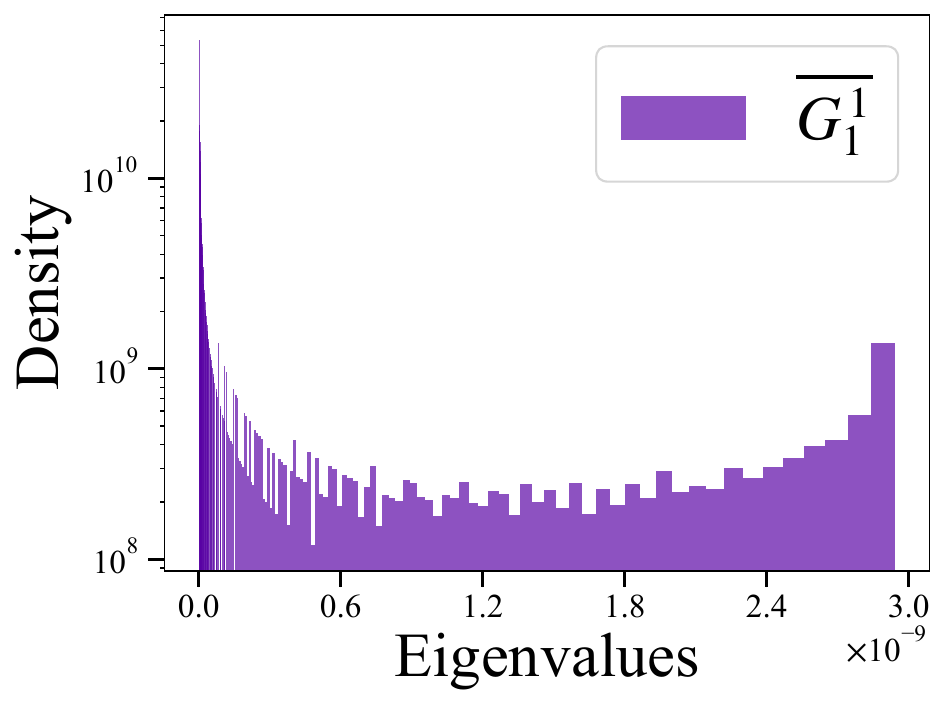}
        \caption{$\overline{\mG_1^1}$  }
    \end{subfigure}
    \hfill
        \begin{subfigure}[t]{0.24\linewidth}
        \centering
        \includegraphics[width=\textwidth]{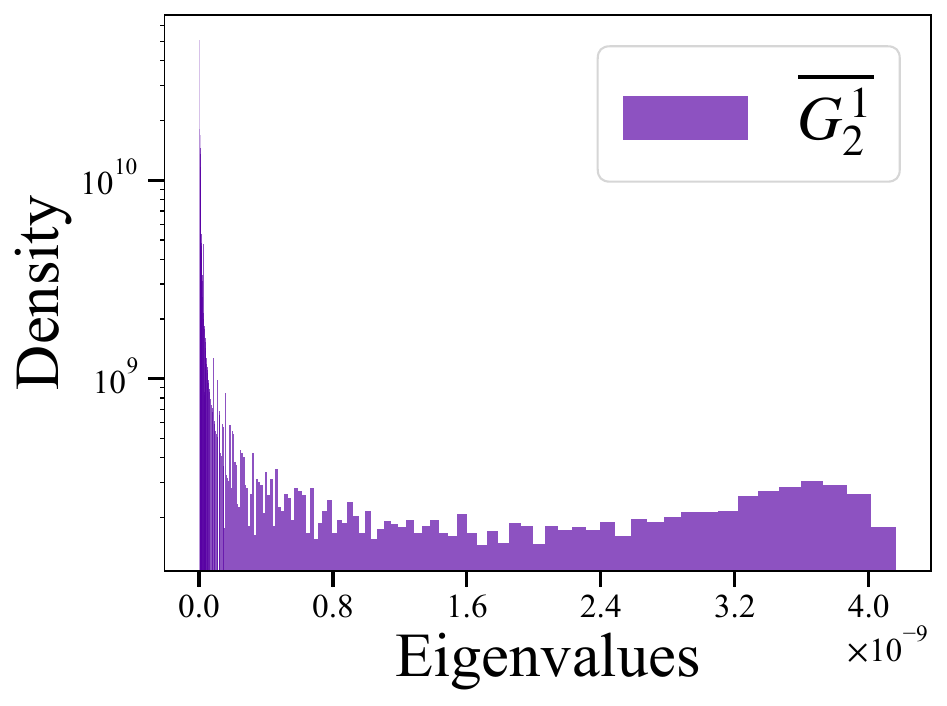}
        \caption{$\overline{\mG_2^1}$  }
    \end{subfigure}
    \hfill
    
    \begin{subfigure}[t]{0.24\linewidth}
        \centering
        \includegraphics[width=\textwidth]{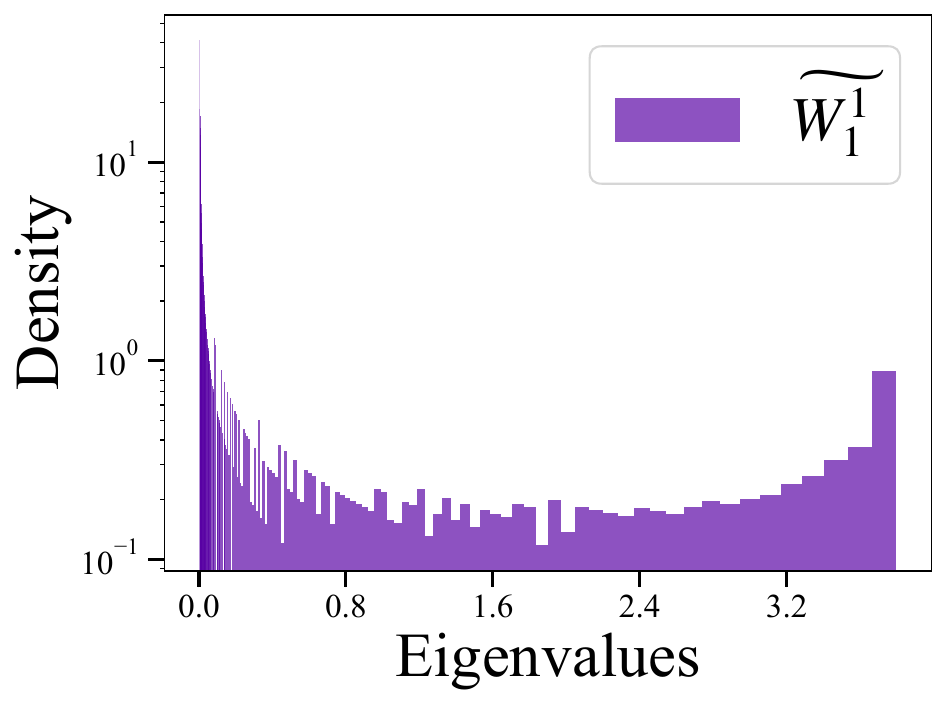}
        \caption{  $\widetilde{\mW_1^1}$   }
    \end{subfigure}
    \hfill
    \begin{subfigure}[t]{0.24\linewidth}
        \centering
        \includegraphics[width=\textwidth]{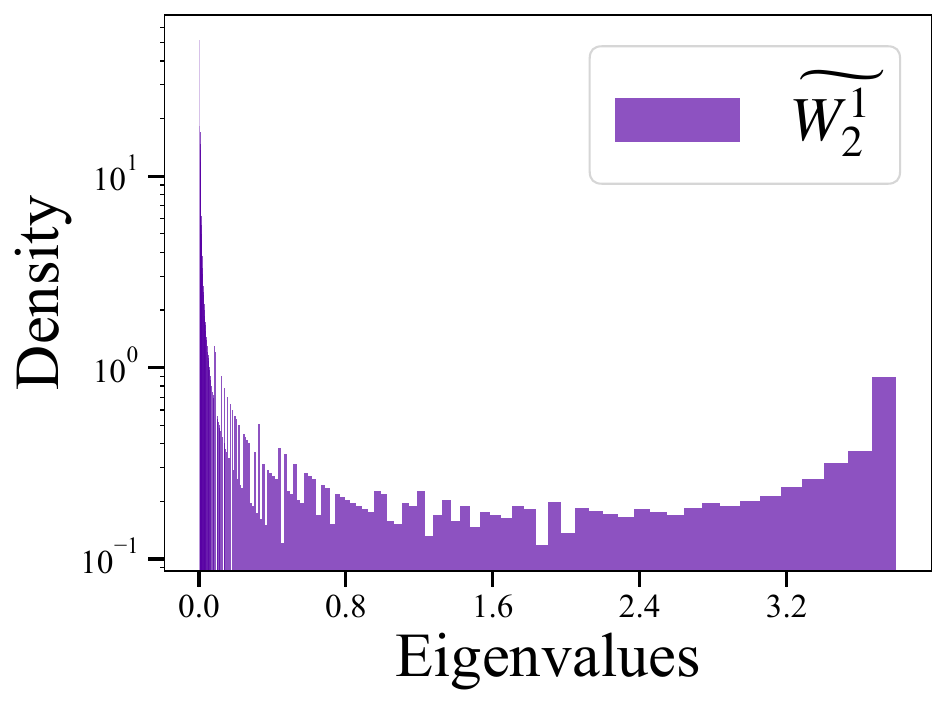}
        \caption{   $\widetilde{\mW_2^1}$ }
    \end{subfigure}
    \hfill
        \begin{subfigure}[t]{0.24\linewidth}
        \centering
        \includegraphics[width=\textwidth]{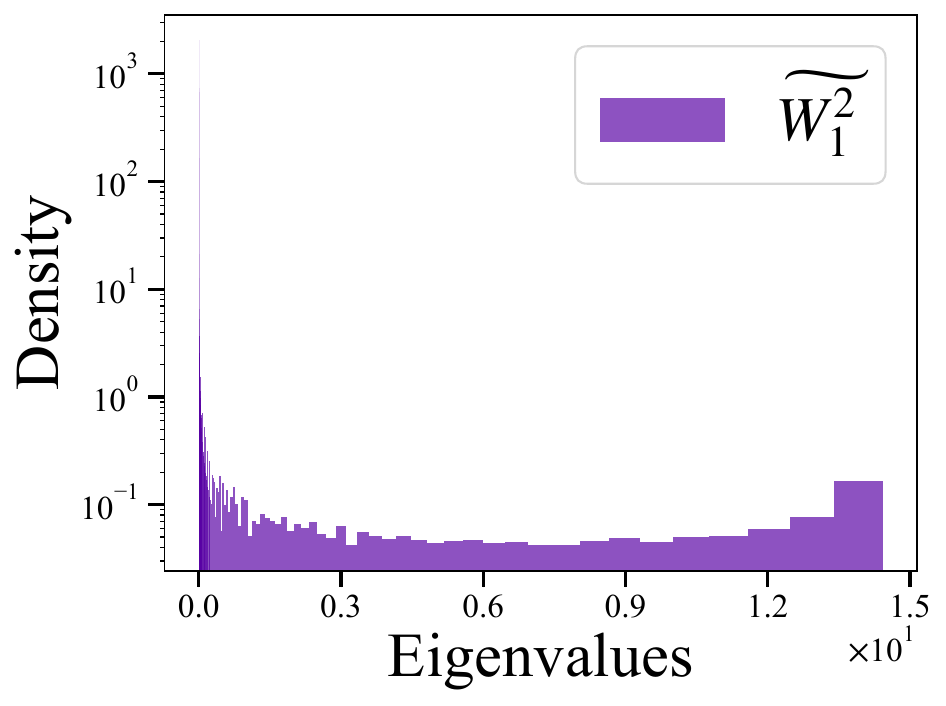}
        \caption{ $\widetilde{\mW_1^2}$}
    \end{subfigure}
    \hfill
        \begin{subfigure}[t]{0.24\linewidth}
        \centering
        \includegraphics[width=\textwidth]{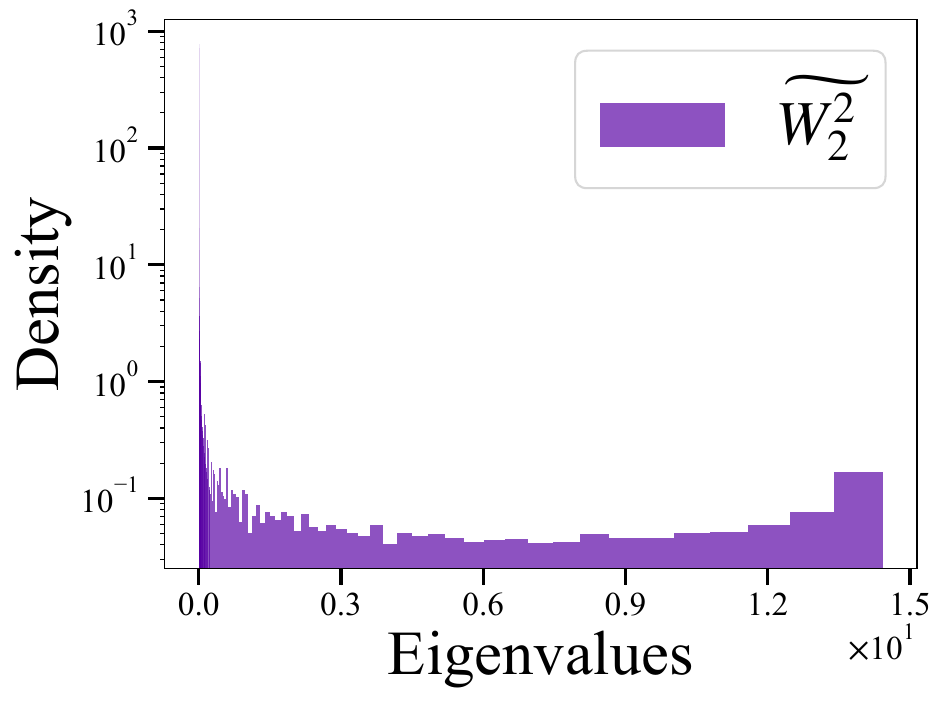}
        \caption{ $\widetilde{\mW_2^2}$ }
    \end{subfigure}
    \hfill
    \begin{subfigure}[t]{0.45\linewidth}
        \centering
        \includegraphics[width=\textwidth]{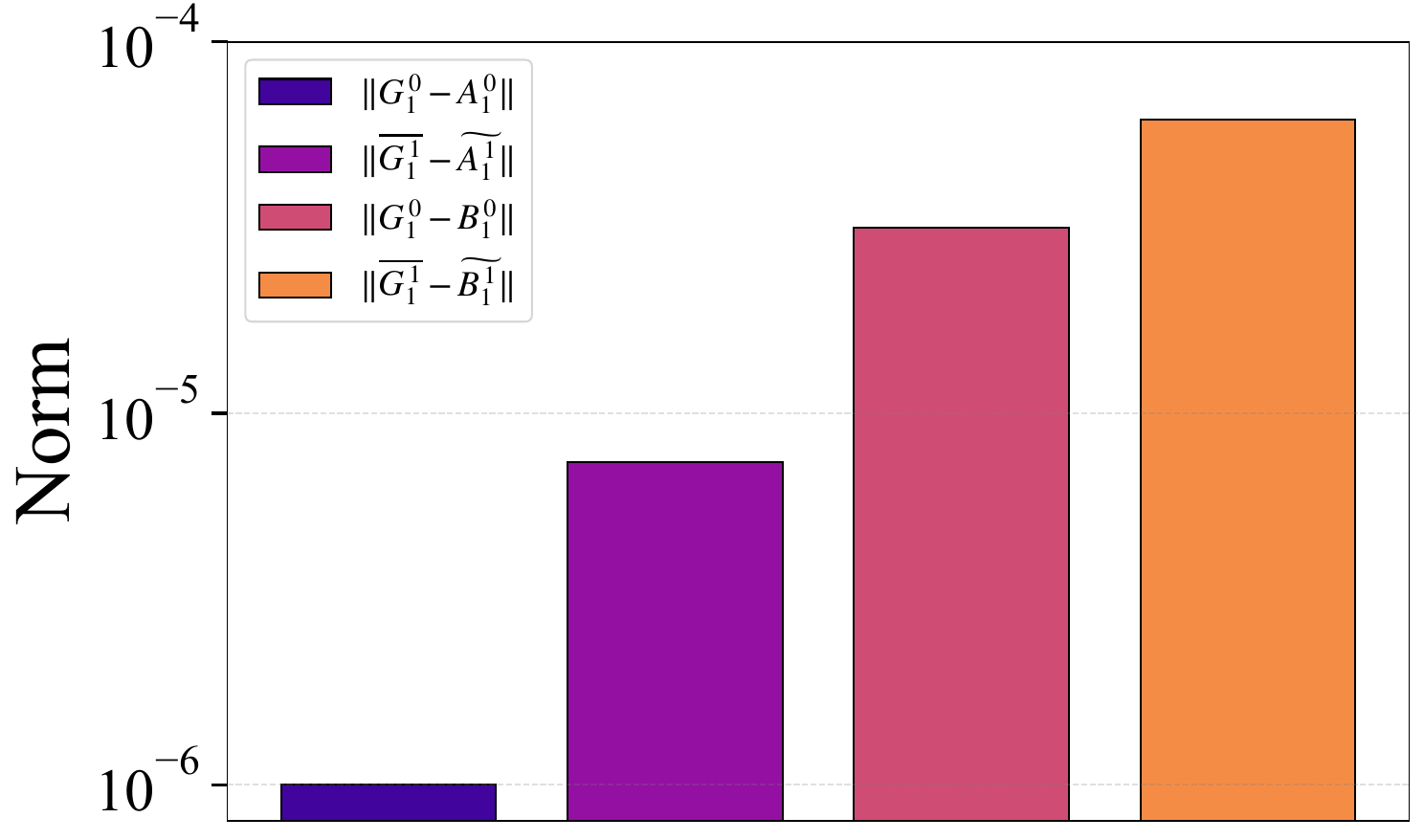}
        \caption{  First Layer Norm Gap }
        \label{fig:First Layer gap}
    \end{subfigure}
    \hfill
        \begin{subfigure}[t]{0.45\linewidth}
        \centering
        \includegraphics[width=\textwidth]{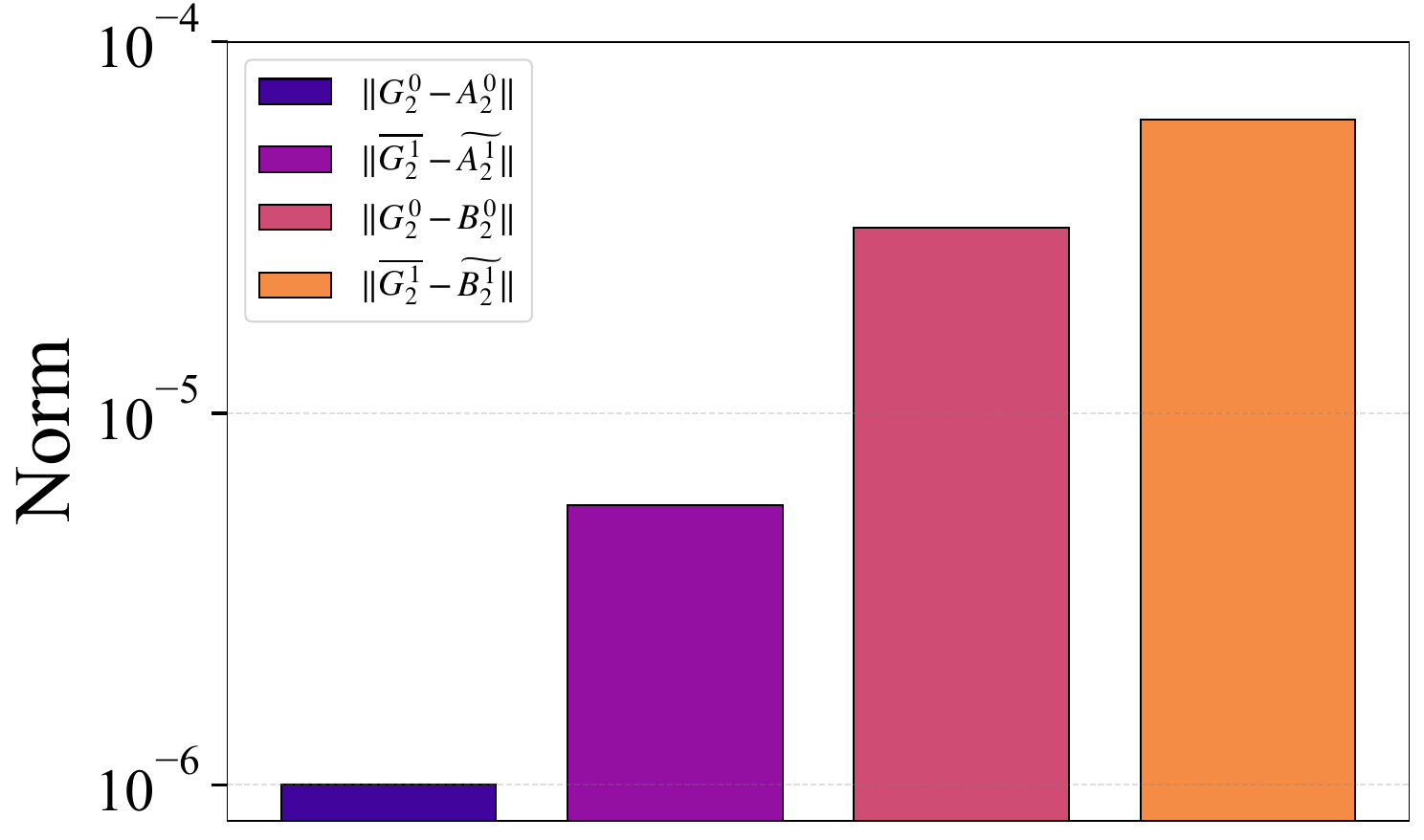}
        \caption{ Second Layer  Norm Gap }
        \label{fig:Second Layer gap}
    \end{subfigure}
    \caption{ We visualize   the ESDs   of gradient matrices and weight matrices   $\{\mA_l^0 \}_{l=1}^2, \{\mB_l^0 \}_{l=1}^2, \{\mG_l^0 \}_{l=1}^2 $,  $\{\widetilde{\mA_l^1} \}_{l=1}^2$, $\{\widetilde{\mB_l^1} \}_{l=1}^2$, $\{\widetilde{\mG_l^1} \}_{l=1}^2$, $\{\widetilde{\mW_l^1} \}_{l=1}^2$ and  $\{\widetilde{\mW_l^2} \}_{l=1}^2$  and norm gap with $\eta_1=\eta_2=h^{\frac{3}{2}}$ and $  h=1000$.
}
\label{fig:spectral analysis}
\end{figure*}

To better understand Proposition~\ref{sec:orthgo-proposition-2-layer}, in this subsection, we perform spectral analysis of the key matrices like $\{\mA_l^0 \}_{l=1}^2, \{\mB_l^0 \}_{l=1}^2, \{\mG_l^0 \}_{l=1}^2 $,  $\{\widetilde{\mA_l^1} \}_{l=1}^2$, $\{\widetilde{\mB_l^1} \}_{l=1}^2$, $\{\widetilde{\mG_l^1} \}_{l=1}^2$ arising after one-step and two-step updates in a two-layer linear neural network under orthogonal initialization.  In Figure~\ref{fig:spectral analysis}, we consider $\eta_1=\eta_2=h^{\frac{3}{2}}$ with $h=1000$, we visualize the empirical spectral densities (ESDs) of the weight matrices, gradient matrices, and the decomposed gradient components  represented as $\mA$ and $\mB$. Take  $\{\mA_l^0 \}_{l=1}^2, \{\mB_l^0 \}_{l=1}^2, \{\mG_l^0 \}_{l=1}^2 $ as examples,  we find that the eigenvalue scales of $\{\mA_l^0\}_{l=1}^2$ and $\{\mG_l^0\}_{l=1}^2$ are comparable, and are larger than those of $\{\mB_l^0\}_{l=1}^2$ by an $O(h)$ factor. This matches our norm analysis in Section~\ref{app:2-NN-orthgonal_one-step_prop}  and ~\ref{app:2-NN-orthgonal_two-step_prop}, showing that $\norm{\mA_l^0}$ exceeds $\norm{\mB_l^0}$) by $O(\sqrt{h})$, since the ESD is computed from the eigenvalues of $\mW^\top \mW$. We visualize the norm gap in Figure~\ref{fig:First Layer gap} and~\ref{fig:Second Layer gap}, which  also confirm that the eigenvalue scales of $\{\mA_l^0\}_{l=1}^2$ and $\{\mG_l^0\}_{l=1}^2$ are comparable, and are larger than those of $\{\mB_l^0\}_{l=1}^2$ in magnitude, we also   find  the eigenvalue scales of $\{\eta_l\mG_l^0\}_{l=1}^2$ and $\{\eta_l\mA_l^0\}_{l=1}^2$ are comparable to $\{\widetilde{\mW_l^1}\}_{l=1}^2$, which matches our Proposition~\ref{sec:orthgo-proposition-2-layer}.   Consequently, the ESDs provide an intuitive explanation for why $\{\mA_l^0\}_{l=1}^2$ and $\{\mG_l^0\}_{l=1}^2$ are close in norm, supporting our use of the approximate gradient when deriving the one-step and two-step exact losses. A similar phenomenon holds for $\{\widetilde{\mA_l^1}\}_{l=1}^2$ and $\{\overline{\mG_l^1}\}_{l=1}^2$, relative to $\{\widetilde{\mB_l^1}\}_{l=1}^2$.


\subsection{Theoretical Simulation}
Here we present more experimental results.

\paragraph{Orthogonal initialization.} In Figure~\ref{fig:2-NN-more-steps-orthgo} and~\ref{fig:3-NN-more-steps-orthog},  Under orthogonal initialization, we set $h=1000$ and conducted experiments for \text{steps} $\in \{1,2,4,8 \}$ under the constraint $\eta_1+\eta_2 \le h^{\frac{3}{2}}$. Consistent with our earlier findings: after two updates the model exhibits local optimality at balanced layer-wise learning rates. We also try different   $h$ and find that when $h$ satisfy the condition on $h$ in Corollary like Corollary~\ref{cor:two-layer-NN}, the balanced learning-rate allocation is  locally optimal, otherwise not. See Figure~\ref{fig:h=5000-2-NN},~\ref{fig:h=5000-3-NN} and~\ref{fig:h=100-2-NN}.

\paragraph{Orthogonal initialization.} We also ran the same set of experiments under Gaussian initialization In Figure~\ref{fig:2-NN-more-steps-gaussian} and~\ref{fig:3-NN-more-steps-gaussian}, with $h=1000$  and $\eta_1+\eta_2 \le h^{\frac{2}{3}}$, again for \text{steps} $\in \{1,2,4,8\}$. The results mirror those under orthogonal initialization: balanced learning rates become locally optimal after two updates, whereas after a single update an asymmetric learning-rate allocation performs better, which is consistent with  the special cases of  Theorem~\ref{theorem for 2-layer} and Theorem~\ref{theorem for 3-layer} for two-layer and three-layer neural networks.

\paragraph{More discussions in Section~\ref{discussion}.} For the question about our paper shows symmetric learning rates are suboptimal for one step but optimal for two steps, which  may be not exactly the same as using asymmetric learning rates early and symmetric ones later. Here we agree that our theory focuses on the result that symmetric learning rates are suboptimal for a single update step but become optimal after two steps. This suggests that 1. asymmetric learning rates may be preferable at the very beginning of training, 2. symmetric learning rates become optimal as cross-layer interactions develop over subsequent steps, even if the initial learning-rate allocation is not optimal. We believe this also points to a practical strategy that use asymmetric learning rates early in training and more symmetric ones later.

To better connect these two regimes, in Figure~\ref{fig:special-NN}(a)(b) we consider a three layer linear network in which the first step uses an asymmetric learning rate allocation by training only the first layer. For the second step, we then search over the test loss as a function of $\eta_1$ under the constraint $\eta_1+\eta_2=C$. We  find that the same transition still appears: from asymmetry at the first step to balance at the second step.

In Figure~\ref{fig:special-NN}(c)(d)(e),   we consider a 3-layer CNN whose  first two layers are convolutional layers and whose final layer is a fixed linear readout layer.  We consider a synthetic binary image classification problem on $16 \times 16$ grayscale images. Each sample belongs to one of two classes: Class $-1$: an image containing a horizontal bar. Class $+1$: an image containing a vertical bar.  We consider two trainable $3\times 3$ convolutional layers with no bias: the first maps from 1 input channel to 8 hidden channels, and the second maps from 8 channels to 8 channels, with a ReLU activation after each convolution. The resulting feature map is then globally average pooled over the spatial dimensions, producing an 8-dimensional representation, which is fed into a fixed random linear readout to produce a single scalar output.

In Figure~\ref{fig:special-NN}(c)(d), for the first step, we consider only updating the first(second) layer, using learning rate $C$. For the second step, we study layer-wise learning-rate allocation under the constraint $\eta_1+\eta_2=C.$ Although the optimum no longer occurs exactly at $\eta_1=\eta_2$ because of the changed architecture and the presence of nonlinearities, the optimal performance is still attained when $\eta_1 \approx \eta_2$. This  indicates that, after an asymmetric first step, learning rates that are approximately symmetric still remain preferable. In Figure~\ref{fig:special-NN}(e), We study a 3-layer CNN in which the second update step uses a symmetric learning-rate allocation, $\eta_1=\eta_2=\frac{C}{2},$ while the first step is optimized under the constraint $\eta_1+\eta_2=C.$ We find training the second layer (corresponding to $\eta_1=0$) at the first step is better than the symmetric allocation. Thus, the CNN experiments lead to similar conclusions as in the 3-layer linear neural network setting, further reinforcing our claim.

\begin{figure}[tb]
    \centering
    \begin{subfigure}[t]{0.24\linewidth}
        \centering
        \includegraphics[width=\textwidth]{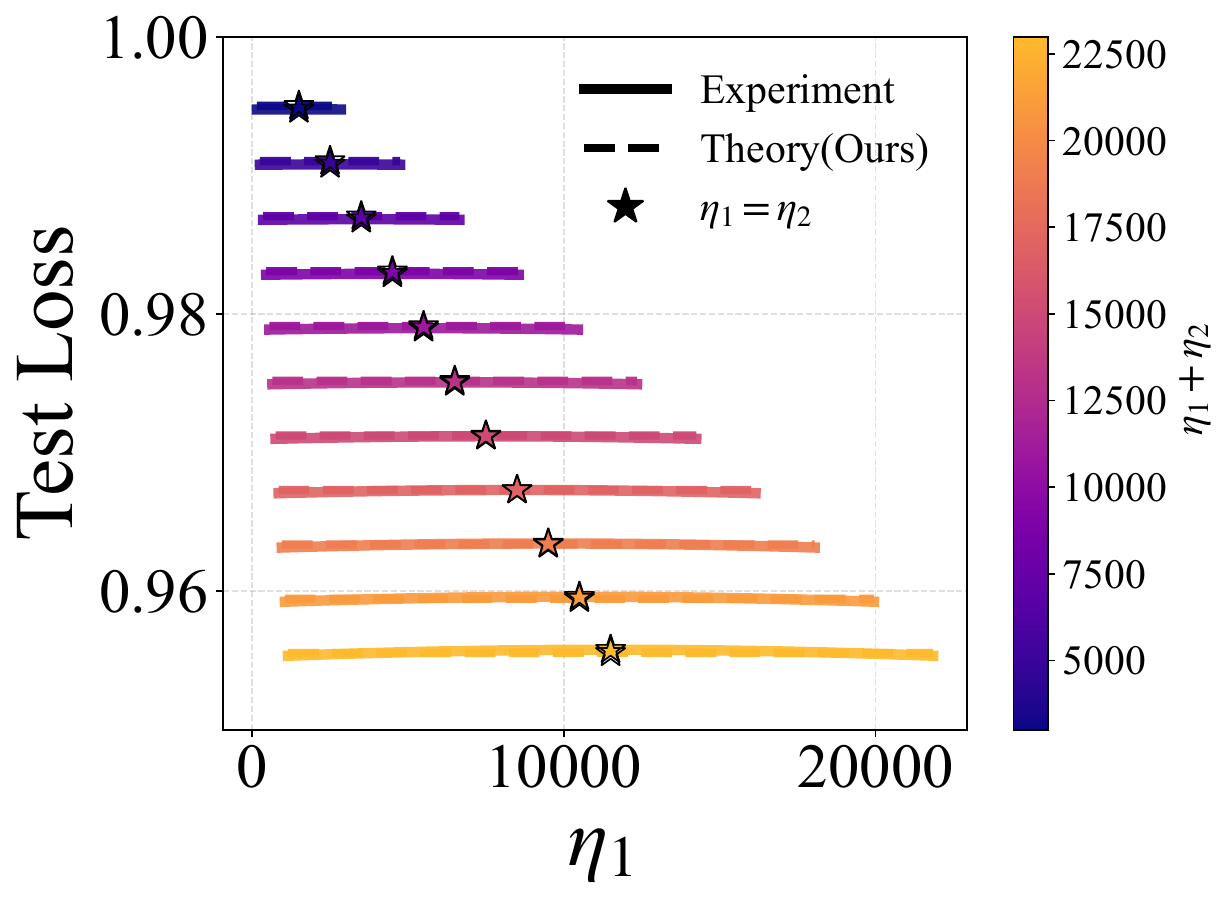}
        \caption{  1-step GD update }
    \end{subfigure}
    \hfill
        \begin{subfigure}[t]{0.24\linewidth}
        \centering
        \includegraphics[width=\textwidth]{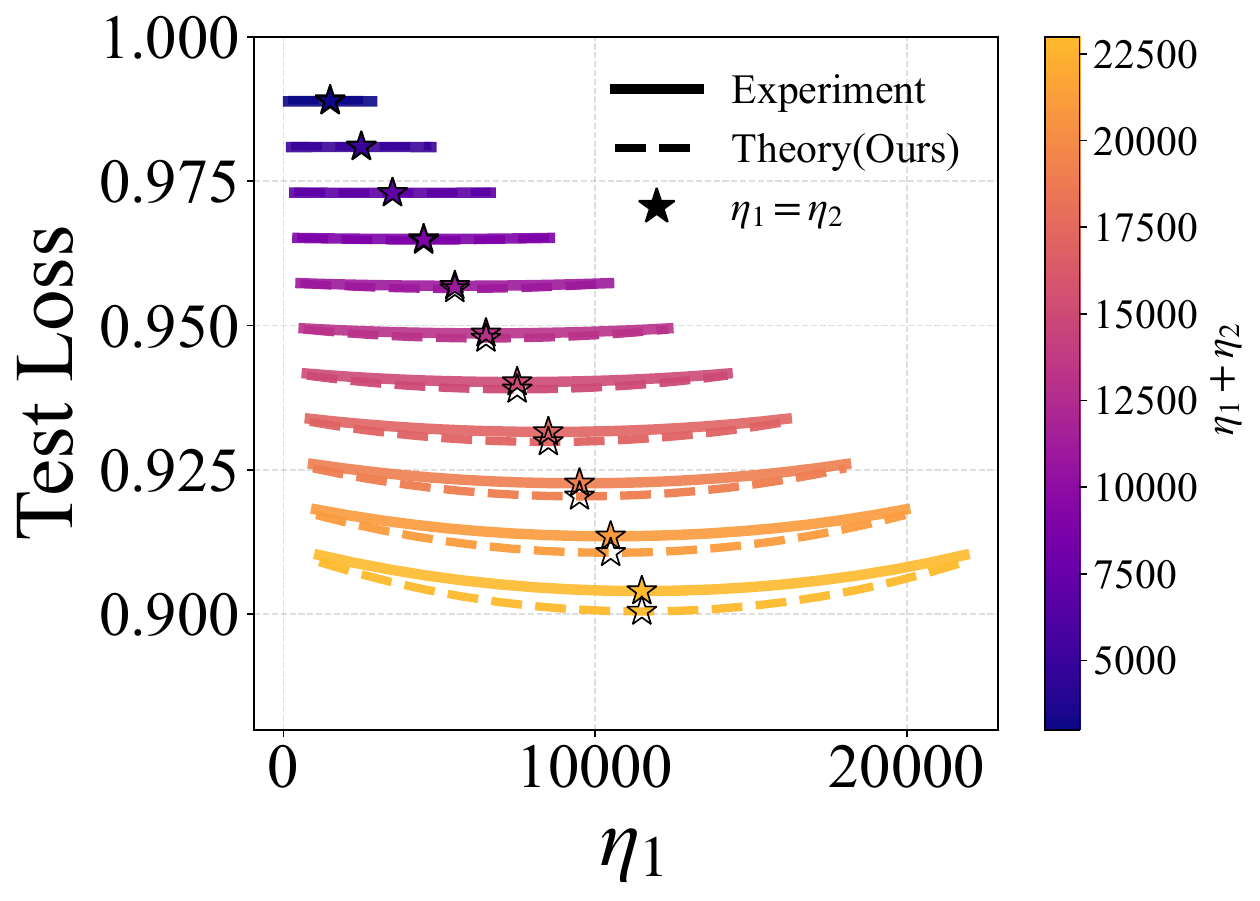}
        \caption{  2-step GD update }
    \end{subfigure}
    \hfill
    \begin{subfigure}[t]{0.24\linewidth}
        \centering
        \includegraphics[width=\textwidth]{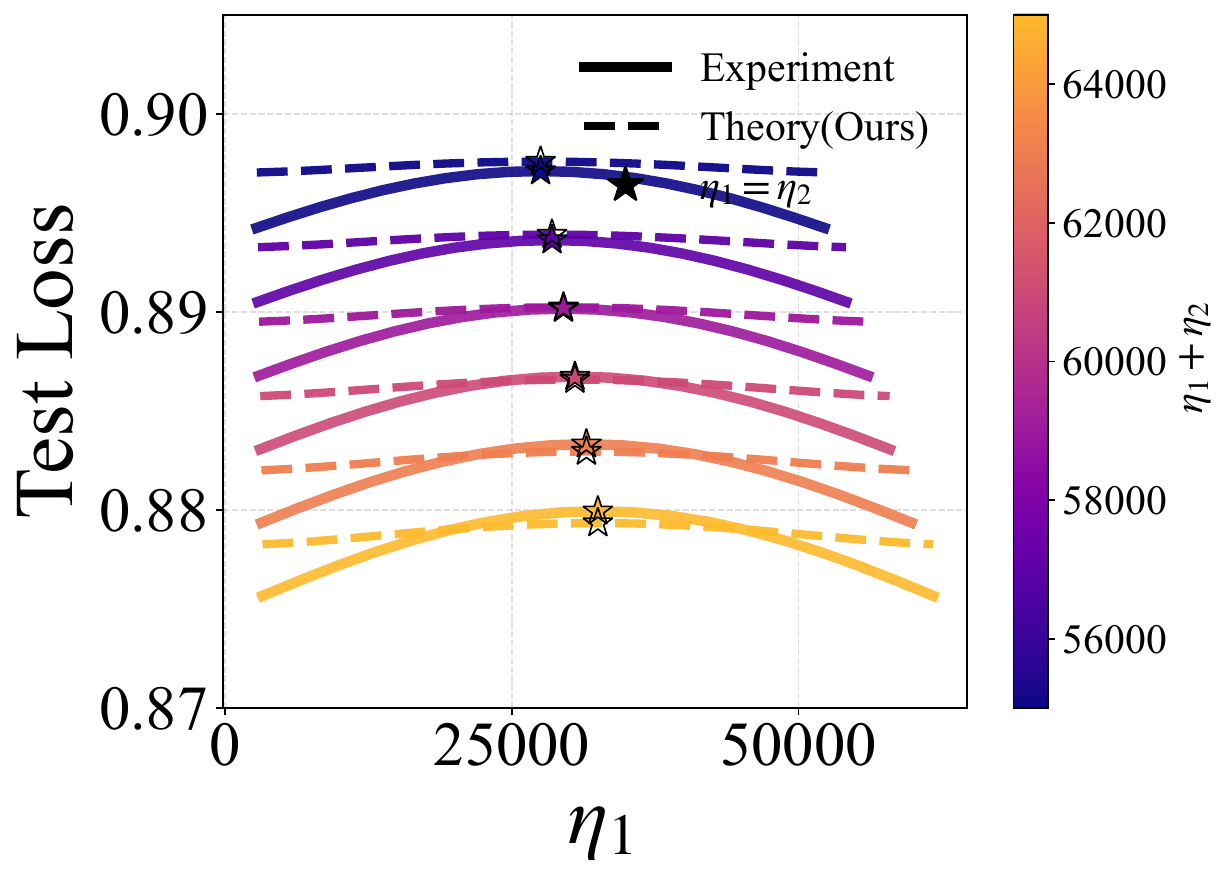}
        \caption{  1-step GD update }
    \end{subfigure}
    \hfill
        \begin{subfigure}[t]{0.24\linewidth}
        \centering
        \includegraphics[width=\textwidth]{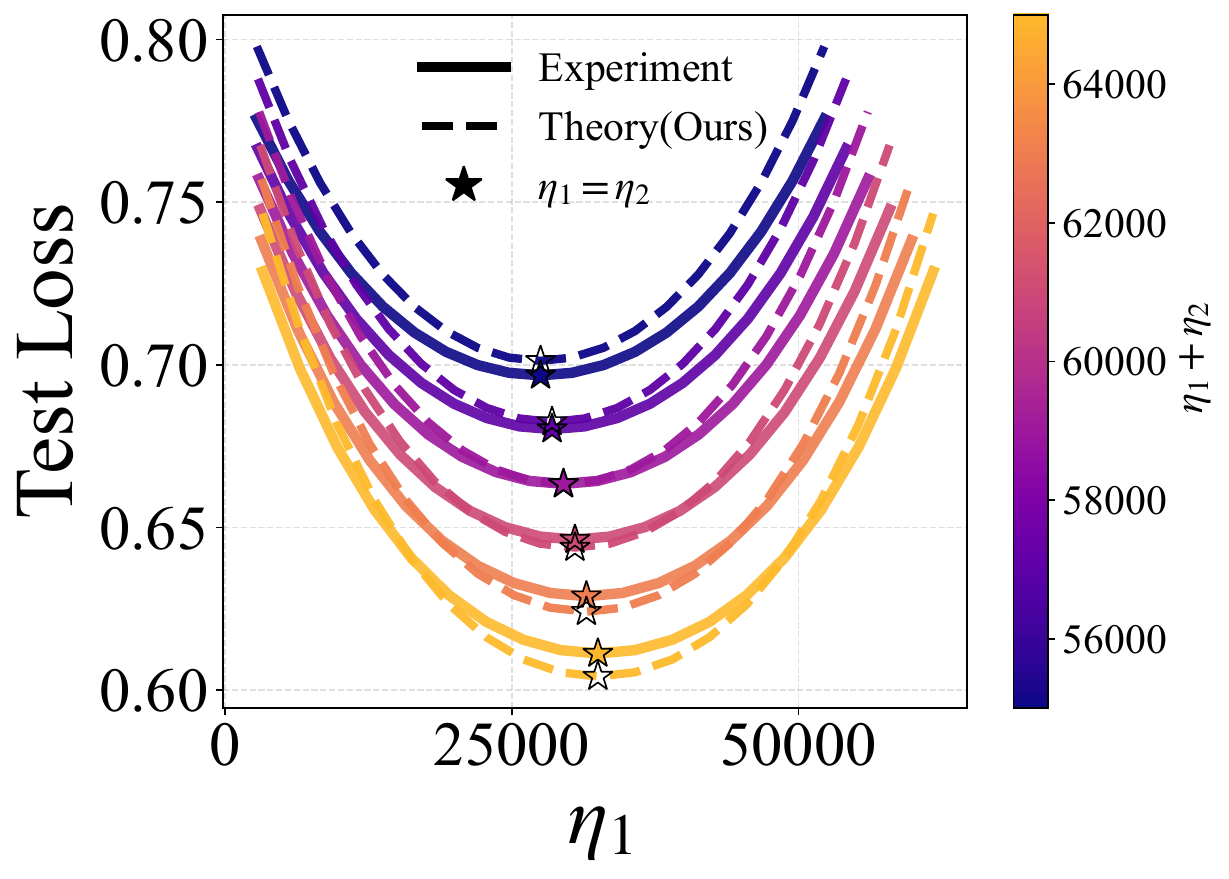}
        \caption{  2-step GD update }
    \end{subfigure}
    \caption{Test loss of a 2-layer NN under orthogonal initialization with width $h=1000$. Here we set $\eta_1+\eta_2\leq O(h^{\frac{3}{2}})$. Our theory accurately predicts the test loss after   1-step and 2-step gradient descent updates with varying learning rates. In particular, we highlight the role of balancing learning rates across layers (i.e $\eta_1=\eta_2$) on the test loss.   \tianyu{ This is a modification to Fig. 2. Compared with Fig.1, Fig.2 considers larger learning rates regime, so I add a small error term $O(1/\sqrt{h}$ for two-step update for fig.3 to our theoretical values to bring them closer to the experimental results. We can consider using Fig. 3 in place of Fig. 2 to illustrate the largest learning-rate regime in which our results hold for the two-layer neural network. For the three-layer network, since we do not provide a rigorous theoretical bound for the two-step case, I plan to keep Fig. 4 as is and place it in the appendix.
    \vigk{Yes, this one looks good. I wonder why the 1-step GD update theory gives a flatter line.}
 } }

\label{fig:2-NN-orthogonal-large-lr}
\end{figure}

\begin{figure*}[!htb]
    \centering
    \begin{subfigure}[t]{0.22\linewidth}
        \centering
        \includegraphics[width=\textwidth]{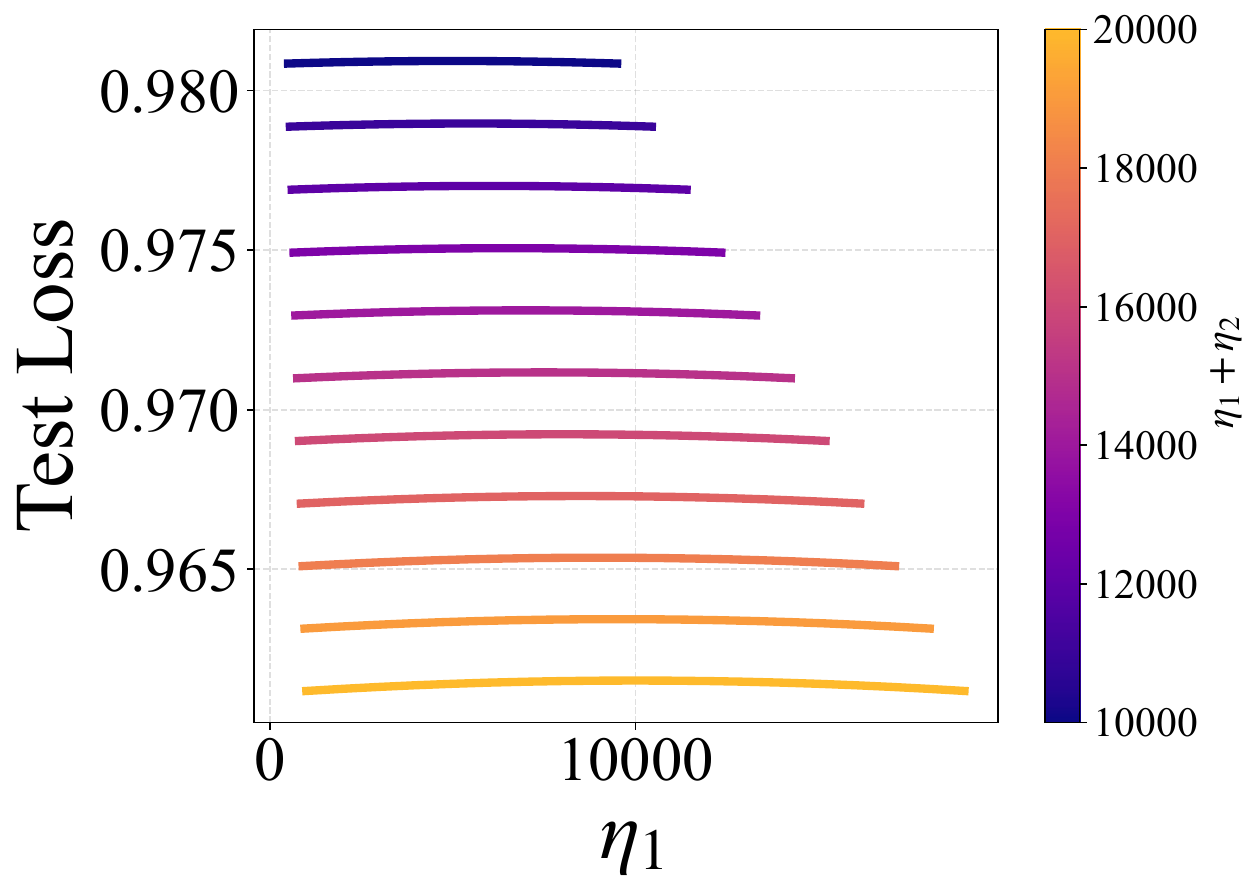}
        \caption{ Step=1 }
    \label{fig:2-layer}
    \end{subfigure}
    \hfill
    \begin{subfigure}[t]{0.21\linewidth}
        \centering
        \includegraphics[width=\textwidth]{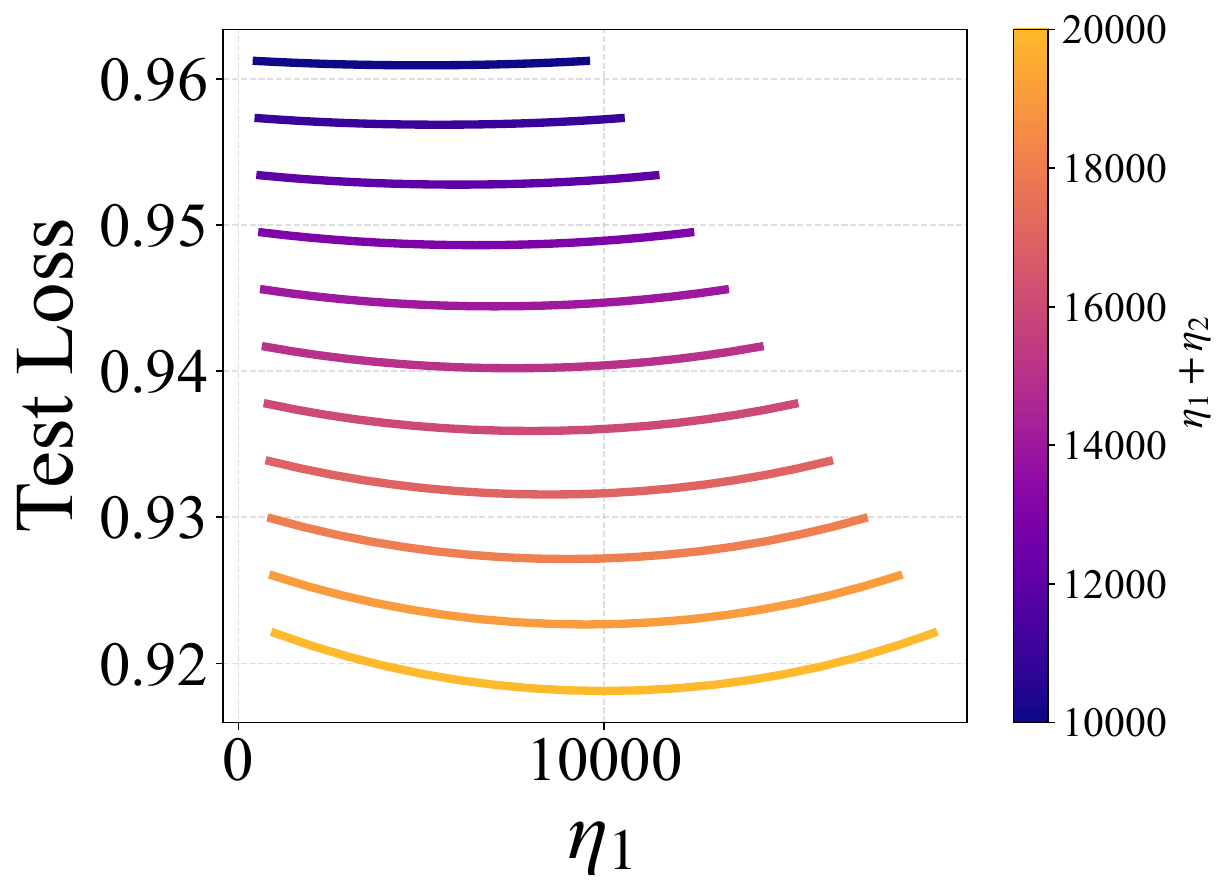}
        \caption{  Step=2 }
    \end{subfigure}
    \hfill
        \begin{subfigure}[t]{0.21\linewidth}
        \centering
        \includegraphics[width=\textwidth]{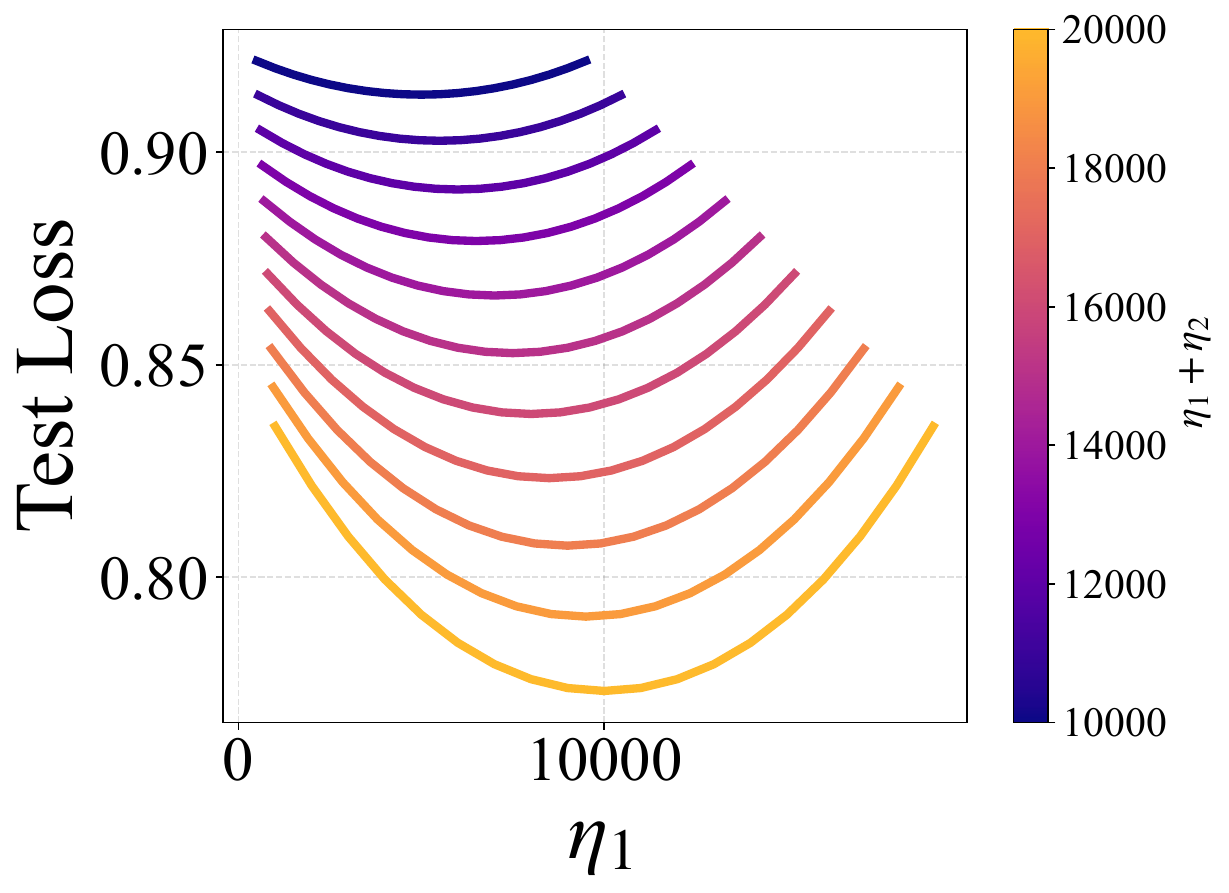}
        \caption{  Step=4}
    \end{subfigure}
    \hfill
        \begin{subfigure}[t]{0.21\linewidth}
        \centering
        \includegraphics[width=\textwidth]{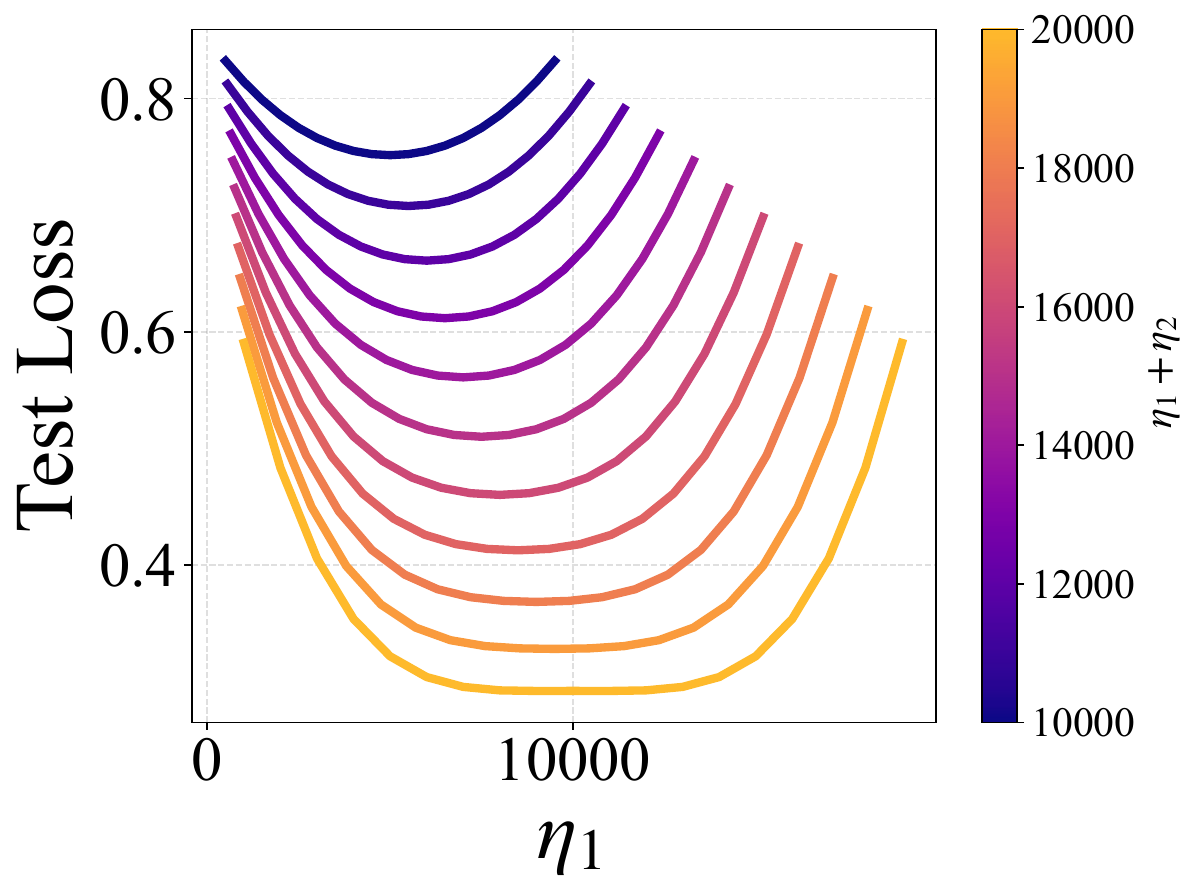}
        \caption{  Step=8}
    \end{subfigure}
    \hfill
    \caption{ \textbf{More-steps-empirical-loss for 2-layer NN  under Orthogonal initialization.} Here we set $\eta_1+\eta_2\leq O(h^{\frac{3}{2}})$  and  $h=1000$. 
}
\label{fig:2-NN-more-steps-orthgo}
\end{figure*}

\begin{figure*}[!htb]
    \centering
    \begin{subfigure}[t]{0.22\linewidth}
        \centering
        \includegraphics[width=\textwidth]{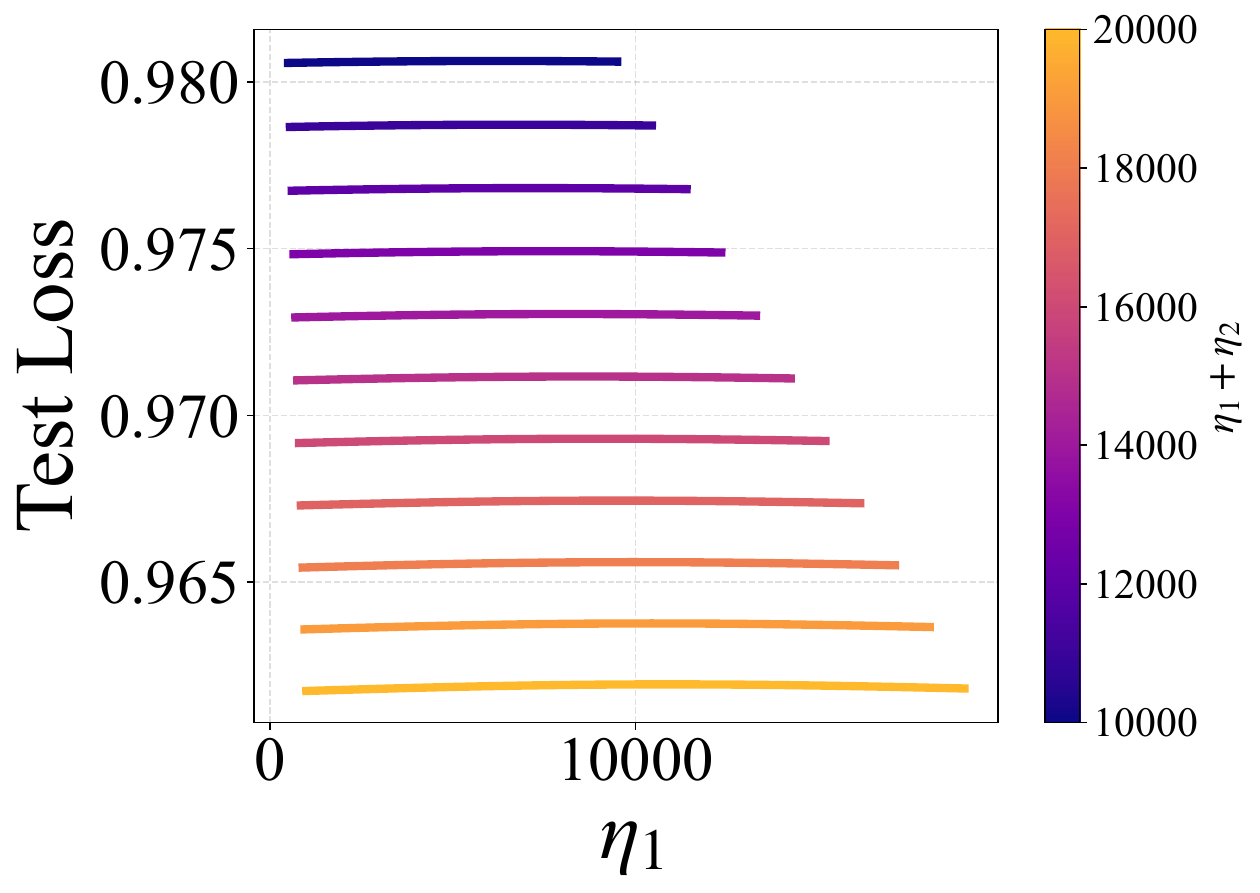}
        \caption{  Step=1 }
    \label{fig:2-layer}
    \end{subfigure}
    \hfill
    \begin{subfigure}[t]{0.21\linewidth}
        \centering
        \includegraphics[width=\textwidth]{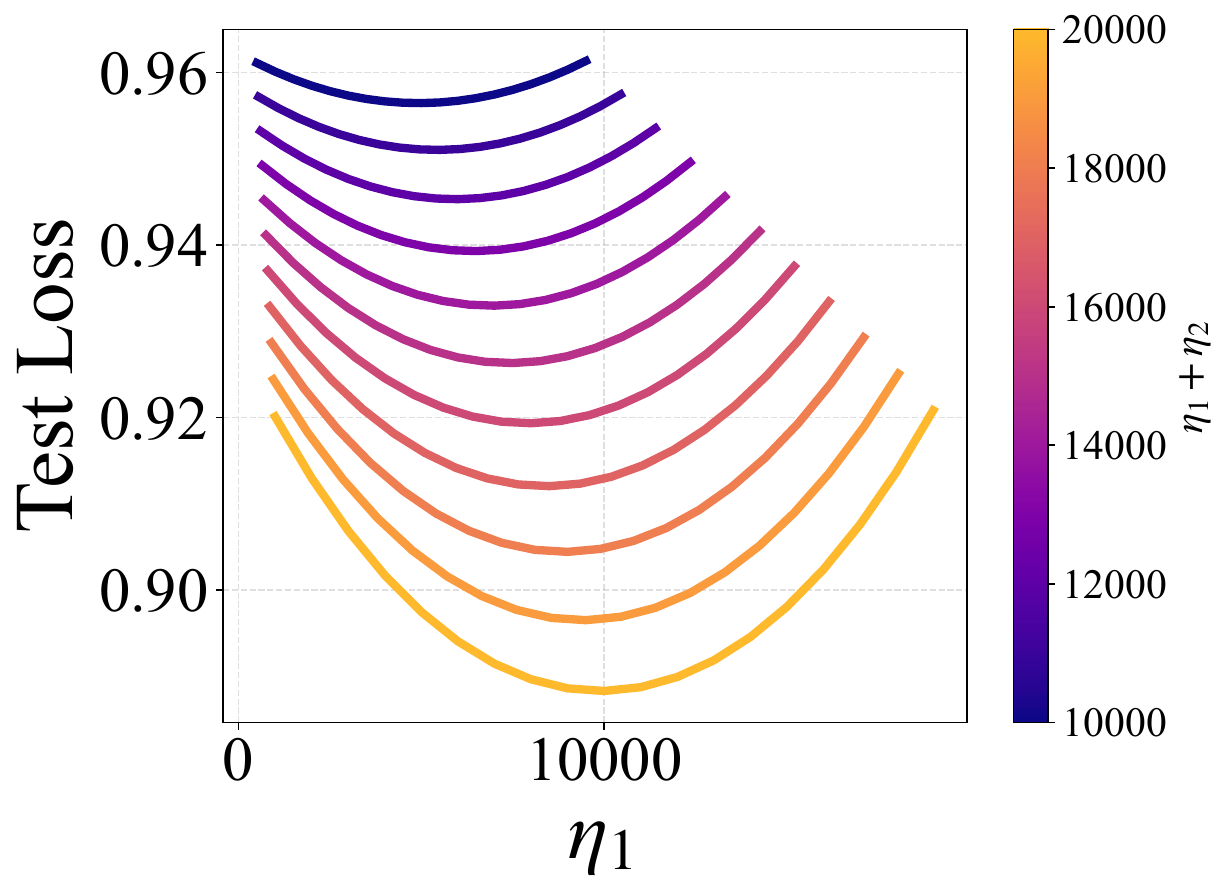}
        \caption{  Step=2 }
    \end{subfigure}
    \hfill
        \begin{subfigure}[t]{0.21\linewidth}
        \centering
        \includegraphics[width=\textwidth]{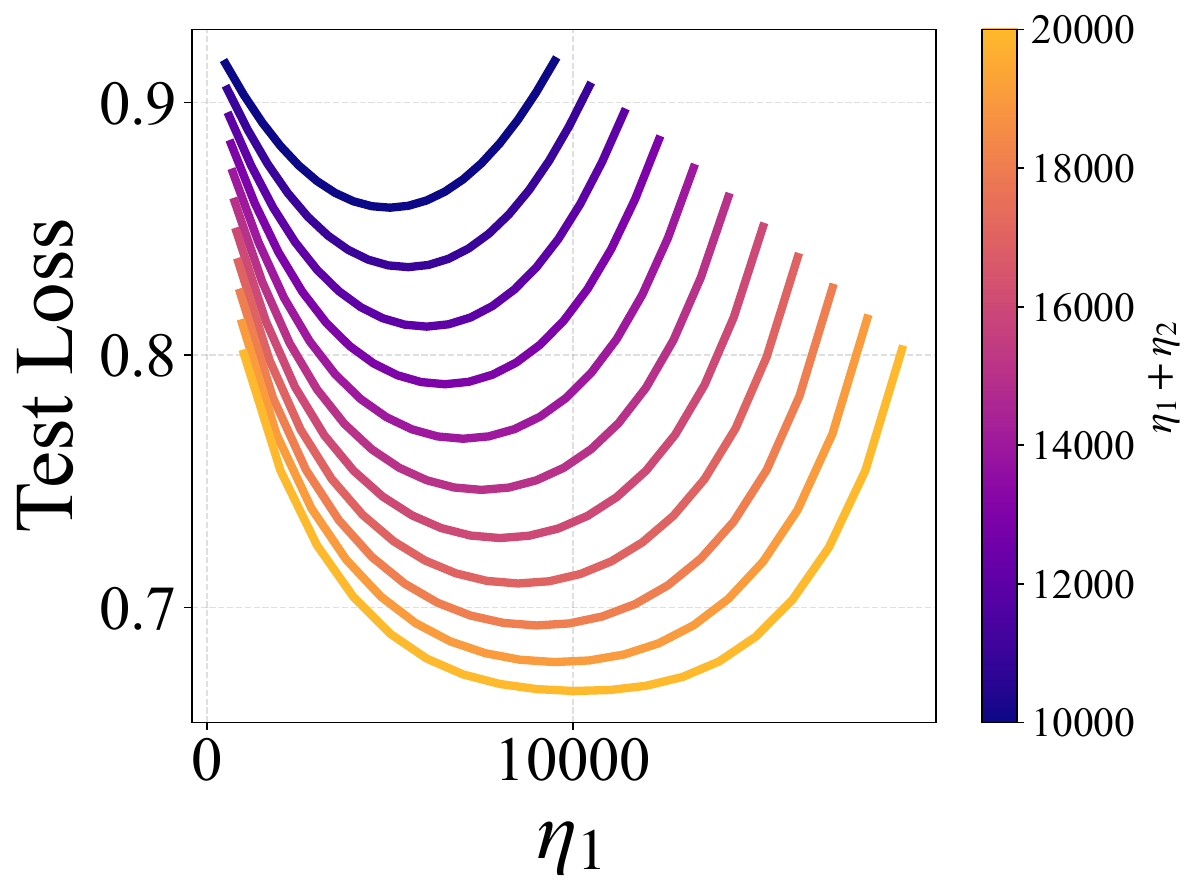}
        \caption{  Step=4}
    \end{subfigure}
    \hfill
        \begin{subfigure}[t]{0.21\linewidth}
        \centering
        \includegraphics[width=\textwidth]{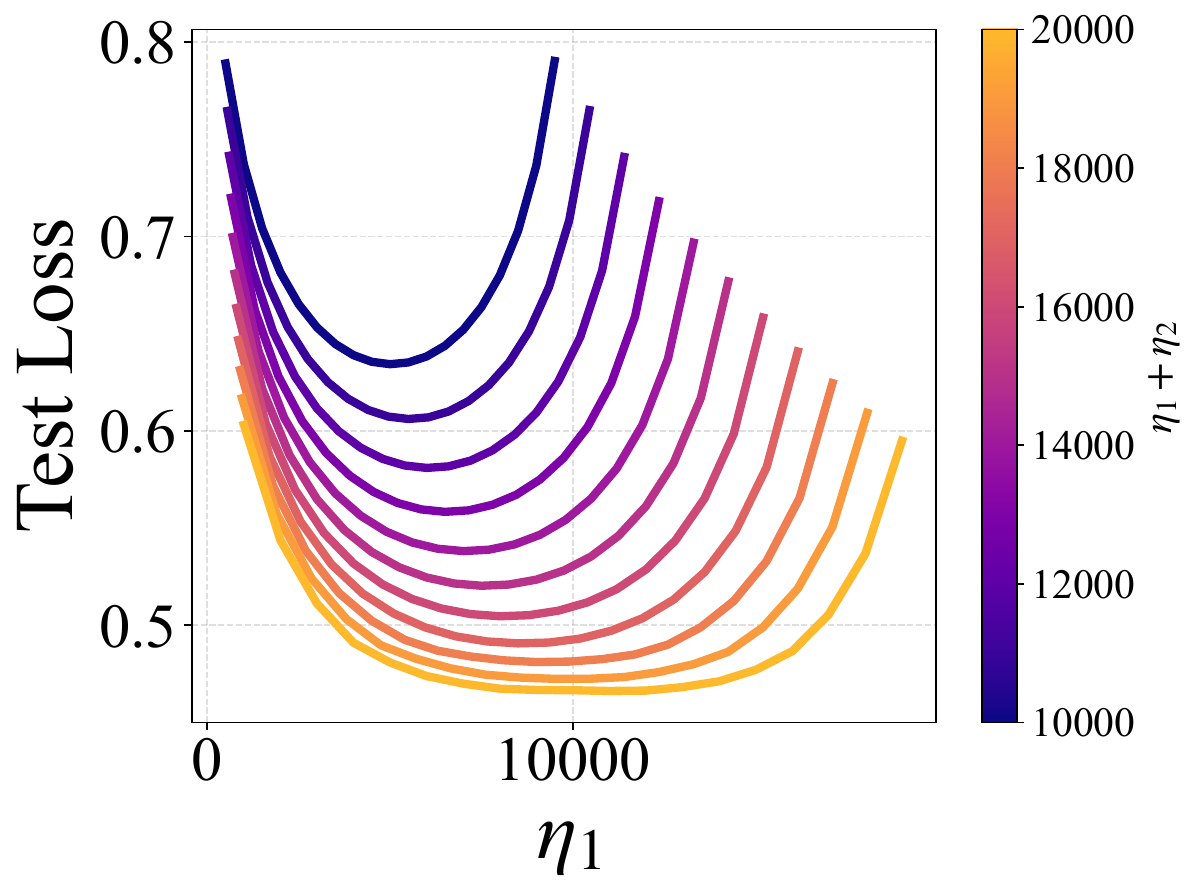}
        \caption{  Step=8}
    \end{subfigure}
    \hfill
    \caption{ \textbf{More-steps-empirical-loss for 2-layer NN  under Gaussian initialization.} 
Here we set $\eta_1+\eta_2\leq O(h^{\frac{3}{2}})$  and  $h=1000$. }
\label{fig:2-NN-more-steps-gaussian}
\end{figure*}

\begin{figure*}[!htb]
    \centering
    \begin{subfigure}[t]{0.22\linewidth}
        \centering
        \includegraphics[width=\textwidth]{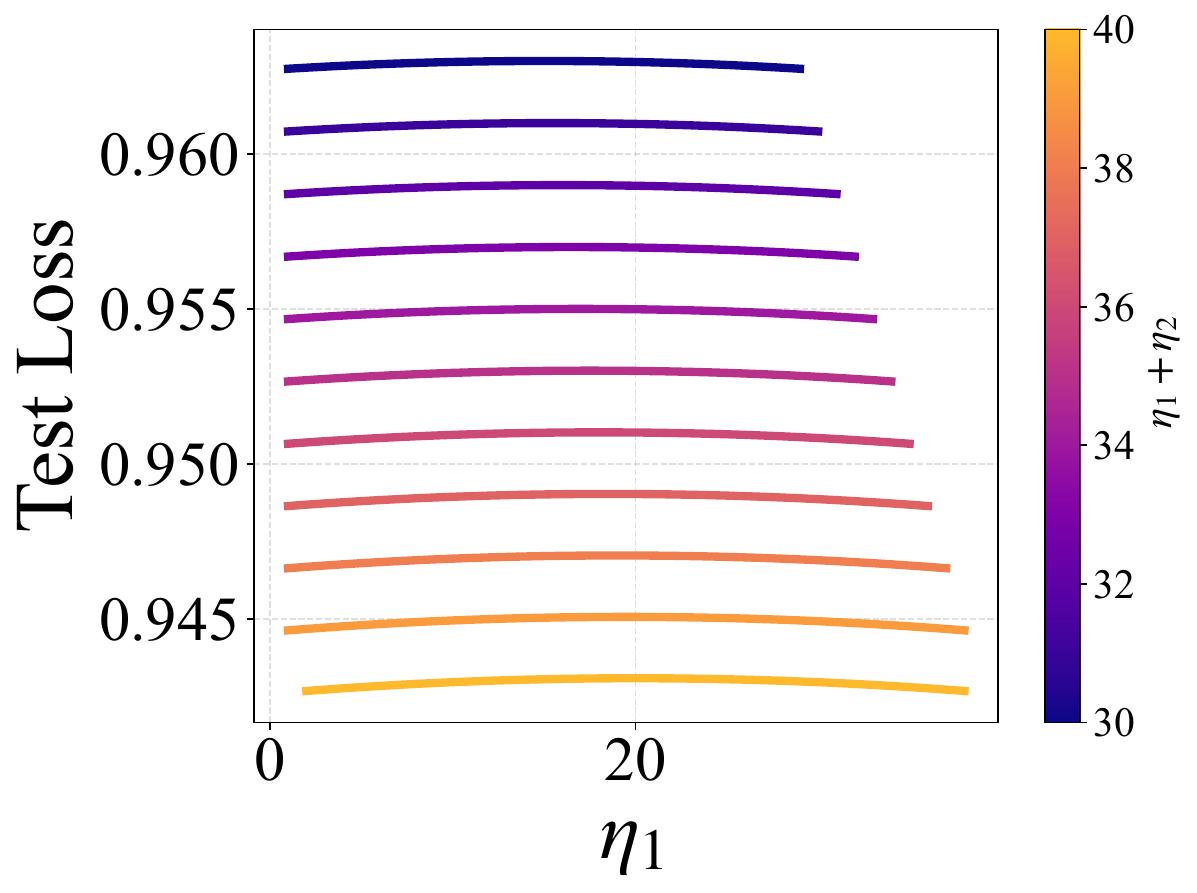}
        \caption{  Step=1 }
    \label{fig:2-layer}
    \end{subfigure}
    \hfill
    \begin{subfigure}[t]{0.21\linewidth}
        \centering
        \includegraphics[width=\textwidth]{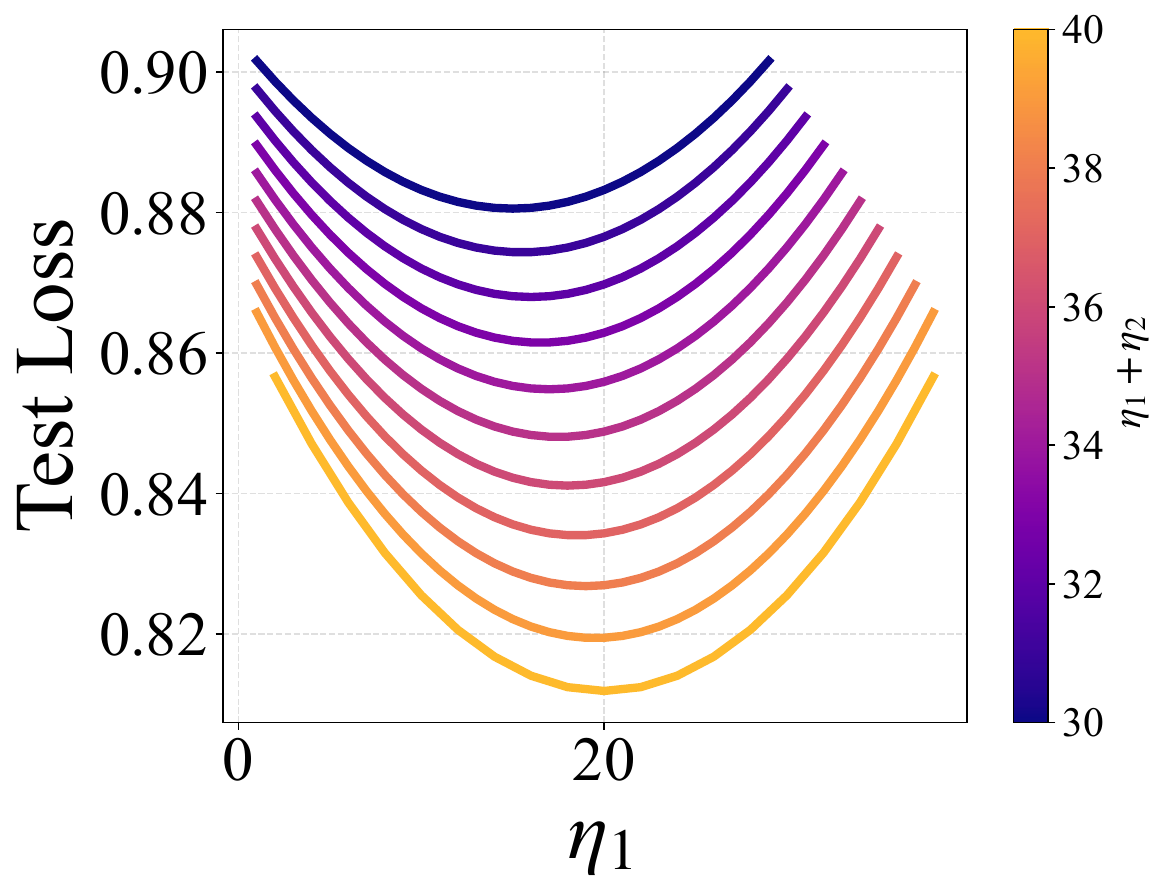}
        \caption{  Step=2 }
    \end{subfigure}
    \hfill
        \begin{subfigure}[t]{0.21\linewidth}
        \centering
        \includegraphics[width=\textwidth]{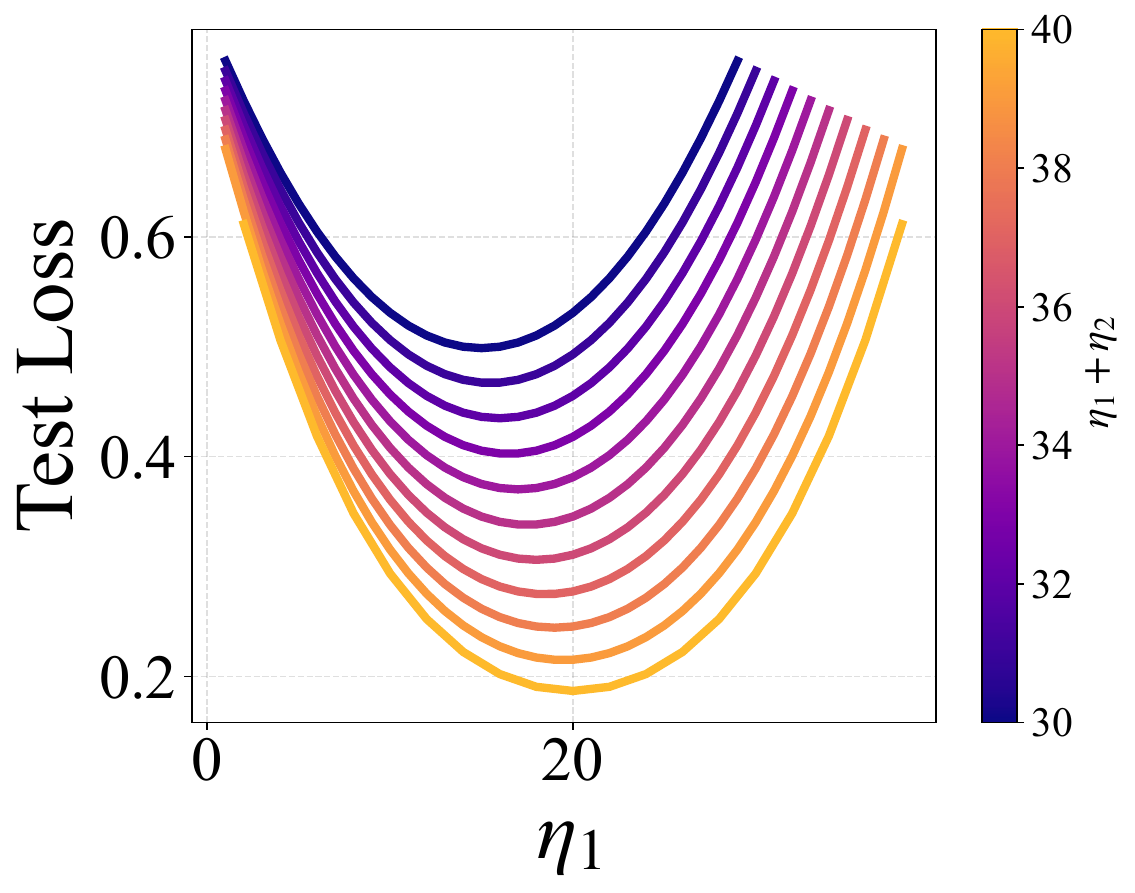}
        \caption{  Step=4}
    \end{subfigure}
    \hfill
        \begin{subfigure}[t]{0.21\linewidth}
        \centering
        \includegraphics[width=\textwidth]{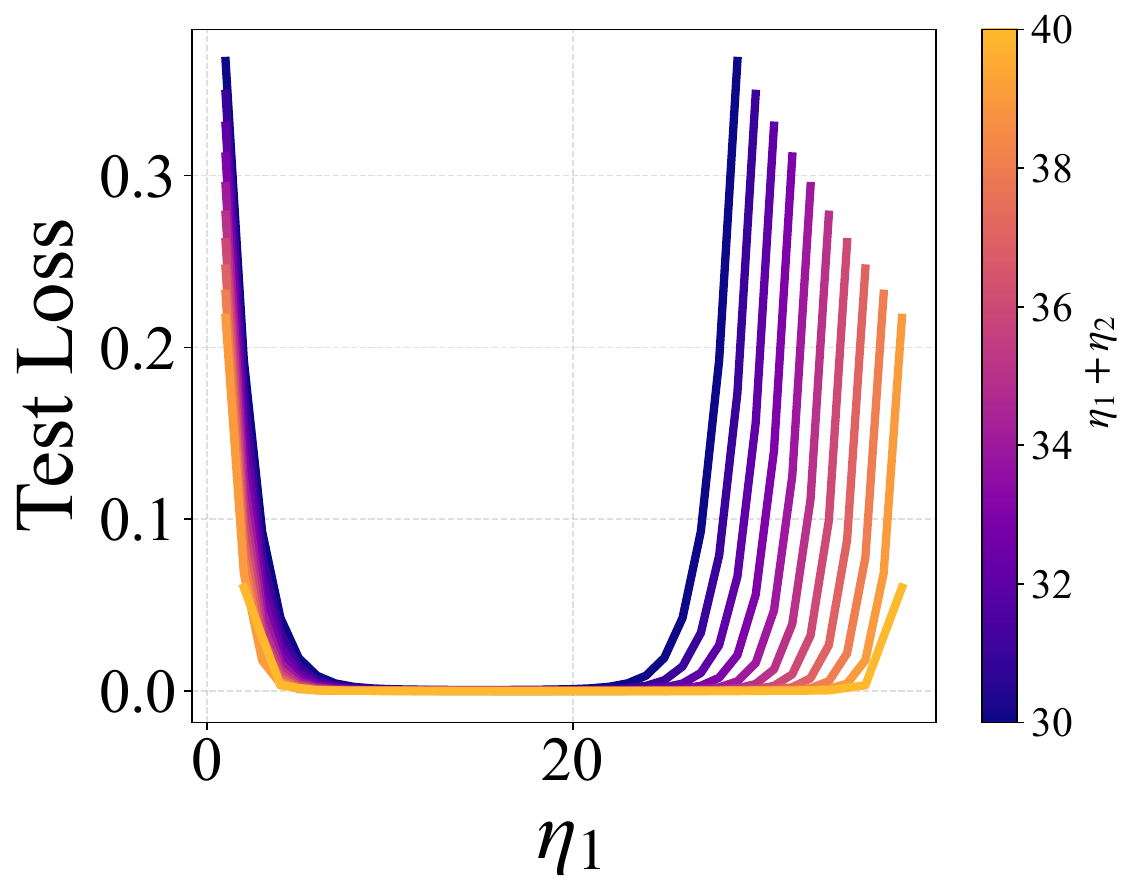}
        \caption{  Step=8}
    \end{subfigure}
    \hfill
    \caption{ \textbf{More-steps-empirical-loss for 3-layer NN  under Orthogonal initialization.} Here we set $\eta_1+\eta_2\leq O(h^{\frac{2}{3}})$  and  $h=1000$.
}
\label{fig:3-NN-more-steps-orthog}
\end{figure*}

\begin{figure*}[!htb]
    \centering
    \begin{subfigure}[t]{0.22\linewidth}
        \centering
        \includegraphics[width=\textwidth]{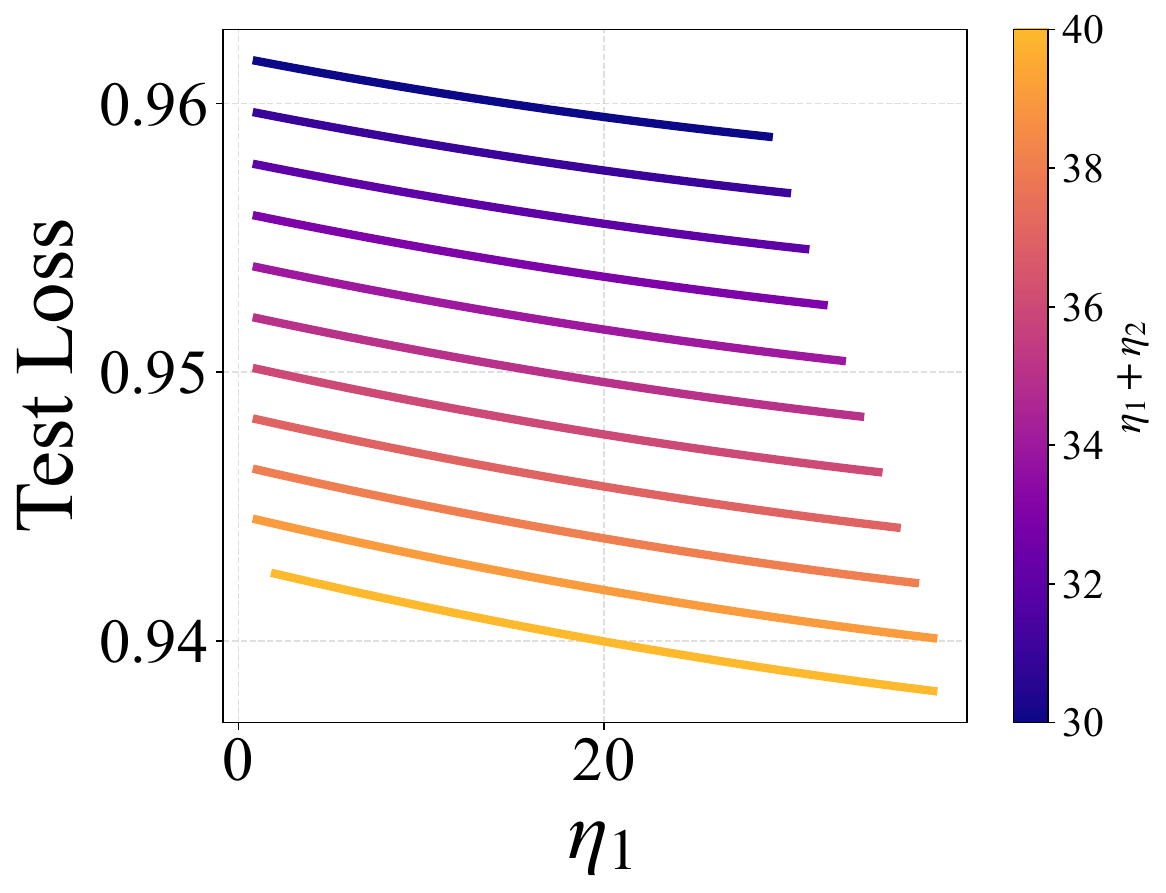}
        \caption{  Step=1 }
    \label{fig:2-layer}
    \end{subfigure}
    \hfill
    \begin{subfigure}[t]{0.21\linewidth}
        \centering
        \includegraphics[width=\textwidth]{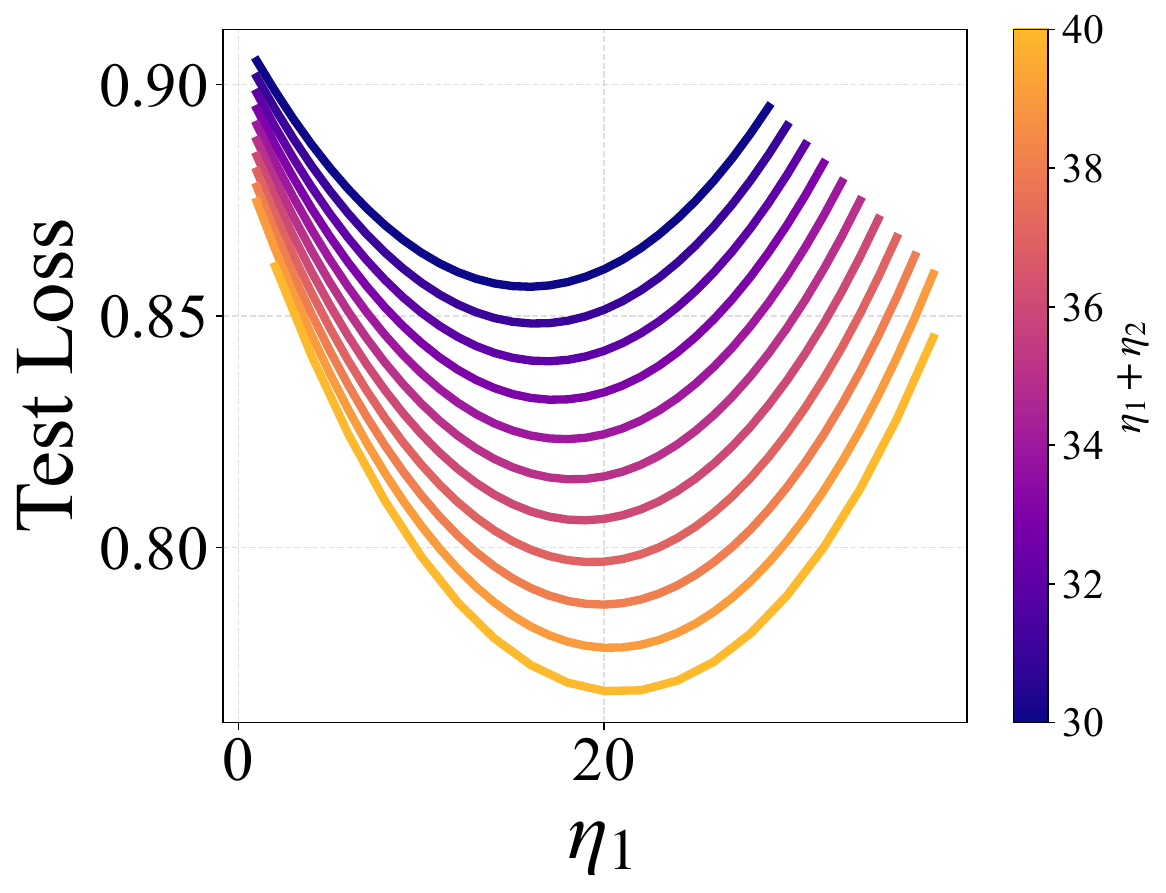}
        \caption{  Step=2}
    \end{subfigure}
    \hfill
        \begin{subfigure}[t]{0.21\linewidth}
        \centering
        \includegraphics[width=\textwidth]{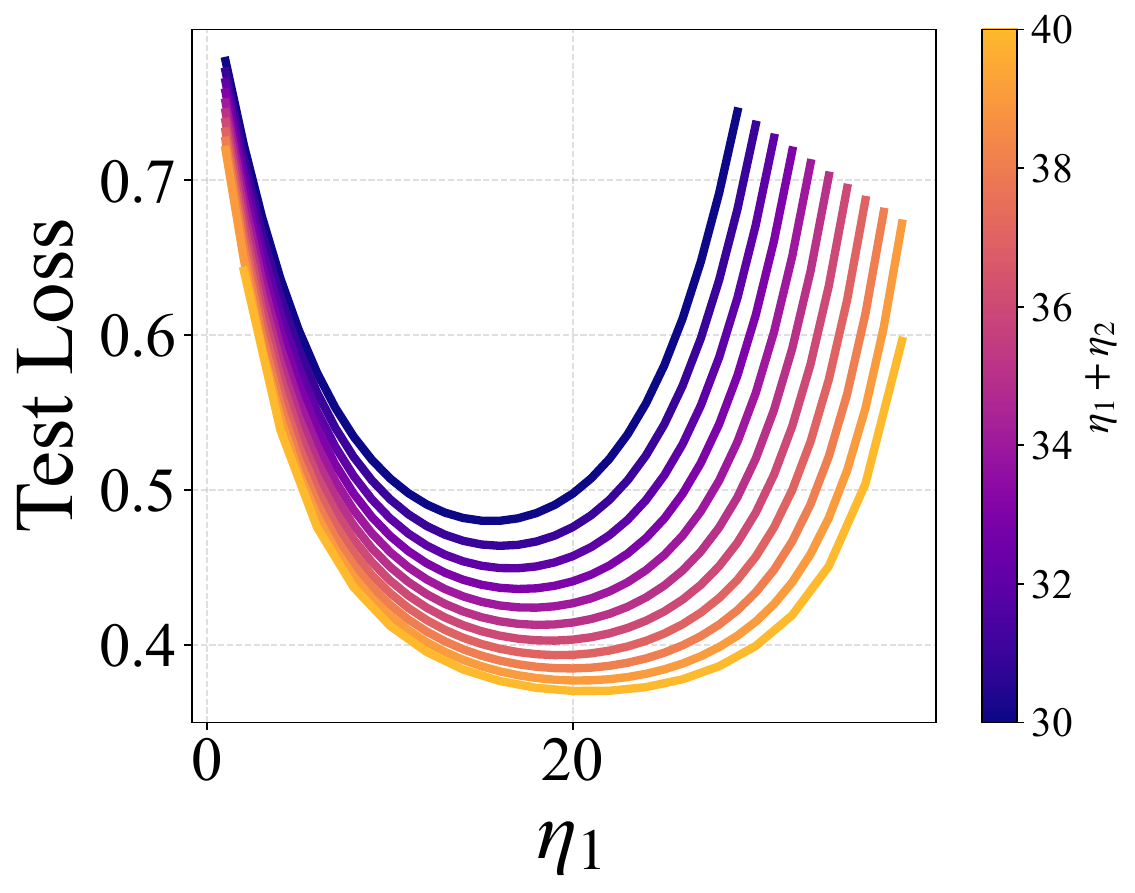}
        \caption{ Step=4}
    \end{subfigure}
    \hfill
        \begin{subfigure}[t]{0.21\linewidth}
        \centering
        \includegraphics[width=\textwidth]{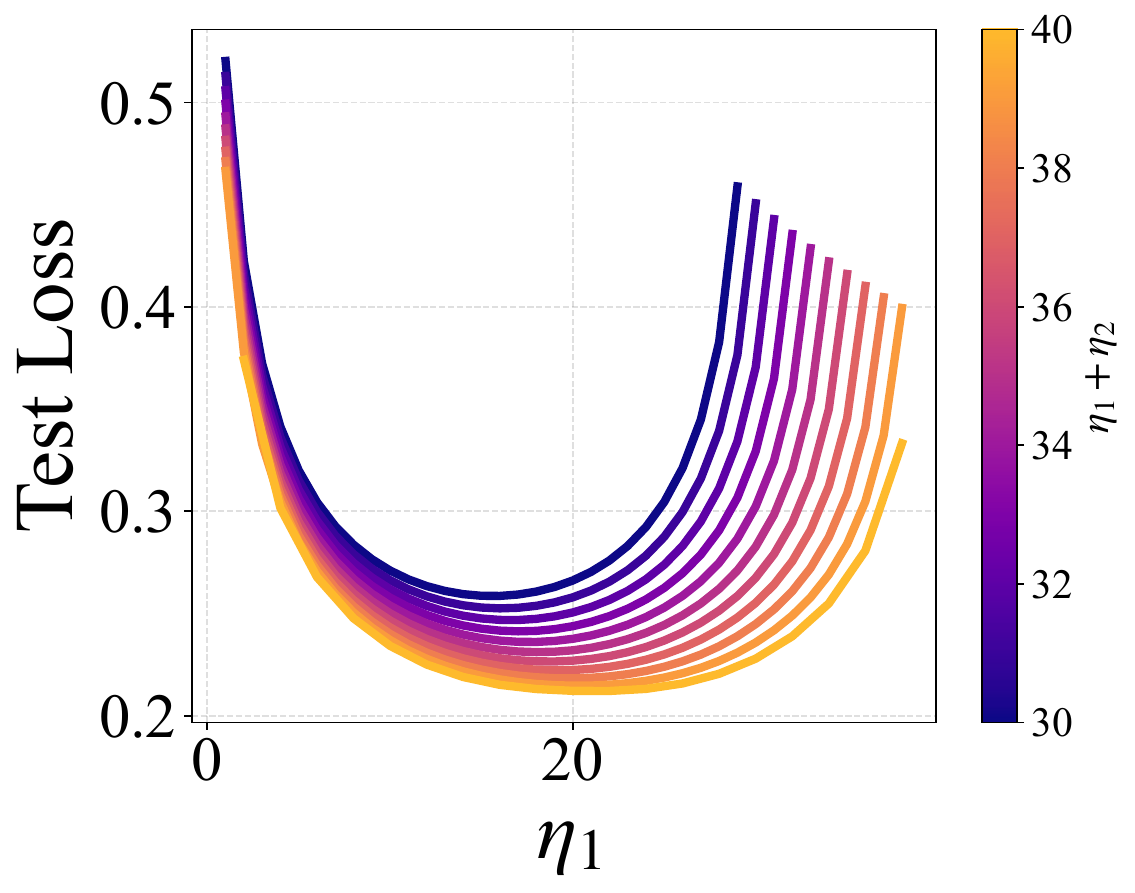}
        \caption{  Step=8}
    \end{subfigure}
    \hfill
    \caption{ \textbf{More-steps-empirical-loss for 3-layer NN  under Gaussian initialization.} 
Here we set $\eta_1+\eta_2\leq O(h^{\frac{2}{3}})$  and  $h=1000$.}
\label{fig:3-NN-more-steps-gaussian}
\end{figure*}

\begin{figure*}[!htb]
    \centering
    \begin{subfigure}[t]{0.23\linewidth}
        \centering
        \includegraphics[width=\textwidth]{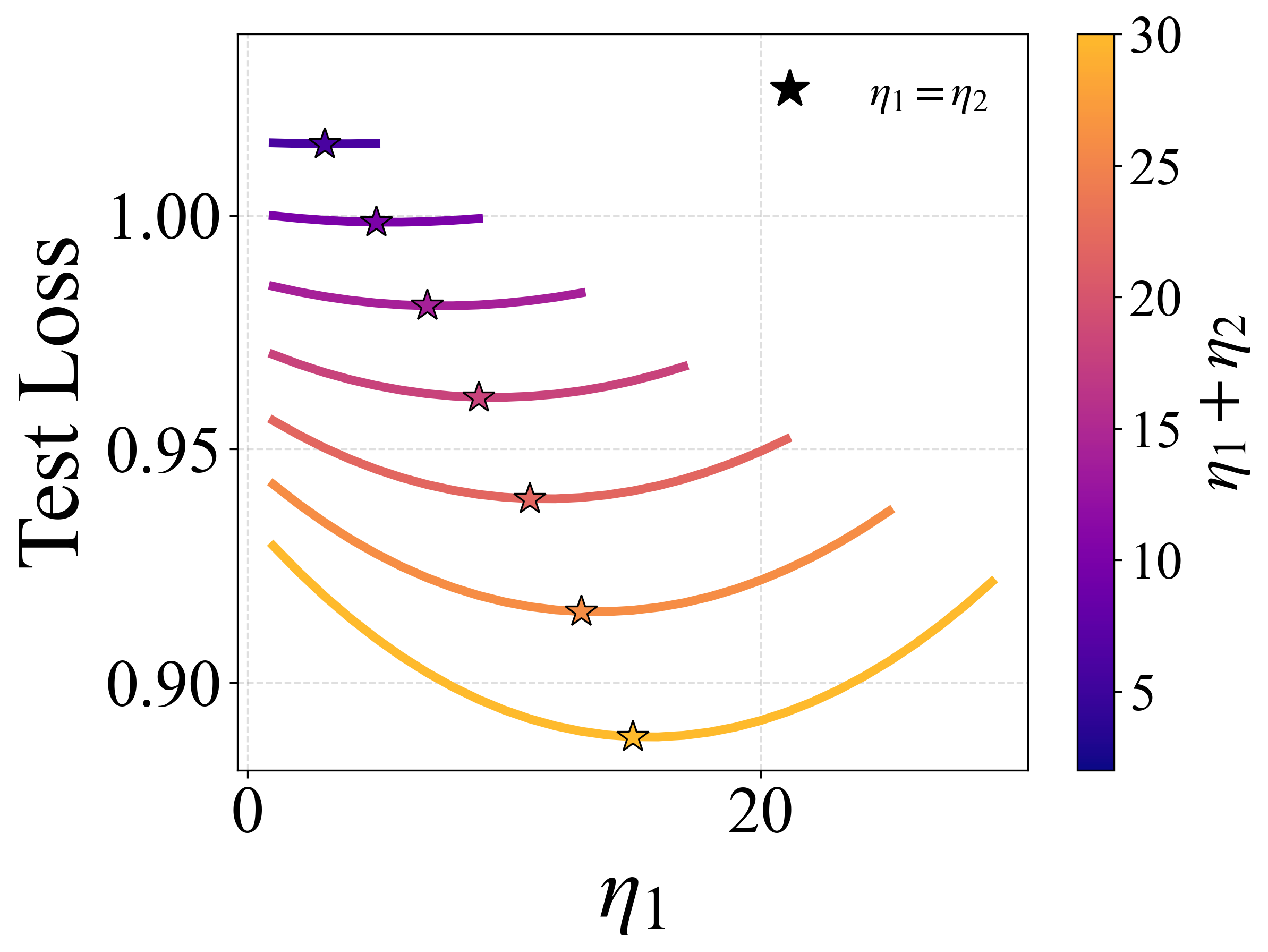}
        \caption{  Step=2 }
    \label{fig:2-layer}
    \end{subfigure}
    \hfill
    \begin{subfigure}[t]{0.23\linewidth}
        \centering
        \includegraphics[width=\textwidth]{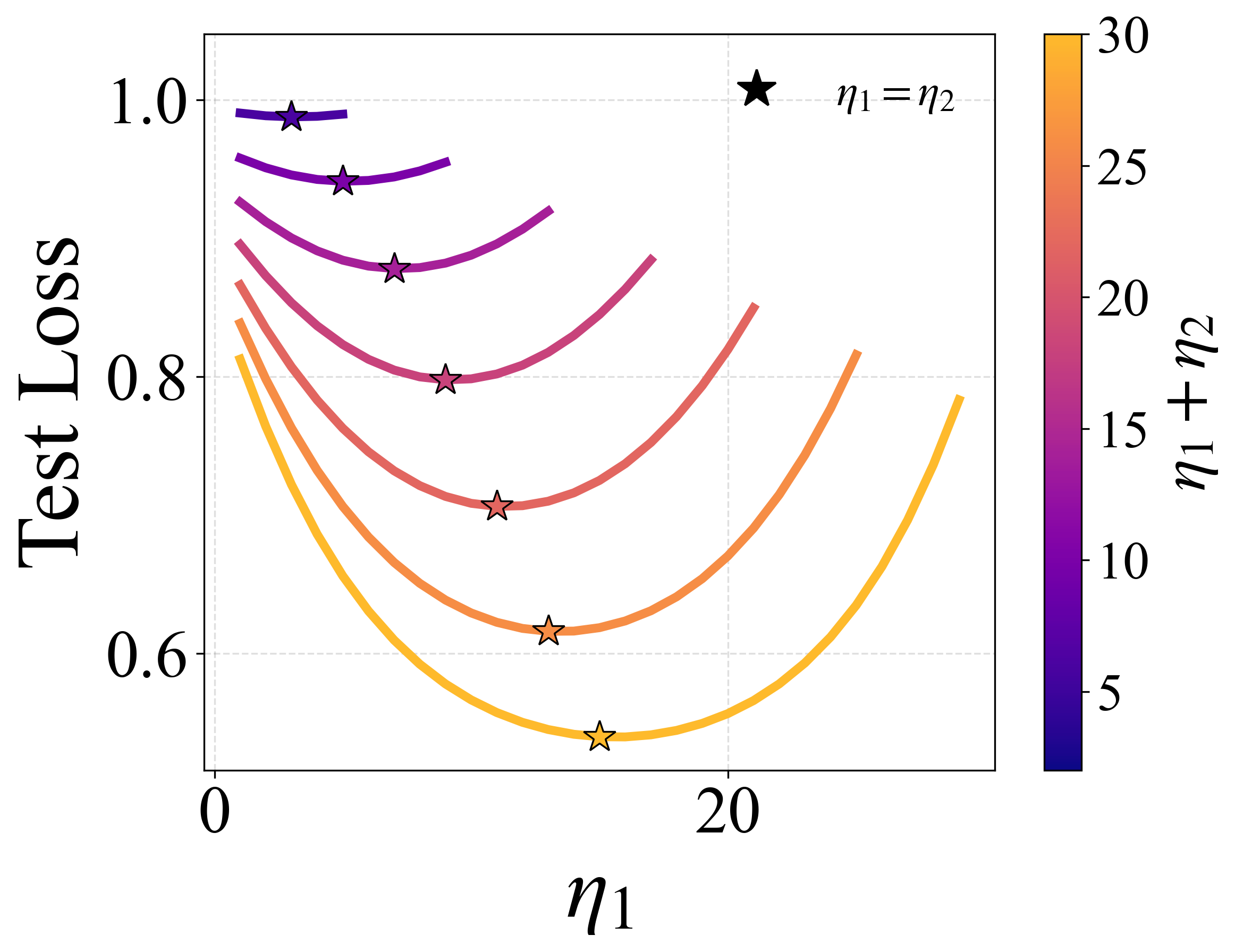}
        \caption{  Step=4}
    \end{subfigure}
    \hfill
        \begin{subfigure}[t]{0.23\linewidth}
        \centering
        \includegraphics[width=\textwidth]{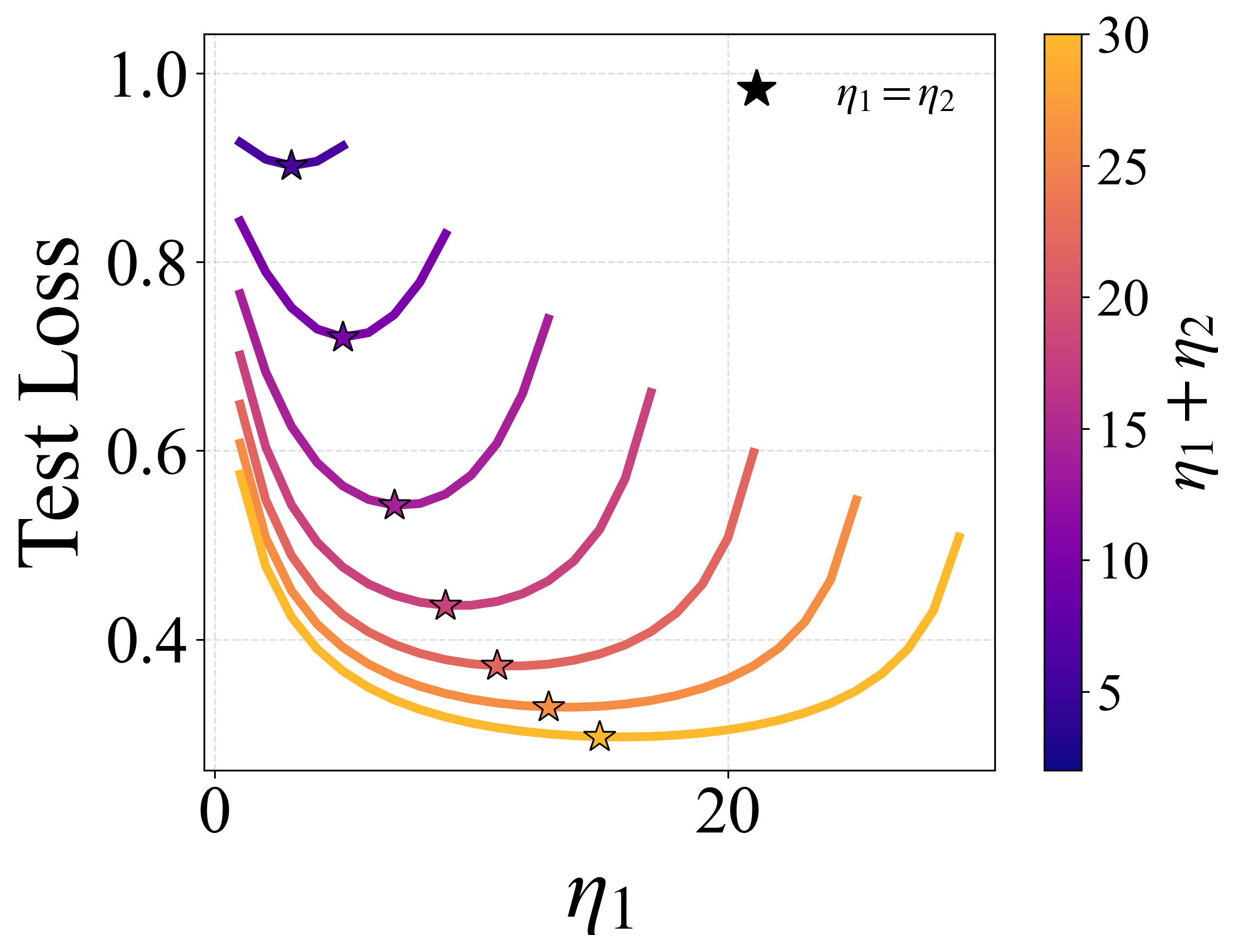}
        \caption{ Step=8}
    \end{subfigure}
    \hfill
        \begin{subfigure}[t]{0.23\linewidth}
        \centering
        \includegraphics[width=\textwidth]{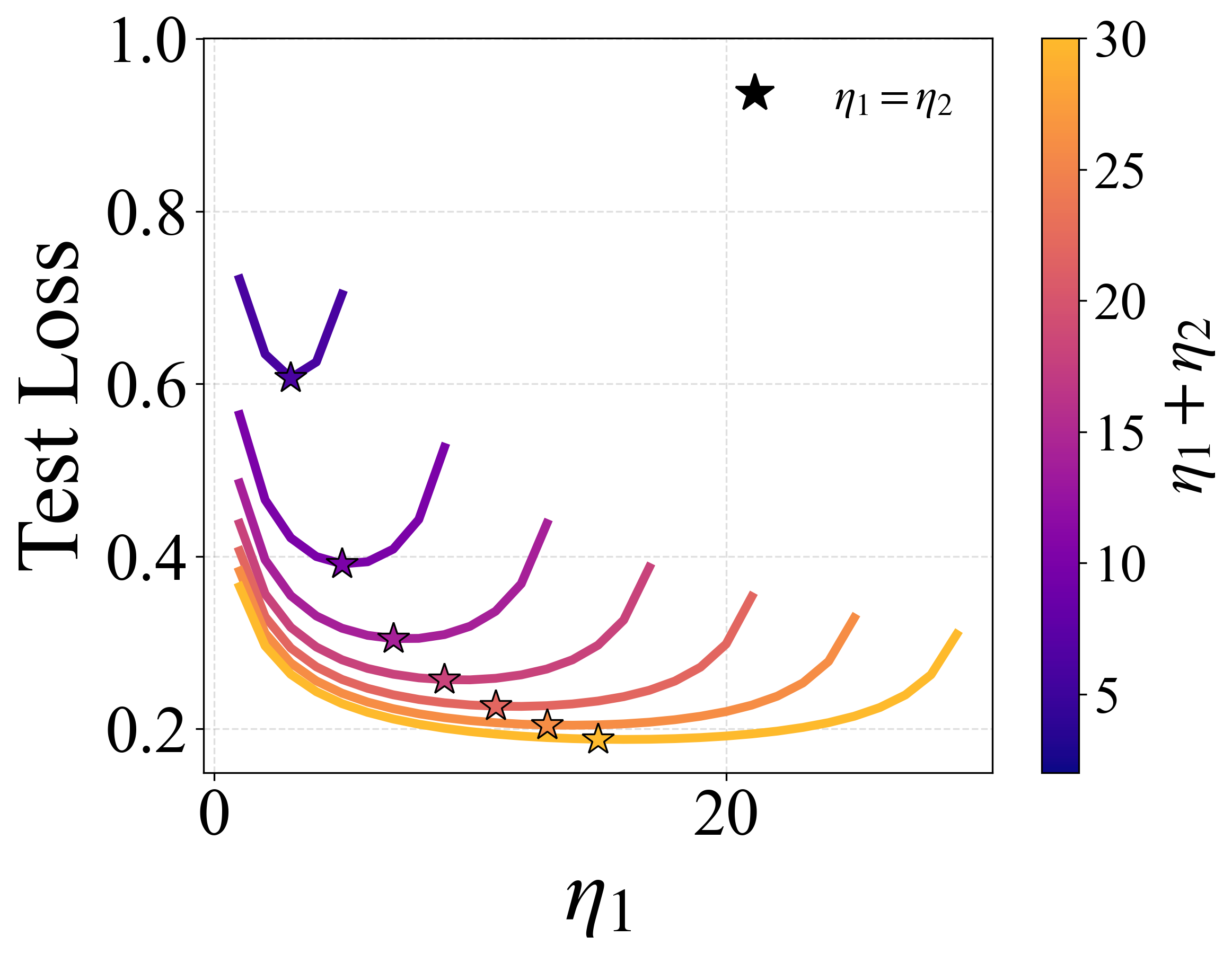}
        \caption{  Step=16}
    \end{subfigure}
    \hfill
    \begin{subfigure}[t]{0.23\linewidth}
        \centering
        \includegraphics[width=\textwidth]{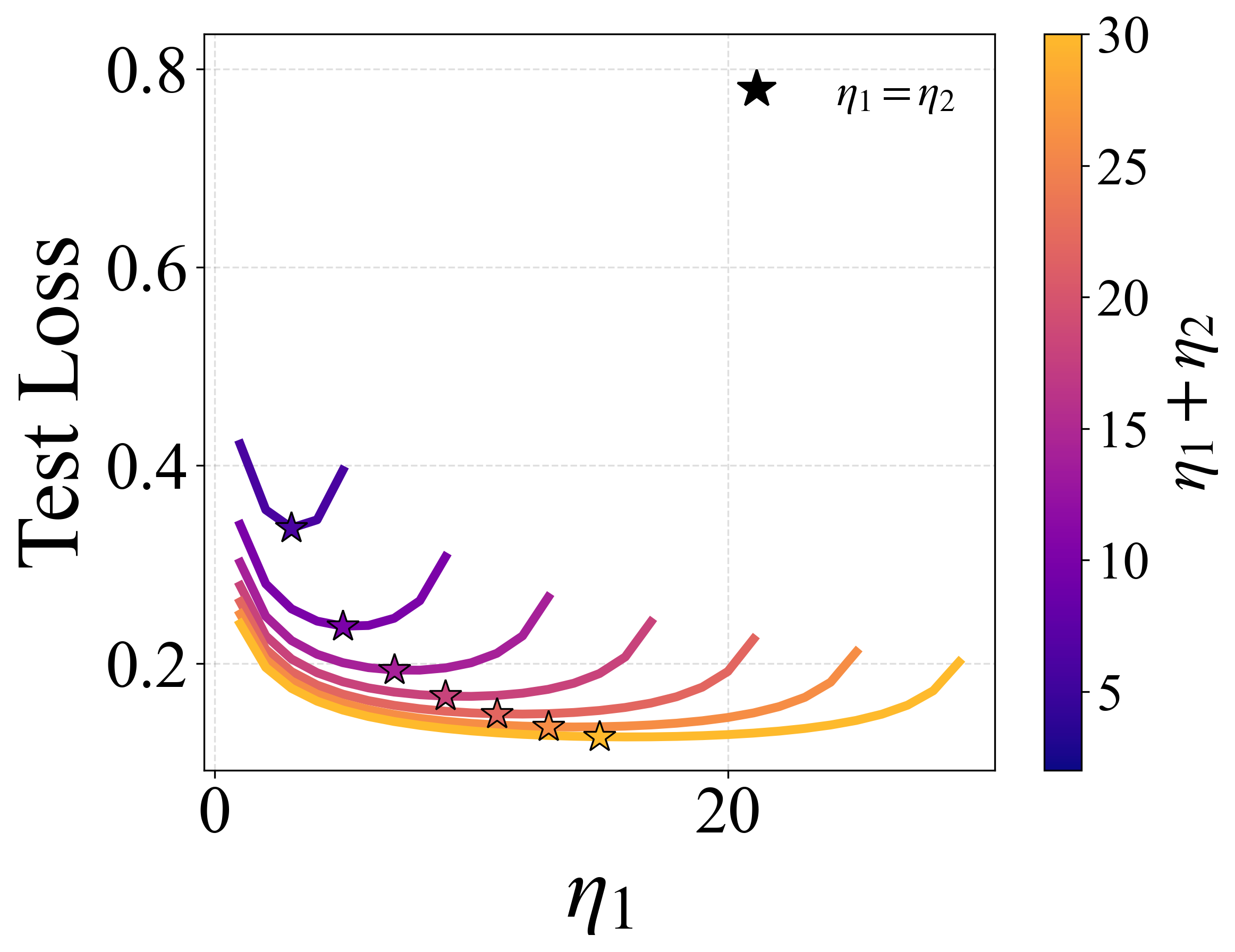}
        \caption{  Step=32 }
    \end{subfigure}
    \hfill
    \begin{subfigure}[t]{0.23\linewidth}
        \centering
        \includegraphics[width=\textwidth]{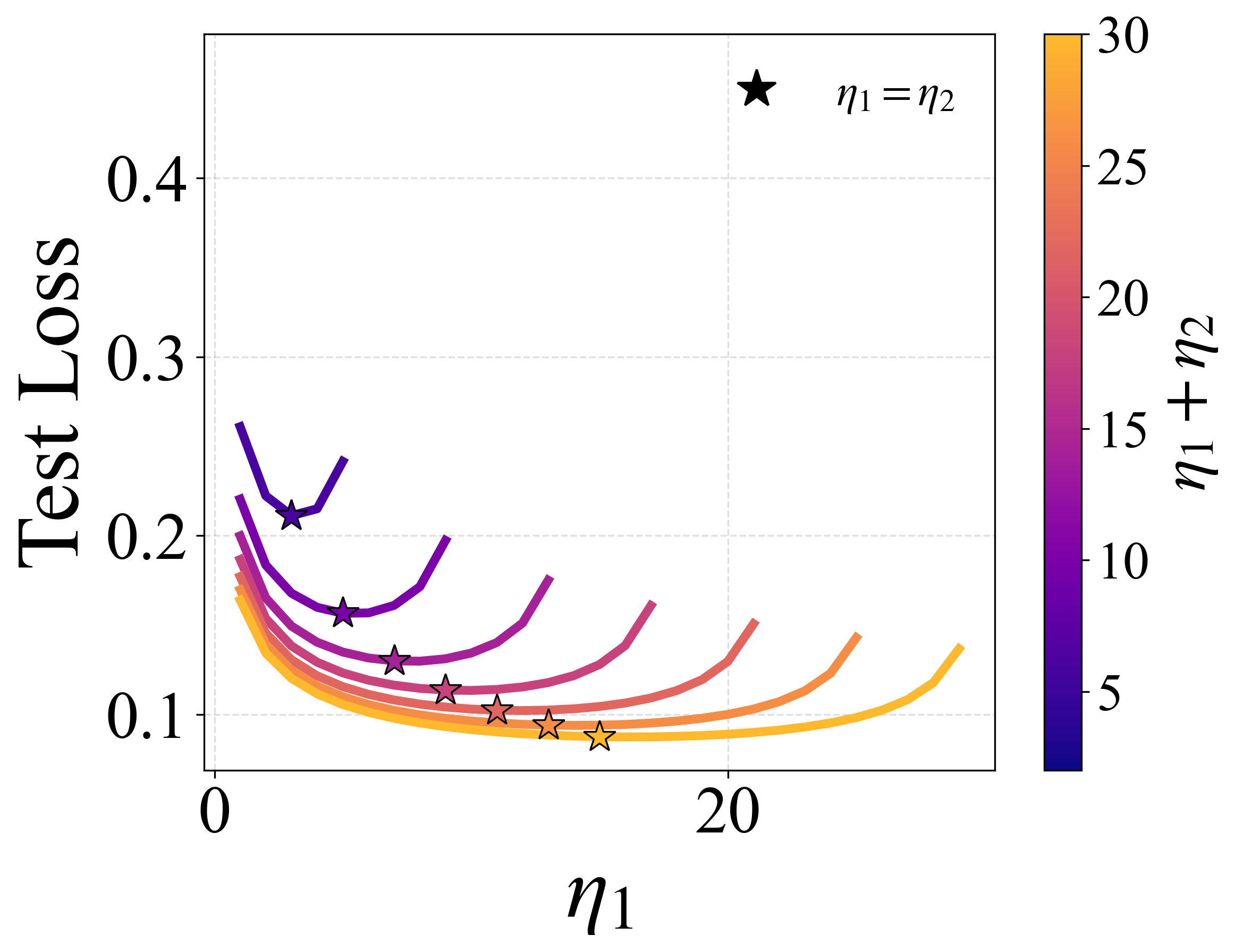}
        \caption{  Step=64}
    \end{subfigure}
    \hfill
        \begin{subfigure}[t]{0.23\linewidth}
        \centering
        \includegraphics[width=\textwidth]{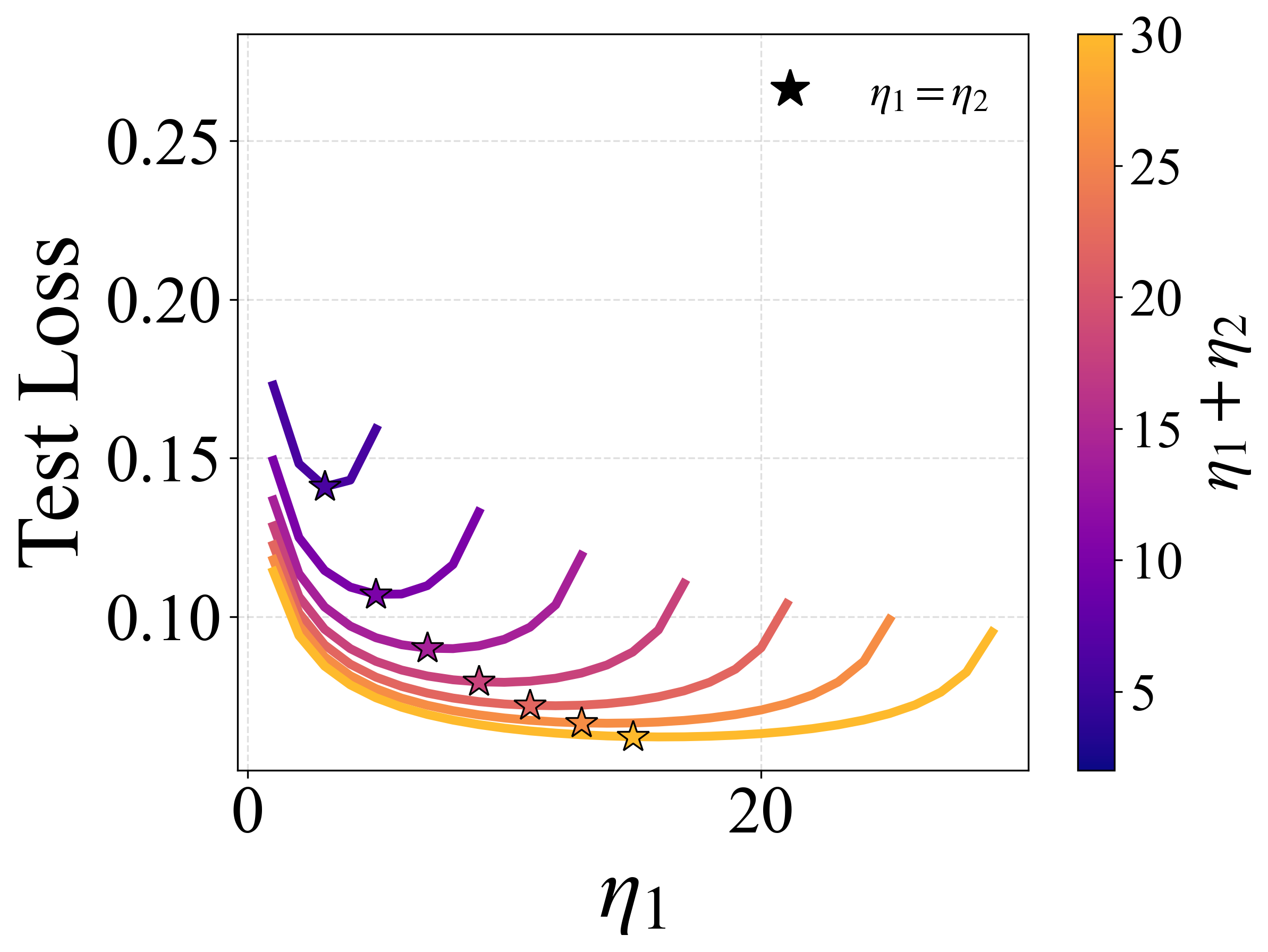}
        \caption{ Step=128}
    \end{subfigure}
    \hfill
        \begin{subfigure}[t]{0.23\linewidth}
        \centering
        \includegraphics[width=\textwidth]{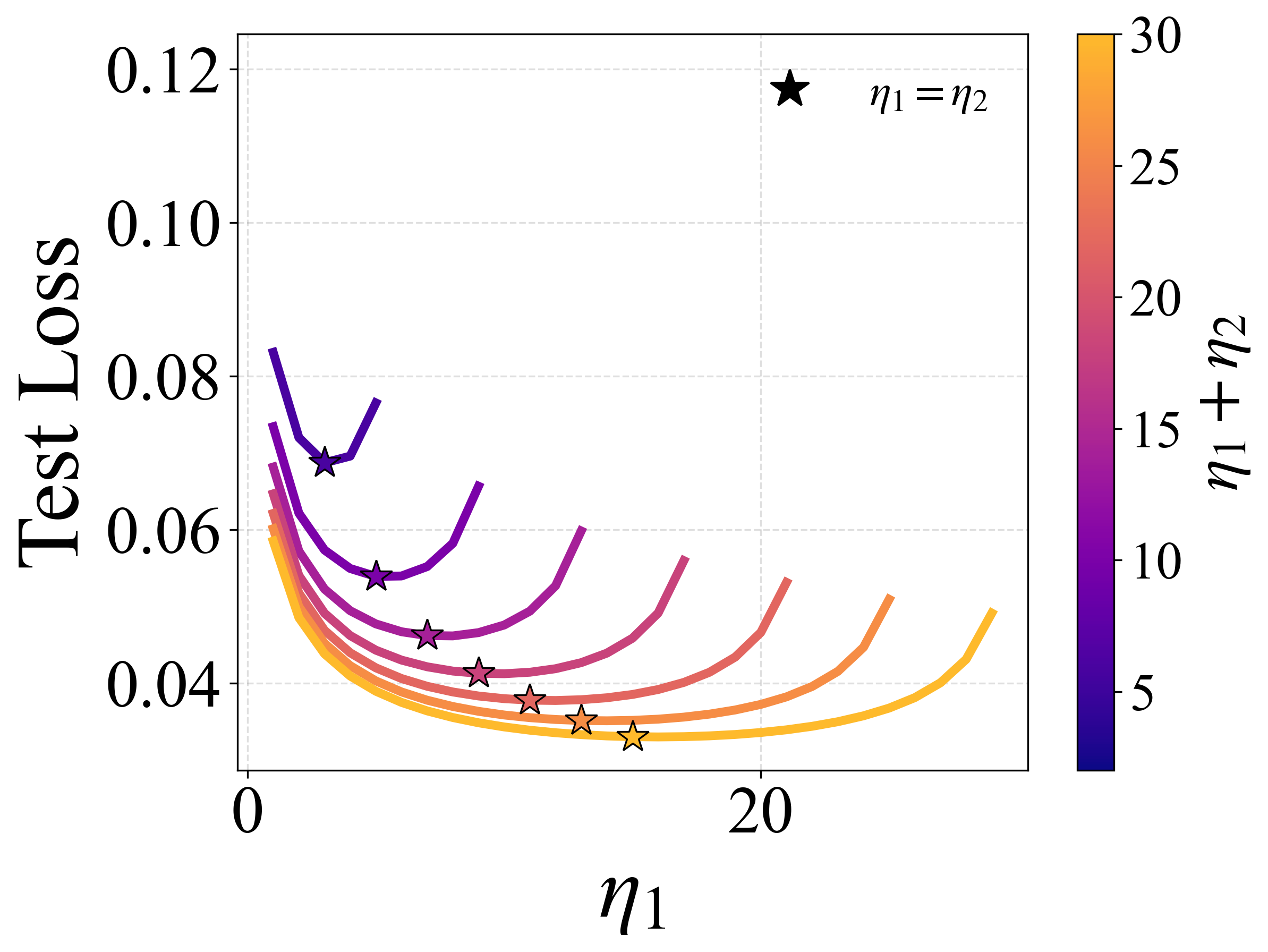}
        \caption{  Step=512}
    \end{subfigure}
    \hfill
    \caption{ \textbf{More-steps-empirical-loss for 3-layer NN  under Orthogonal initialization up to 512 steps.} 
Here we set $\eta_1+\eta_2\leq O(h^{\frac{2}{3}})$  and  $h=1000$.}
\label{fig:3-NN-512-steps-orthog}
\end{figure*}

\begin{figure*}[!htb]
    \centering
    \begin{subfigure}[t]{0.22\linewidth}
        \centering
        \includegraphics[width=\textwidth]{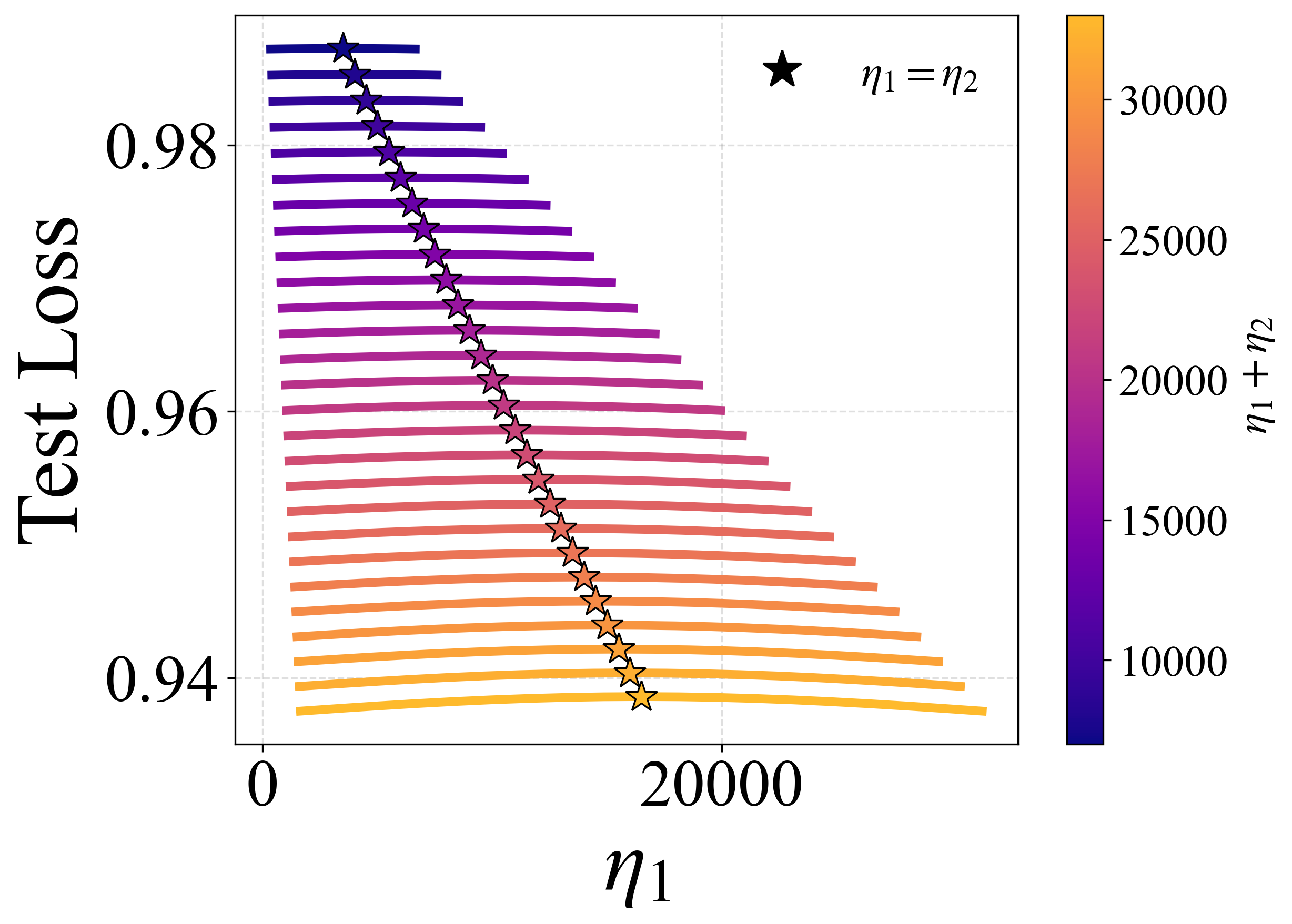}
        \caption{  2-layer NN \\Step=1 }
    \label{fig:2-layer}
    \end{subfigure}
    \hfill
    \begin{subfigure}[t]{0.21\linewidth}
        \centering
        \includegraphics[width=\textwidth]{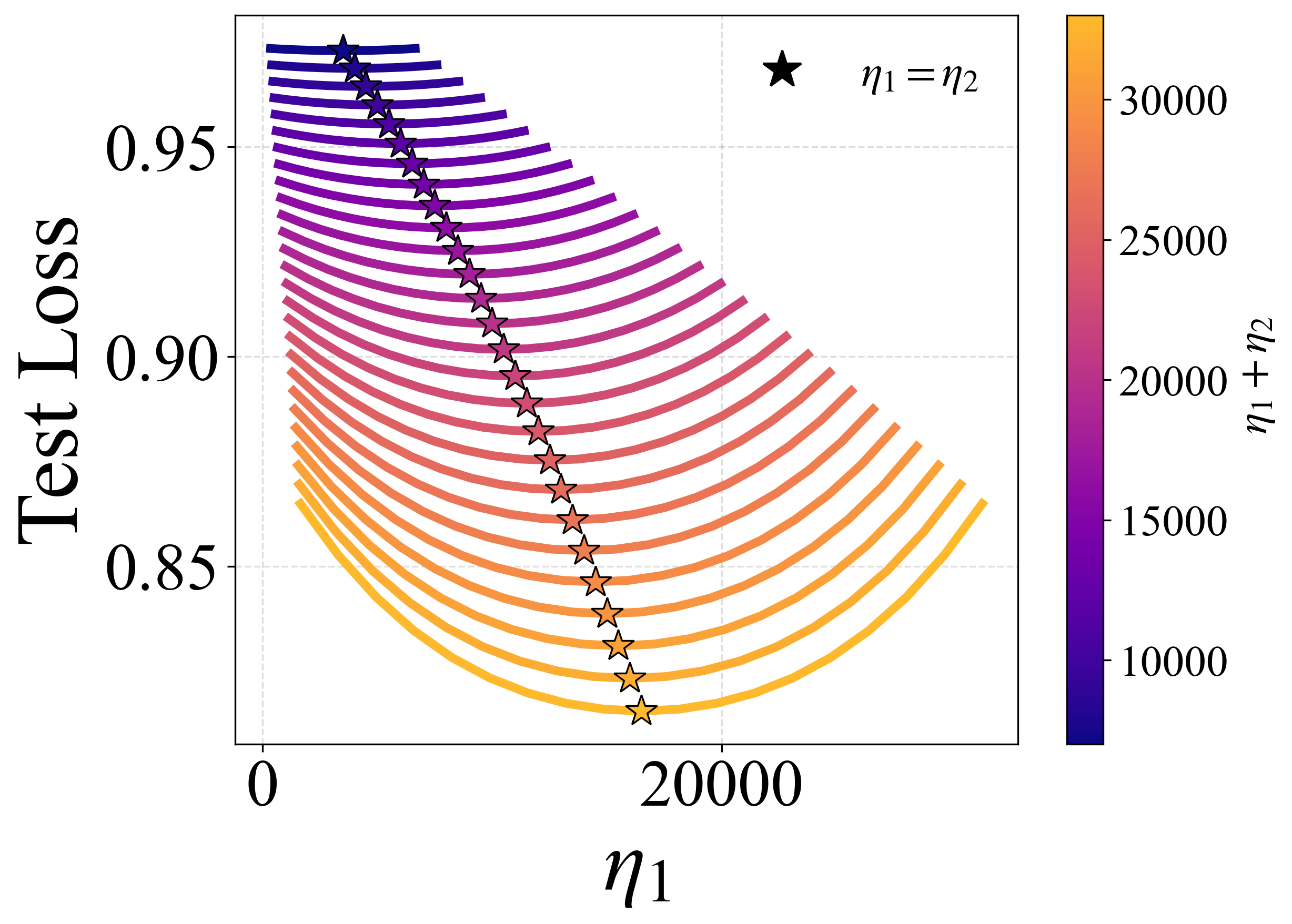}
        \caption{  2-layer NN \\Step=2 }
    \end{subfigure}
    \hfill
        \begin{subfigure}[t]{0.21\linewidth}
        \centering
        \includegraphics[width=\textwidth]{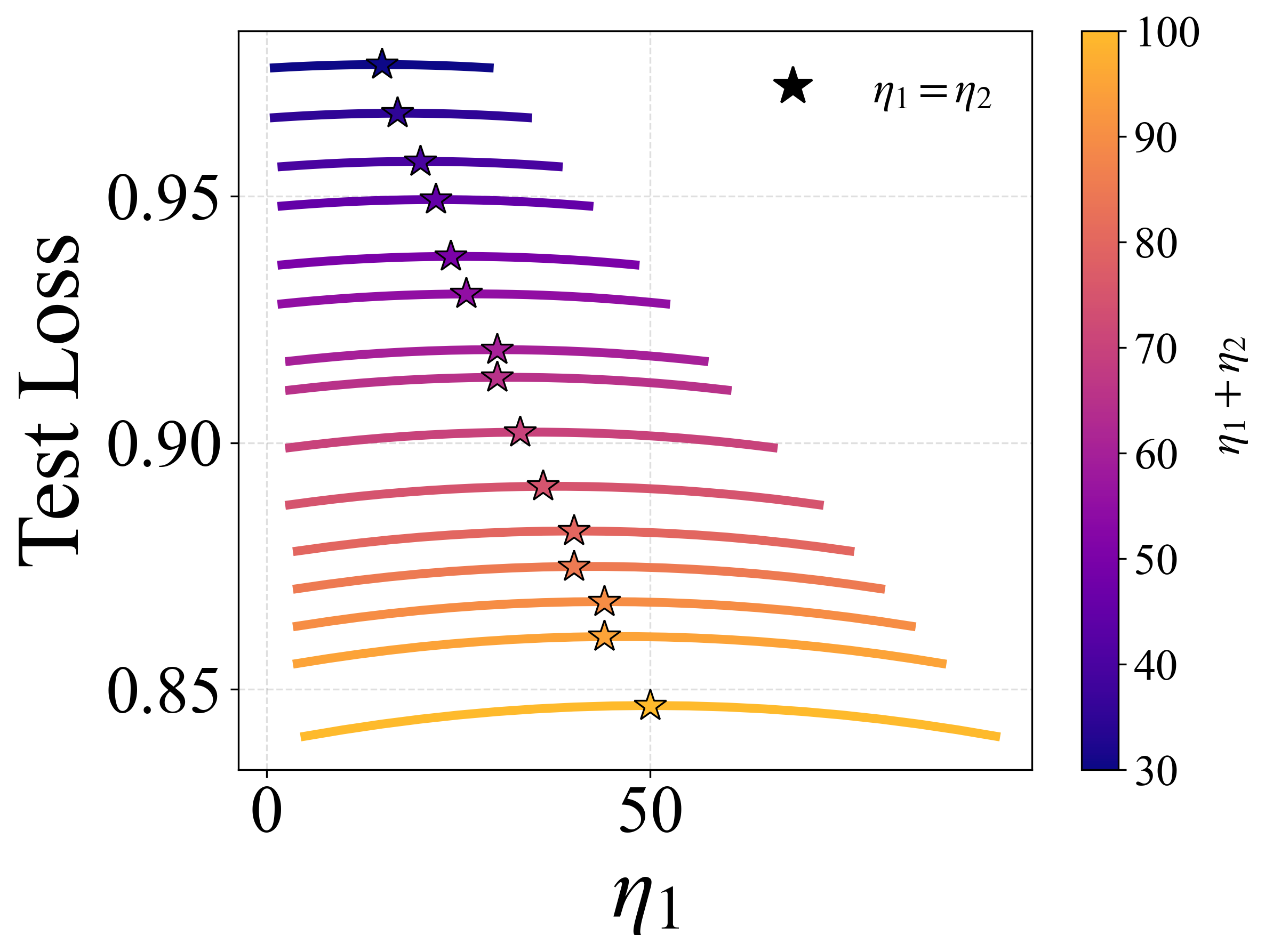}
        \caption{   3-layer NN \\Step=1}
    \end{subfigure}
    \hfill
        \begin{subfigure}[t]{0.21\linewidth}
        \centering
        \includegraphics[width=\textwidth]{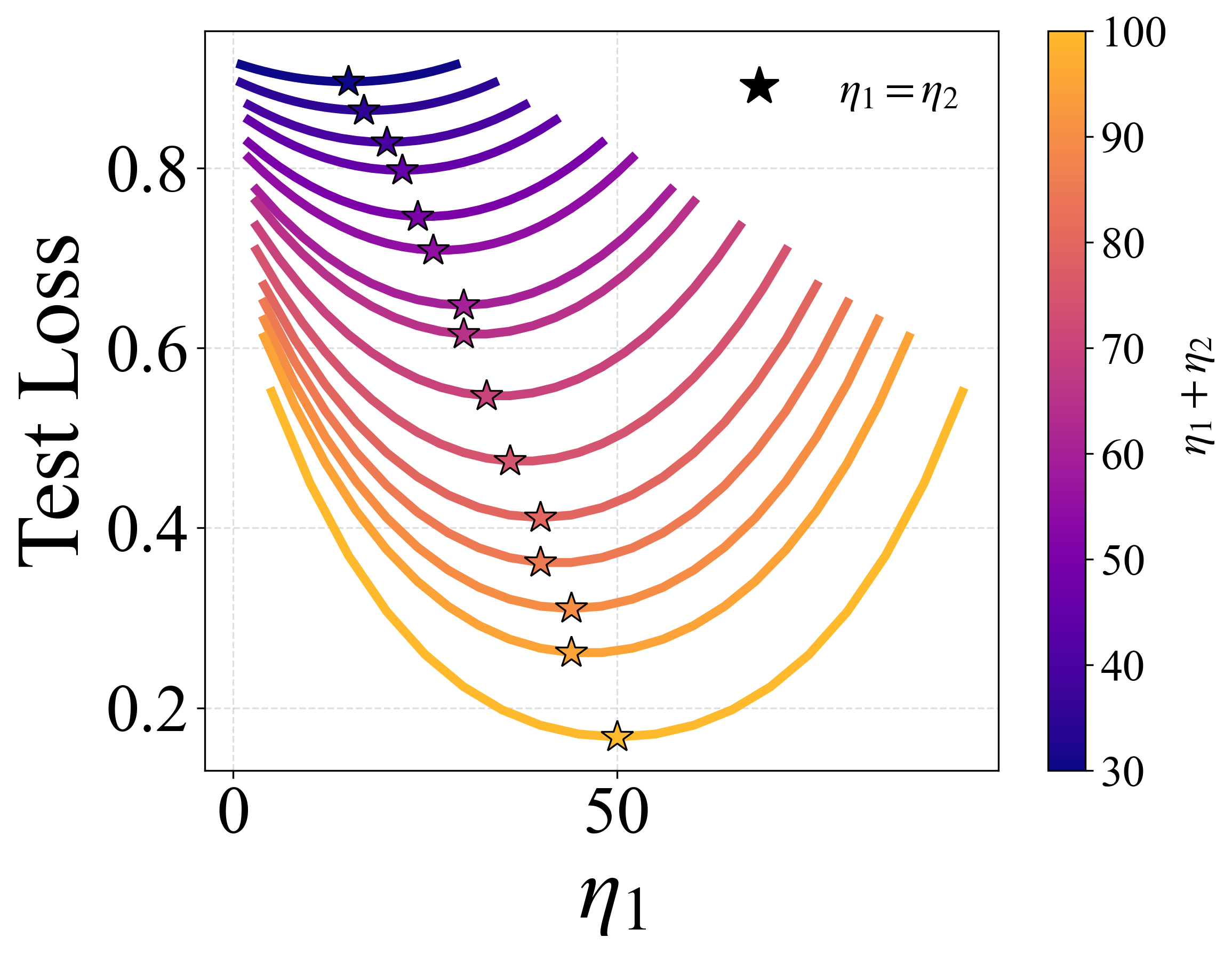}
        \caption{  3-layer NN \\Step=2 }
    \end{subfigure}
    \hfill
    \caption{ (a)(b)\textbf{2-layer NN with label noise $ \xi \in \mathcal{N}(0, \rho)$ under orthogonal initialization.} Here we set $\eta_1 + \eta_2 \leq O(h^{\frac{3}{2}})$ with $h=1000$ and $\rho=0.001$.  (c)(d)\textbf{3-layer NN with label noise $ \xi \in \mathcal{N}(0, \rho)$ under orthogonal initialization.} Here we set $\eta_1 + \eta_2 \leq O(h^{\frac{2}{3}})$ with $h=1000$ and $\rho=0.001$. 
}
\label{fig:NN-noisy-orthgonal-initialization}
\end{figure*}

\begin{figure}[tb]
    \centering
    \begin{subfigure}[t]{0.24\linewidth}
        \centering
        \includegraphics[width=\textwidth]{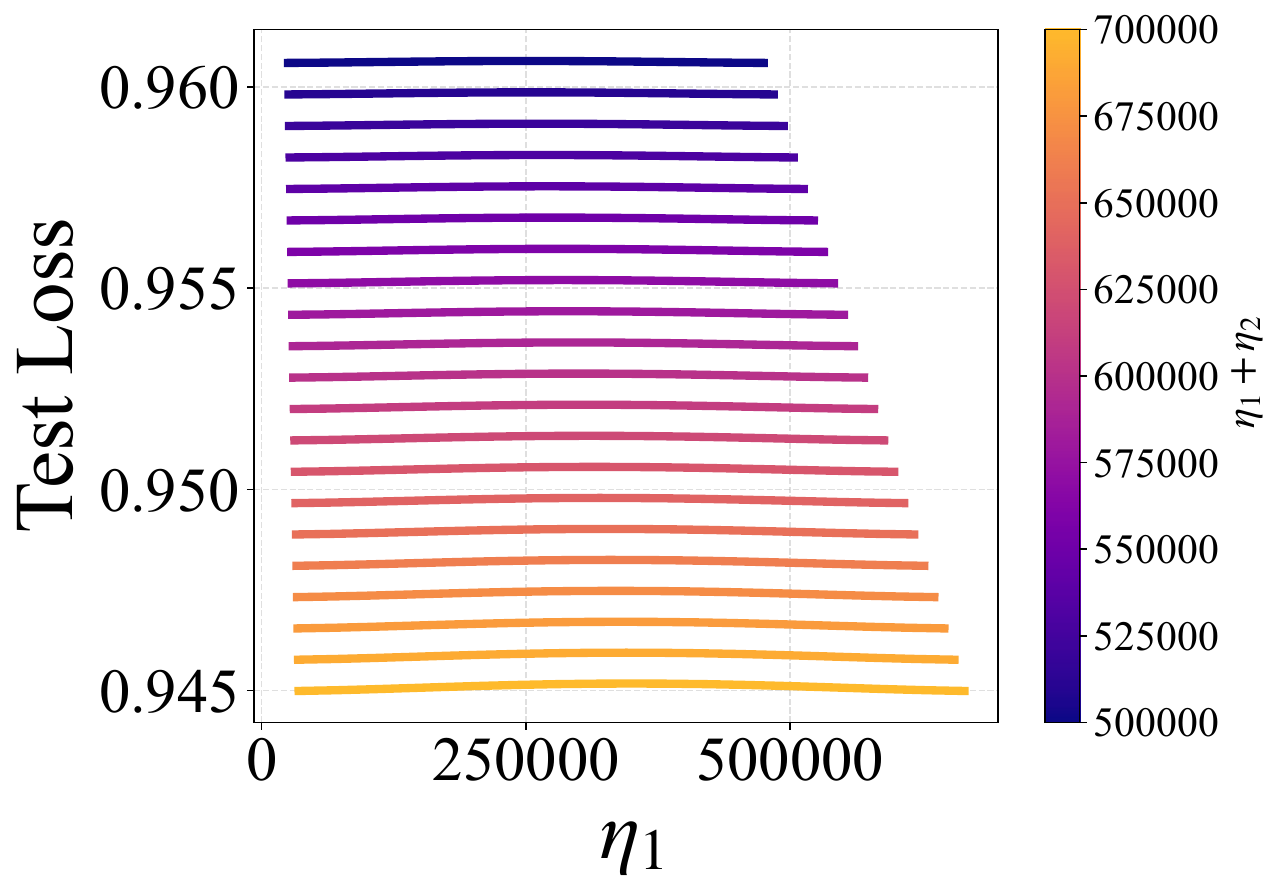}
        \caption{  1-step (theory) }
    \end{subfigure}
    \hfill
        \begin{subfigure}[t]{0.24\linewidth}
        \centering
        \includegraphics[width=\textwidth]{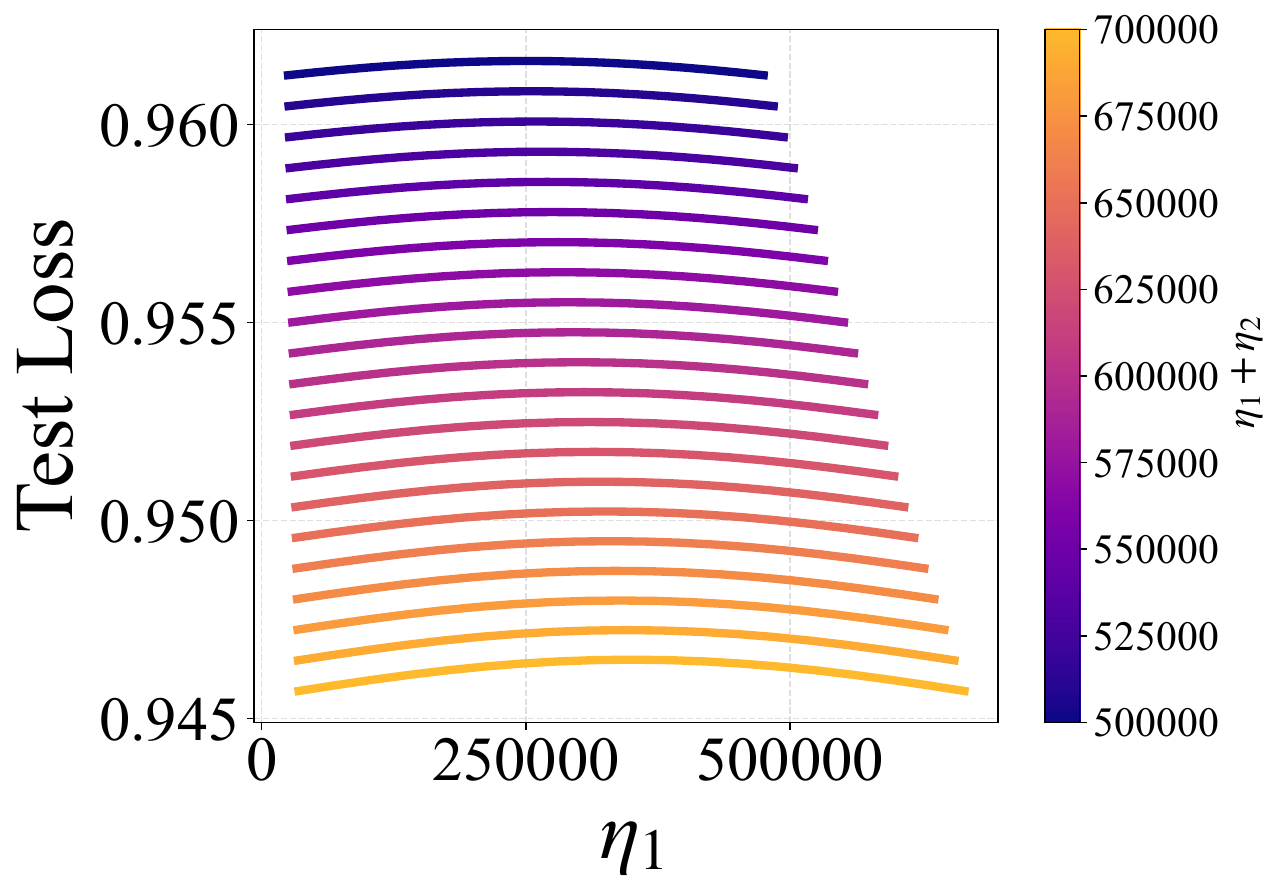}
        \caption{  1-step (experiment) }
    \end{subfigure}
        \hfill
    \begin{subfigure}[t]{0.24\linewidth}
        \centering
        \includegraphics[width=\textwidth]{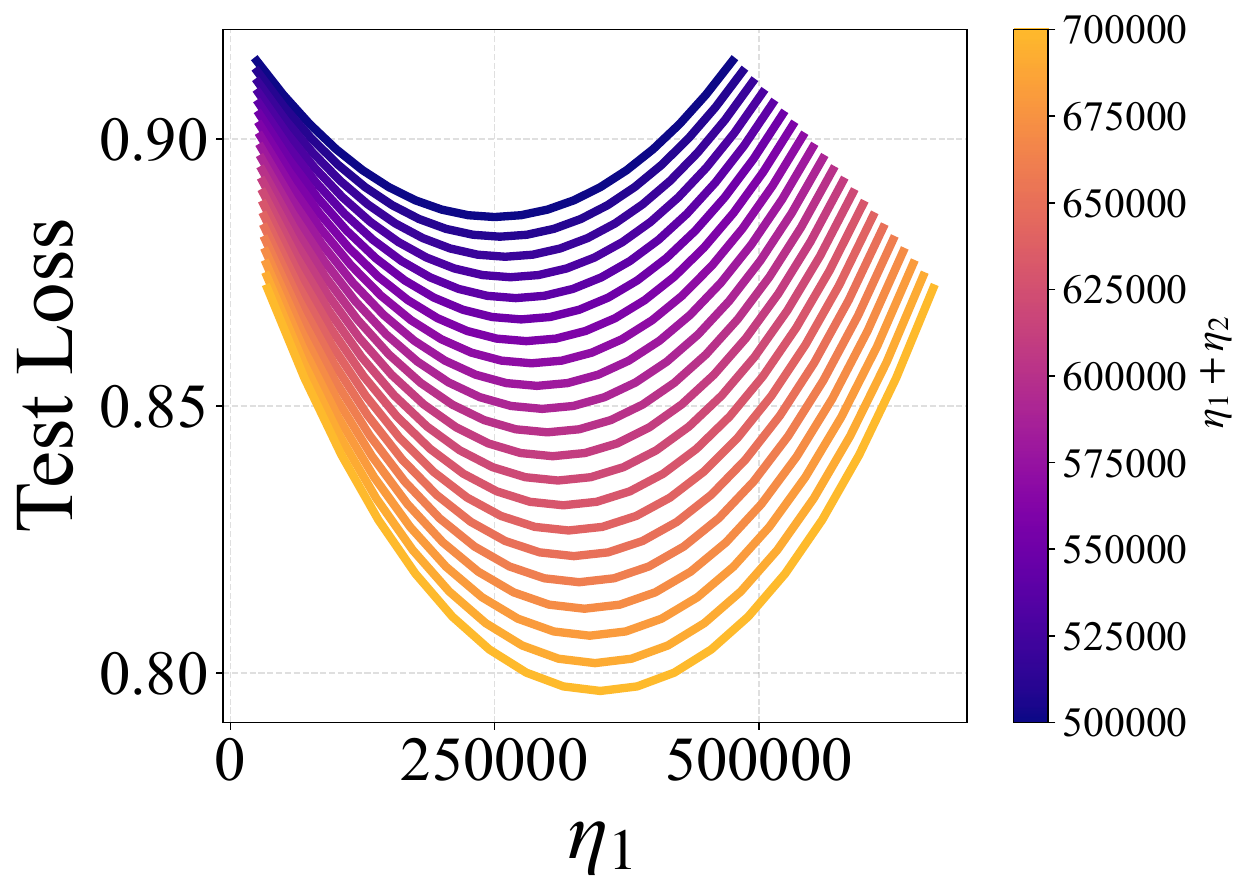}
        \caption{   2-step (theory)  }
    \end{subfigure}
    \hfill
    \begin{subfigure}[t]{0.24\linewidth}
        \centering
        \includegraphics[width=\textwidth]{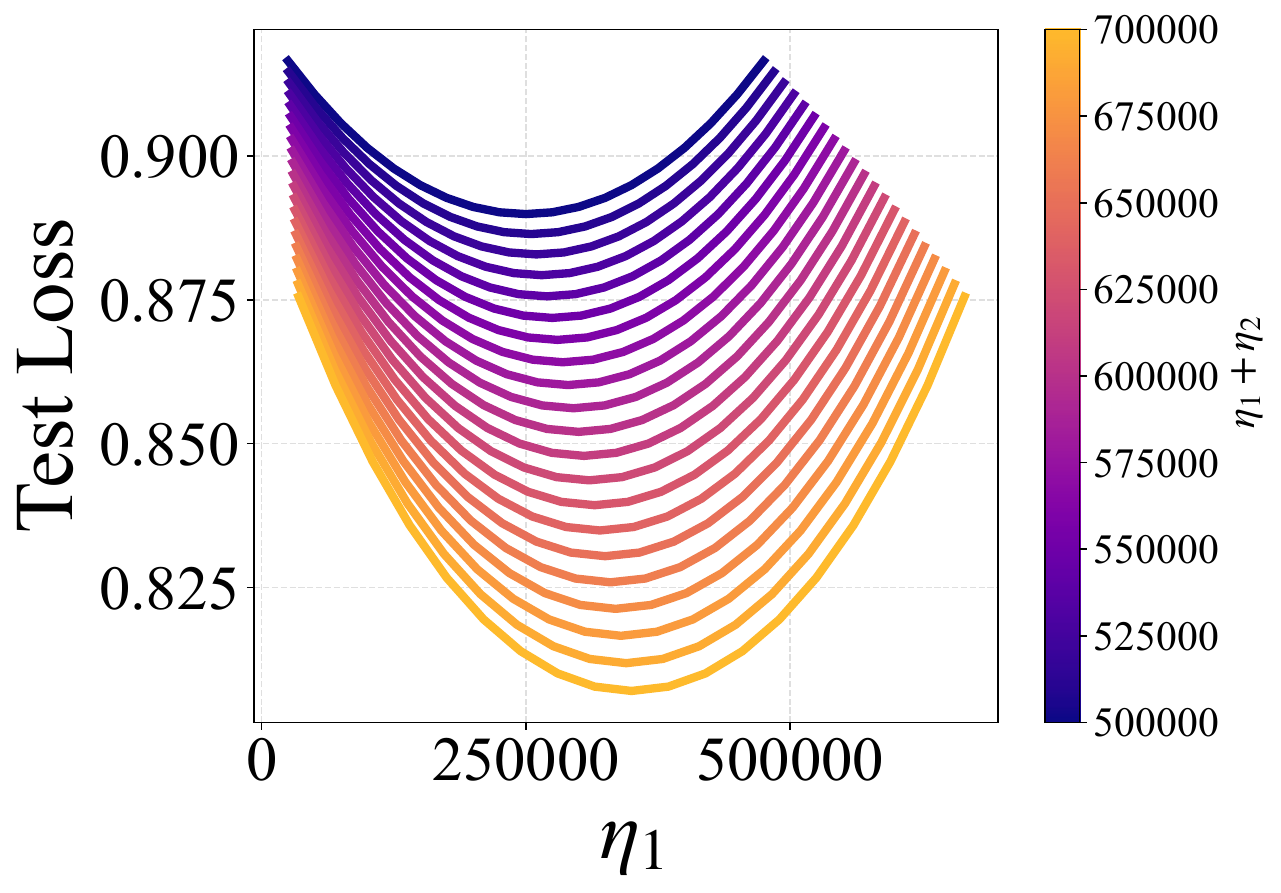}
        \caption{  2-step (experiment)  }
    \end{subfigure}
    \caption{ \textbf{ 2-layer NN under orthogonal initialization.} Here we set $\eta_1+\eta_2\leq O(h^{\frac{3}{2}})$  and  $h=5000$.
}
\label{fig:h=5000-2-NN}
\end{figure}

\begin{figure}[tb]
    \centering
    \begin{subfigure}[t]{0.24\linewidth}
        \centering
        \includegraphics[width=\textwidth]{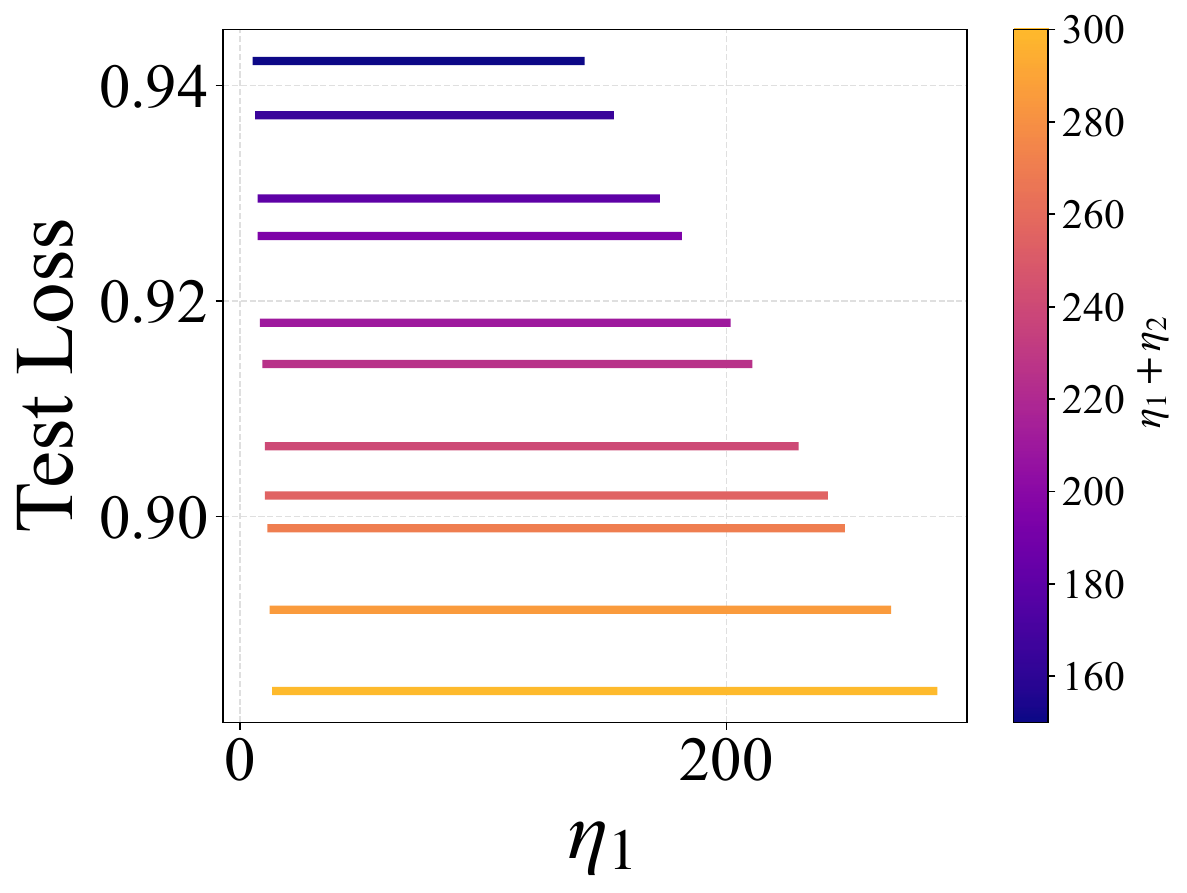}
        \caption{  1-step (theory) }
    \end{subfigure}
    \hfill
    \begin{subfigure}[t]{0.24\linewidth}
        \centering
        \includegraphics[width=\textwidth]{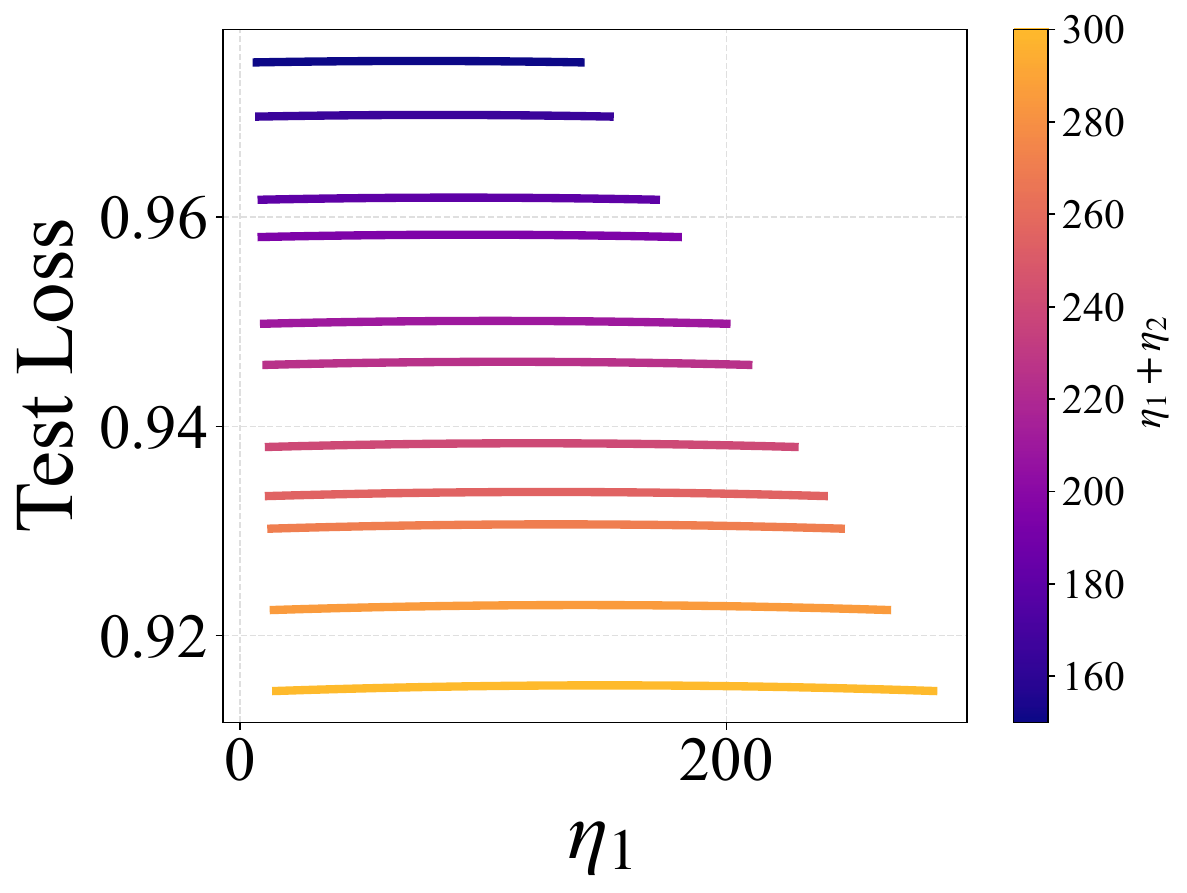}
        \caption{  1-step (experiment) }
    \end{subfigure}
    \hfill
    \begin{subfigure}[t]{0.24\linewidth}
        \centering
        \includegraphics[width=\textwidth]{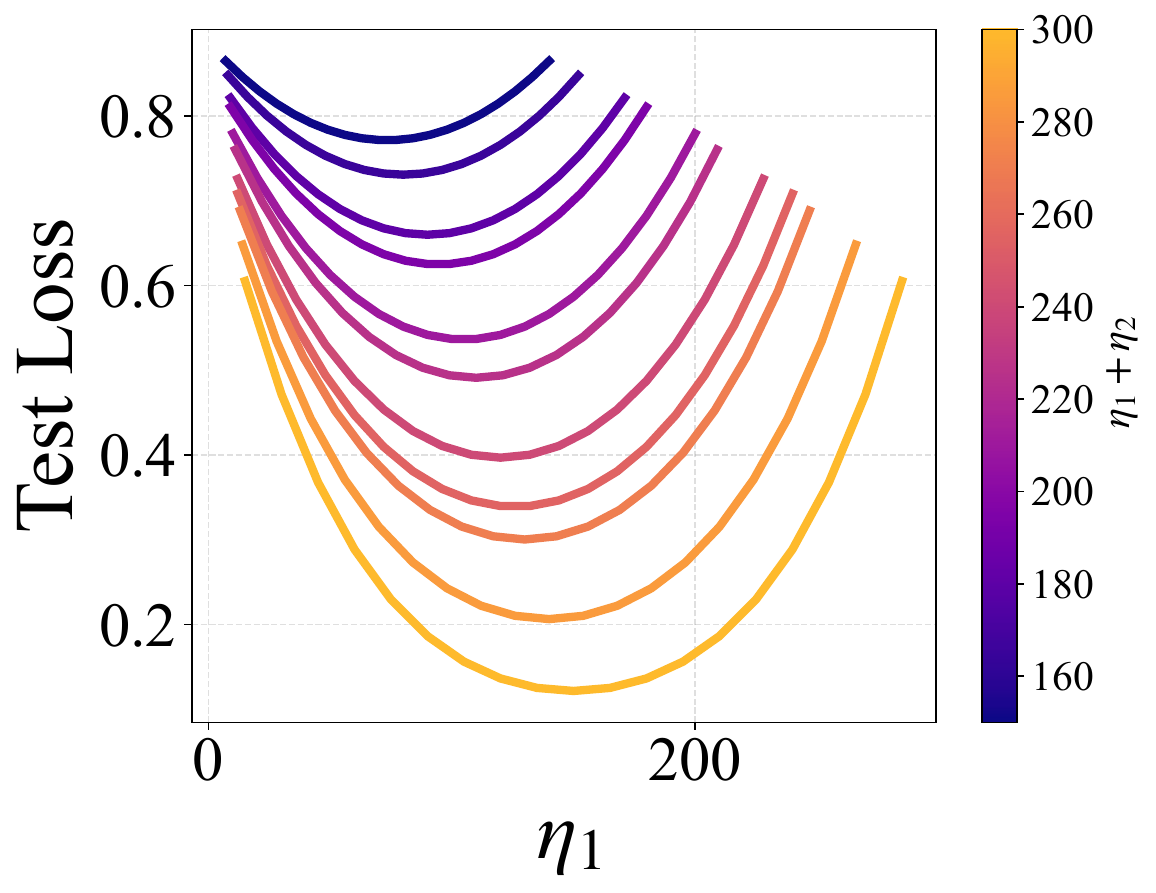}
        \caption{  2-step (theory) }
    \end{subfigure}
    \hfill
        \begin{subfigure}[t]{0.24\linewidth}
        \centering
        \includegraphics[width=\textwidth]{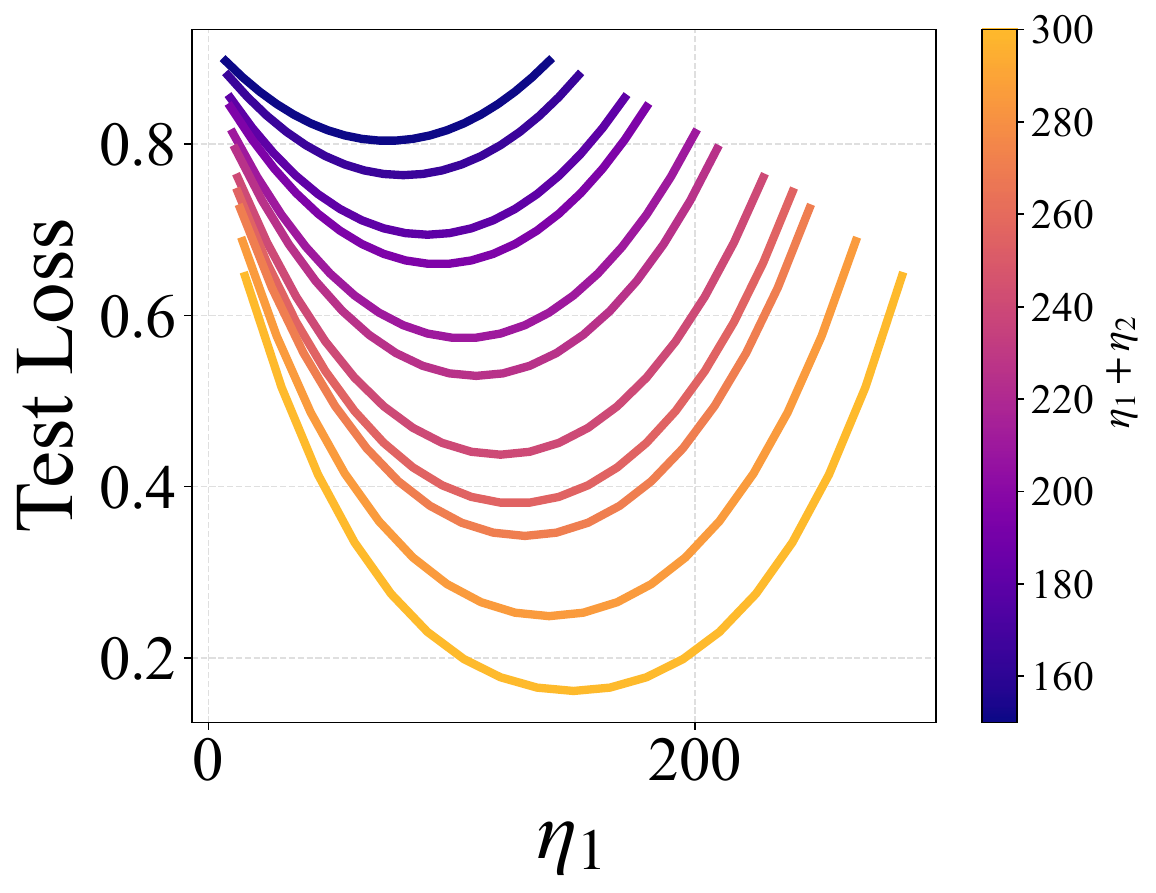}
        \caption{  2-step (experiment) }
    \end{subfigure}
\caption{ \textbf{ 3-layer NN under orthogonal initialization.} Here we set $\eta_1+\eta_2\leq O(h^{\frac{2}{3}})$  and  $h=5000$.}
\label{fig:h=5000-3-NN}
\end{figure}

\begin{figure}[tb]
    \centering
    \begin{subfigure}[t]{0.24\linewidth}
        \centering
        \includegraphics[width=\textwidth]{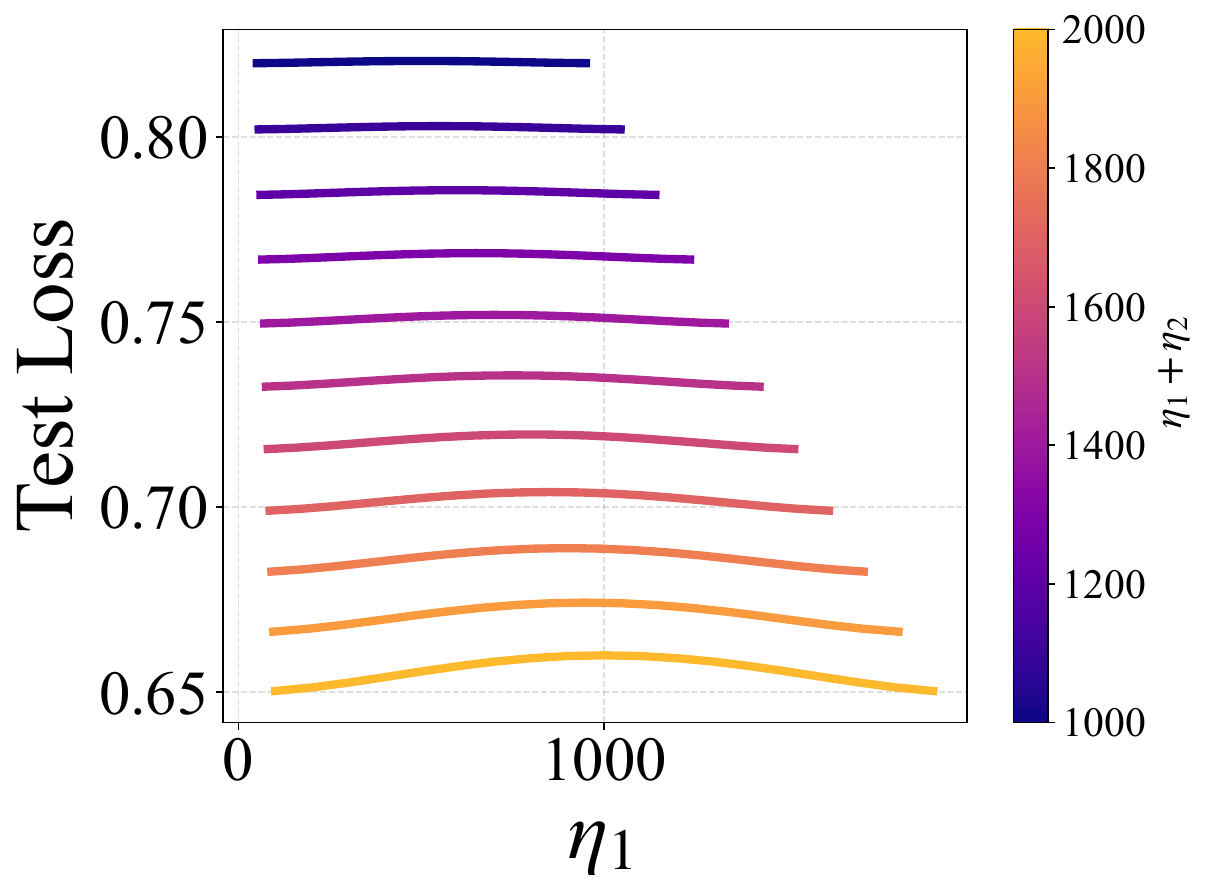}
        \caption{  1-step (theory) }
    \end{subfigure}
    \hfill
        \begin{subfigure}[t]{0.24\linewidth}
        \centering
        \includegraphics[width=\textwidth]{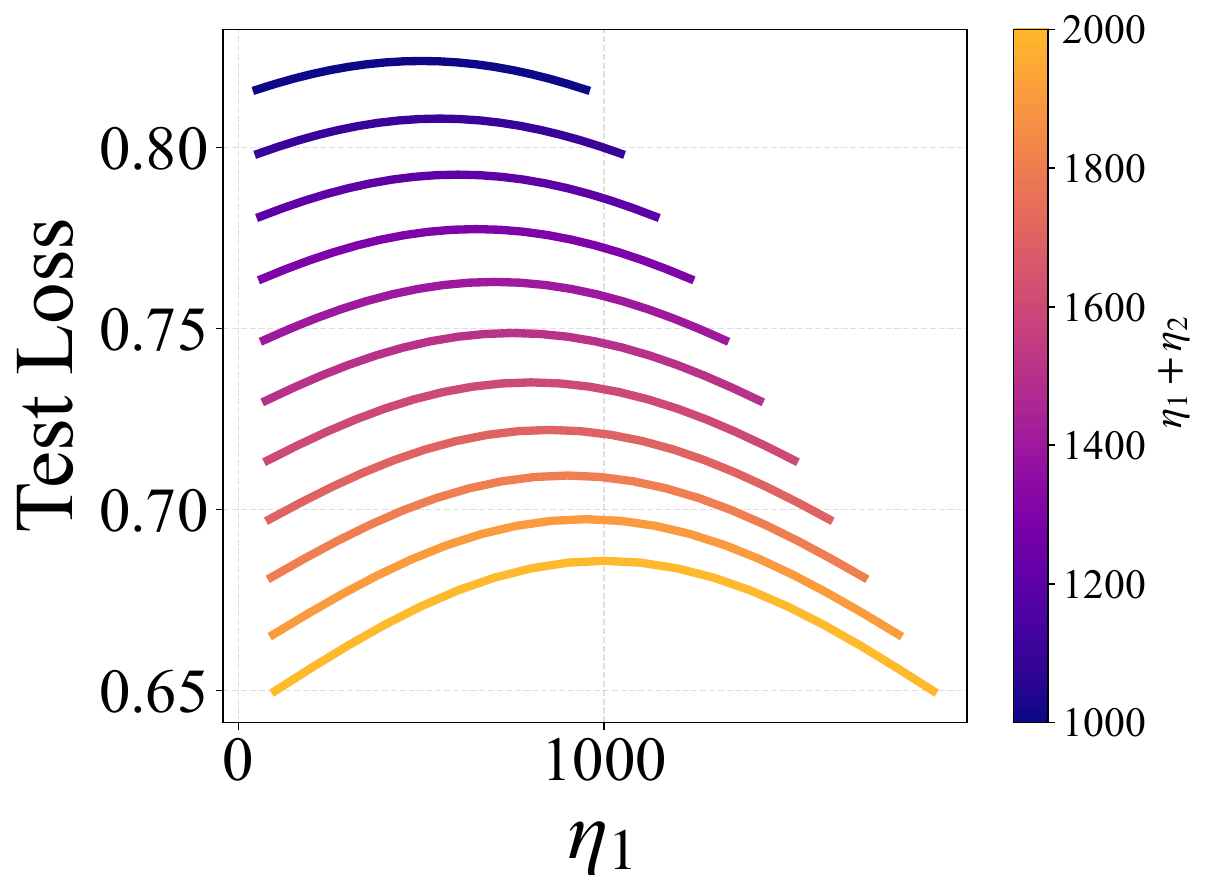}
        \caption{  1-step (experiment) }
    \end{subfigure}
        \hfill
    \begin{subfigure}[t]{0.24\linewidth}
        \centering
        \includegraphics[width=\textwidth]{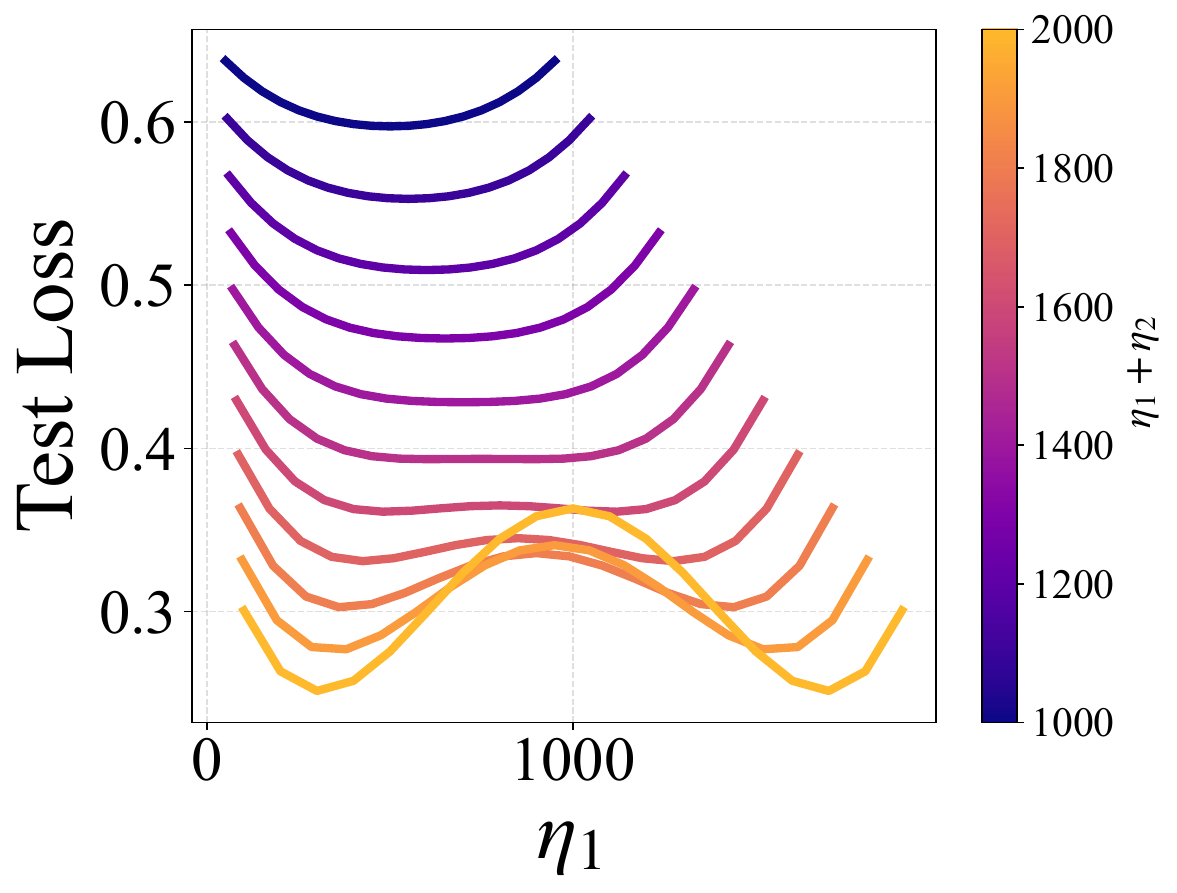}
        \caption{   2-step (theory)  }
    \end{subfigure}
    \hfill
    \begin{subfigure}[t]{0.24\linewidth}
        \centering
        \includegraphics[width=\textwidth]{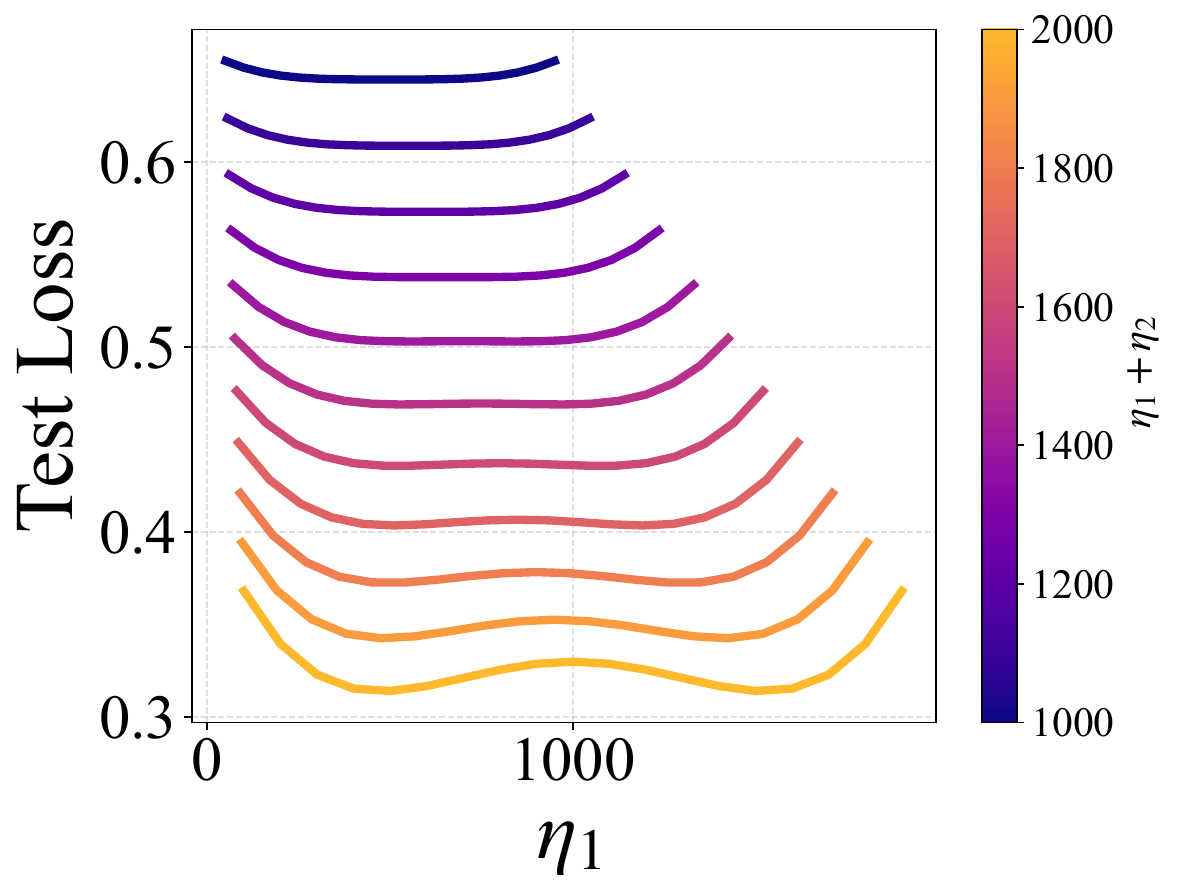}
        \caption{  2-step (experiment)  }
    \end{subfigure}
    \caption{ \textbf{ 2-layer NN under orthogonal initialization.} Here we set $\eta_1+\eta_2\leq O(h^{\frac{3}{2}})$  and  $h=100$. We can see since $h$ does not satisfy the condition on $h$ in Corollary~\ref{cor:two-layer-NN}, the balanced learning-rate allocation is not locally optimal.
}
\label{fig:h=100-2-NN}
\end{figure}

\begin{figure}[tb]
    \centering
    \begin{subfigure}[t]{0.24\linewidth}
        \centering
        \includegraphics[width=\textwidth]{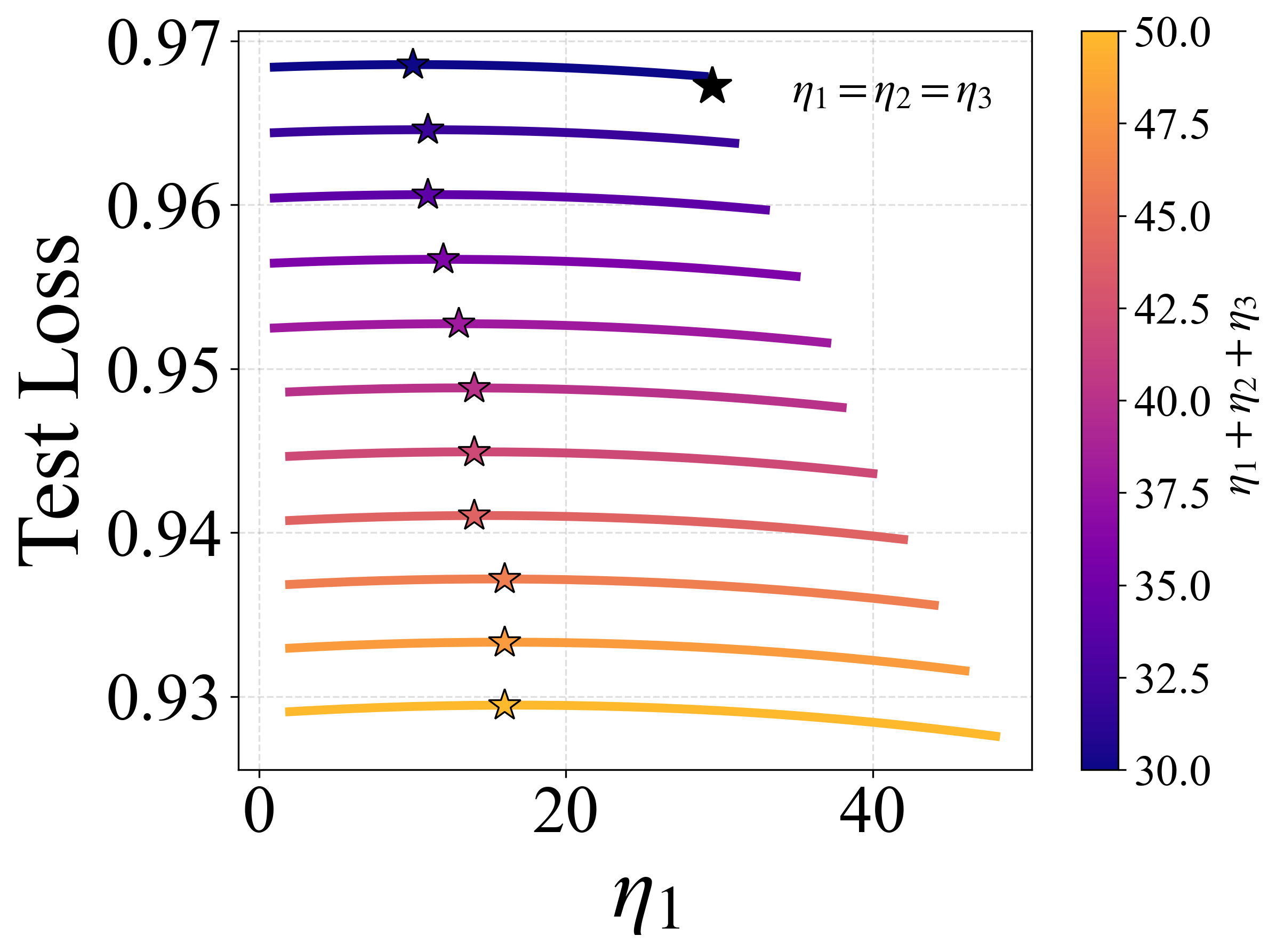}
        \caption{  4-Layer 1-step }
    \end{subfigure}
    \hfill
        \begin{subfigure}[t]{0.24\linewidth}
        \centering
        \includegraphics[width=\textwidth]{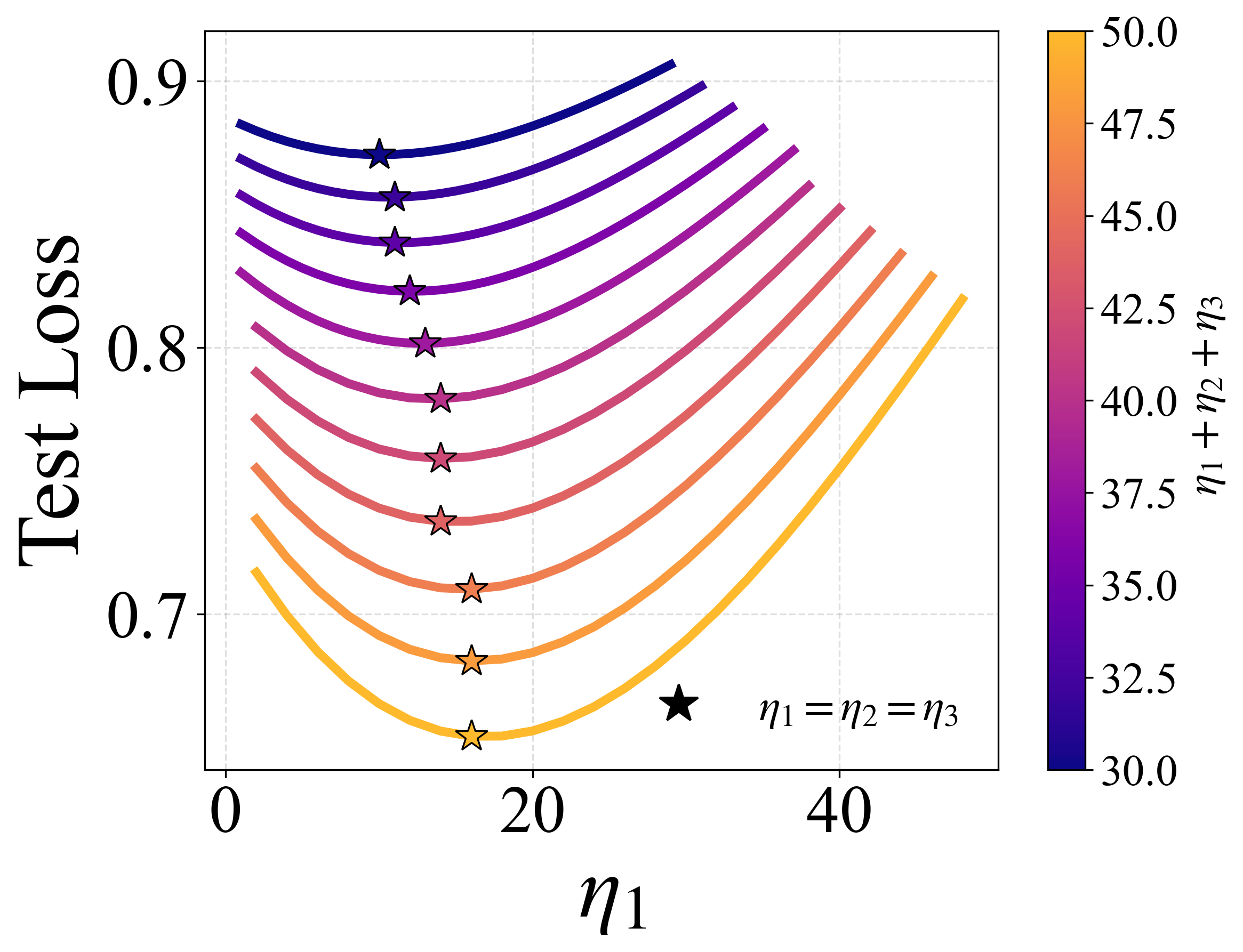}
        \caption{  4-Layer 2-step }
    \end{subfigure}
        \hfill
    \begin{subfigure}[t]{0.24\linewidth}
        \centering
        \includegraphics[width=\textwidth]{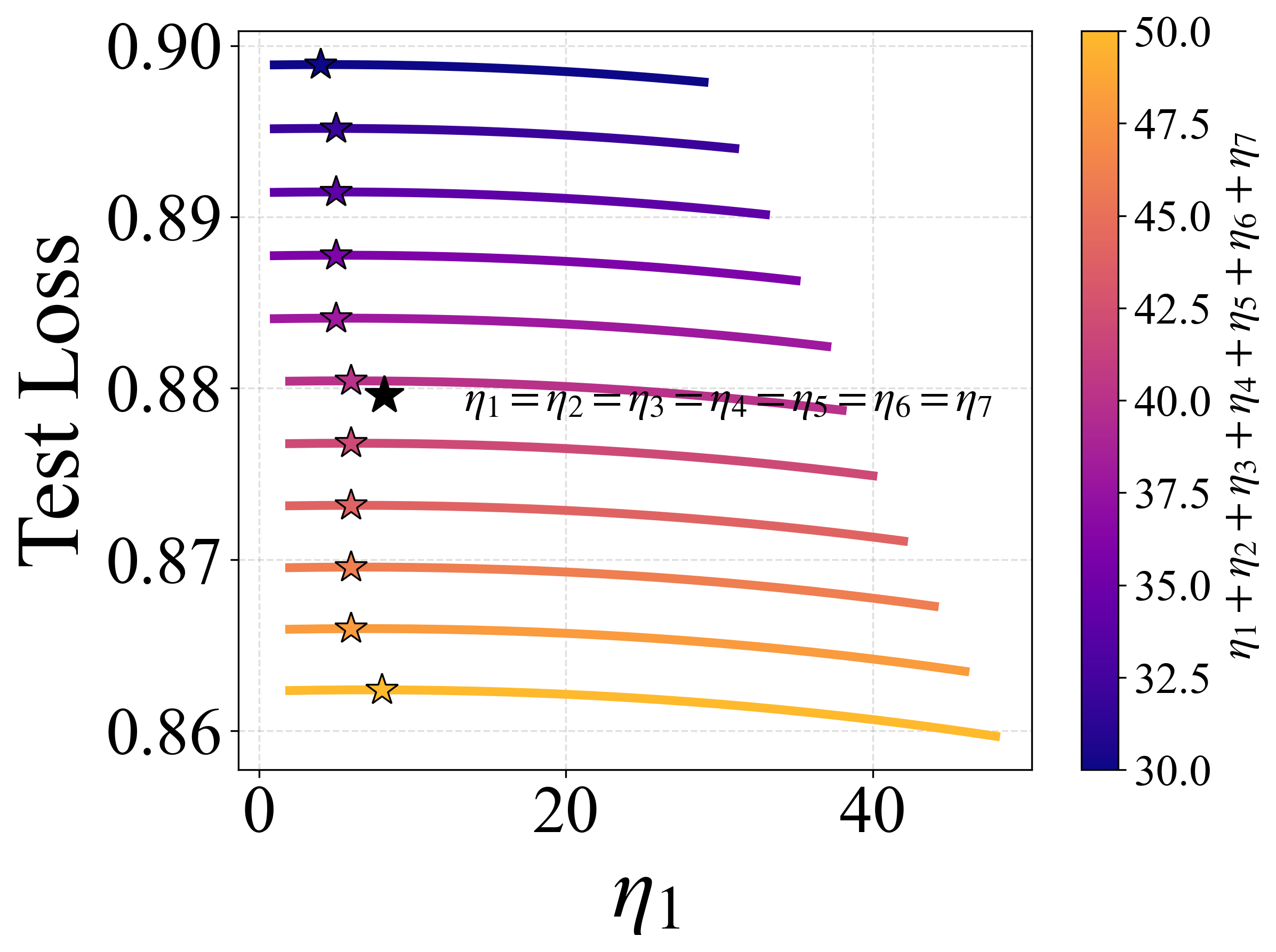}
        \caption{   8-Layer 1-step  }
    \end{subfigure}
    \hfill
    \begin{subfigure}[t]{0.24\linewidth}
        \centering
        \includegraphics[width=\textwidth]{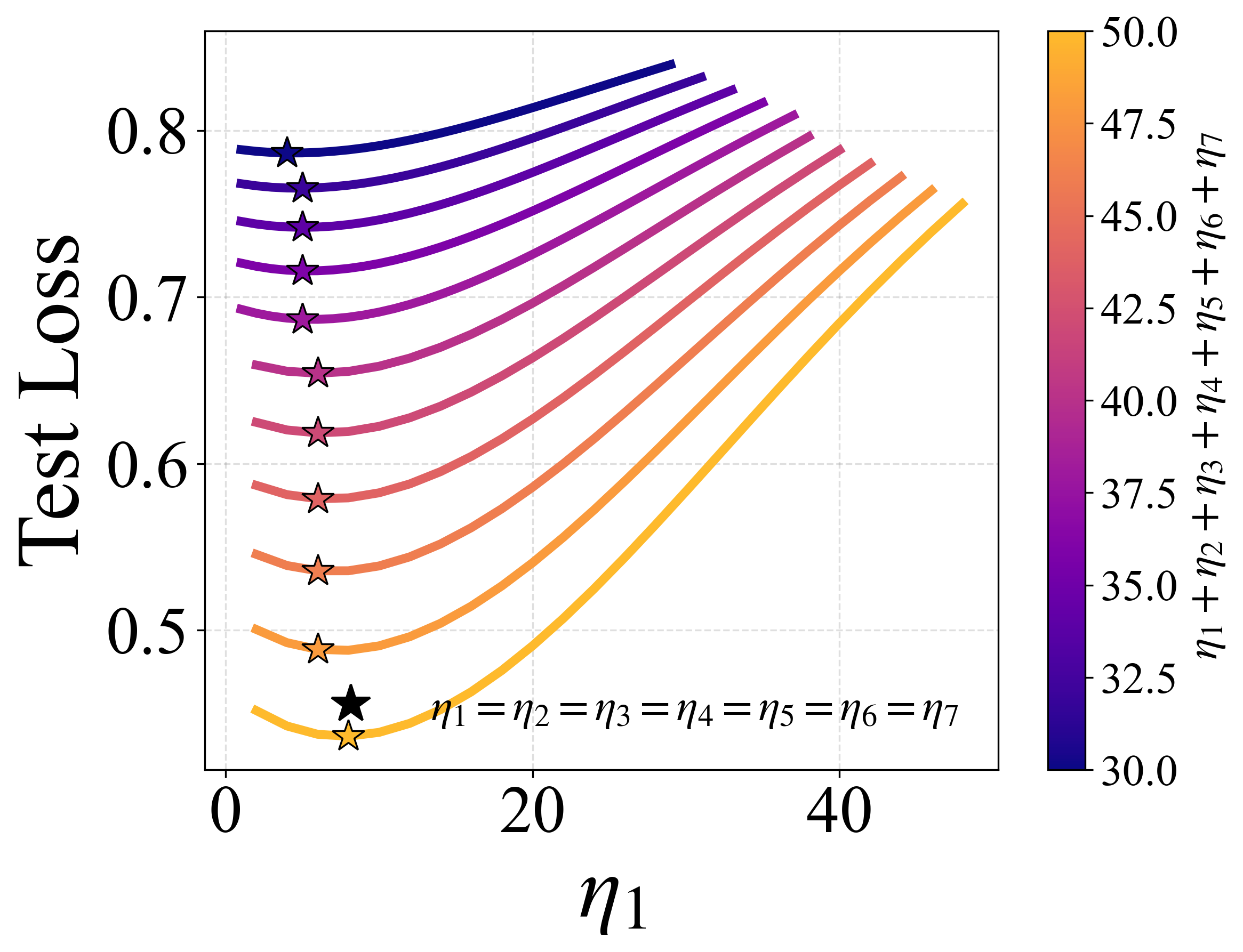}
        \caption{  8-Layer 2-step  }
    \end{subfigure}
    \caption{ \textbf{ 4-layer and 8-layer NN under orthogonal initialization for 1 and 2-step updates.} For 4-NN, we set $\eta_1+\eta_2+\eta_3=C \leq O(h^{\frac{2}{3}})$ with $h=1000$ and we set $\eta_2=\eta_3=\frac{C-\eta_1}{2}.$ For 8-NN, we set $\eta_1+\eta_2+\eta_3+\eta_4+\eta_5+\eta_6+\eta_7=C\leq O(h^{\frac{2}{3}})$ with $h=1000$ and we set $\eta_2=\eta_3=\eta_4=\eta_5=\eta_6=\eta_7=\frac{C-\eta_1}{6}.$ 
}
\label{fig:deep-linear-NN}
\end{figure}


\begin{figure}[tb]
    \centering
    \begin{subfigure}[t]{0.48\linewidth}
        \centering
        \includegraphics[width=\textwidth]{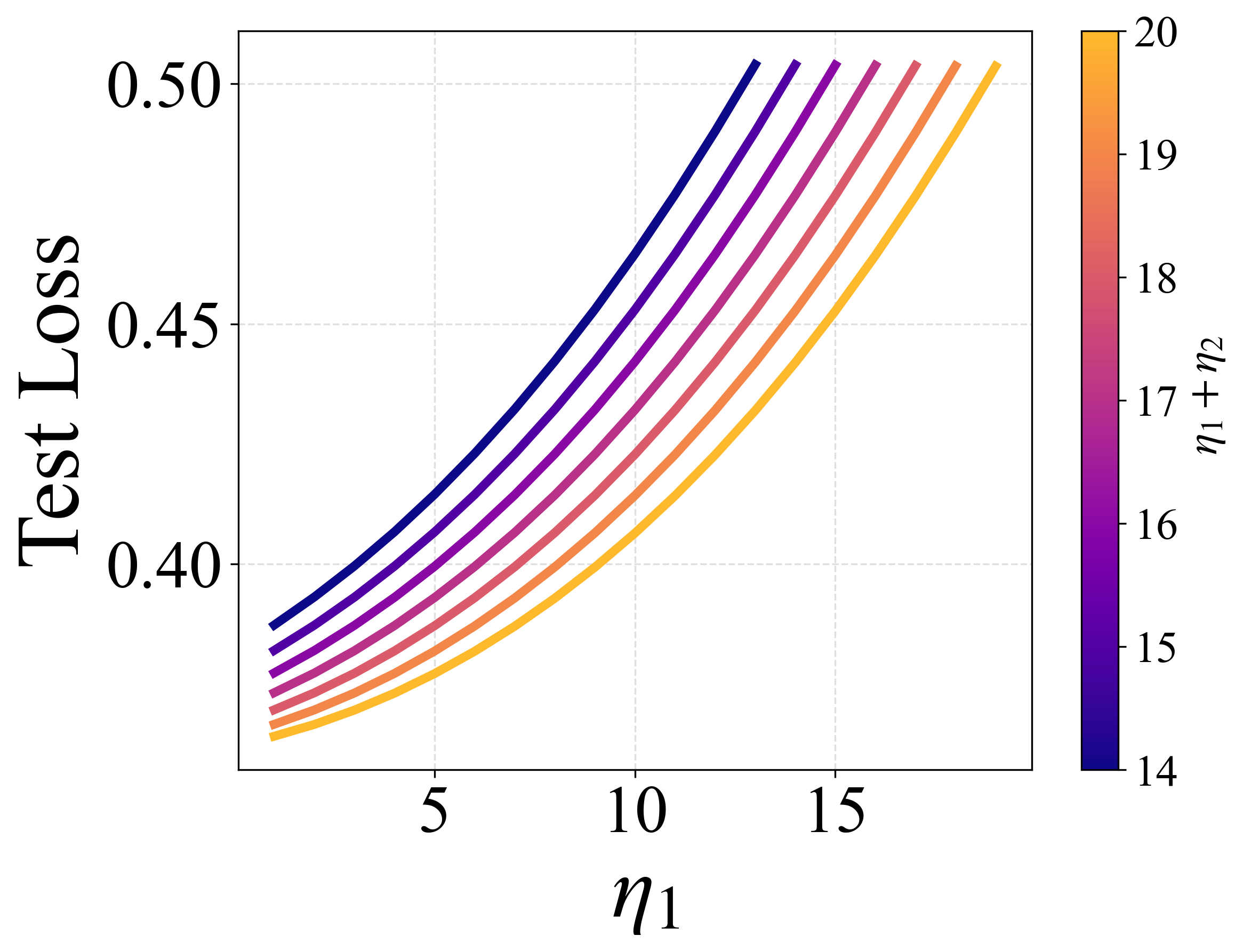}
        \caption{ 1-step }
    \end{subfigure}
    \hfill
        \begin{subfigure}[t]{0.48\linewidth}
        \centering
        \includegraphics[width=\textwidth]{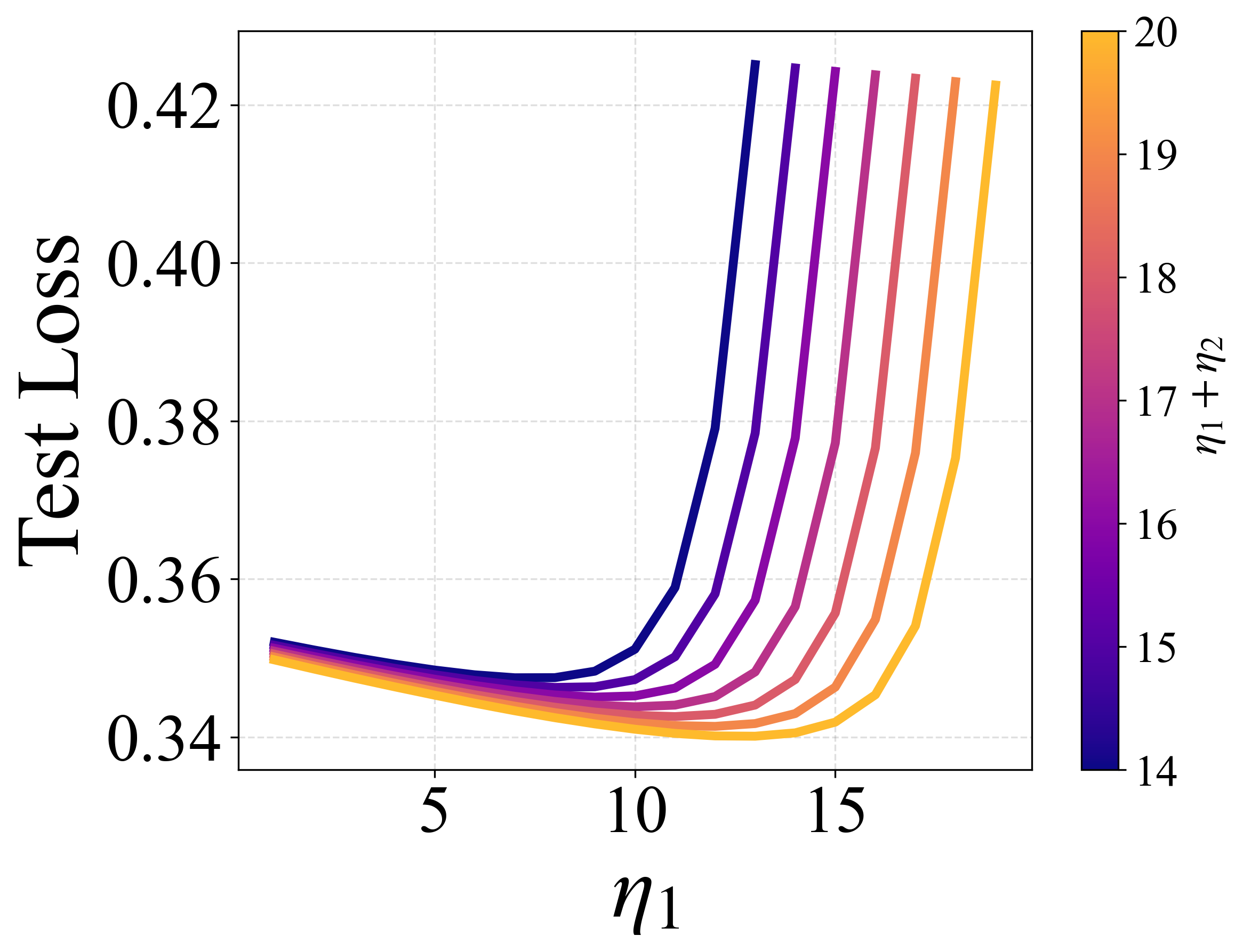}
        \caption{ 2-step }
    \end{subfigure}
\caption{ \textbf{ 3-NN nonlinear under orthogonal initialization for 1 and 8-step updates.} Here we consider student model is $f(\vx_i)=\frac{1}{\sqrt{h}}\sigma(\sigma(\vx_i^{\top}\mW_1)\mW_2)\va,$ and the teacher model is $\vy_i=\sigma({\vbeta^*}^{\top}\vx_i),$ with $\sigma$ being the ReLU activation. 
}
\label{fig:nonlinear-NN}
\end{figure}


\begin{figure}[tb]
    \centering
        \begin{subfigure}[t]{0.45\linewidth}
        \centering
        \includegraphics[width=\textwidth]{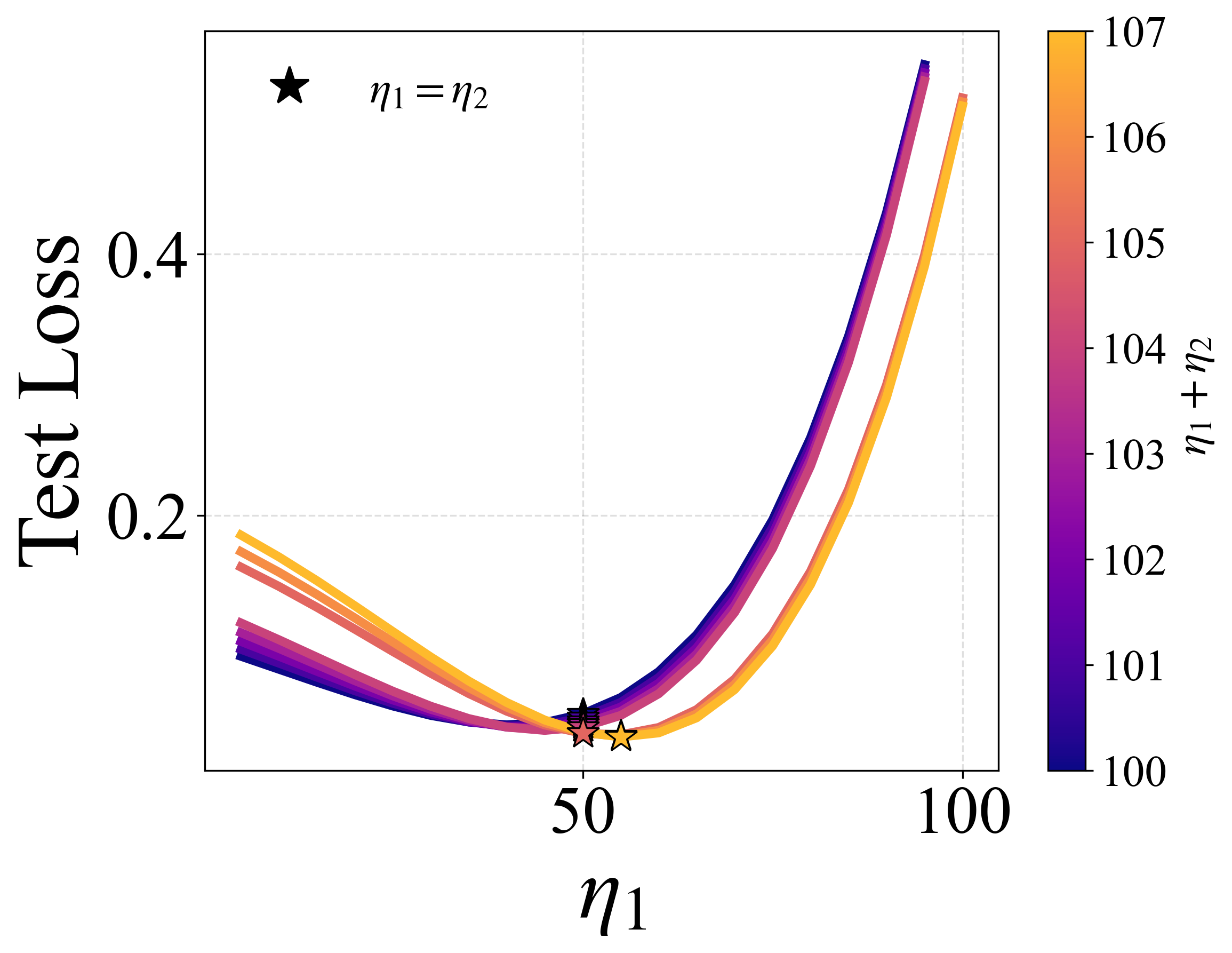}
        \caption{3-layer NN, step 1: first-layer-only update}
    \end{subfigure}
    \hfill
        \begin{subfigure}[t]{0.45\linewidth}
        \centering
        \includegraphics[width=\textwidth]{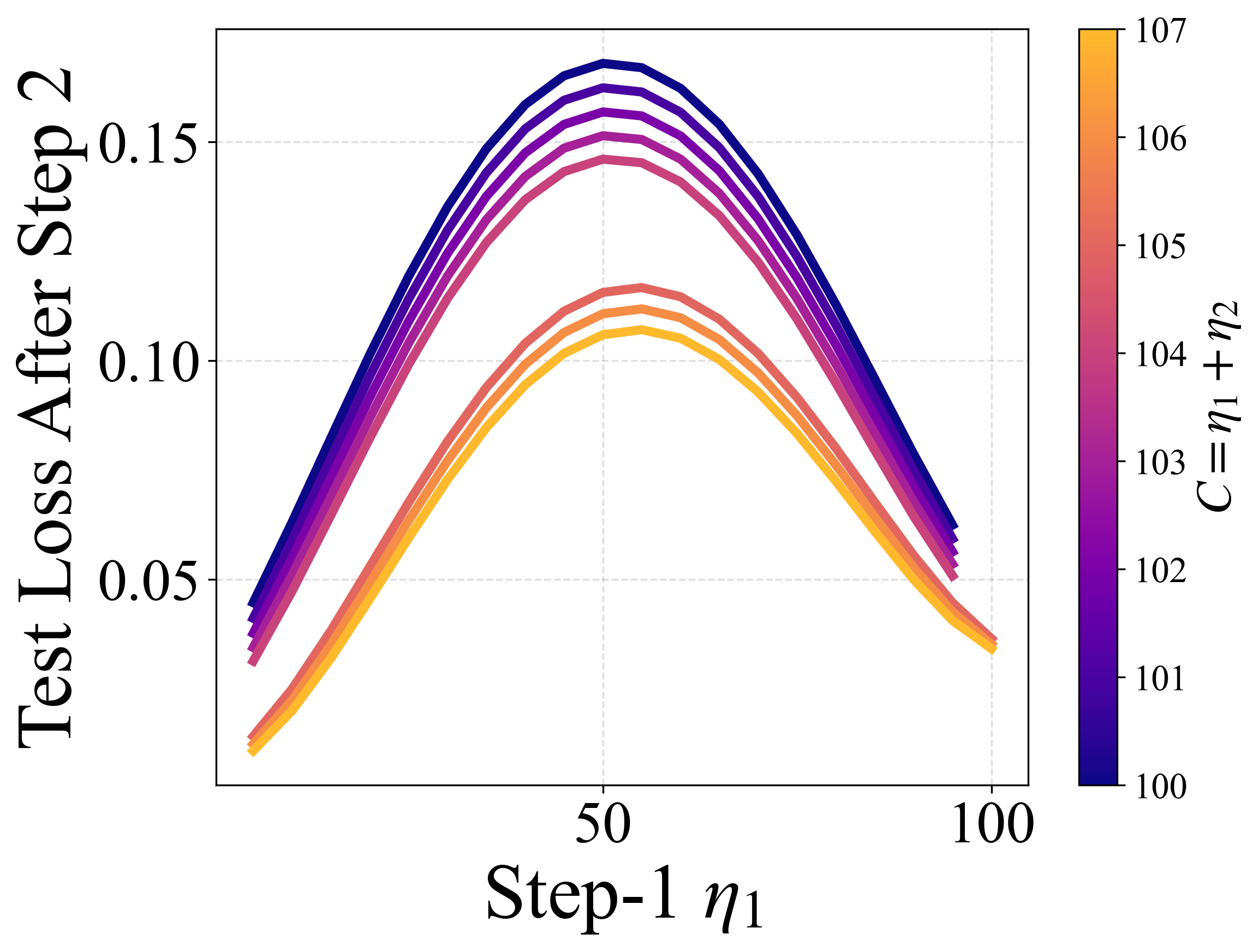}
        \caption{3-layer NN, step 2: symmetric two-layer update}
    \end{subfigure}
    \hfill
     \begin{subfigure}[t]{0.31\linewidth}
        \centering
        \includegraphics[width=\textwidth]{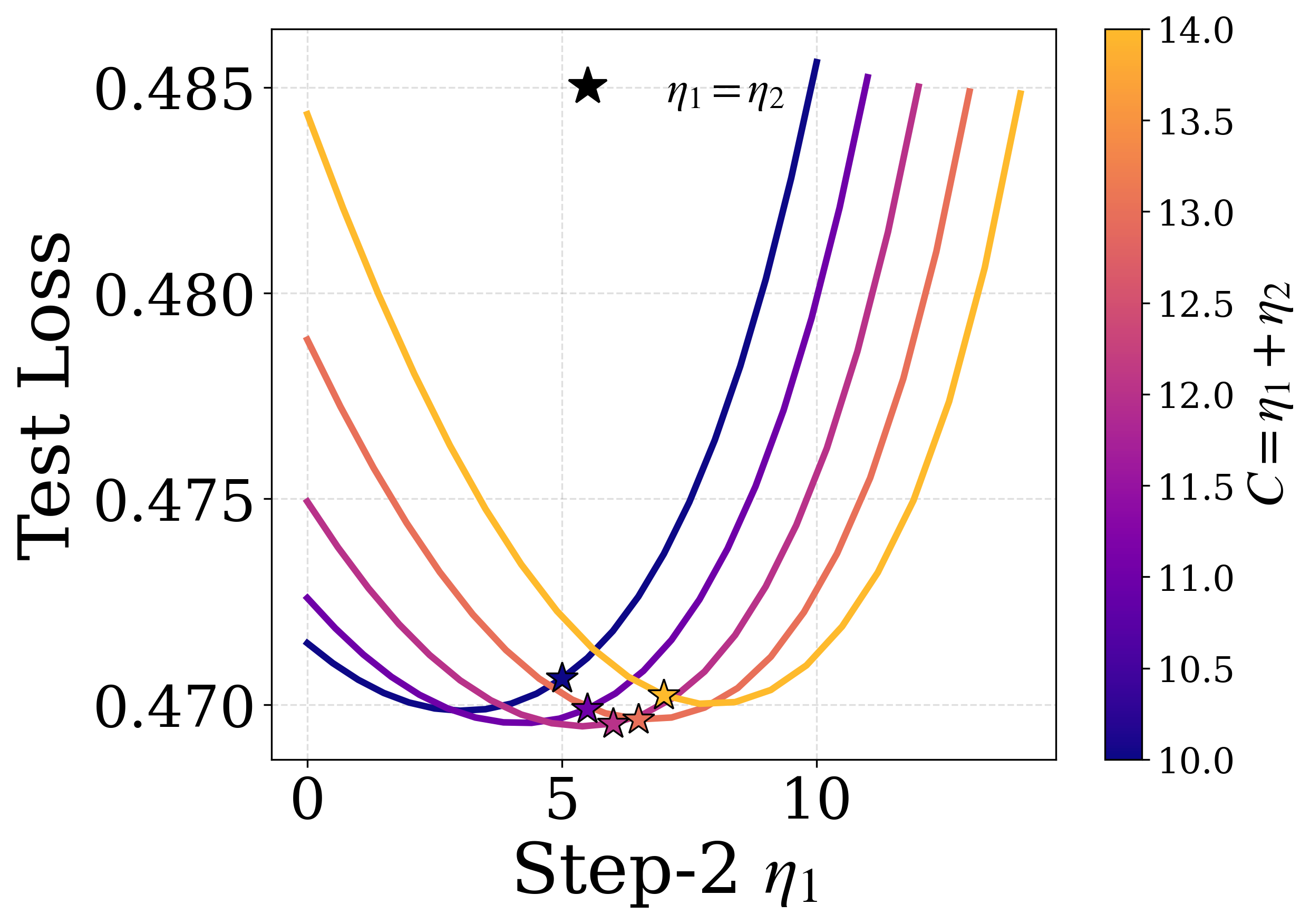}
        \caption{3-layer CNN \\ step 1: first-layer-only update}
    \end{subfigure}
    \hfill
        \begin{subfigure}[t]{0.31\linewidth}
        \centering
        \includegraphics[width=\textwidth]{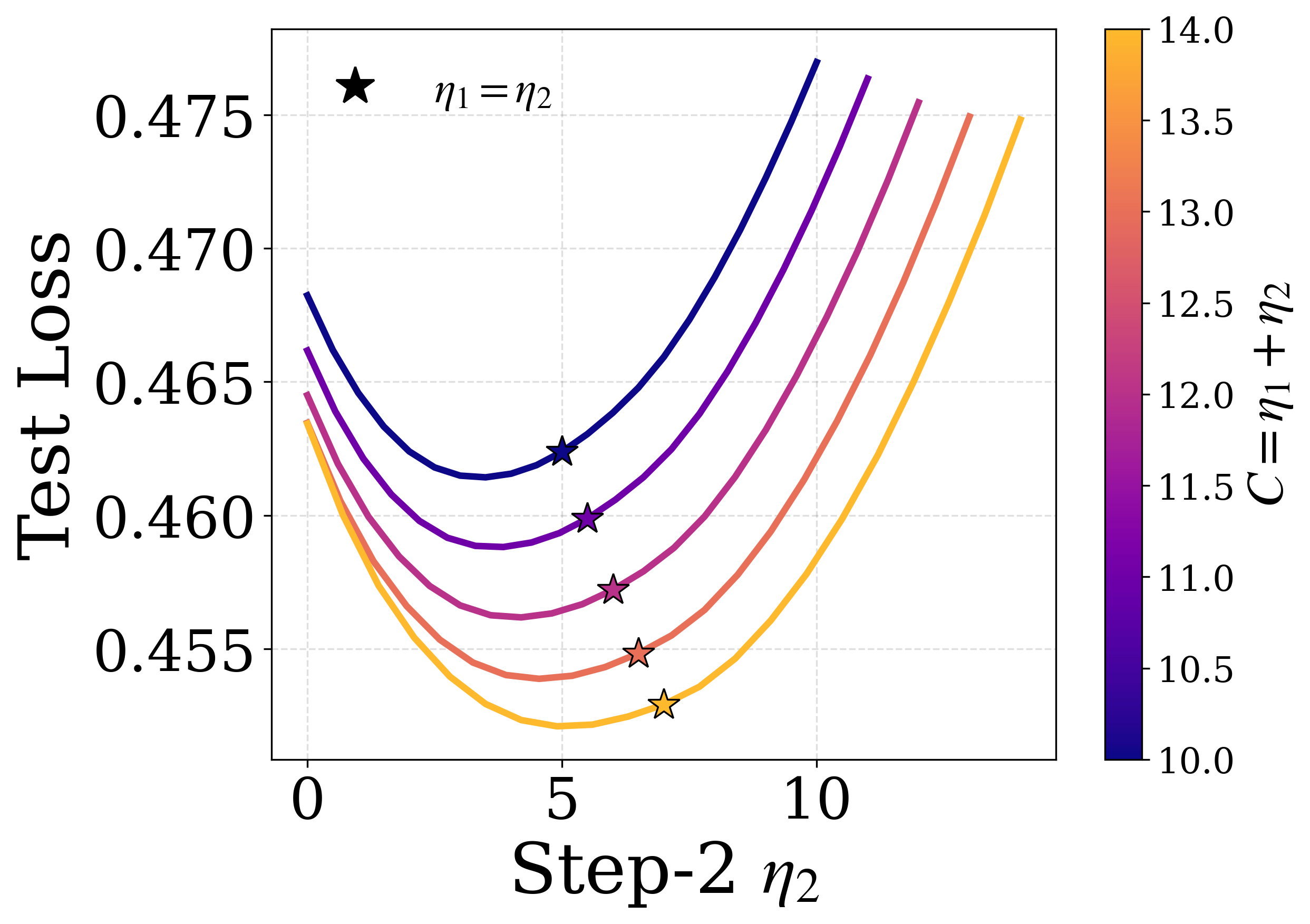}
        \caption{3-layer CNN \\ step 1: second-layer-only update}
    \end{subfigure}
            \begin{subfigure}[t]{0.31\linewidth}
        \centering
        \includegraphics[width=\textwidth]{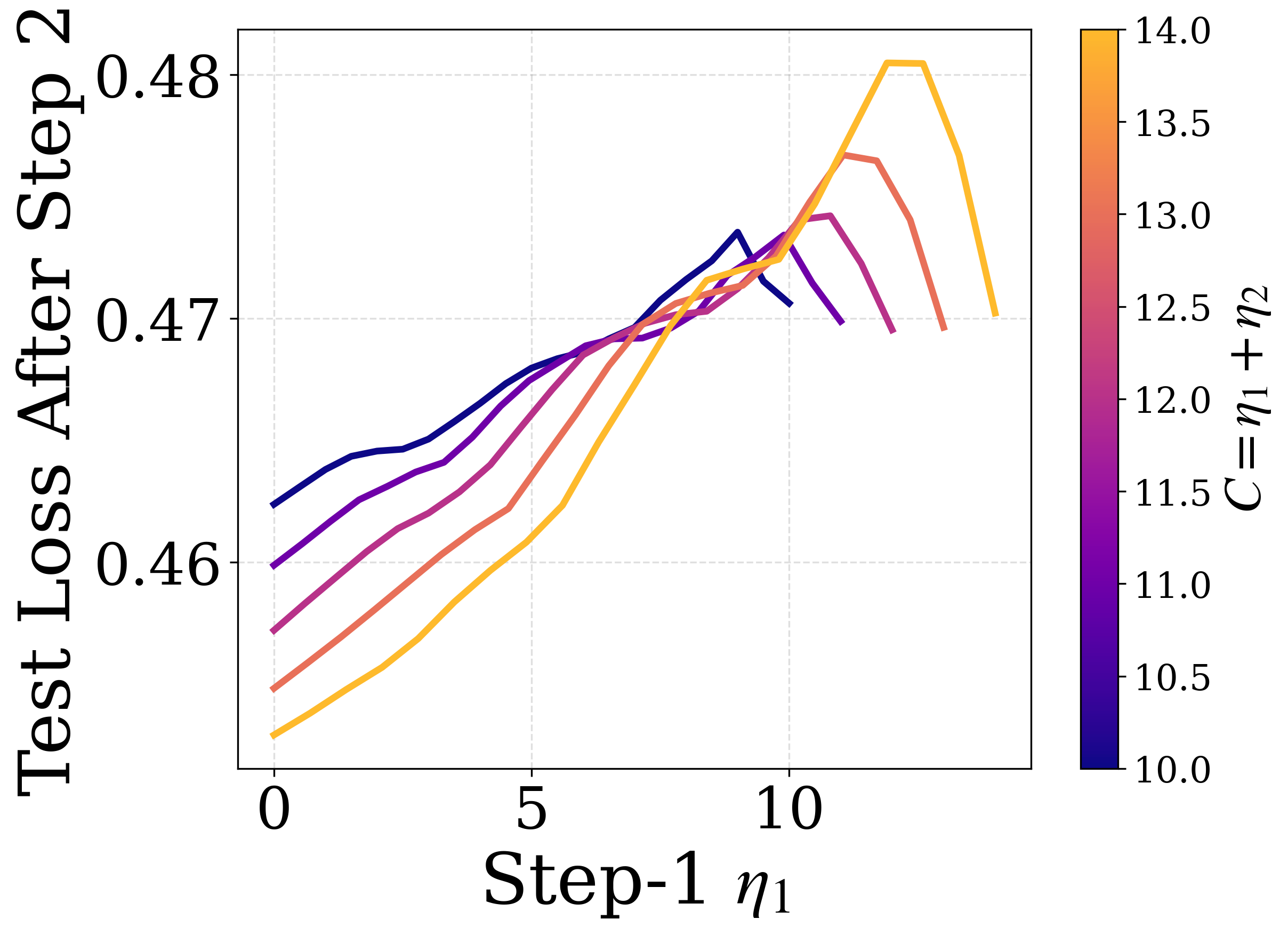}
        \caption{3-layer CNN \\ step 2: symmetric two-layer update}
    \end{subfigure}
\caption{ \textbf{(a)} 3-layer NN under orthogonal initialization for special 2-step update. Here we set $\eta_1+\eta_2=C \leq O(h^{\frac{2}{3}})$ with $h=1000.$ We set the first step update step to use an asymmetric learning-rate allocation: the first layer is updated with learning rate C, while the second layer is not trained. For the second update step, we then optimize under the constraint $\eta_1+\eta_2=C.$ \textbf{(b)} 3-layer NN under orthogonal initialization for special 2-step update. Here we set $\eta_1+\eta_2=C$.  We set the second update step to use an symmetric learning-rate allocation: $\eta_1=\eta_2=\frac{C}{2}$. For the first update step, we optimize under the constraint $\eta_1+\eta_2=C.$ \textbf{(c)} 3-layer CNN for special 2-step update. Here we set $\eta_1+\eta_2=C.$ We set the first update step to use an asymmetric learning-rate allocation: the first layer is updated with learning rate C, while the second layer is not trained. For the second update step, we then optimize under the constraint  $\eta_1+\eta_2=C.$ \textbf{(d)} 3-layer CNN for special 2-step update. Here we set $\eta_1+\eta_2=C.$ We set the first update step to use an asymmetric learning-rate allocation: the second layer is updated with learning rate C, while the first layer is not trained. For the second update step, we then optimize under the constraint $\eta_1+\eta_2=C$. \textbf{(e)} 3-layer CNN for special 2-step update. Here we set $\eta_1+\eta_2=C.$ We set the second update step to use an symmetric learning-rate allocation: $\eta_1=\eta_2=\frac{C}{2}$. For the first update step, we optimize under the constraint   $\eta_1+\eta_2=C.$
}
\label{fig:special-NN}
\end{figure}


\begin{figure}[tb]
    \centering
        \begin{subfigure}[t]{0.24\linewidth}
        \centering
        \includegraphics[width=\textwidth]{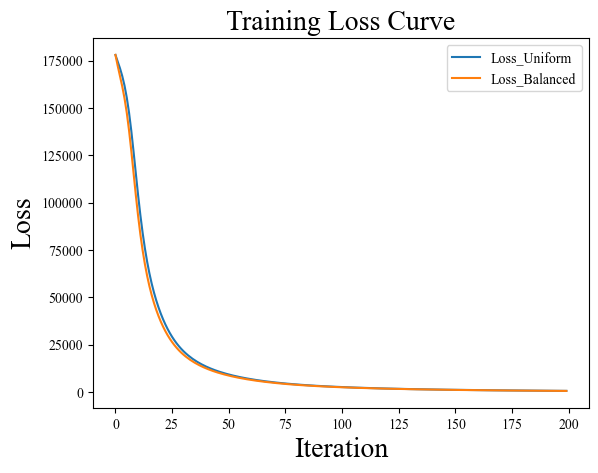}
        \caption{train loss(lin-lin)}
    \end{subfigure}
    \hfill
        \begin{subfigure}[t]{0.24\linewidth}
        \centering
        \includegraphics[width=\textwidth]{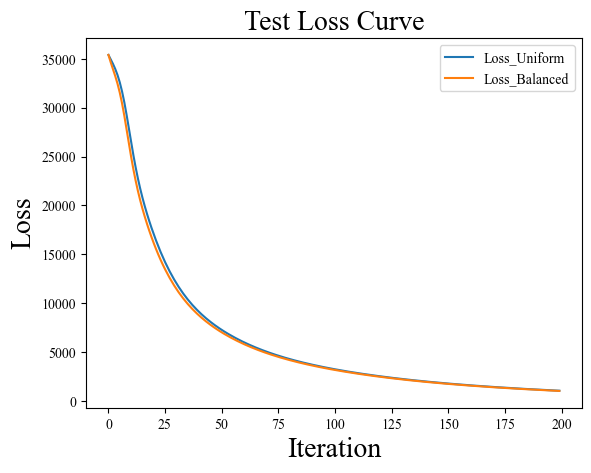}
        \caption{test loss(lin-lin)}
    \end{subfigure}
    \hfill
    \begin{subfigure}[t]{0.24\linewidth}
        \centering
        \includegraphics[width=\textwidth]{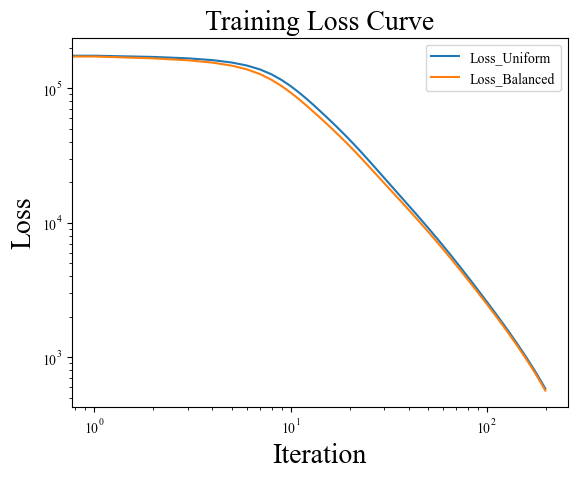}
        \caption{train loss(log-log)}
    \end{subfigure}
    \hfill
        \begin{subfigure}[t]{0.24\linewidth}
        \centering
        \includegraphics[width=\textwidth]{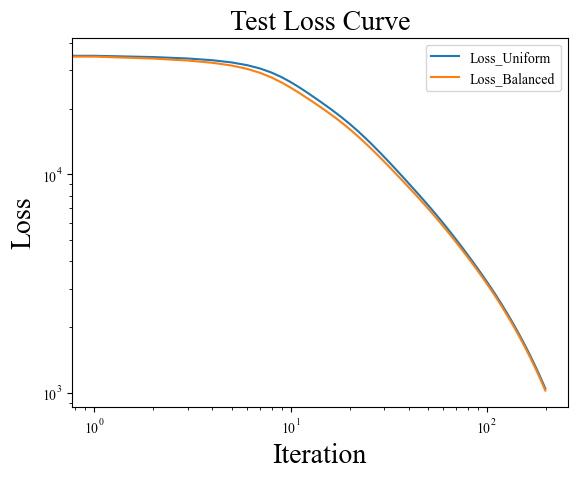}
        \caption{test loss(log-log)}
    \end{subfigure}
    \hfill
     \begin{subfigure}[t]{0.33\linewidth}
        \centering
        \includegraphics[width=\textwidth]{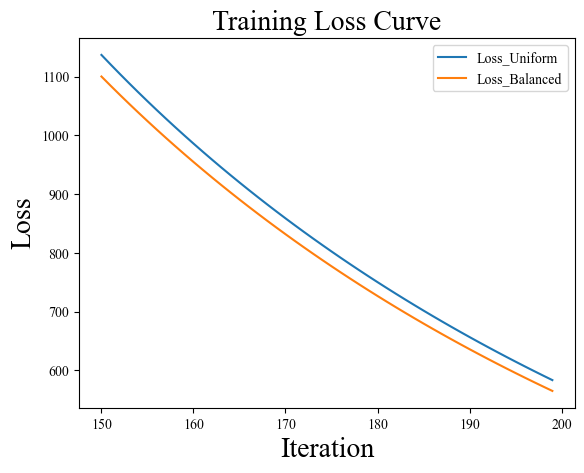}
        \caption{train loss(later stage)}
    \end{subfigure}
    \hfill
        \begin{subfigure}[t]{0.33\linewidth}
        \centering
        \includegraphics[width=\textwidth]{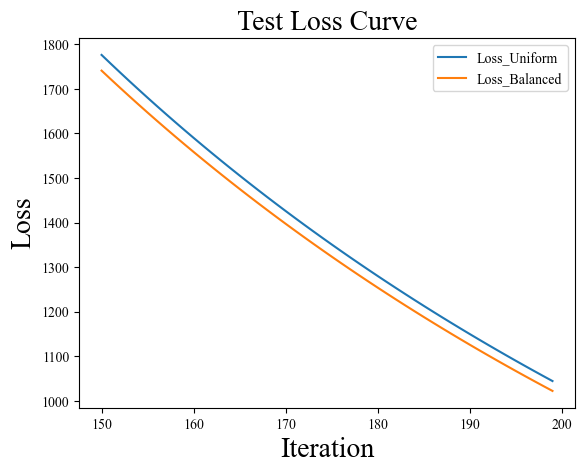}
        \caption{test loss(later stage)}
    \end{subfigure}
            \begin{subfigure}[t]{0.33\linewidth}
        \centering
        \includegraphics[width=\textwidth]{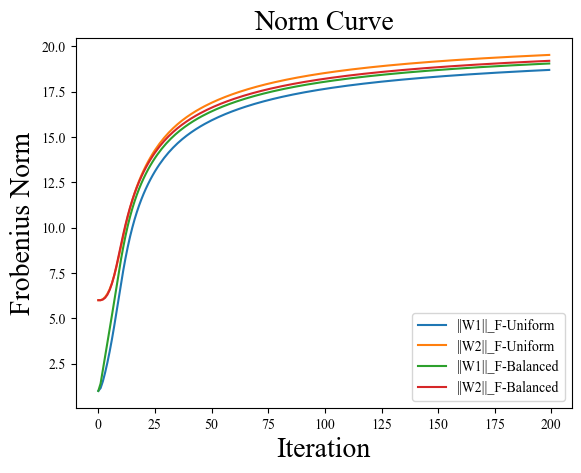}
        \caption{Frobenius norm Gap}
    \end{subfigure}
\caption{ \textbf{Insights for designing layer-wise lr scheduler.} Here we condier $\mW_1 \in \mathbb{R}^{60\times100}, \mW_2 \in \mathbb{R}^{100\times 60}$ and $\mM \in   \mathbb{R}^{60\times 60}$, with $\norm{\mW_1}_F=1$ and  $\norm{\mW_2}_F=6$ at initialization. We use 100 training samples and 20 test samples, base $lr=0.0001$, adopt the MSE loss, and train the model using gradient descent for 200 iterations.
}
\label{fig:fro-balance}
\end{figure}

\end{document}